\documentclass[final,authoryear,12pt]{elsarticle}

\usepackage{graphicx}

\usepackage{amssymb}

\begin{document}

\begin{frontmatter}


\title{Principal motion components for \\ gesture recognition using a single-example\footnote{This work was partially supported by ChaLearn: \emph{The Challenges in Machine Learning Organization,} \tt{http://www.chalearn.org/}, \normalfont{ whose directors are gratefully acknowledged. }}}
\author[label2]{Hugo Jair Escalante}\ead{hugojair@inaoep.mx}
\author[label3]{Isabelle Guyon}\ead{guyon@clopinet.com}
\author[label4]{Vassilis Athitsos}\ead{athitsos@uta.edu}
\author[label4]{Pat Jangyodsuk}\ead{pat.jangyodsuk@mavs.uta.edu}
\author[label5]{Jun Wan}\ead{09112088@bjtu.edu.cn}
\address[label2]{Computer Science Department,\\ Instituto Nacional de Astrof\'isica, \'Optica y Electr\'onica,\\
Luis Enrique Erro 1, Puebla, 72840, Mexico}
\address[label3]{ChaLearn\\
955 Creston road, Berkeley, CA 94708, USA}
\address[label4]{Computer Science and Engineering Department, \\
    University of Texas at Arlington,\\
    701 S. Nedderman Drive, Arlington, TX 76019, USA}
\address[label5]{Institute of Information Science, \\
    Beijing Jiaotong University and Beijing Key Laboratory of \\
    Advanced Information Science and Network Technology,\\
    Beijing 100044, China}
\begin{abstract}
This paper introduces \emph{principal motion components} (PMC), a new method for one-shot gesture recognition. In the considered scenario a single training-video is available for each gesture to be recognized, which limits the application of traditional techniques (e.g., HMMs). In PMC, a 2D  map of motion energy is obtained per each pair of consecutive frames in a video. Motion maps associated to a video are processed to obtain a PCA model, which is used for recognition under a reconstruction-error approach.
The main benefits of the proposed approach are its simplicity, easiness of implementation, competitive  performance and efficiency. We report experimental results in one-shot gesture recognition using the ChaLearn Gesture Dataset; a benchmark comprising more than $50,000$ gestures, recorded as both RGB and depth video with a Kinect$^{TM}$ camera.  Results obtained with PMC are competitive with alternative methods proposed for the same data set.
\end{abstract}
\end{frontmatter}


\section{Introduction}
\label{sec:intro}

Gestures are a form of non-verbal communication, which is highly intuitive and very effective. Gestures are used in a wide diversity of domains from verbal (e.g., to facilitate oral communication) and non-verbal communication, to precision surgery, security, entertainment and sports, among many others. Because of its relevance, automated gesture recognition is a research topic with a growing popularity in computer science, see e.g., \citep{HAR,gestrecogsurvey}. The availability of novel sensors, like Kinect$^{TM}$, has made even more attractive this task to researchers, as one can reliably obtain the location of body-parts in real time.

Traditional approaches for the automated recognition of gestures learn a model (e.g., a hidden Markov model~\citep{rabiner}) from a set of sample videos including the gestures of interests; where, commonly, the variation of spatial positions from body-parts (e.g., hands) across time are used as inputs for the models. In general, the more examples we have for building a model, the better its performance is in new data~\citep{arriaga2011,hmm1,hmm2,geseg2}. However, in many domains gathering examples of gestures is a time consuming and expensive process. Hence, gesture recognition methods that can learn from few examples are needed. On the other hand, it is also desirable that gesture recognition methods do not rely on specialized sensors to estimate body-part positions; or on the output of techniques for associated problems like hand-detection/tracking or pose estimation~\citep{ETH}, as these techniques may introduce noise into the data acquisition process. Undoubtedly, methods that can be trained from very few examples and using unspecialized equipment
would make the applicability of gesture recognition more widespread: e.g., anyone with access to a webcam would be able to build gesture recognizers.

In this paper we approach the problem of gesture recognition by using a single example of each gesture to be recognized.
This task, called one-shot gesture recognition, was proposed in the context of the Chalearn gesture challenge~\citep{guyon_grc,guyonwdai}. The target for this type of methods are user adaptive applications that require the recognition of gestures from arbitrary and user-defined vocabularies; domains where gestures can change with time and models need to be modified periodically; and scenarios where gathering data is too expensive or users are not willing to spend time collecting large amounts of data.

For each gesture to be recognized the only information we have for building a model is a single video recorded with a Kinect$^{TM}$ camera, where both RGB and depth videos are available. Despite the fact that Kinect$^{TM}$ can record additional data (e.g., skeleton information) it was disregarded in the ChaLearn gesture challenge. This favored the development of new methods not relying on a first step of skeleton extraction, which is often not robust to occlusions, and requires spatial and temporal resolutions not available in many application settings. The problem is restricted to single user gesture recognition, there is little variation in the background and the user is placed right in front the sensor. On the other hand, the problem is very challenging as a single example is available for each gesture, thus, traditional recognition methods (e.g., those based on HMMs) cannot be applied directly. Also, the gestures in the database used in this work were performed by different users with different skill-levels to perform the gesture; there is a wide diversity of domains of gestures, ranging from highly dynamic (e.g., \emph{``aircraft-landing''} signals) to static (e.g., \emph{``Chinese letters''}) and some body-parts may be occluded~\citep{guyon_cgd}. Additionally, the sampling rate is low (of the order of 12fps). Clearly, standard gesture recognition methods are not directly applicable, and even though the problem has been simplified, it remains a difficult task.

We propose a simple and efficient method, yet very effective, for one-shot gesture recognition called principal motion components (PMC). The main goal of PMC was to act as a strong baseline for the Chalearn gesture challenge~\citep{guyon_grc,guyonwdai,guyon_cgd} and it has inspired several of the top ranking entries, see e.g.,~\cite{wushen}. The proposed method is based on a motion map representation that is obtained by processing the sequence of frames in a video. Motion maps are used in combination with principal component analysis (PCA) under a reconstruction-error classification approach. The proposed method was evaluated in a large database with $54,000$ gestures used in Chalearn gesture challenge~\citep{guyon_grc,guyonwdai,guyon_cgd}. We compare the performance of PMC to a wide variety of techniques. Experimental results show that the proposed method is competitive with alternative methods. In particular, we found that the proposed method resulted very effective for recognizing highly-dynamic gestures, although it is less effective when static gestures are analyzed. The proposed method can be improved in several ways and it can be used in combination with other approaches, see e.g.,~\citep{wushen,prlet}. The main contributions of this work are threefold.
\begin{itemize}
\item The introduction of a \emph{new representation for motion} in video, where we capture motion in successive frames through 2D motion maps. The proposed representation can be seen as a \emph{bag-of-frames} formulation, where each video is characterized by the (orderless) set of motion maps it contains. The representation can be used with other methods for gesture recognition and it can be used for other tasks, e.g., for segmentation purposes using motion detection.

\item The proposal of a \emph{new one-shot gesture recognition approach based on PCA}. Motion maps for a video (in a bag-of-frames representation) are used to generate a PCA model. The reconstruction error of the PCA models is used as criterion for gesture recognition. Our proposal is capable of building a predictive PCA model from a single video without using any temporal information.

\item The \emph{evaluation of the proposed method in a large-scale heterogeneous database} and a comparison of it with a variety of alternative techniques. We show that the proposed method is effective for highly dynamic gestures. Several variants of the bag-of-frames representation (including representations based on HOG, HOF, STIP features) and different recognition techniques (classifiers and template methods) are considered in our study.

\end{itemize}

The rest of this paper is organized as follows. The next section reviews work closely related to our proposal. Section~\ref{sec:mpca} introduces the principal motion components method. Section~\ref{sec:experiments} describes the experimental settings adopted in this work and Section~\ref{sec:results} reports experimental results. Finally, Section~\ref{sec:conclusions} presents the conclusions derived from this paper and outlines future work directions.

\section{Related work}
\label{sec:rw}

This section reviews related work on two key components of the proposed approach: motion-based representations and PCA-based recognition.

\subsection{Motion-based representations}
When it is not possible to track body-parts across a sequence of images, motion-based representations have been used for gesture recognition. Different approaches have been proposed, mainly based on template matching (\cite{HAR,reviewHM}). The seminal work of~\cite{MHI} used motion history images (MHIs) to represent videos, where MHIs are obtained by accumulatively adding (thresholded) binary-difference images, this type of templates reveal information about the history of motion in a video (i.e., how movement happened). Statistical moments obtained from the MHIs were used for recognition.
\cite{hierarchMotion} extended the MHI representation to generate histograms of motion orientation.
The MHI is obtained for each video and the resulting template is divided into spatial regions. Gradients from motion values are obtained on each region separately, a histogram is generated per each region using as bins a set of predefined orientations over the gradients.
Per-region histograms are concatenated to obtain a 1D representation for each video. A similarity-based approach was used for recognition in that work.

\cite{lowlevelMotion} represented a sequence of images by a spatiotemporal template. As preprocessing, the object of interest is isolated from the rest of the scene. Then, the sequence of cropped frames is processed to obtain optical flow fields. Flow frames are divided into a spatial grid and motion magnitudes are added in each cell. \cite{motionVectors} obtained motion vectors (correspondences between blocks of pixels in adjacent frames) for successive frames and generated a 2D motion histogram, in which the occurrence of motion vectors is quantized. They used this representation for gesture recognition in a very small data set of \emph{easy} gestures.
\cite{PCRM} proposed a representation, called Pixel Change Ration Map (PCRM), based on motion histograms that account for the occurrence of specific values of motion in the video sequence. That is, the bins correspond to different (normalized) motion values. This approach is very similar to \cite{hierarchMotion}. However, under PCRM, the average of motion energy in cells of the grid are used instead of the orientation of gradients. The representation proved to be very effective for video retrieval, clustering and classification. \cite{keyframe} proposed a method for key frame extraction from video shots. The core of the method is a representation based on motion histograms. Optical flow fields are obtained for each frame, a subset of different combinations of magnitude and direction of motion values are used as the bins of the motion histogram. Motion histograms, one per-frame, are then processed to extract representative frames of the sequence.

Other approaches define motion histograms in terms of symbols derived from optical flow analysis (\cite{HoFPRL}); build classification models using motion histograms over voxels as features (\cite{boostmotion}); and generate histograms of gradient orientations for static gesture recognition (\cite{OriHistHand}).

In most of the above described approaches, a single template based on motion histograms is obtained to represent a whole sequence of frames. In our proposed representation, a motion map, accounting for the spatial distribution of motion across successive frames, is obtained per each difference image. This can be thought of as a relatively low resolution 2D map, each location accounting for the amount of motion at a given position, at a given time. However, we discard the time ordering of the various maps and time is only taken into account by the fact that the maps are based on consecutive frame differences. Thus, by analogy to bag-of-words representations in text recognition that ignore word ordering in text, we can talk of a ``bag-of-frame'' type of representation, which is neither a template nor a time ordered sequence of features. In this way, we have a set of observations (motion maps) associated with a single-gesture, which can be used for the induction of classifiers. To the best of our knowledge none of the above described methods has been evaluated in one-shot gesture recognition~\cite{guyon_grc,guyonwdai}.

In the context of one-shot learning gesture recognition, template-based methods have been popular.
A simple average template approach was the first baseline proposed by the organizers of the gesture recognition challenge, and it remained a difficult baseline to beat during the first weeks of the competition~\citep{guyon_grc}. \cite{mahbu} proposed a template matching approach for one-shot learning gesture recognition, where three ways of generating templates were proposed (2D standard-deviation, Fourier-transform and MHIs). For recognition the authors used the correlation coefficient to compare templates and testing videos. \cite{wuetal} proposed an extended MHI that incorporates gait energy information and inverse recording, although the method obtained very good performance, it is difficult to assess the contribution of the sole recognition approach as several pre-processing steps were performed beforehand (the authors mention that pre-processing improves the performance of their method by about $9\%$).
Other methods have been proposed in the context of the gesture recognition challenge, including probabilistic graphical models~\citep{manavender} and techniques from manifold learning~\cite{YML}, these and other methods are summarized by~\cite{guyon_grc,guyonwdai}. In Section~\ref{sec:results} we compare the performance of our proposal to these methods.

\subsection{PCA for gesture recognition}
\label{sec:recogpca}
The second component of our proposal is a PCA-based method for gesture recognition. PCA has been widely used in many computer vision tasks, including gesture recognition (\cite{HAR,lowlevelMotion,Depthsillouthetes}). In most of the times, PCA has been used to reduce the dimensionality of the representation or to eliminate noisy and redundant information, see e.g.,~\citep{aslpca}; in fact, this is a common preprocessing step when facing any machine learning task~\citep{guyon_fs}.

Some authors have used PCA for recognition~\citep{eigenfaces,jemartin,Gomez,PCAOBDET}. The most used approach consists of estimating the reconstruction error obtained after projecting the data into a PCA model as a measure of the likelihood that an instance belongs to a class. This recognition method was first reported in the seminal work of~\cite{eigenfaces} for face recognition. A similar approach was adopted by~\cite{jemartin} to classify hand postures to be used for gesture recognition by a high-level approach. \cite{jemartin} used a large data set of images with diverse hand postures and used the PCA-reconstruction approach to classify hand postures. This approach has proved to be very effective in other domains as well (e.g., spam filtering,~\citep{Gomez}, and pedestrian detection,~\citep{PCAOBDET}). The reconstruction approach based on PCA has been also used for one-class classification and outlier detection~\citep{Tax,kpca}.

The motivation behind using a reconstruction-error approach for one-shot recognition stems from the fact that we do not know what are the underlying motion dimensions associated to a particular gesture, and we would like PCA to automatically determine what are those dimensions and to use such information for recognition. One should note that previous work has used the PCA-reconstruction approach considering a data set of labeled instances, where many instances are available per each class. In our proposal, we have multiple observations taken from a single instance associated to a class (the bag-of-frames for a gesture). Another way of thinking of our model is that it acts as a single state hidden Markov model for each gesture, the PCA model representing an i.i.d. generating process.
To the best of our knowledge PCA has not been used similarly for recognition, not even for other tasks than gesture recognition.

\section{Principal motion components}
\label{sec:mpca}
In general terms, the proposed principal motion components (PMC) approach to one-shot gesture recognition is as follows.
The set of motion maps, i.e., the bag-of-frames representation, associated to a video is processed to generate a PCA model per each gesture. When a new gesture needs to be recognized, its motion maps are extracted and projected into the PCA model for each gesture, the projected data are reconstructed back and we measure the average of reconstruction error.
The underlying idea is that a PCA model can capture the important motion variation across frames, and therefore, motion maps obtained from the same gesture will be better reconstructed than with models associated with different gestures.
Unintentional movements that are not directly related to the user will vanish with the reconstruction performed by PCA.

\subsection{Representation: motion maps (bag-of-frames)}
\label{sec:mothis}
Let $\mathcal{V}$ be a video composed of $N$ frames, $\mathcal{V} = \{I_1, \ldots, I_N\}$, where $I_i \in \mathbb{R}^{w \times h}$ is the $i^{th}$ frame, $w$ and $h$ being the width and height of the image, respectively.
We represent a video by a set of motion energy maps, $H_1, \ldots, H_{N-1}$, $H_j \in \mathbb{R}^{N_b}$, one per each frame.
Each map accounts for the movement taking place in consecutive frames on fixed spatial locations of the frames.

\begin{figure}[!htb]
    \includegraphics[width=3cm,height=3cm]{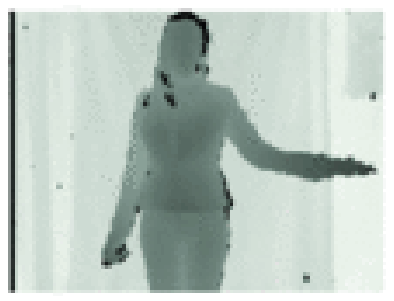}
    \includegraphics[width=3cm,height=3cm]{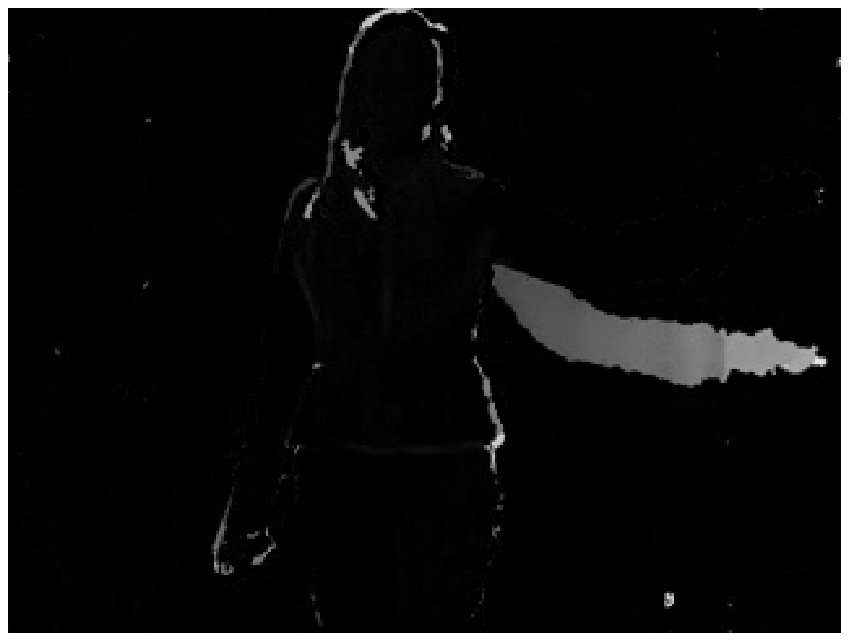}
    \includegraphics[width=3cm,height=3cm]{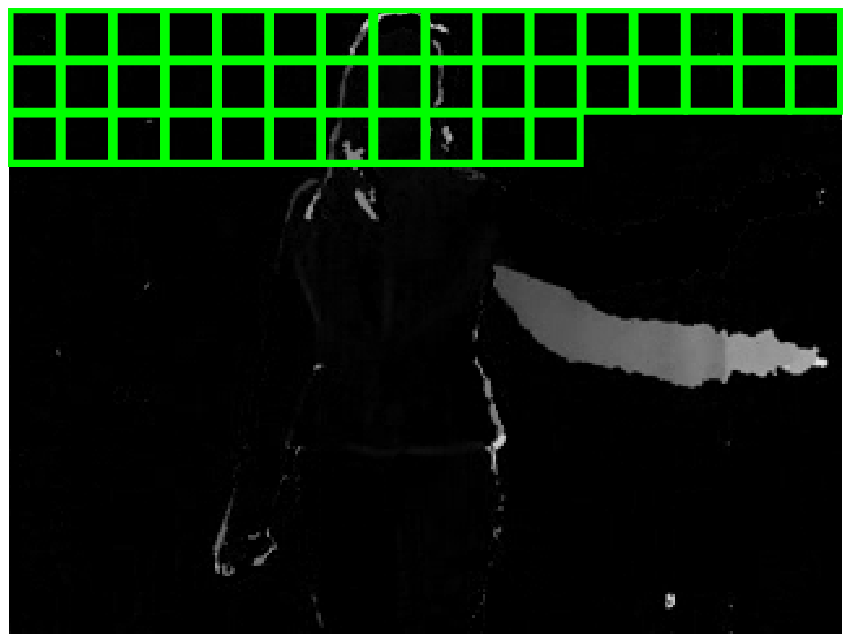}
    \includegraphics[width=3.5cm,height=3cm]{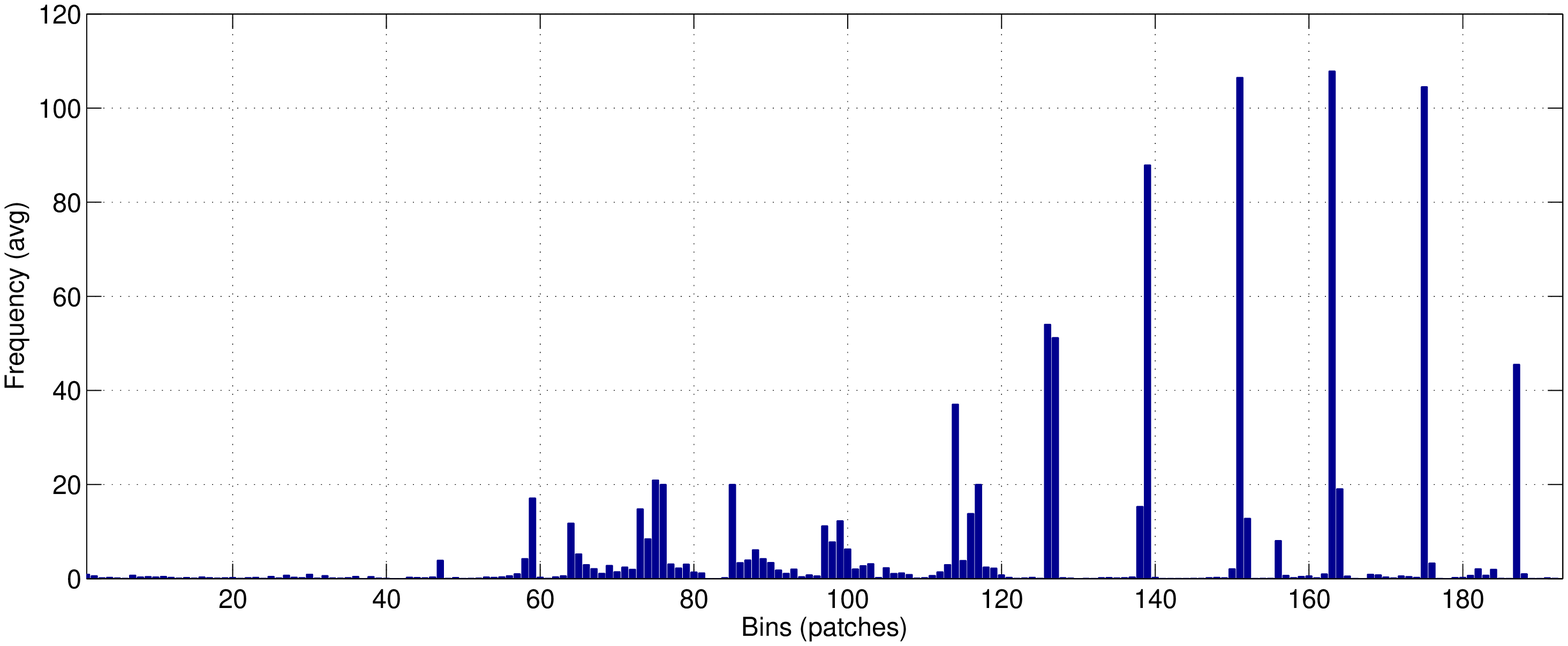}
    \caption{Extraction of motion maps: from left to right, (1) a frame from a video, (2) difference image of consecutive frames, (3) part of the grid defined over the difference image, (4) motion map associated to the difference image (each patch from the grid corresponds to a bar in this plot, the value depicted by each bar is the average of motion in the corresponding patch).}\label{fig:grid}
\end{figure}

\begin{figure}[!htb]
  \includegraphics[width=2.1cm]{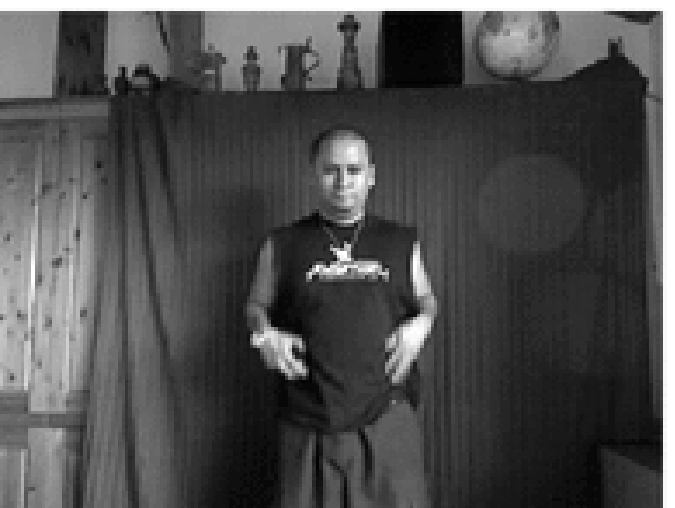}
    \includegraphics[width=2.1cm]{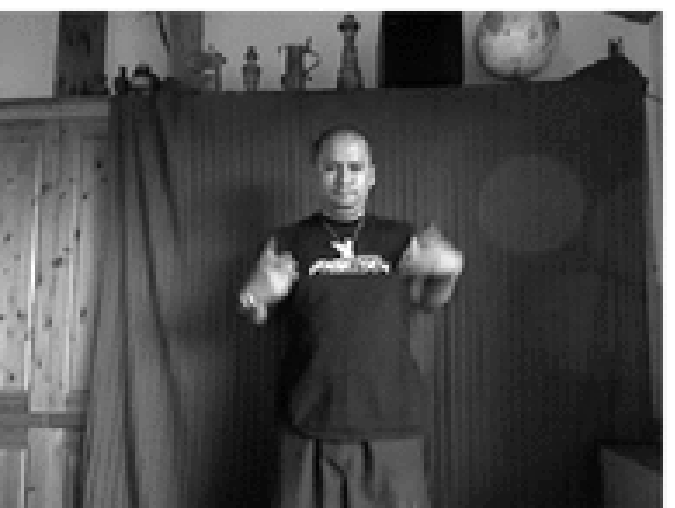}
    \includegraphics[width=2.1cm]{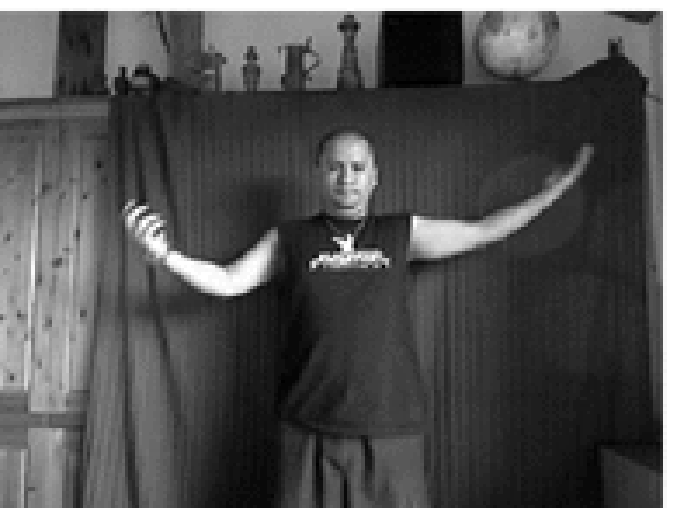}
          \includegraphics[width=2.1cm]{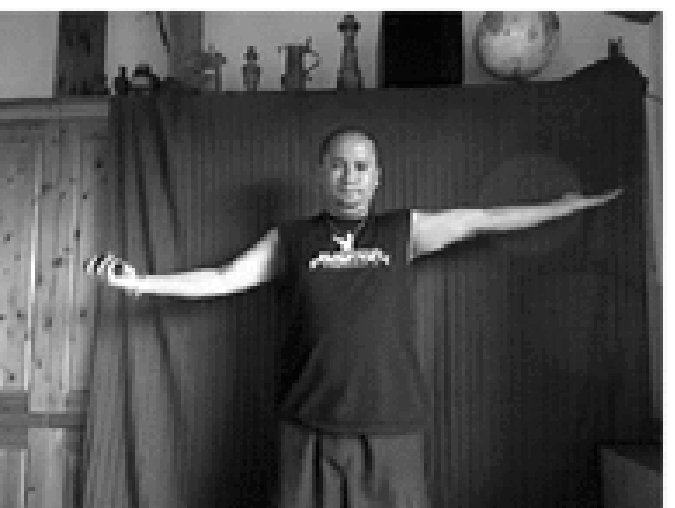}
   \includegraphics[width=2.1cm]{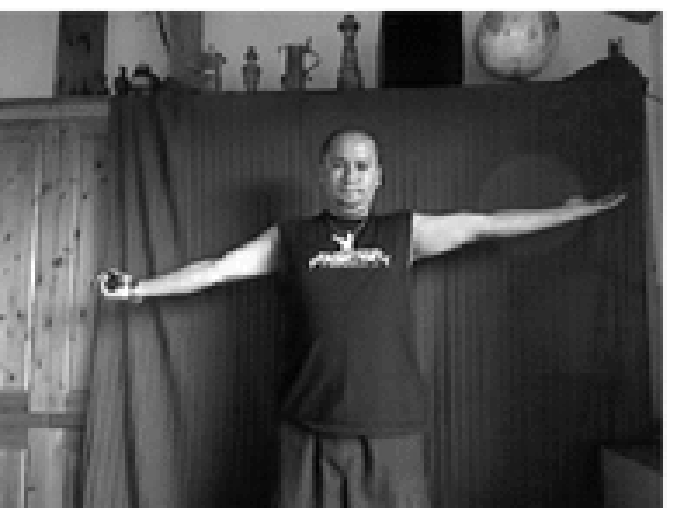}
         \includegraphics[width=2.1cm]{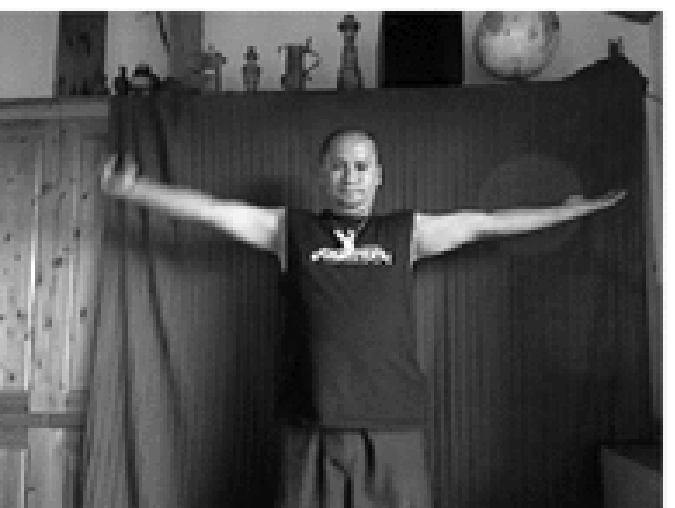}\\
  \includegraphics[width=2.1cm]{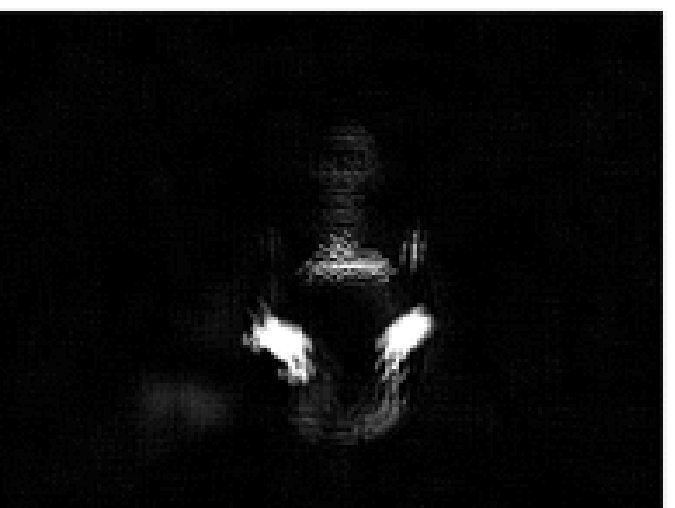}
    \includegraphics[width=2.1cm]{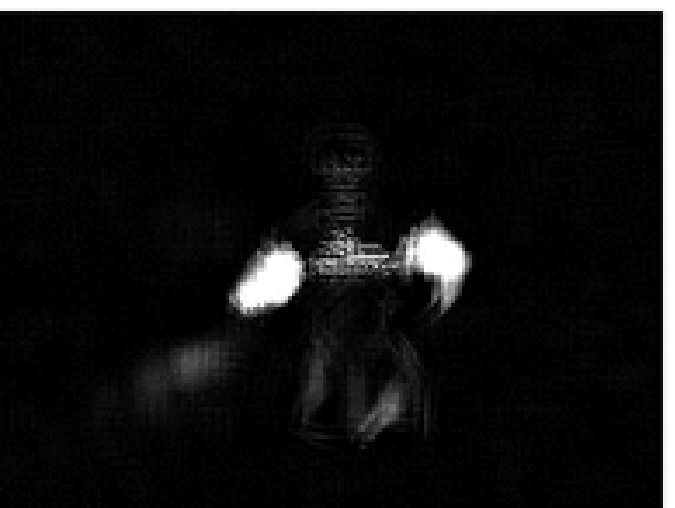}
    \includegraphics[width=2.1cm]{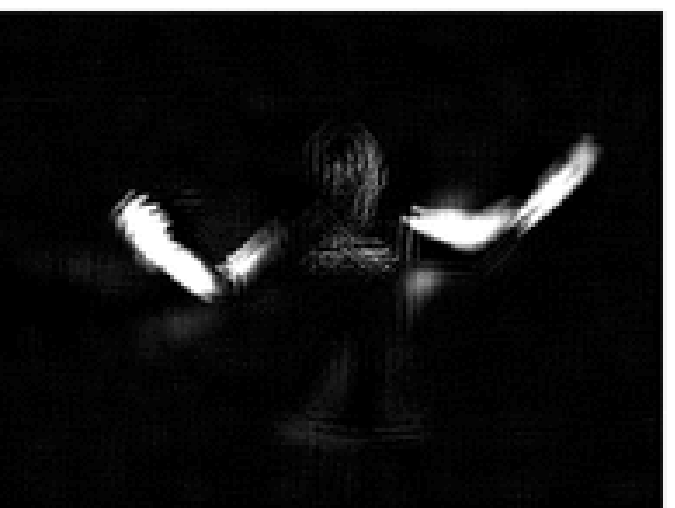}
          \includegraphics[width=2.1cm]{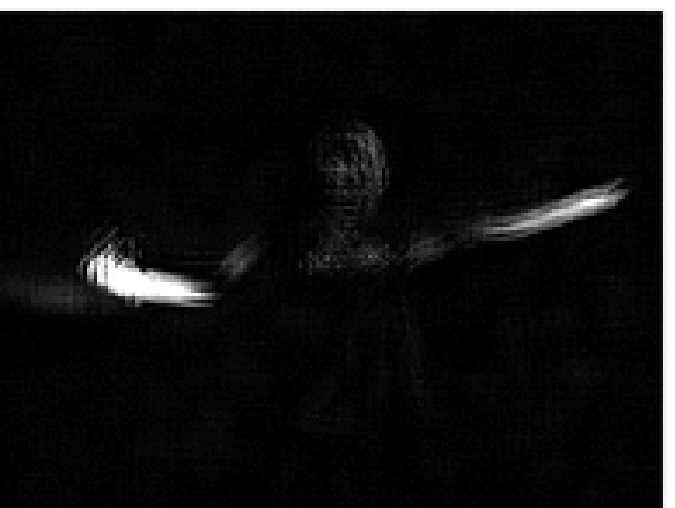}
   \includegraphics[width=2.1cm]{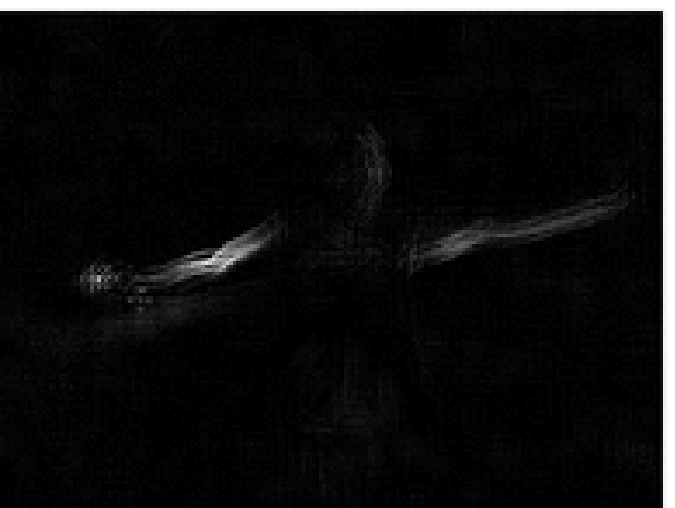}
         \includegraphics[width=2.1cm]{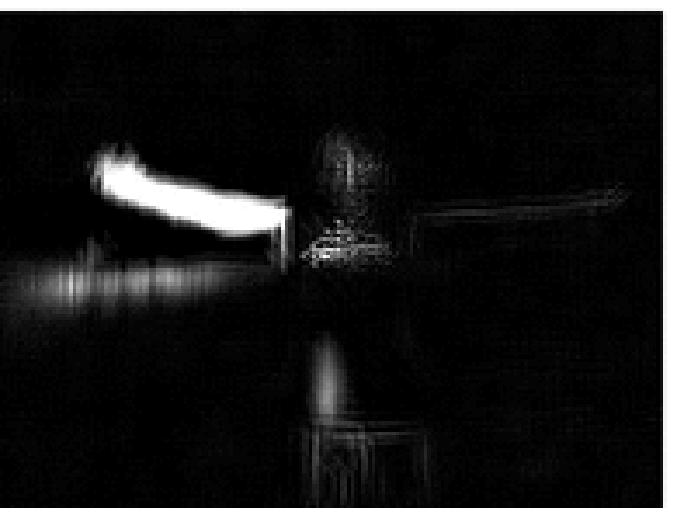}\\
                    \includegraphics[width=2.1cm]{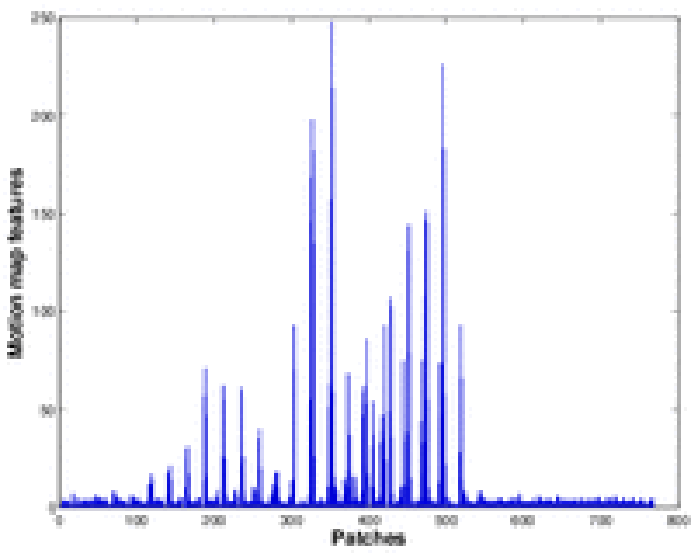}
    \includegraphics[width=2.1cm]{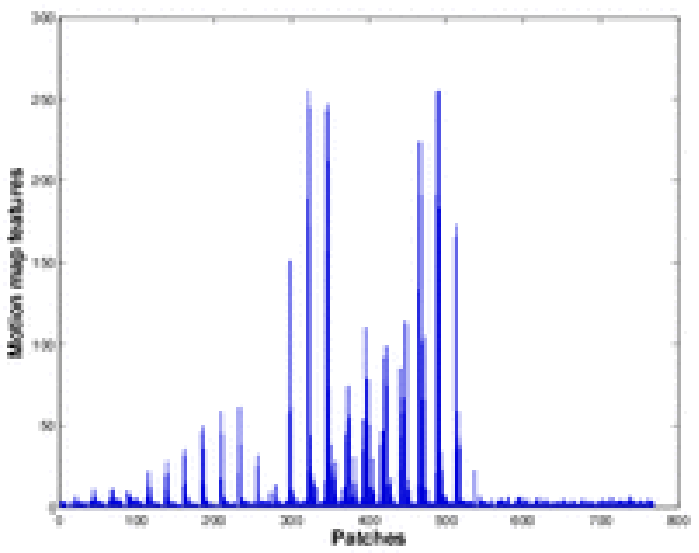}
    \includegraphics[width=2.1cm]{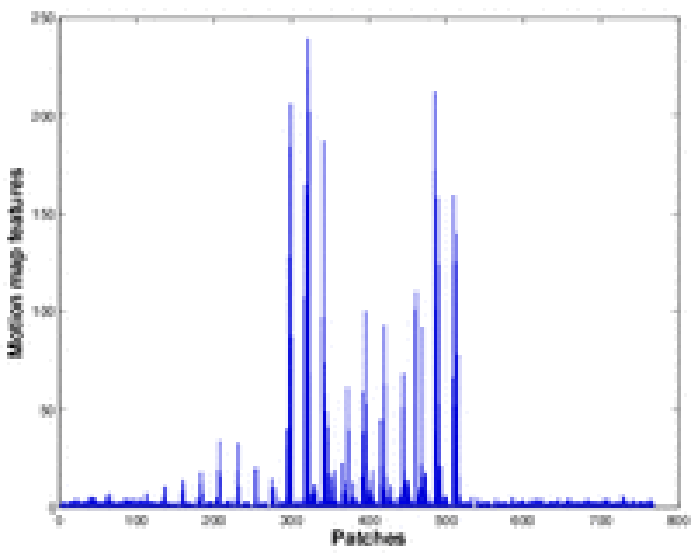}
          \includegraphics[width=2.1cm]{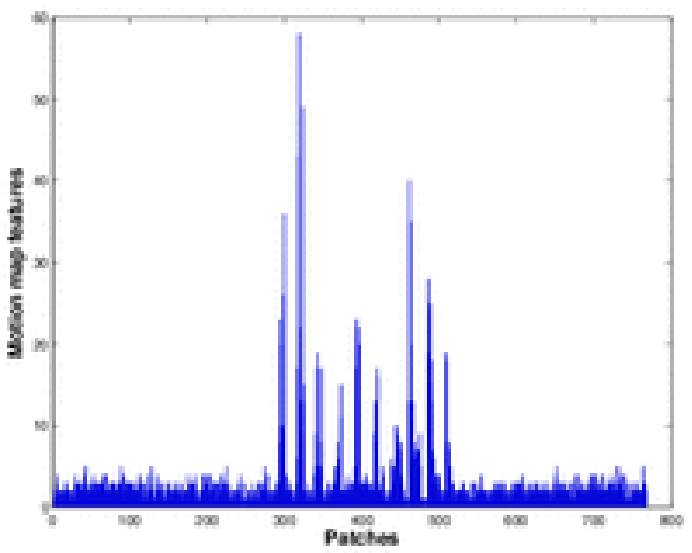}
   \includegraphics[width=2.1cm]{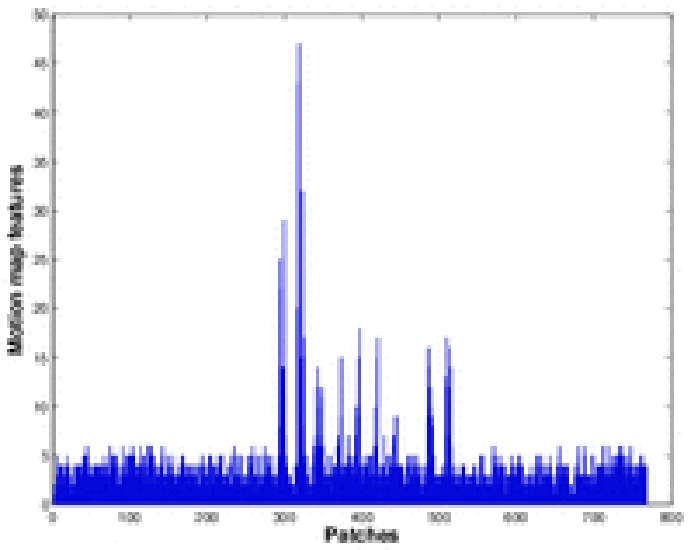}
         \includegraphics[width=2.1cm]{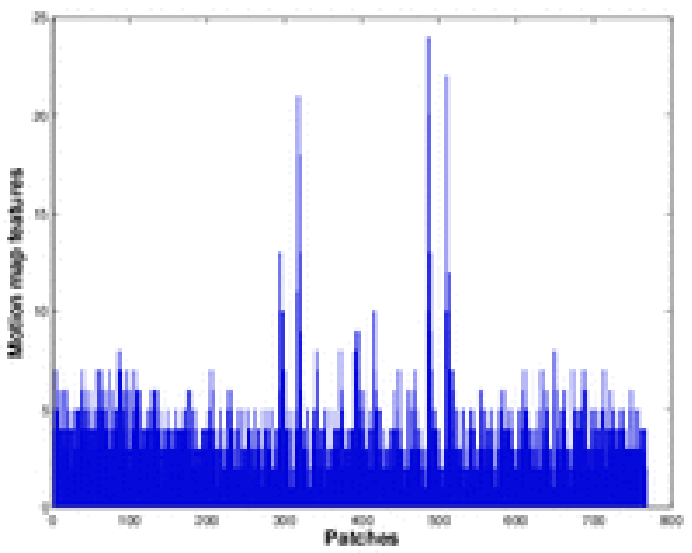}\\
  \caption{Extraction of motion maps for a video. From top to bottom: frames from the video; difference images; 1D motion maps. The matrix of all motion maps for a video can be seen in the left images of Figure~\ref{fig:mohist}.}\label{fig:extrafig}
\end{figure}

For obtaining motion maps, we first generate motion energy images by subtracting consecutive frames in the video: $D_{i} = I_{i+1} - I_{i}$, $i = \{2, \ldots, N-1\}$ (we set $D_{1}=0$ to have the same number of difference images as frames in the video). Next, a grid of equally spaced patches is defined over the difference images.  The size of the patches is the same for all of the images, see Figure~\ref{fig:grid}. We denote with $N_b$ the number of patches in the grid. We estimate for each difference image $D_i$, 
the average motion energy in each of the patches of the grid; this is done by averaging motion values for pixels within each patch. That is, we obtain a 2D motion map for each difference image, where each element of the map accounts for the average motion energy in the image in the corresponding 2D location. The 2D maps are transformed into a 1D vector $H_i \in \mathbb{R}^{N_b}$. Hence, each video $\mathcal{V}_i$ is associated to a matrix $\textbf{H}_i$ of dimensions $N-1 \times N_b$, with one row per frame and one column per patch. We call $\textbf{H}_i$ the bag-of-frames representation for the video, under the motion maps characterization. Figure~\ref{fig:extrafig} shows motion maps for a subset of frames in a video. In the figure motion maps are shown in temporal order, although, in the proposed approach, order of motion maps is not taken into account.

For the implementation we adopted a more efficient approach to generate motion maps. Each motion energy image $D_{i}$, $i=\{1, \ldots, N-1\}$ is downsized (e.g., via \emph{cubic interpolation}) up to a specified scale $\gamma$. Motion maps are obtained by concatenating the rows from the downsized images.

One should note that as the proposed representation captures motion in fixed spatial locations, translation variations may have a negative impact into the motion maps representation. The extreme case is when considering a large number of patches (e.g., when having one bin per pixel), resulting in a fine-grained map for which translation variance is a critical issue. In order to overcome this problem, we expand motion information in each difference image $D_i$ as follows: $D_i = D_i + D^l_i + D^r_i + D^u_i + D^d_i$.  Where $D^l_i$,  $D^r_i$,  $D^u_i$,  and $D^d_i$ are difference images $D_i$ translated by a gap of $\tau-$pixels to the left, right, up, and down directions, respectively. Basically, were are growing the region of motion to make the representation less dependant on the position of the user with respect to the camera.

\begin{figure}[!htb]
\includegraphics[height=3.4cm]{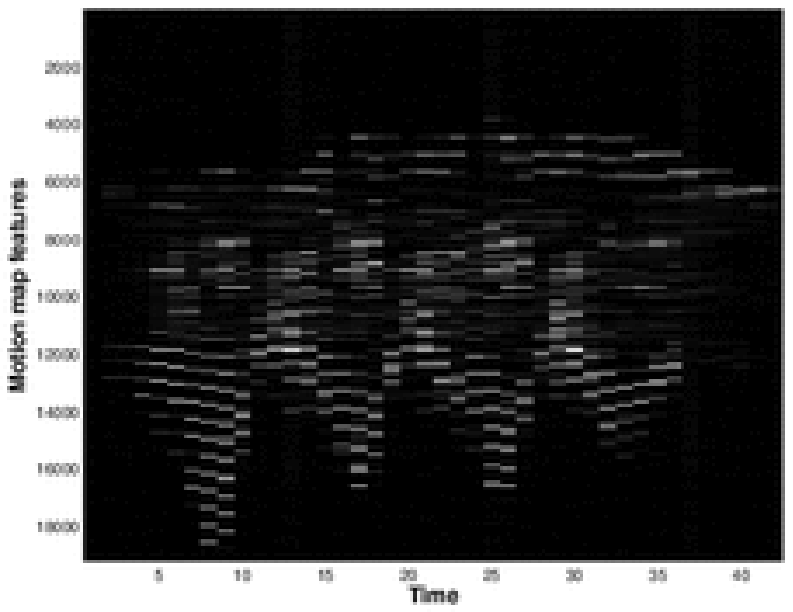}\includegraphics[height=3.4cm]{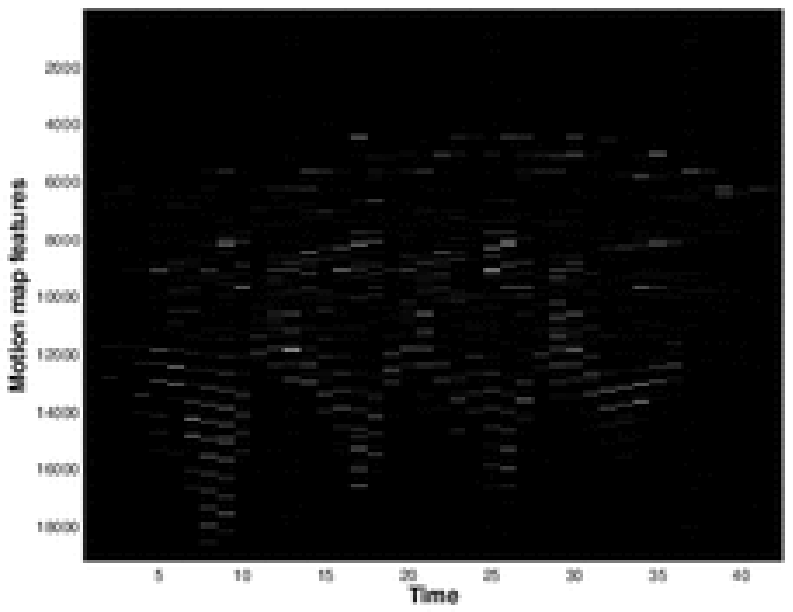}\includegraphics[height=3.4cm]{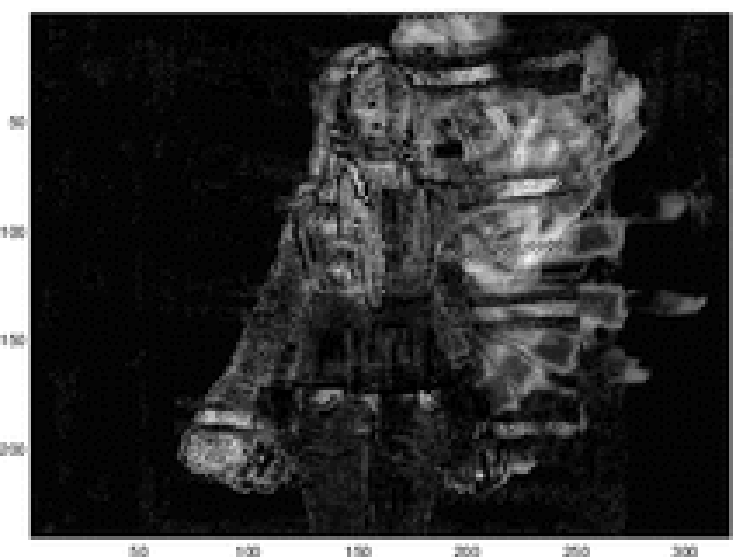}
\includegraphics[height=3.4cm]{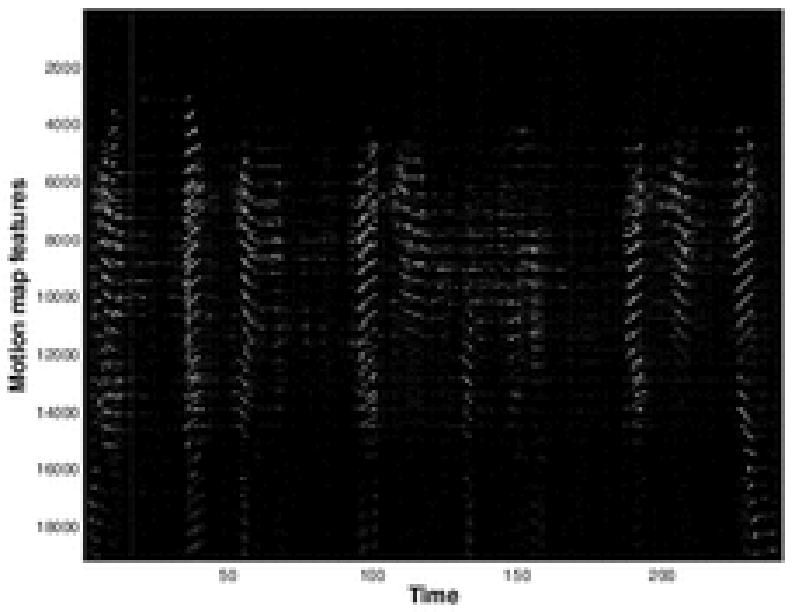}\includegraphics[height=3.4cm]{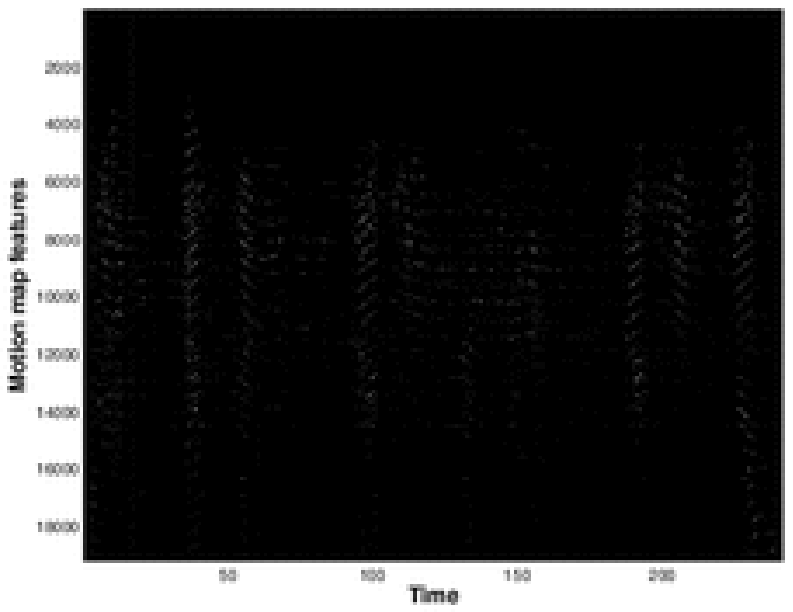}\includegraphics[height=3.4cm]{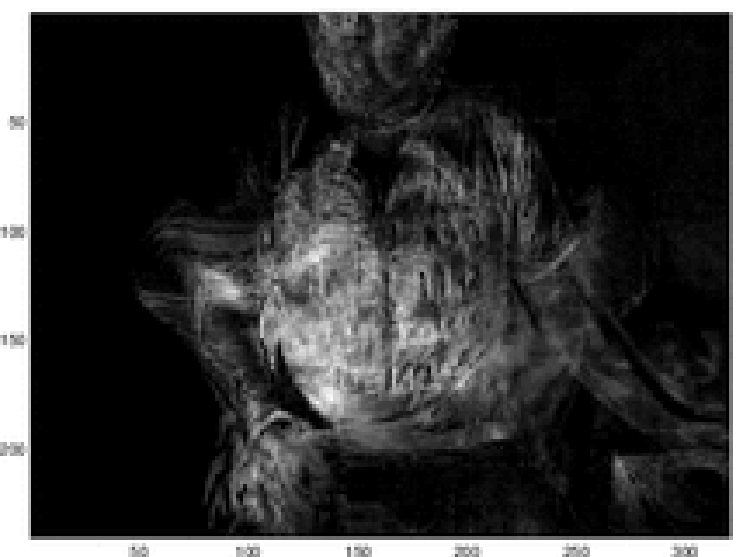}
\caption{Motion maps for selected gestures taken from two different vocabularies. We show motion maps with (left) and without motion expansion (center), together with the corresponding MHI (right). The $x-$axis of the images show the motion maps for the different frames.
The domains of the depicted gestures are: \emph{``Canada-aviation ground circulation''} (top), and \emph{``Gang hand-signals''} (bottom).
}\label{fig:mohist}
\end{figure}

Figure~\ref{fig:mohist} shows motion maps extracted from videos depicting different gestures and performed by different persons; row 1 shows a very dynamic gesture, whereas row 2 shows a static one. We can see that motion information is effectively captured by the proposed representation, as expected, the more dynamic the gesture (as depicted in the accompanying MHIs) the higher the values of the motion map. It is interesting that even the representation for the static gesture shows high motion energy values, which can be due to unintentional movement from the user that is not related to the gesture. The PCA model is expected to capture the main dimensions of motion and to limit the contribution of such noisy movements. From Figure~\ref{fig:mohist} we can also see that the motion expansion emphasizes motion energy in neighboring patches (compare the leftmost and center images), which makes the representation more robust against variance in translation.

\subsection{Recognition: PCA-based reconstruction}

For recognition we consider a reconstruction-error approach based on PCA. 
Consider a training video representing a single gesture. We first compute a bag-of-frames representation $H_1, \ldots, H_{N-1}$, (alternatively denoted by matrix $\textbf{H}_i$), as explained in the previous section. Here $n = 1, \ldots, N-1$ does NOT represent a time index and the frames representing motion (converted in feature vectors) can be arbitrarily re-ordered. The modeling approach then consists in treating the $H_n$ feature vectors as training examples of a PCA model, globally representing the frames of that gesture. The principal components can be thought of as ``principal motions''. Given now a new video also in a bag-of-frames representation, its similarity to the training video can be assessed by the average reconstruction error of the frames of the video under the PCA model.

Let $\mathbb{V} = \{\mathcal{V}_1, \ldots, \mathcal{V}_K\}$ be the set of videos corresponding the a gesture vocabulary (e.g., \emph{``diving signals''}), where each video corresponds to a gesture (e.g., \emph{``out of air''} gesture).  We apply PCA to each of the bag-of-frames representations $\textbf{H}_1, \ldots, \textbf{H}_K$ associated to the different training videos in $\mathbb{V}$.  We center each matrix $\textbf{H}_i$ and apply singular value decomposition: $\textbf{H}_i=\textbf{USV}$, we store the top $c$ singular values $\textbf{S}_c$ from $\textbf{S}$ together with the corresponding eigenvectors $\textbf{V}_c$ (i.e., the principal components), where $\textbf{V}_c$ is the matrix formed by the first $c-$ columns of $\mathbf{V}$.
Hence for each gesture in the vocabulary we obtain a PCA model represented by the pair $(\textbf{S}_c, \textbf{V}_c)_{\{1, \ldots, K\}}$.
\begin{figure}[!htb]
  \includegraphics[width=2.1cm]{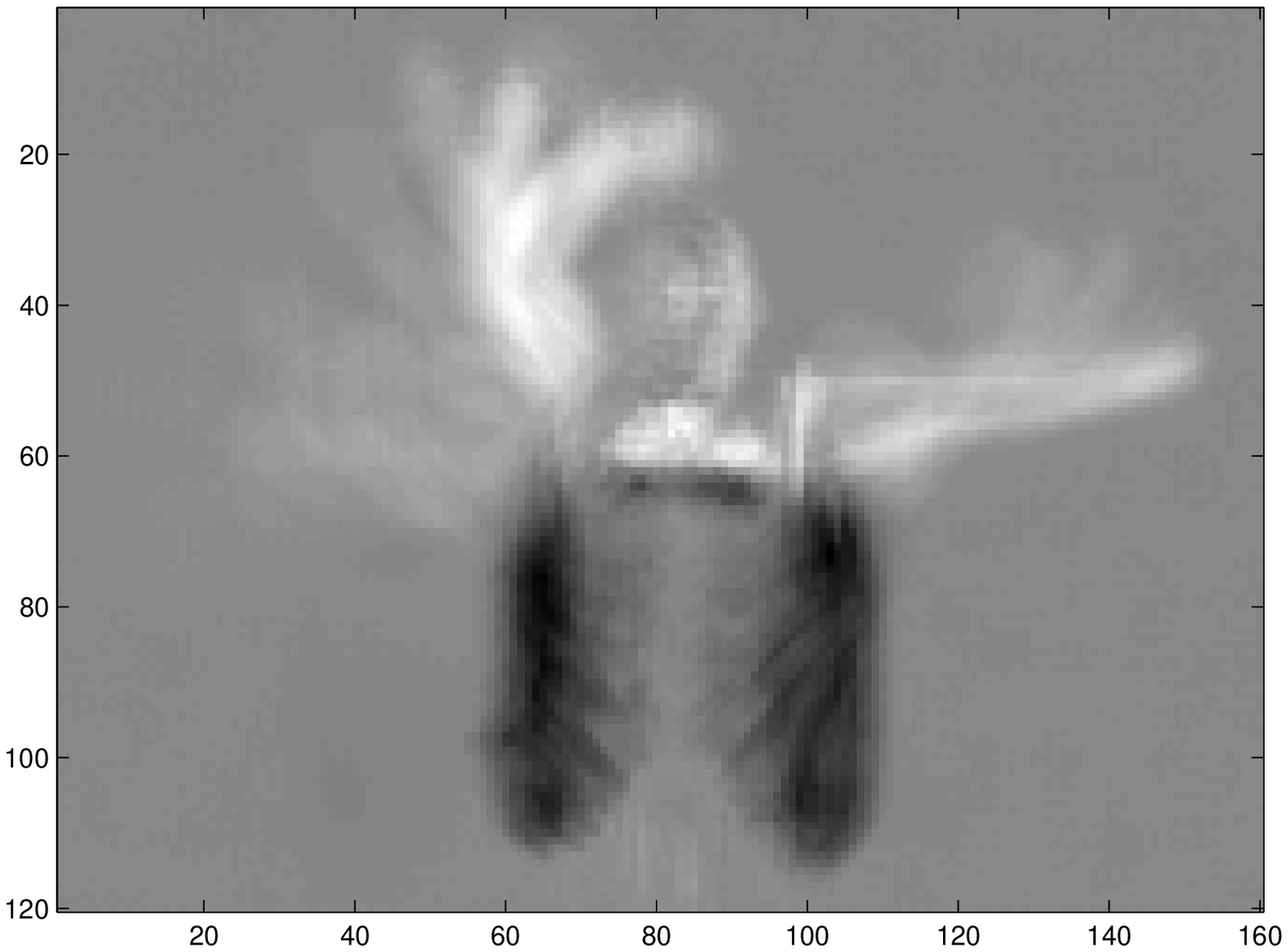}
  \includegraphics[width=2.1cm]{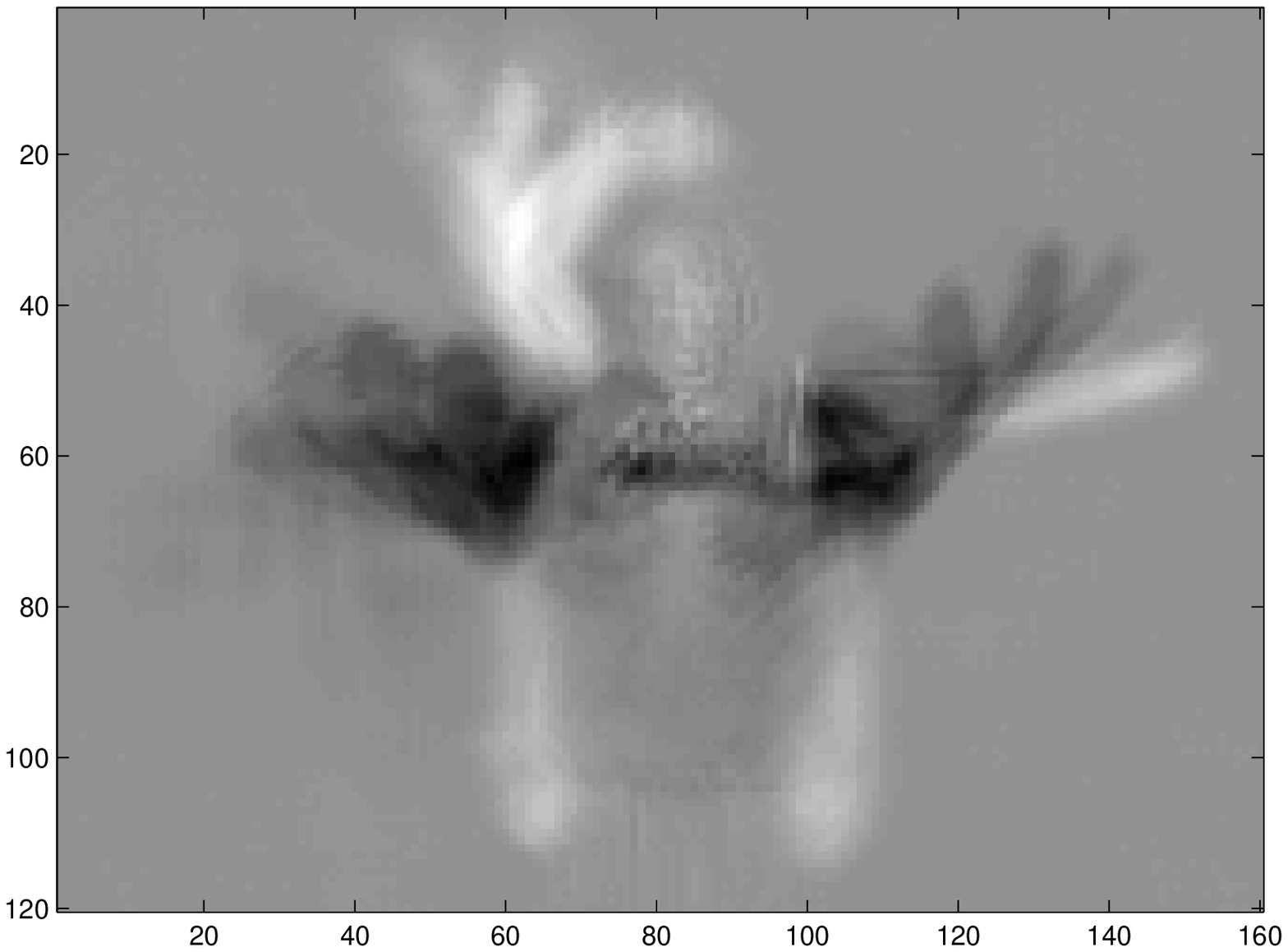}
  \includegraphics[width=2.1cm]{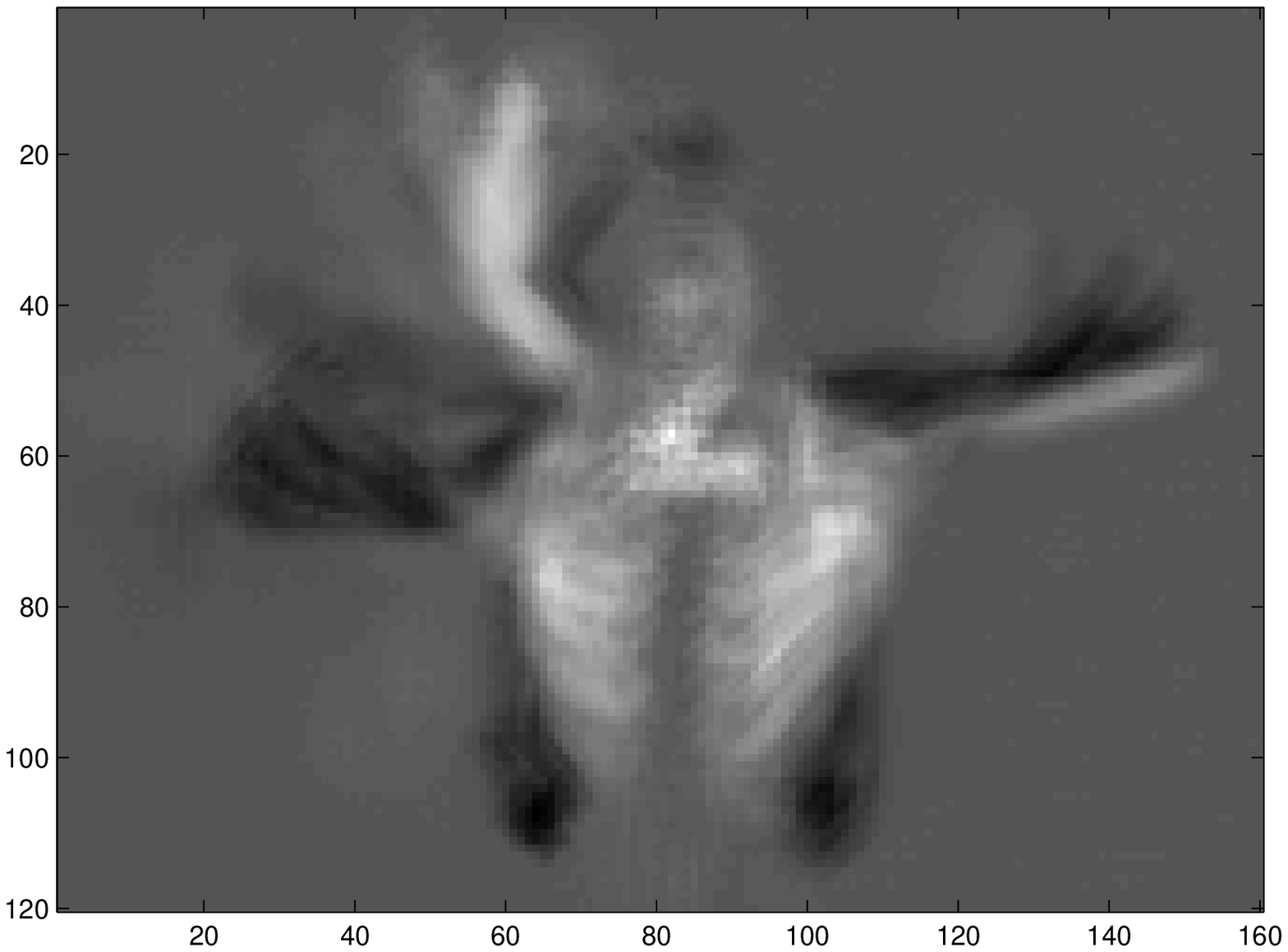}
  \includegraphics[width=2.1cm]{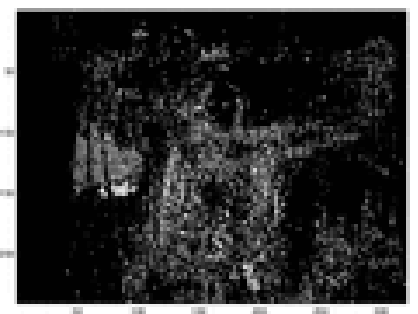}
    \includegraphics[width=2.1cm]{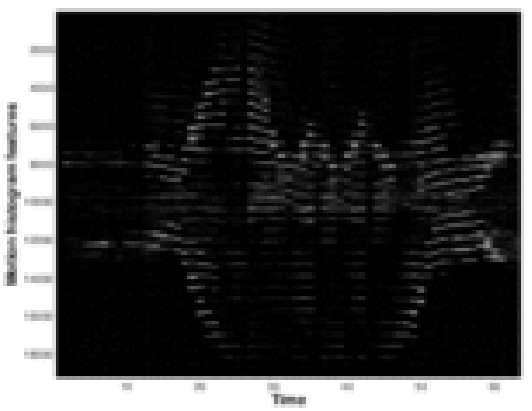}
   \includegraphics[width=2.1cm]{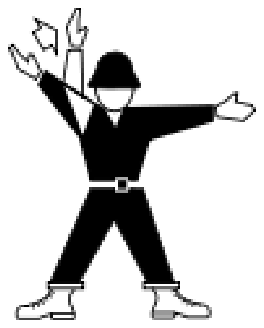}\\
   \includegraphics[width=2.1cm]{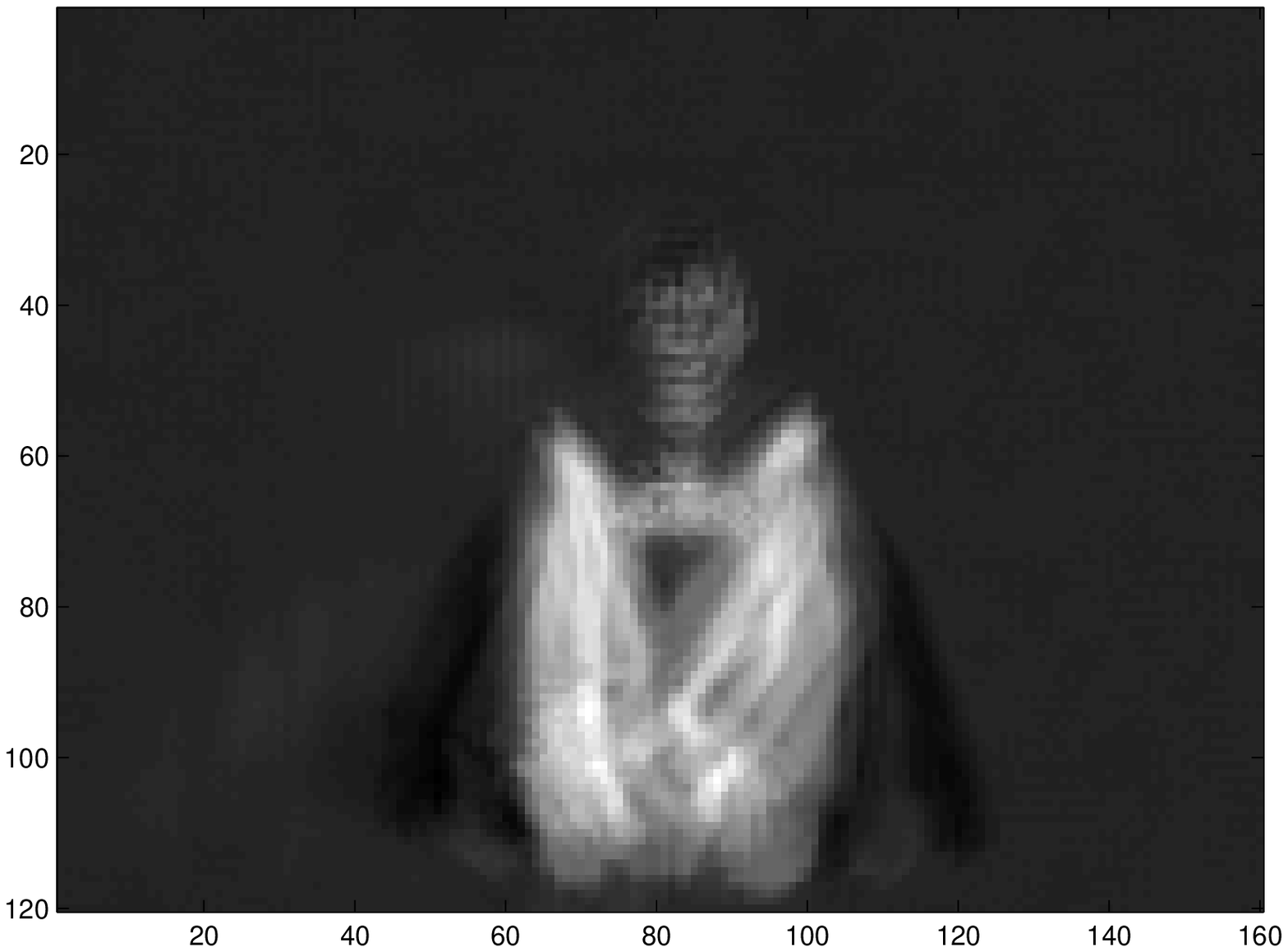}
  \includegraphics[width=2.1cm]{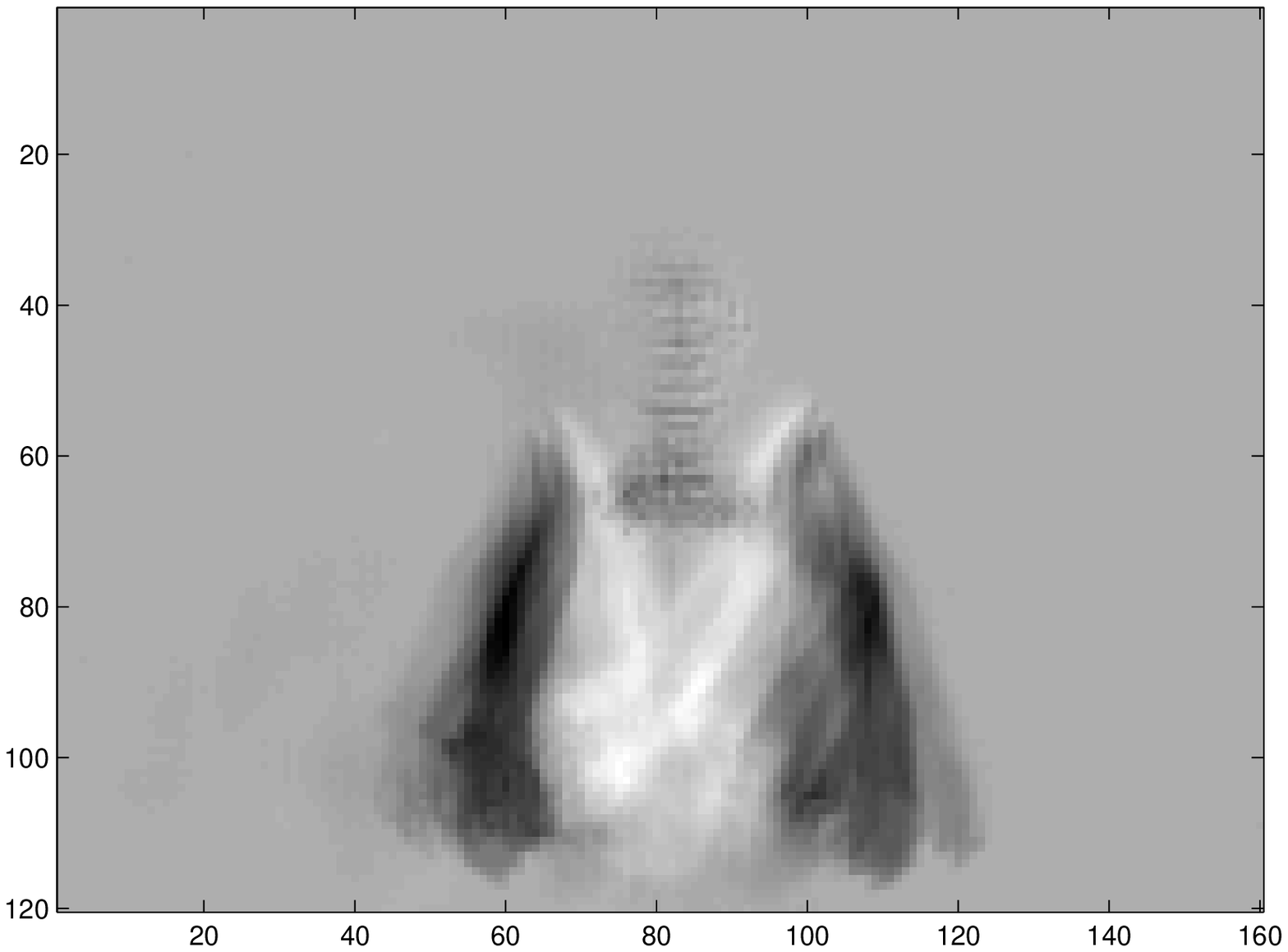}
  \includegraphics[width=2.1cm]{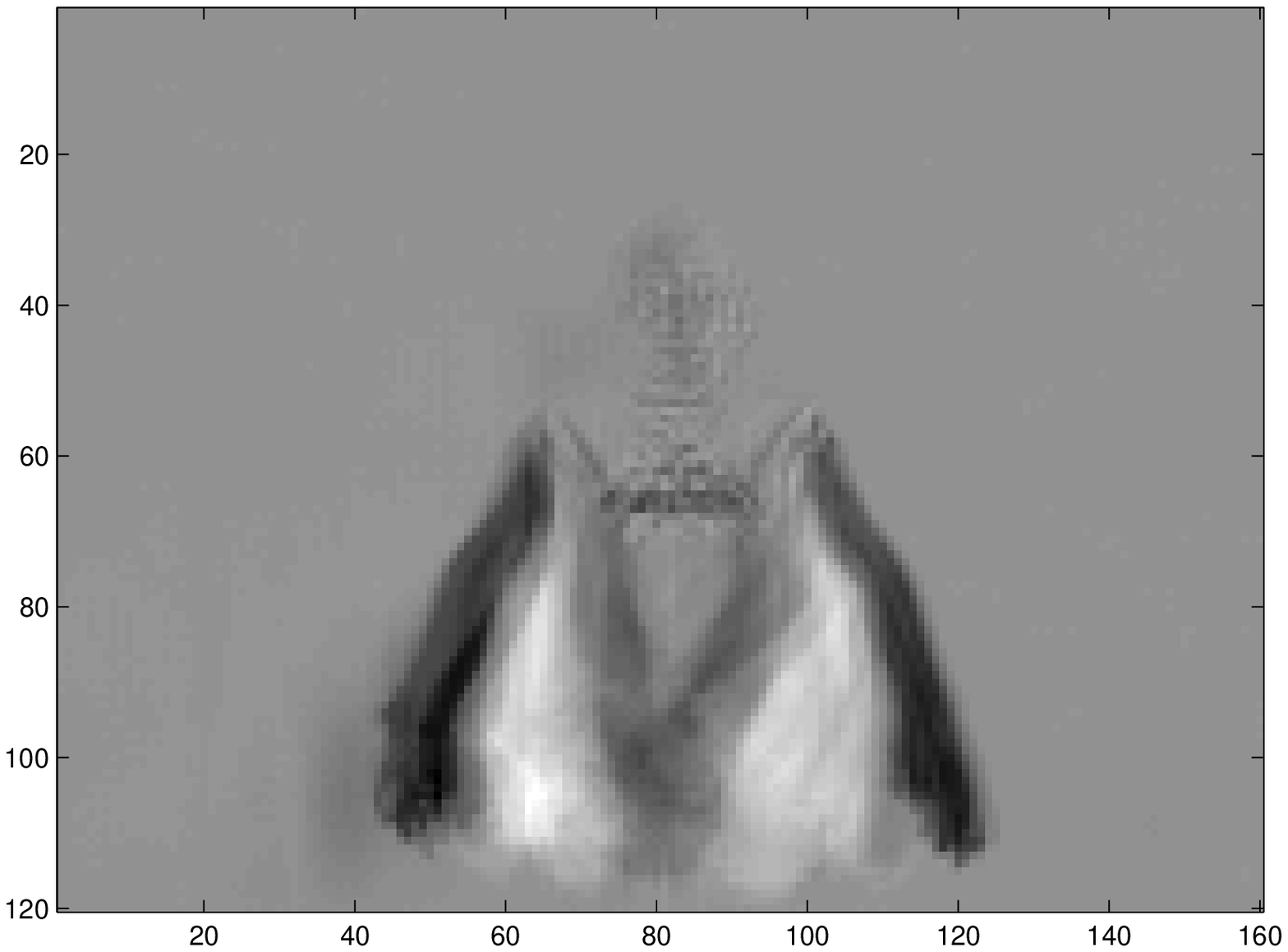}
    \includegraphics[width=2.1cm]{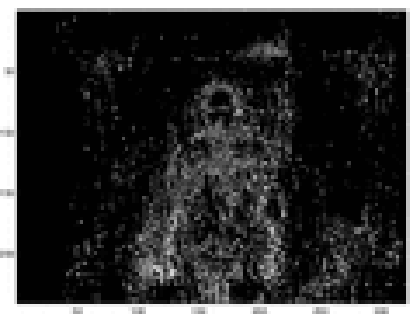}
    \includegraphics[width=2.1cm]{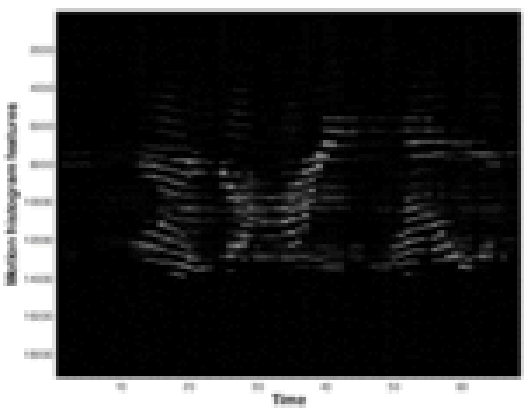}
    \includegraphics[width=2.1cm]{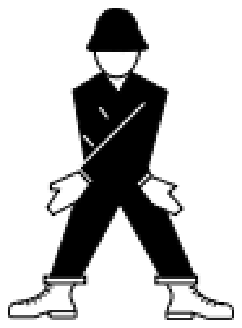}\\
   \includegraphics[width=2.1cm]{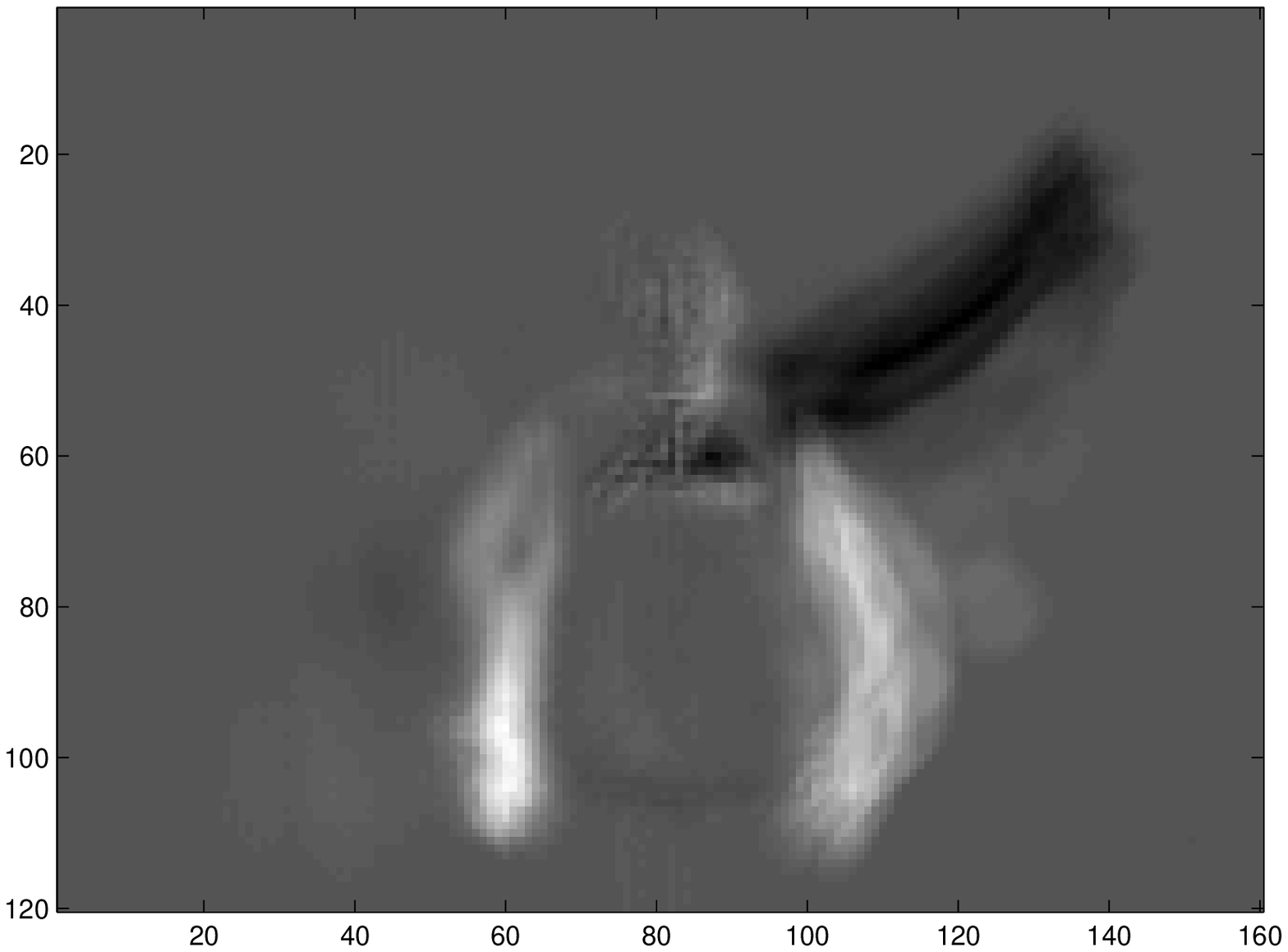}
  \includegraphics[width=2.1cm]{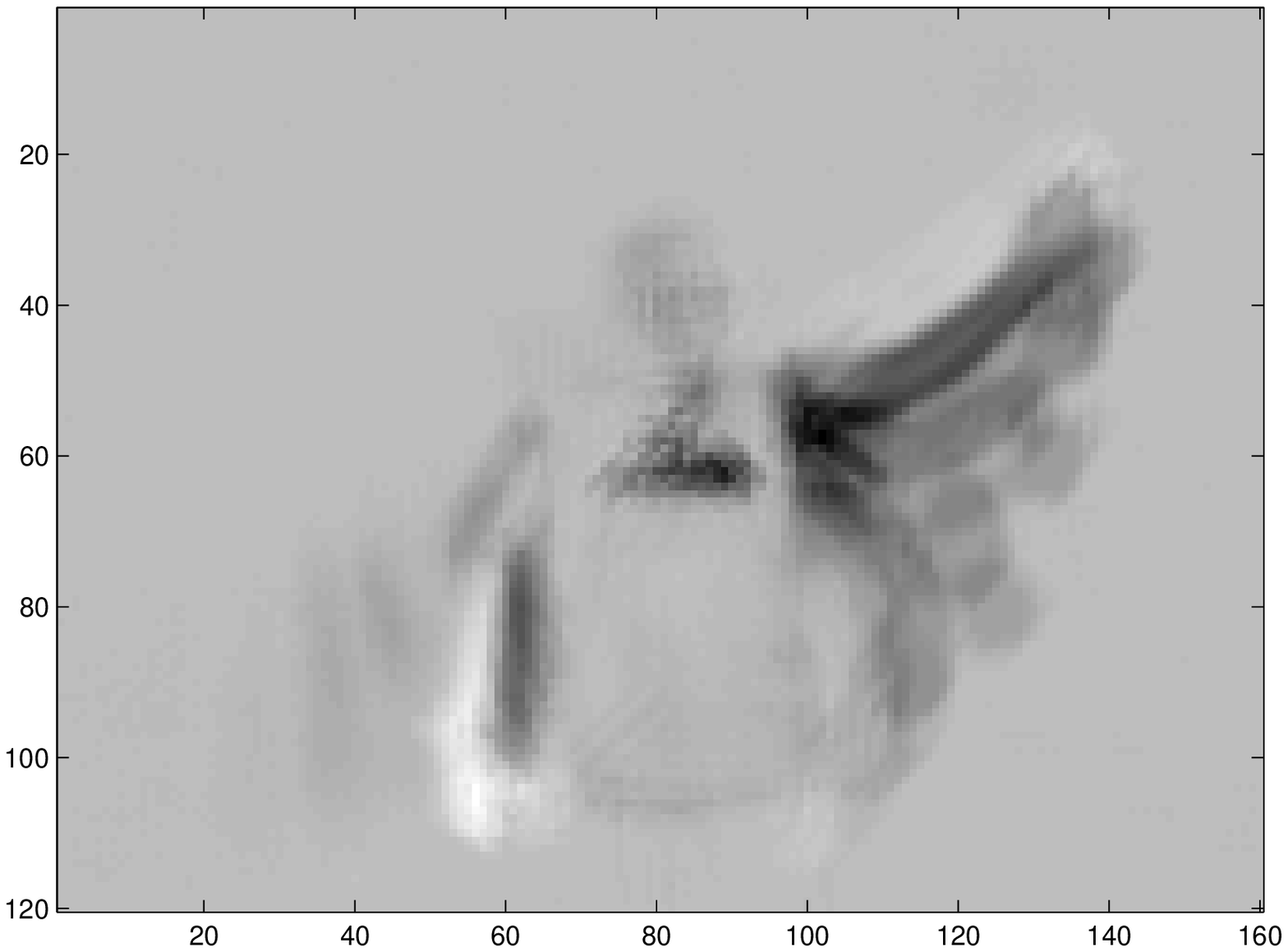}
  \includegraphics[width=2.1cm]{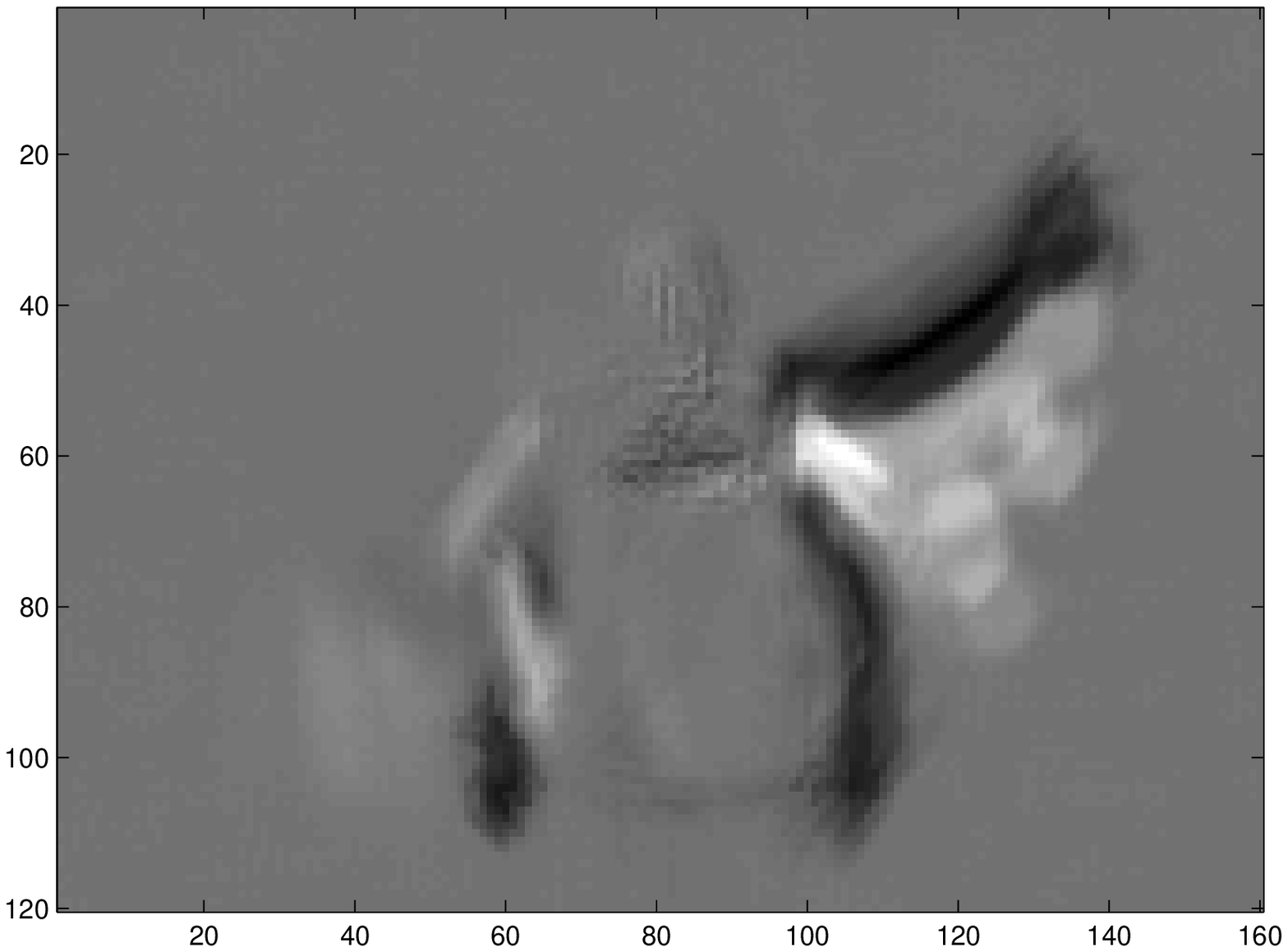}
    \includegraphics[width=2.1cm]{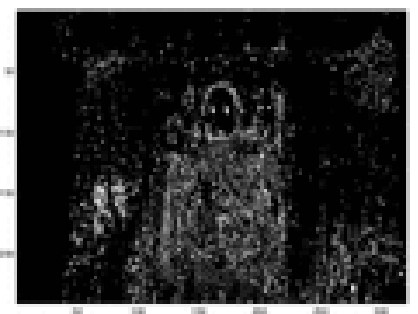}
    \includegraphics[width=2.1cm]{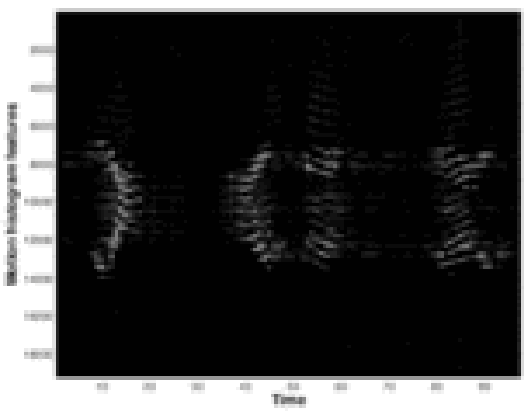}
  \includegraphics[width=2.1cm]{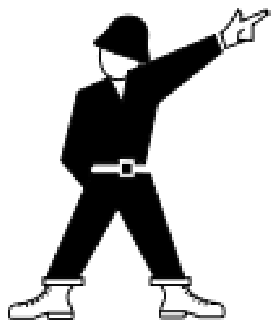}\\
   \includegraphics[width=2.1cm]{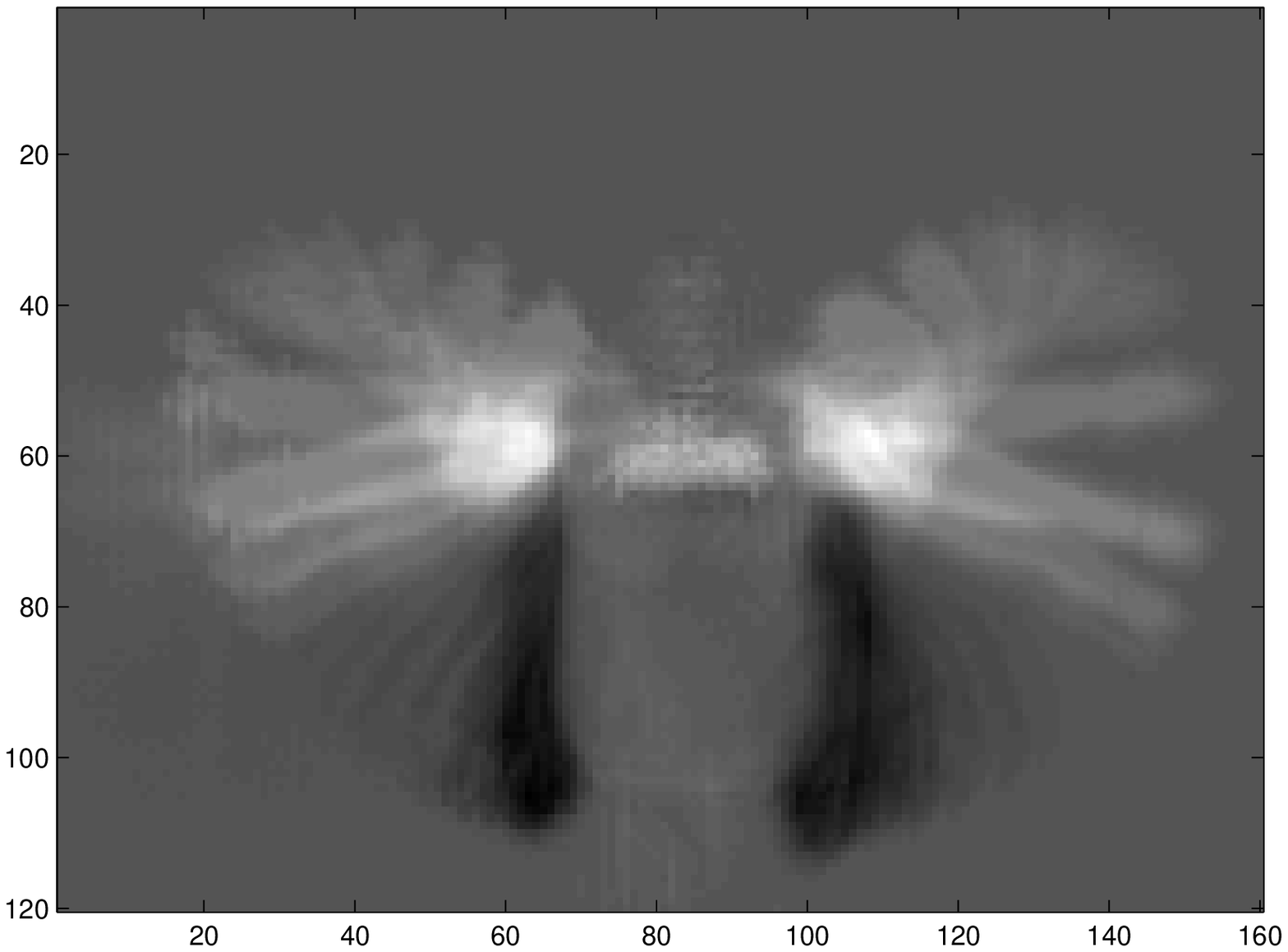}
  \includegraphics[width=2.1cm]{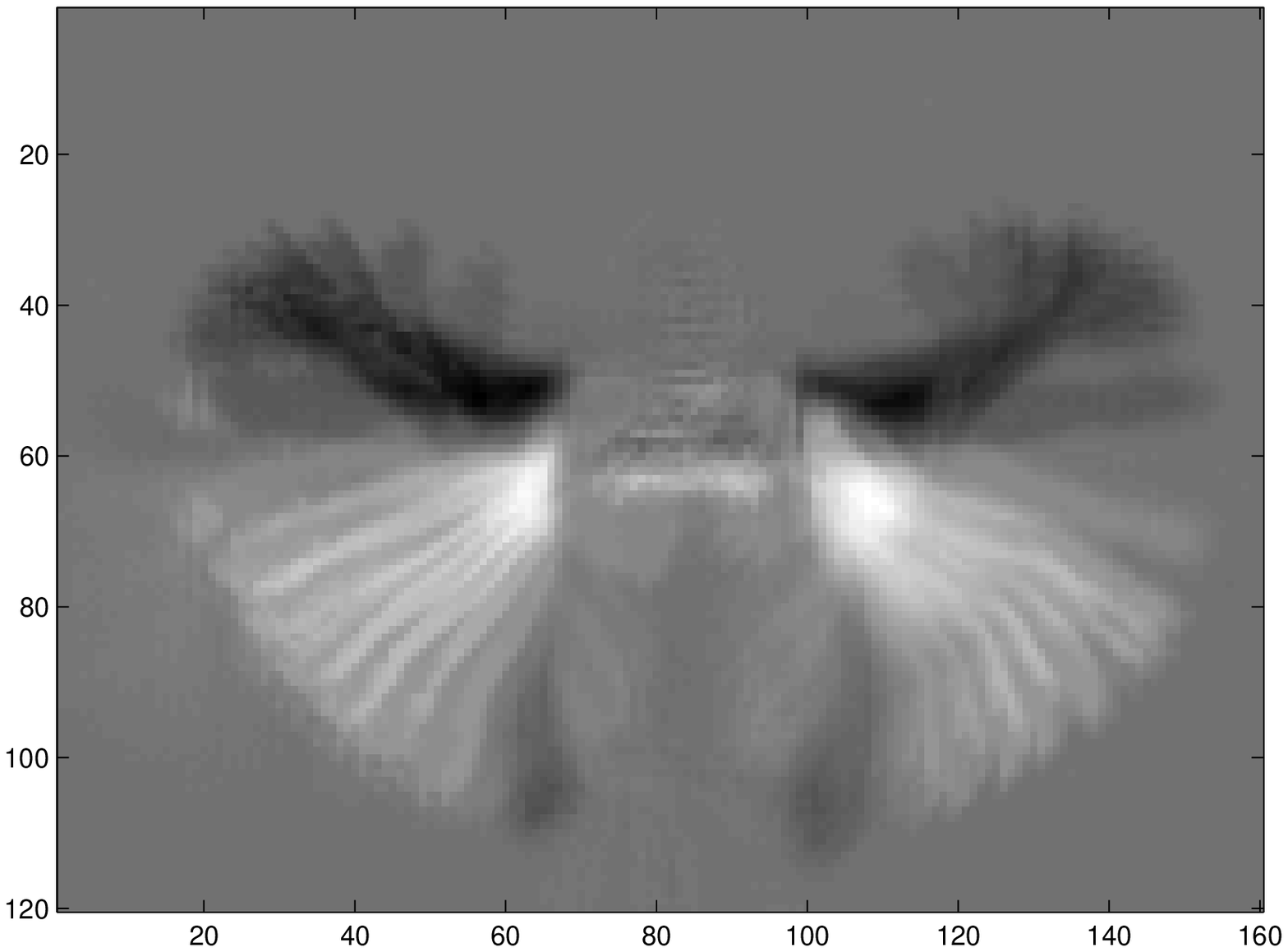}
  \includegraphics[width=2.1cm]{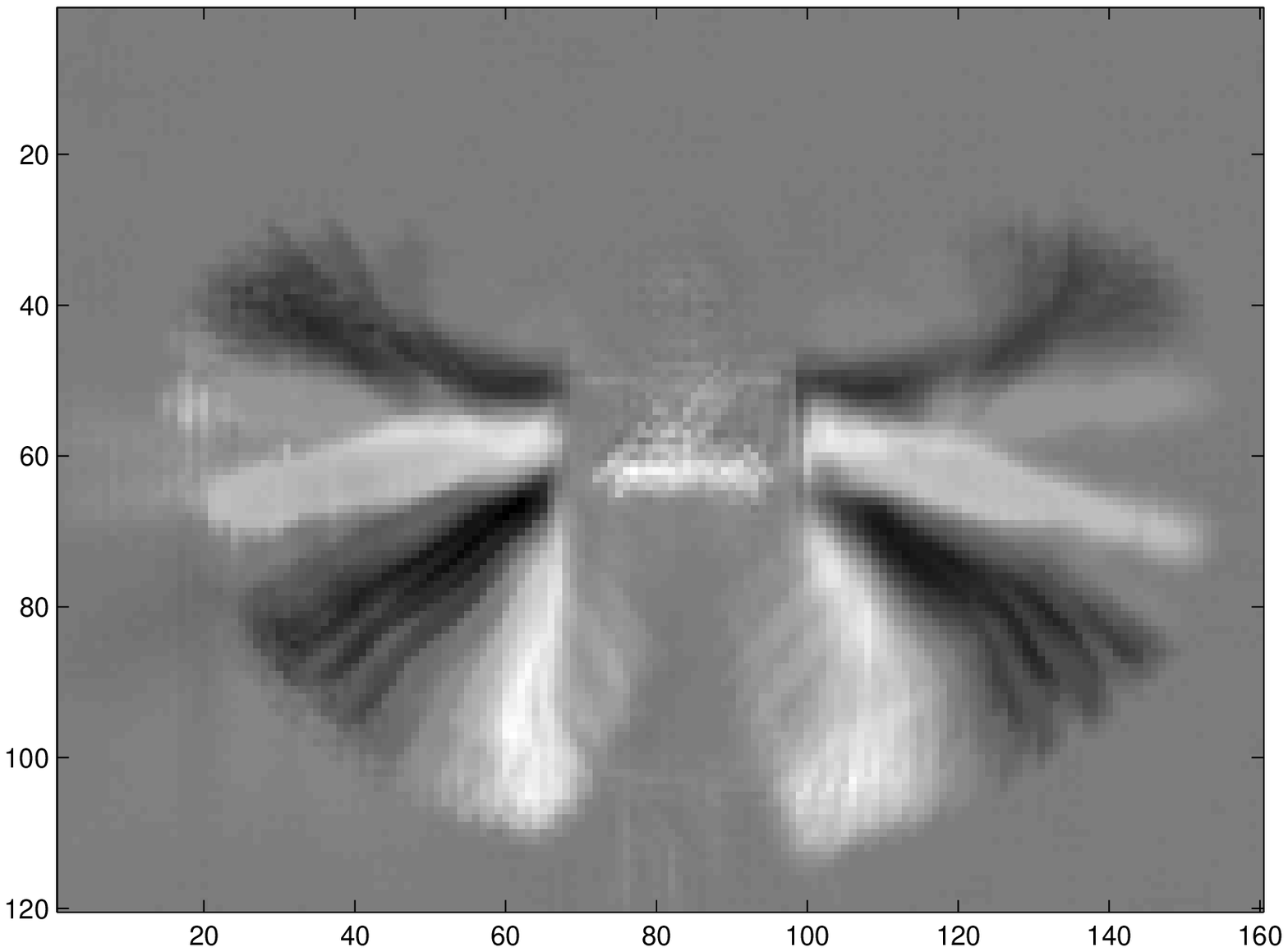}
    \includegraphics[width=2.1cm]{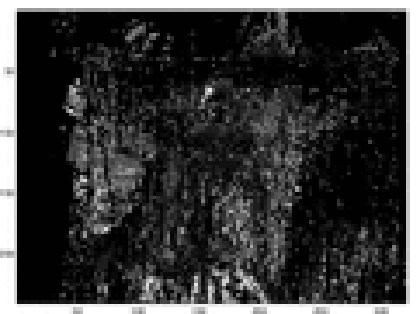}
    \includegraphics[width=2.1cm]{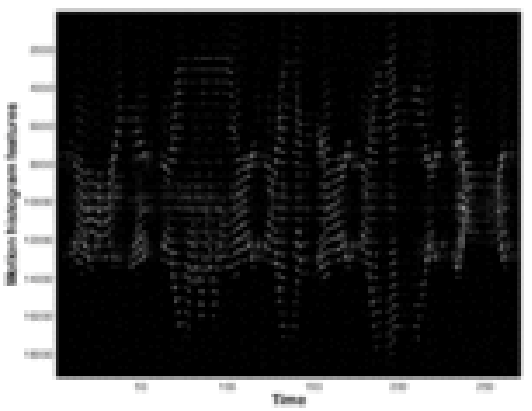}
  \includegraphics[width=2.1cm]{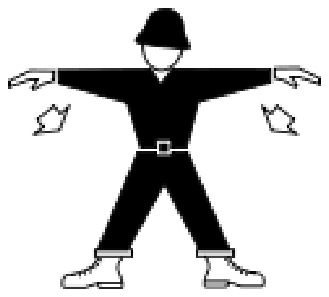}\\
     \includegraphics[width=2.1cm]{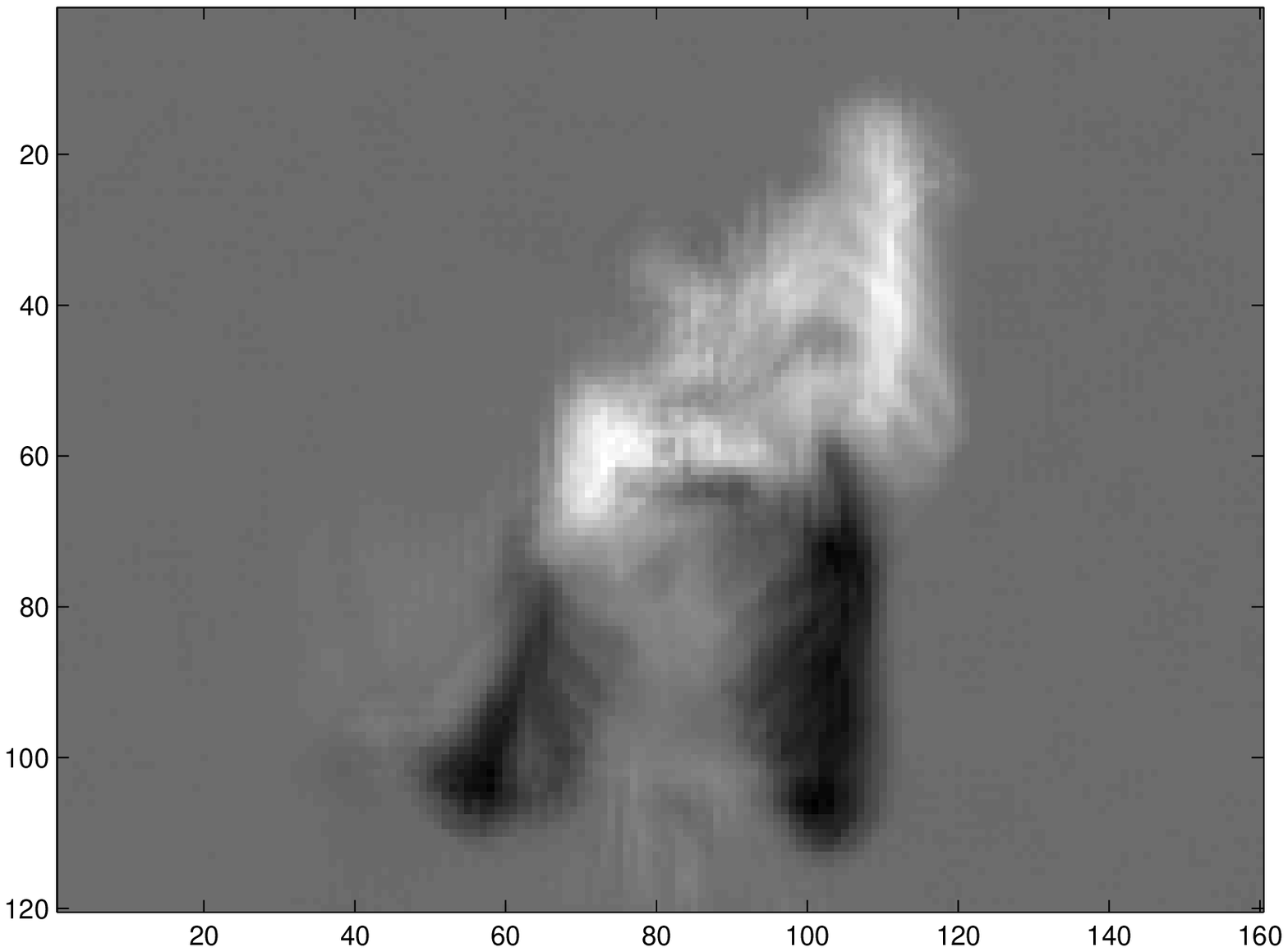}
  \includegraphics[width=2.1cm]{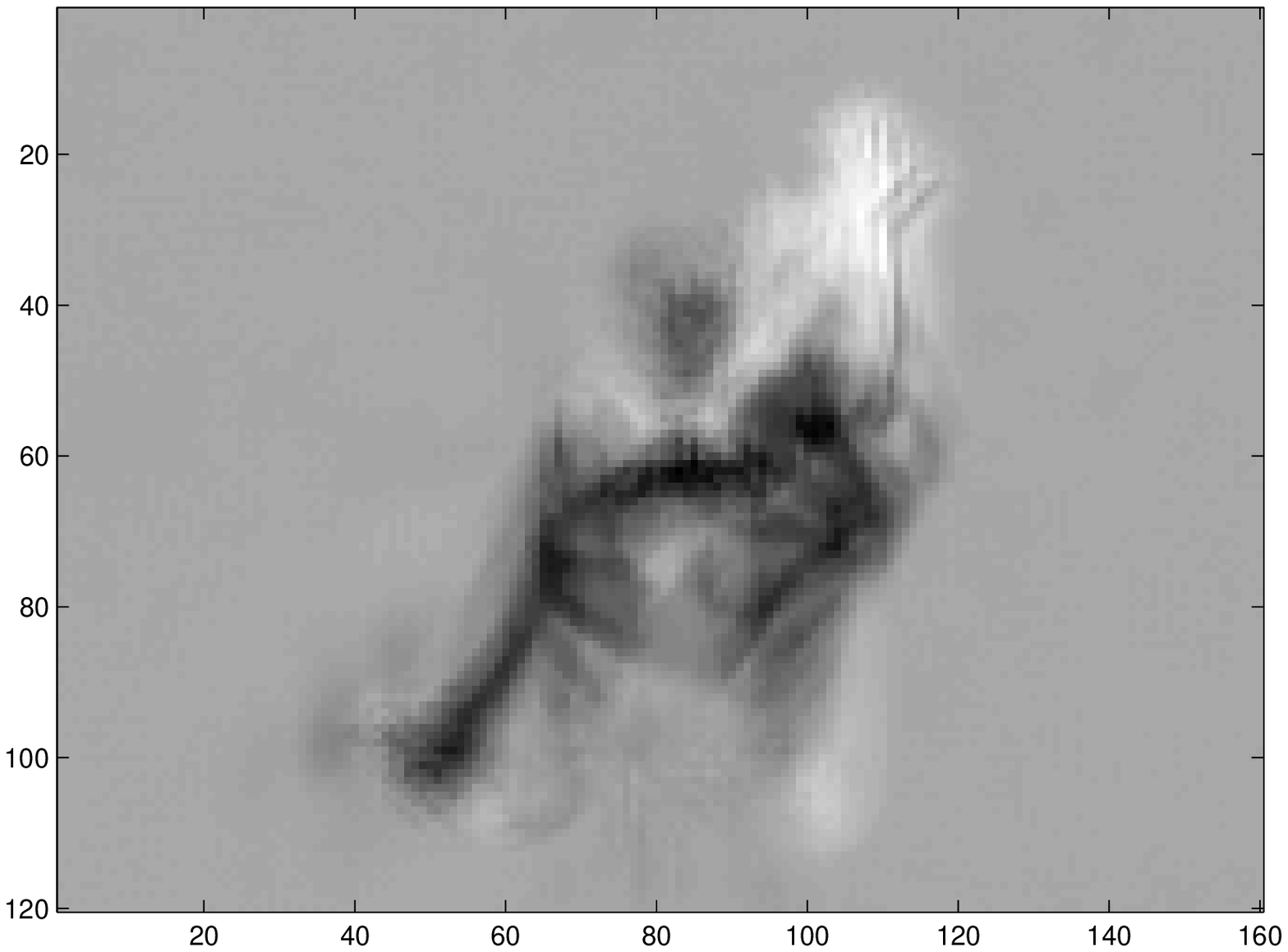}
  \includegraphics[width=2.1cm]{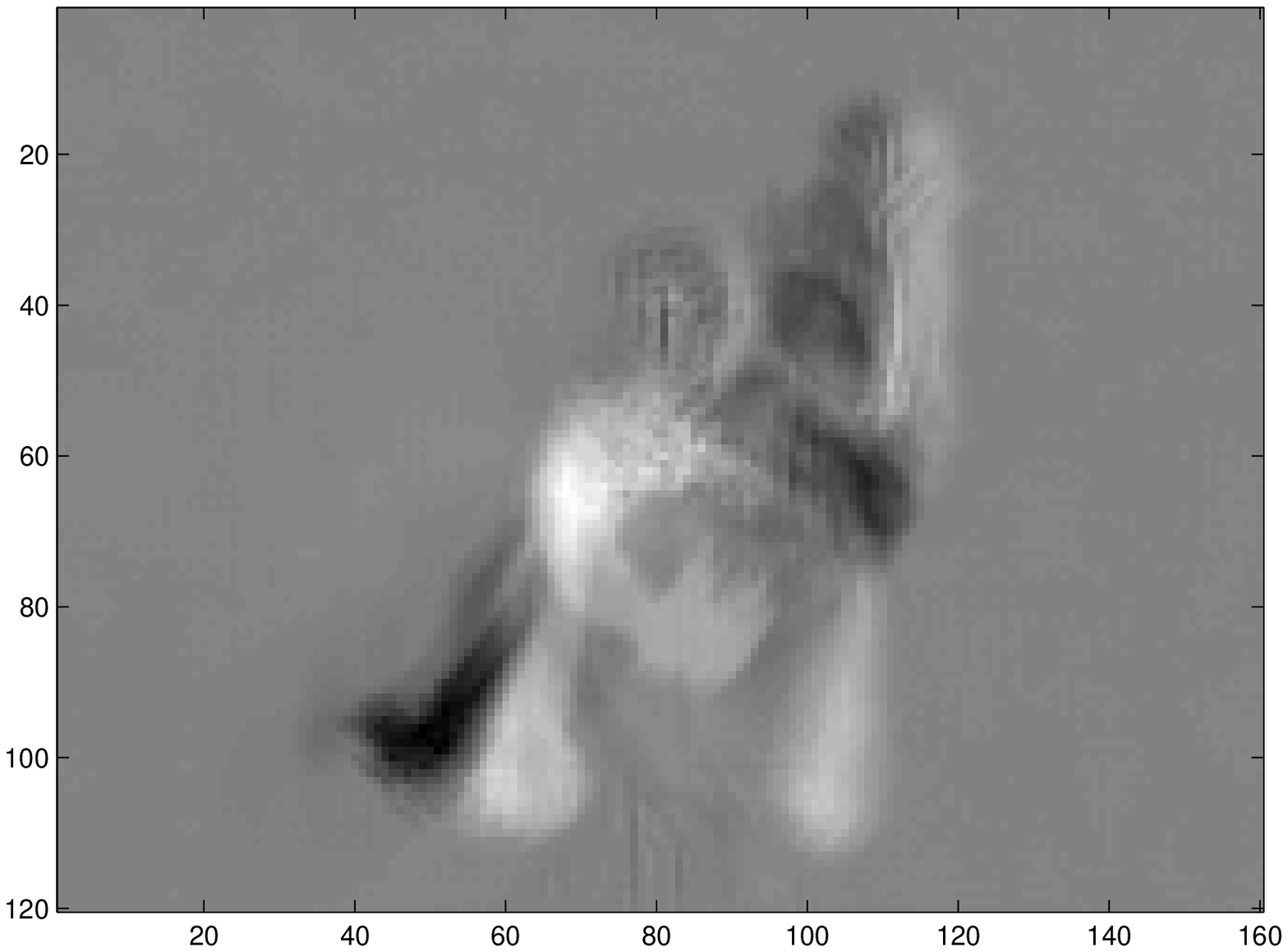}
    \includegraphics[width=2.1cm]{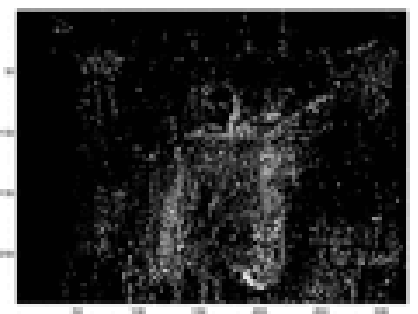}
    \includegraphics[width=2.1cm]{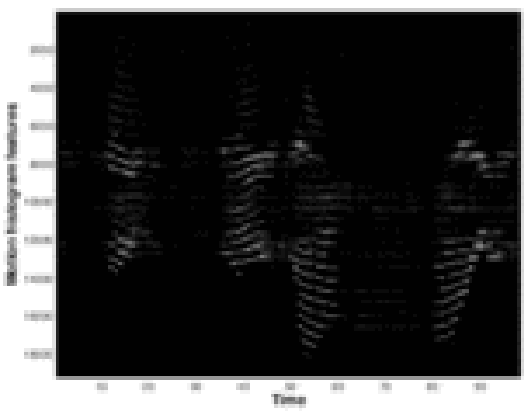}
  \includegraphics[width=2.1cm]{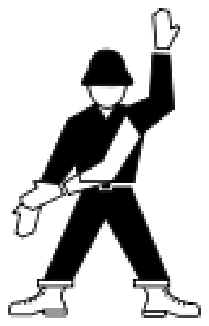}\\
       \includegraphics[width=2.1cm]{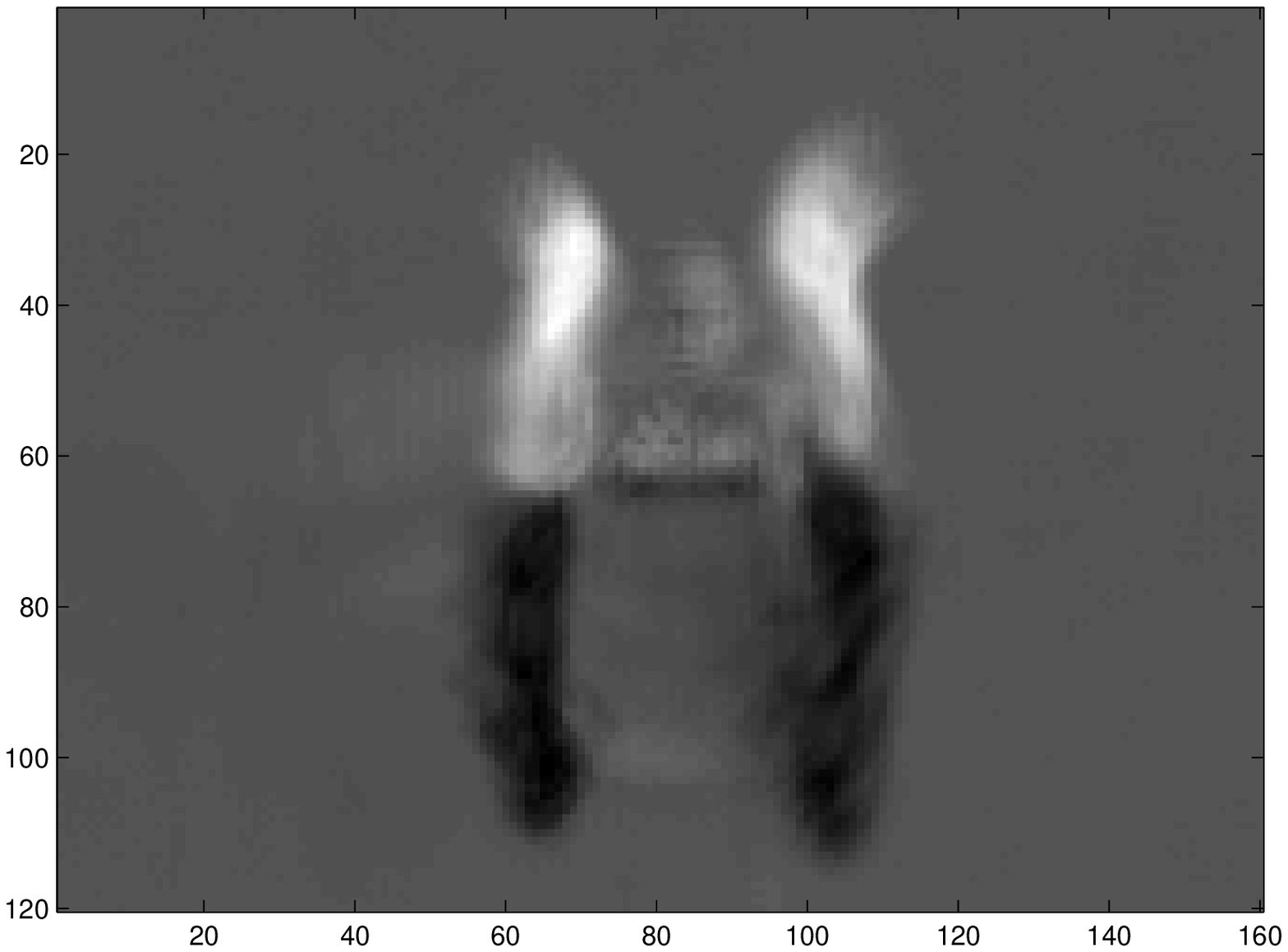}
  \includegraphics[width=2.1cm]{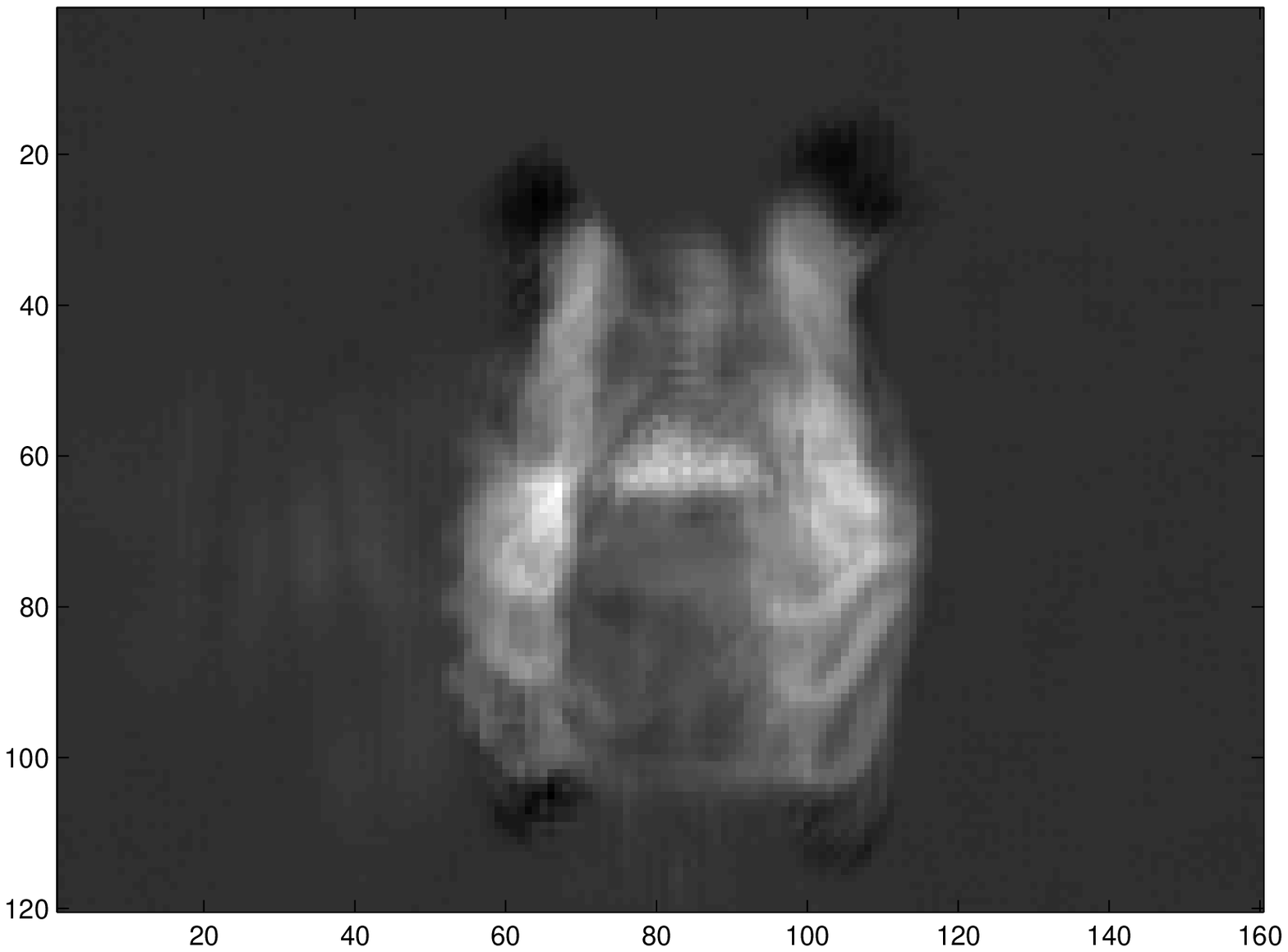}
  \includegraphics[width=2.1cm]{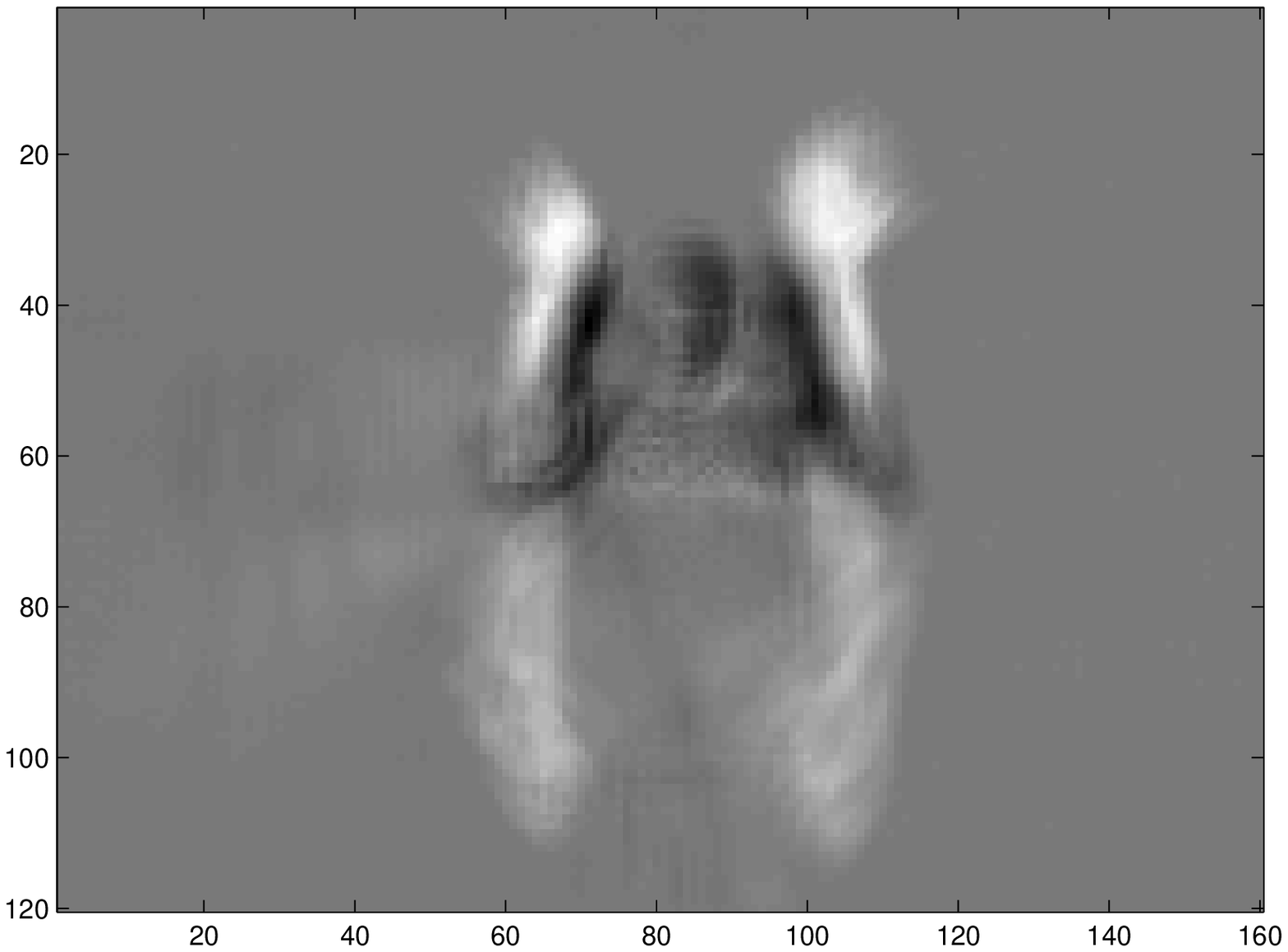}
    \includegraphics[width=2.1cm]{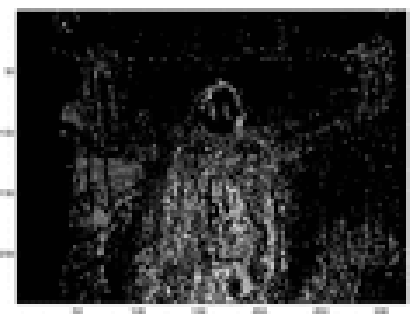}
    \includegraphics[width=2.1cm]{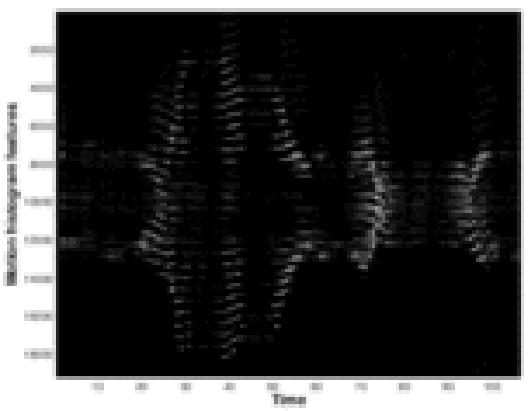}
  \includegraphics[width=2.1cm]{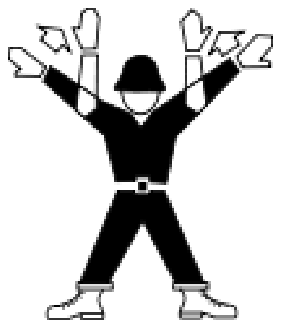}\\
         \includegraphics[width=2.1cm]{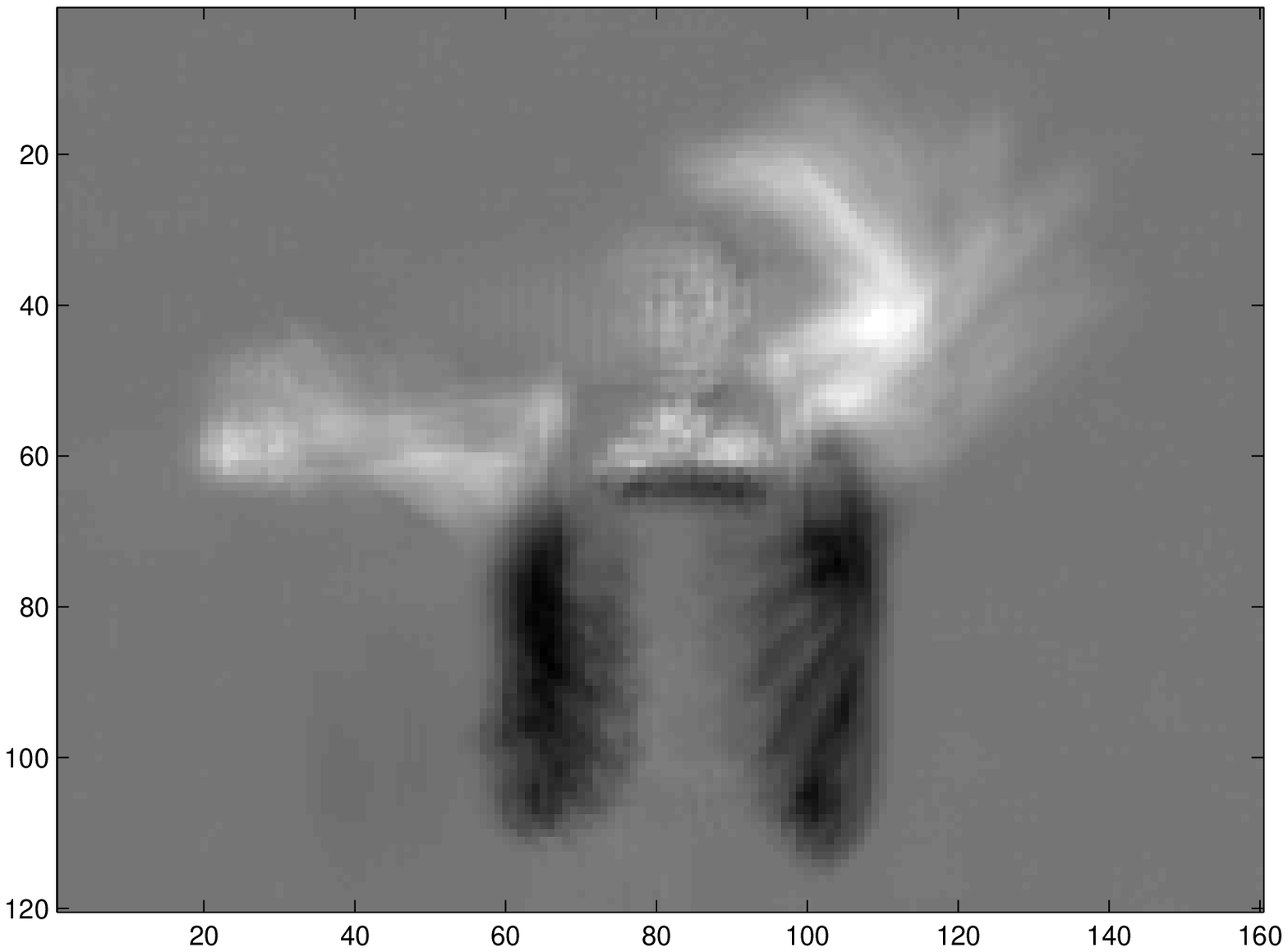}
  \includegraphics[width=2.1cm]{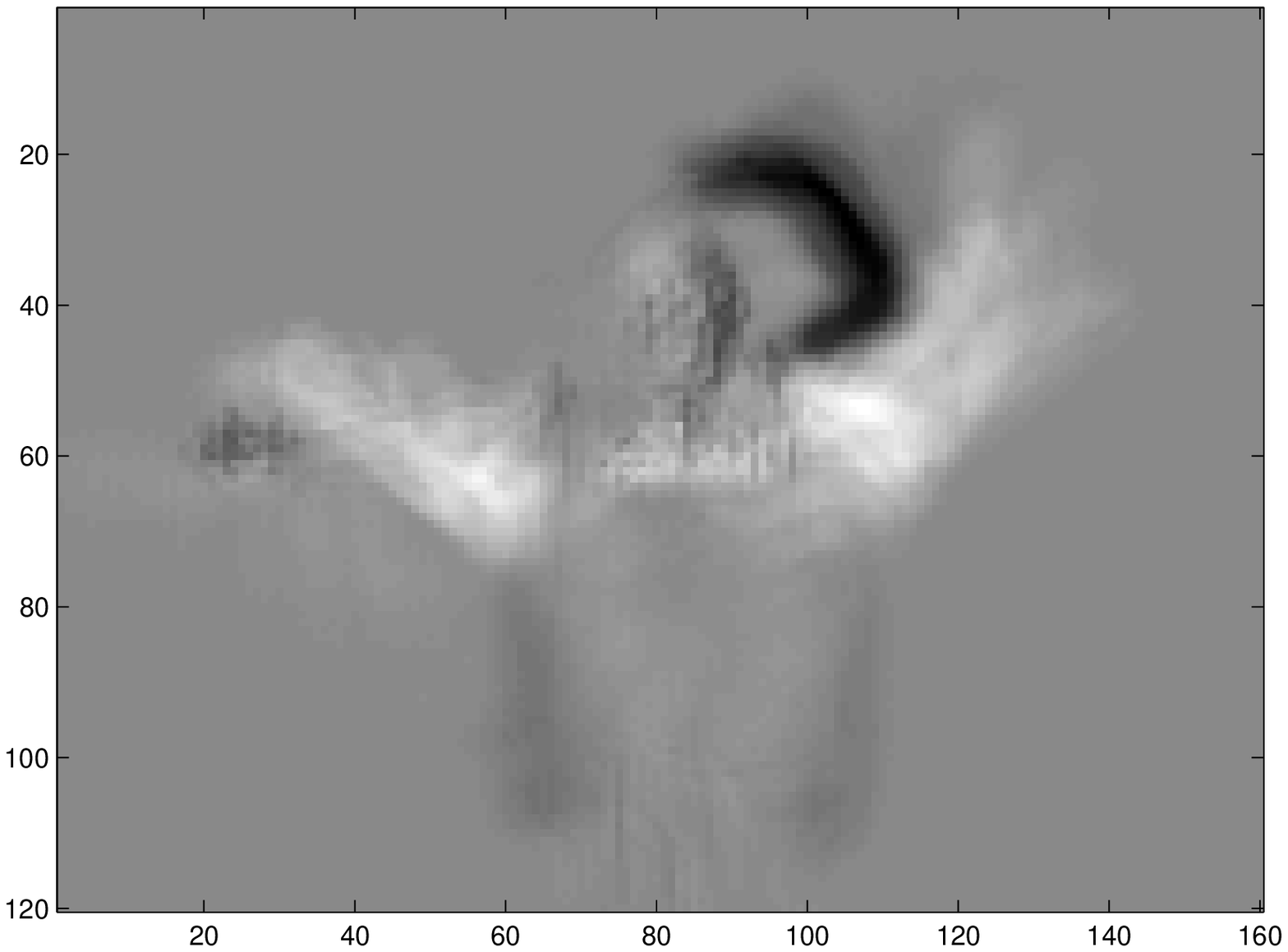}
  \includegraphics[width=2.1cm]{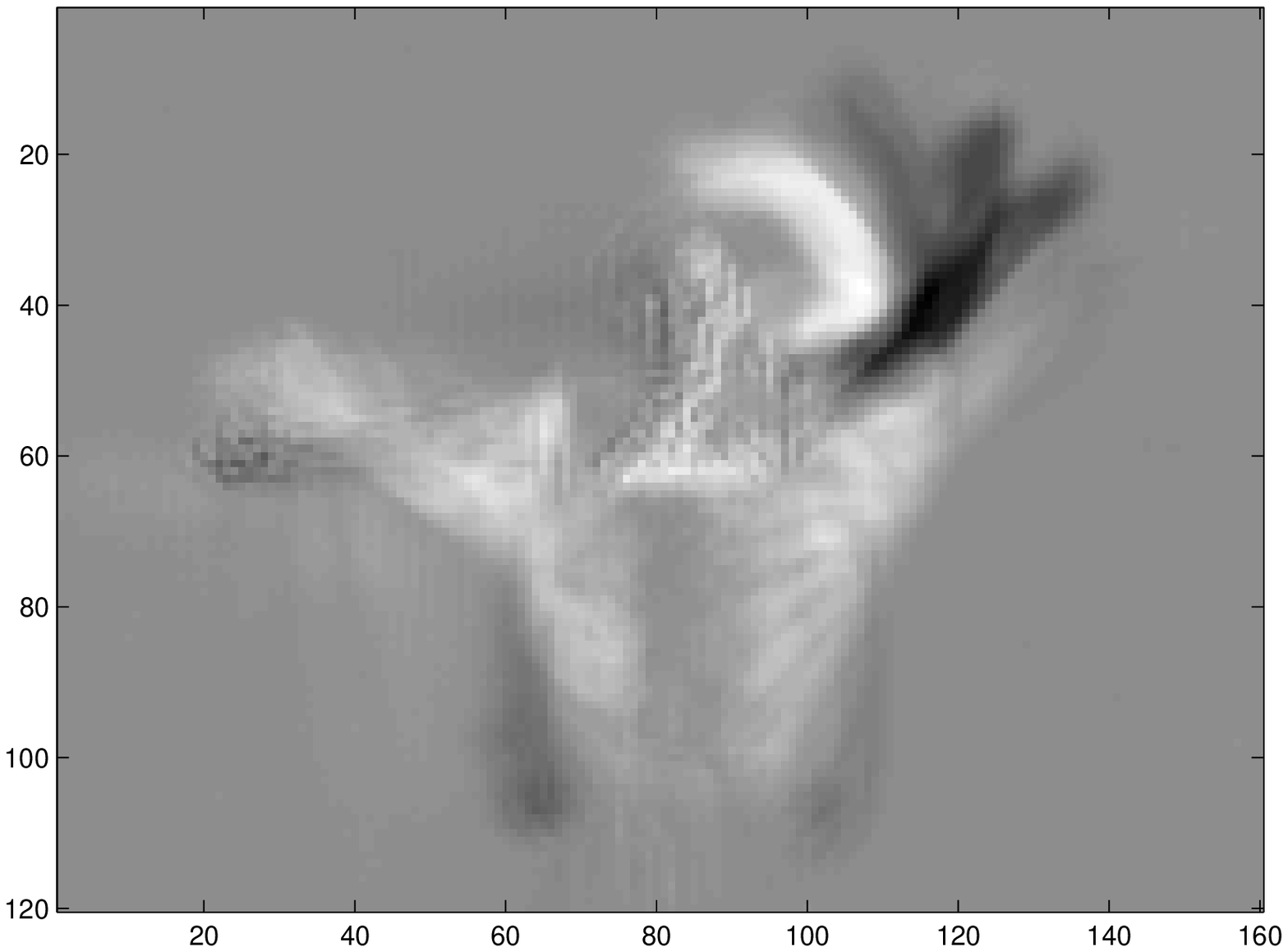}
    \includegraphics[width=2.1cm]{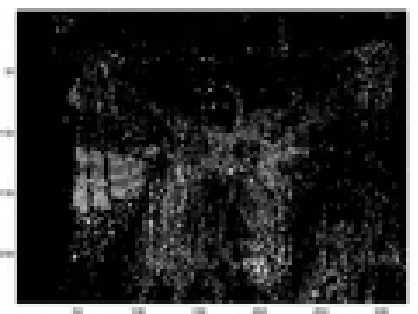}
    \includegraphics[width=2.1cm]{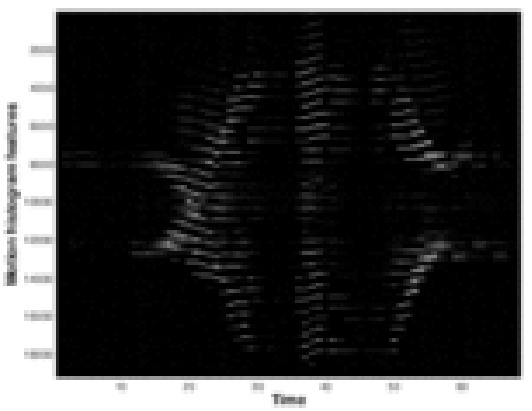}
  \includegraphics[width=2.1cm]{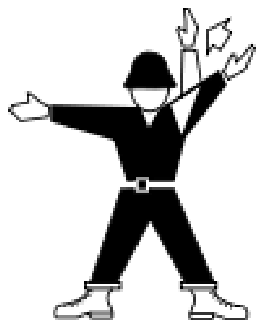}\\
         \includegraphics[width=2.1cm]{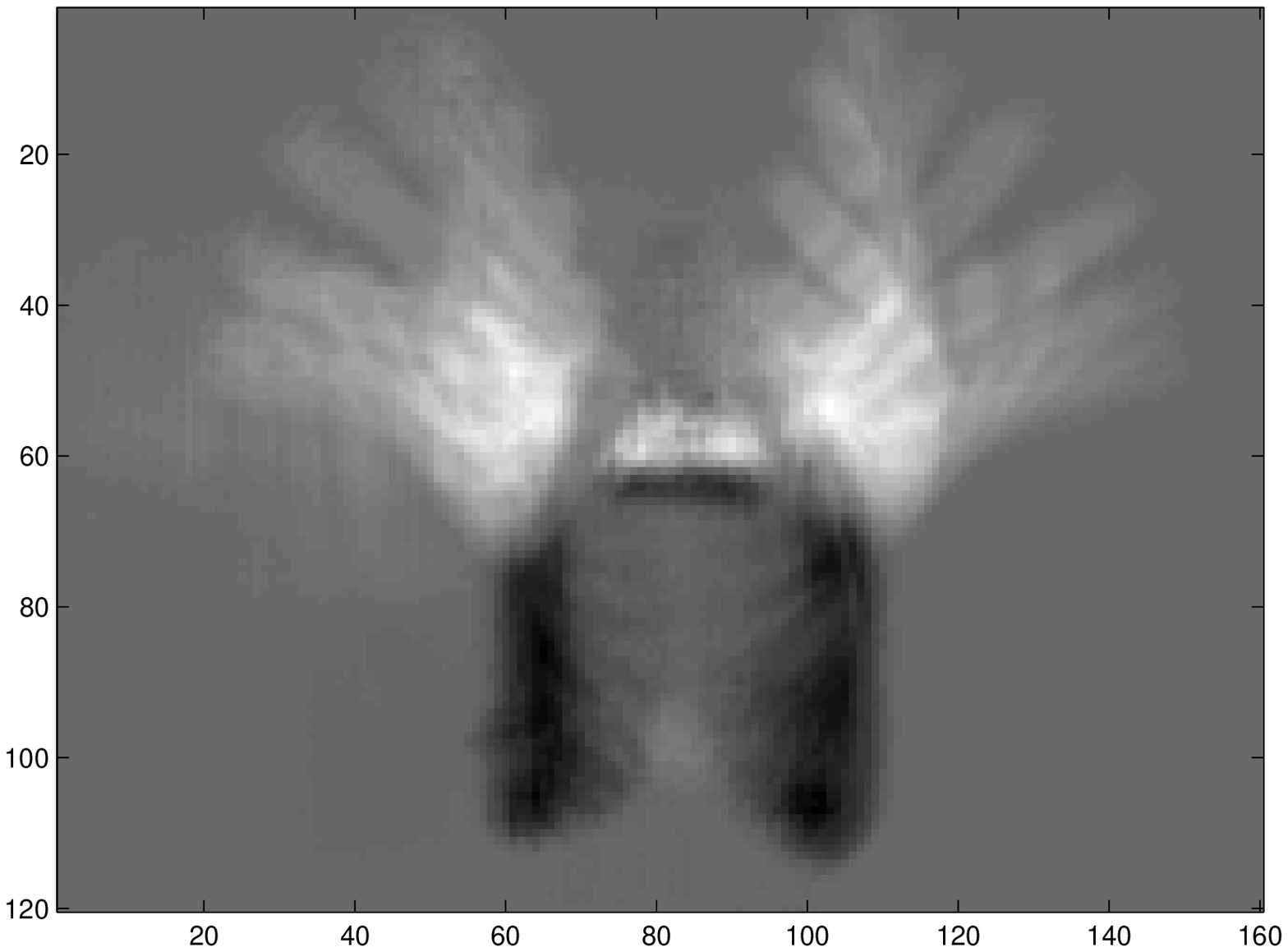}
  \includegraphics[width=2.1cm]{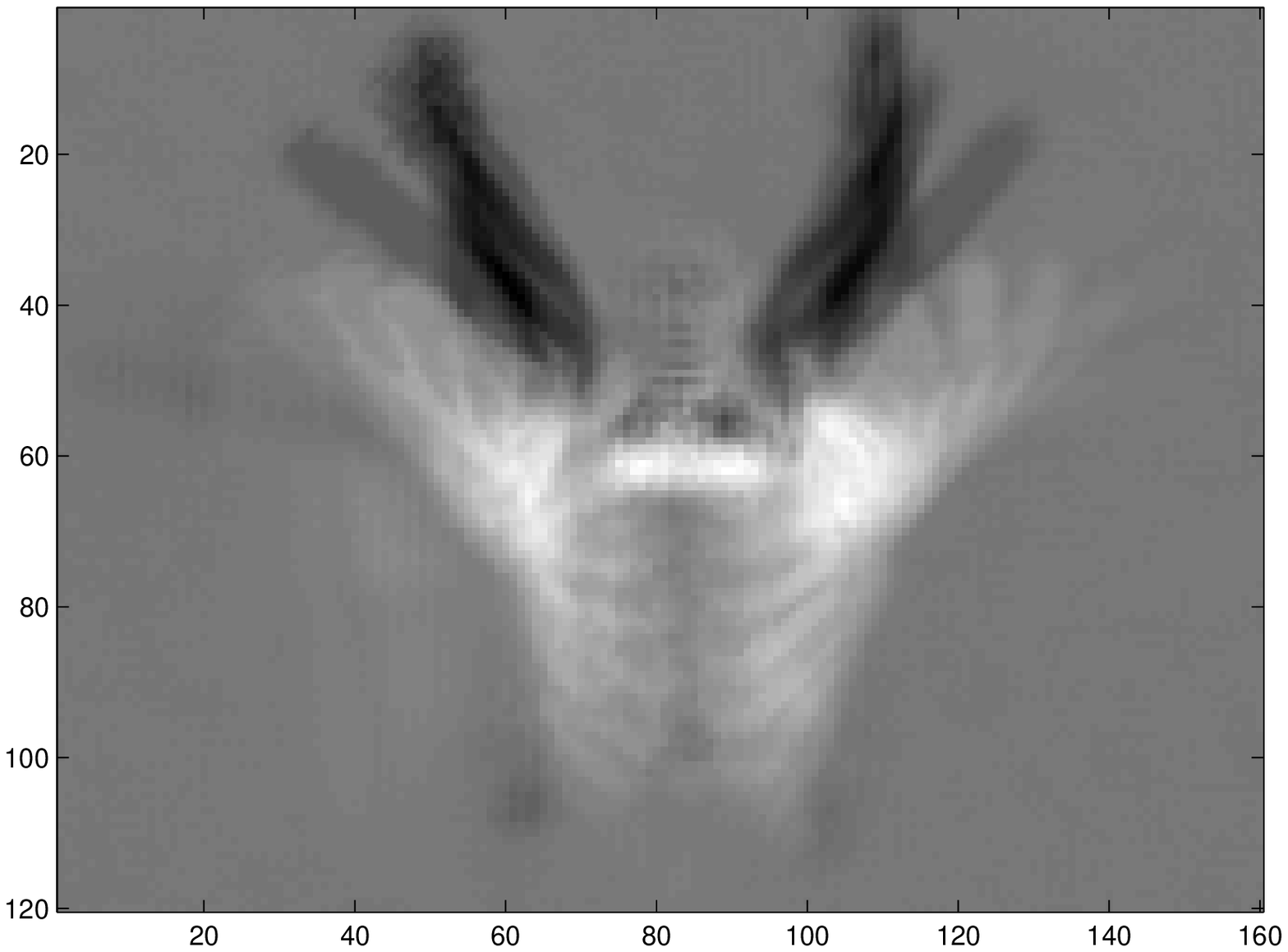}
  \includegraphics[width=2.1cm]{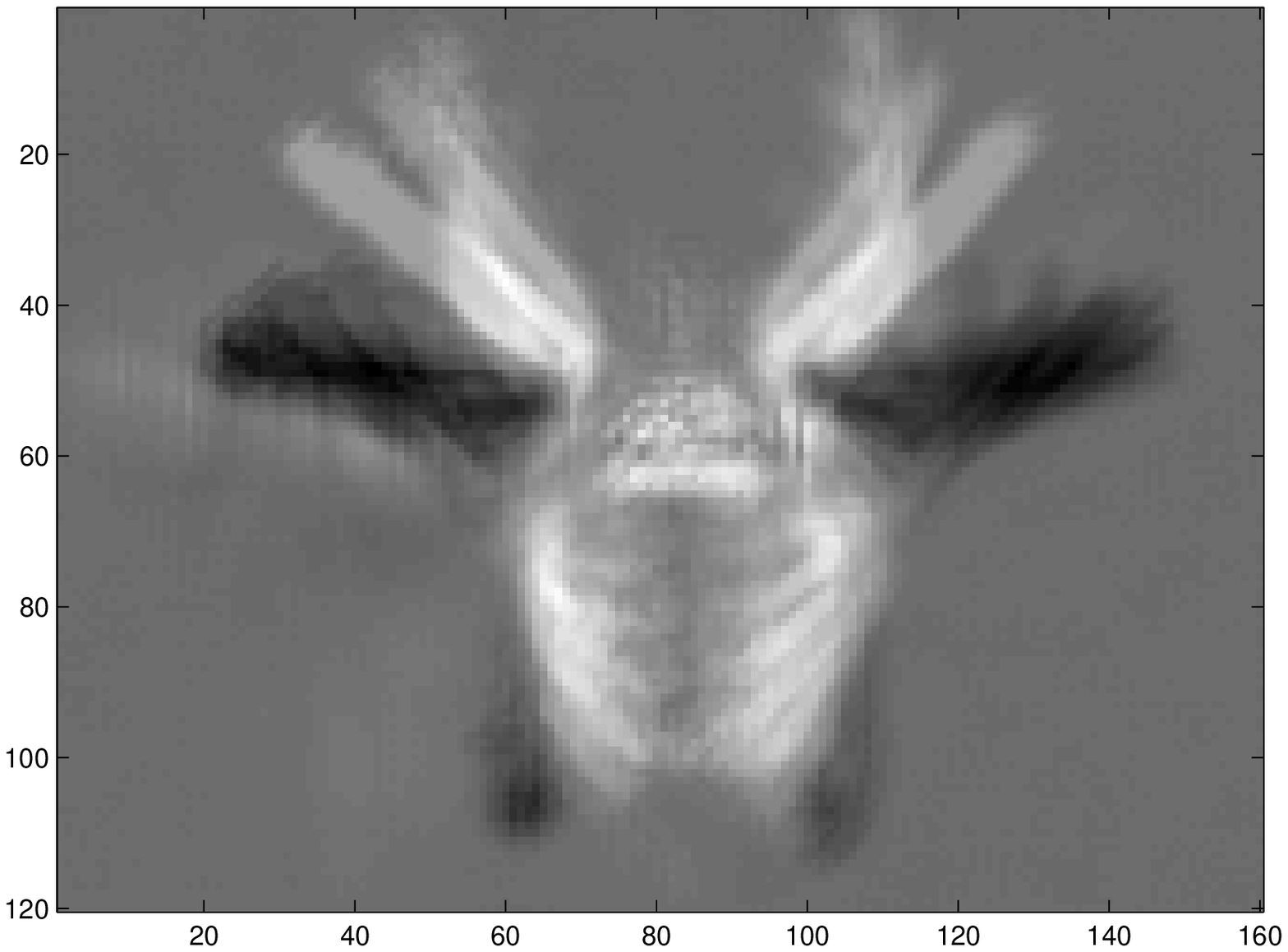}
    \includegraphics[width=2.1cm]{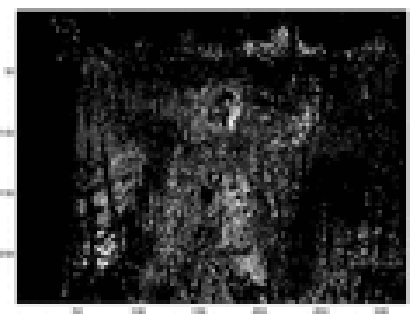}
    \includegraphics[width=2.1cm]{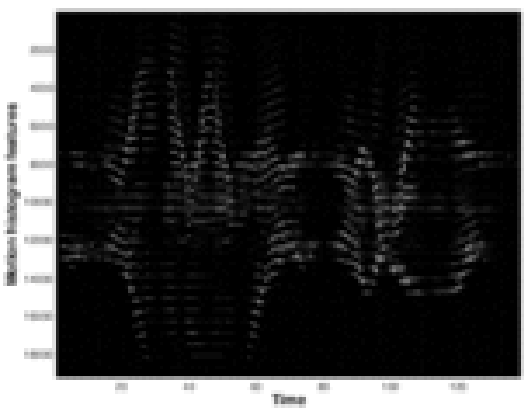}
  \includegraphics[width=2.1cm]{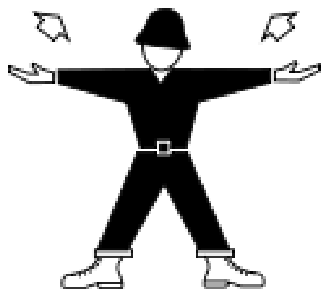}\\
         \includegraphics[width=2.1cm]{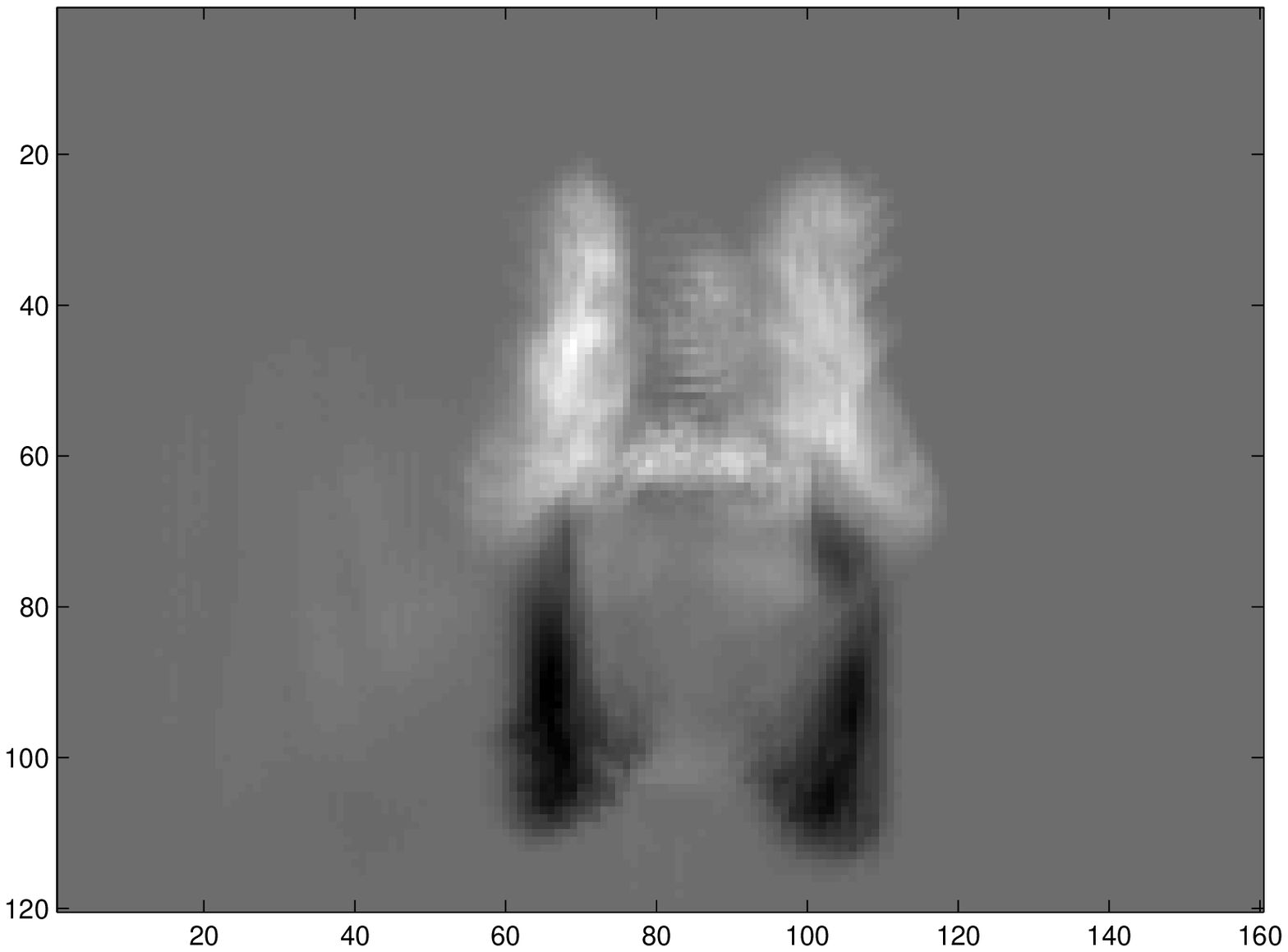}
  \includegraphics[width=2.1cm]{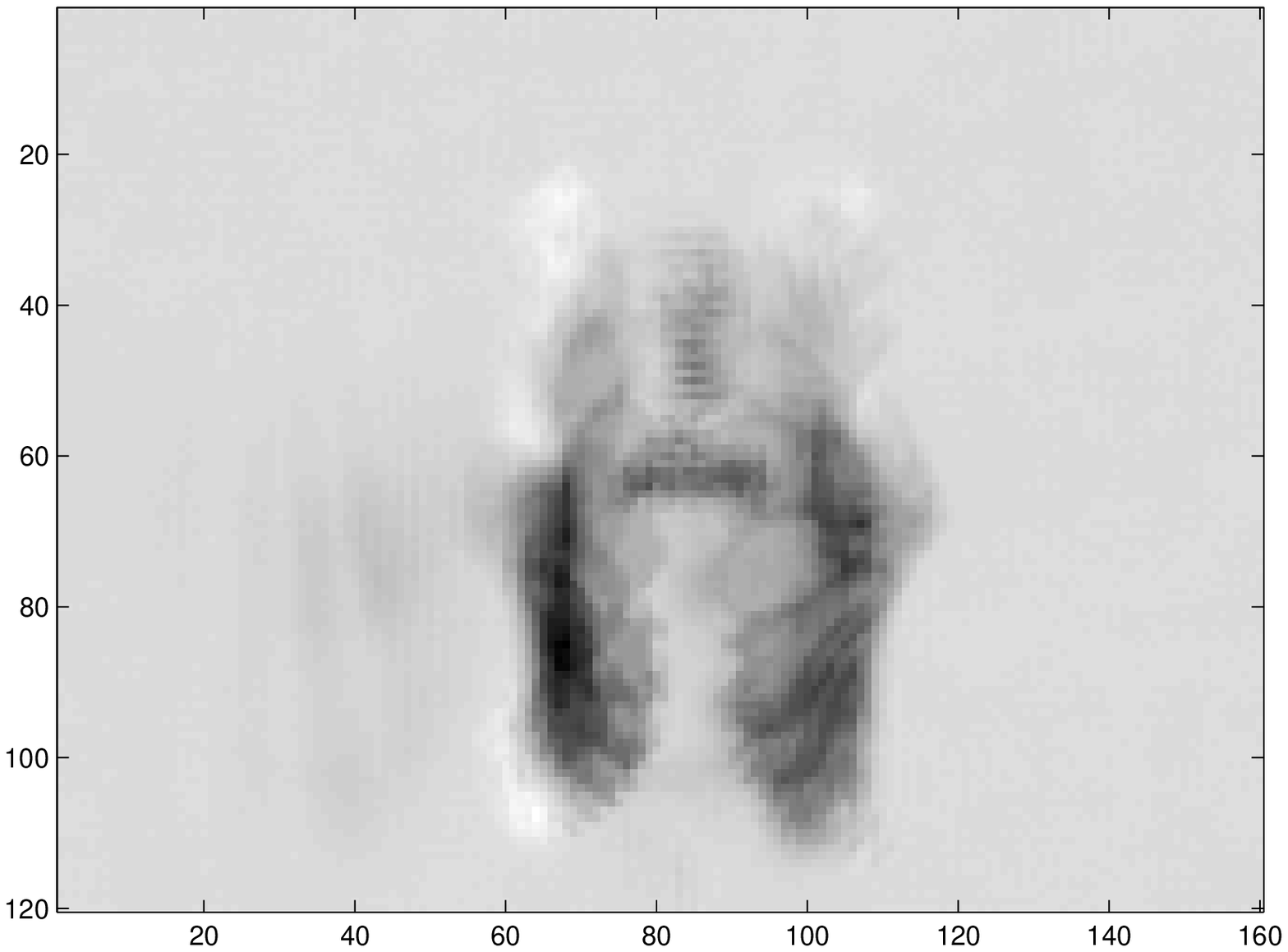}
  \includegraphics[width=2.1cm]{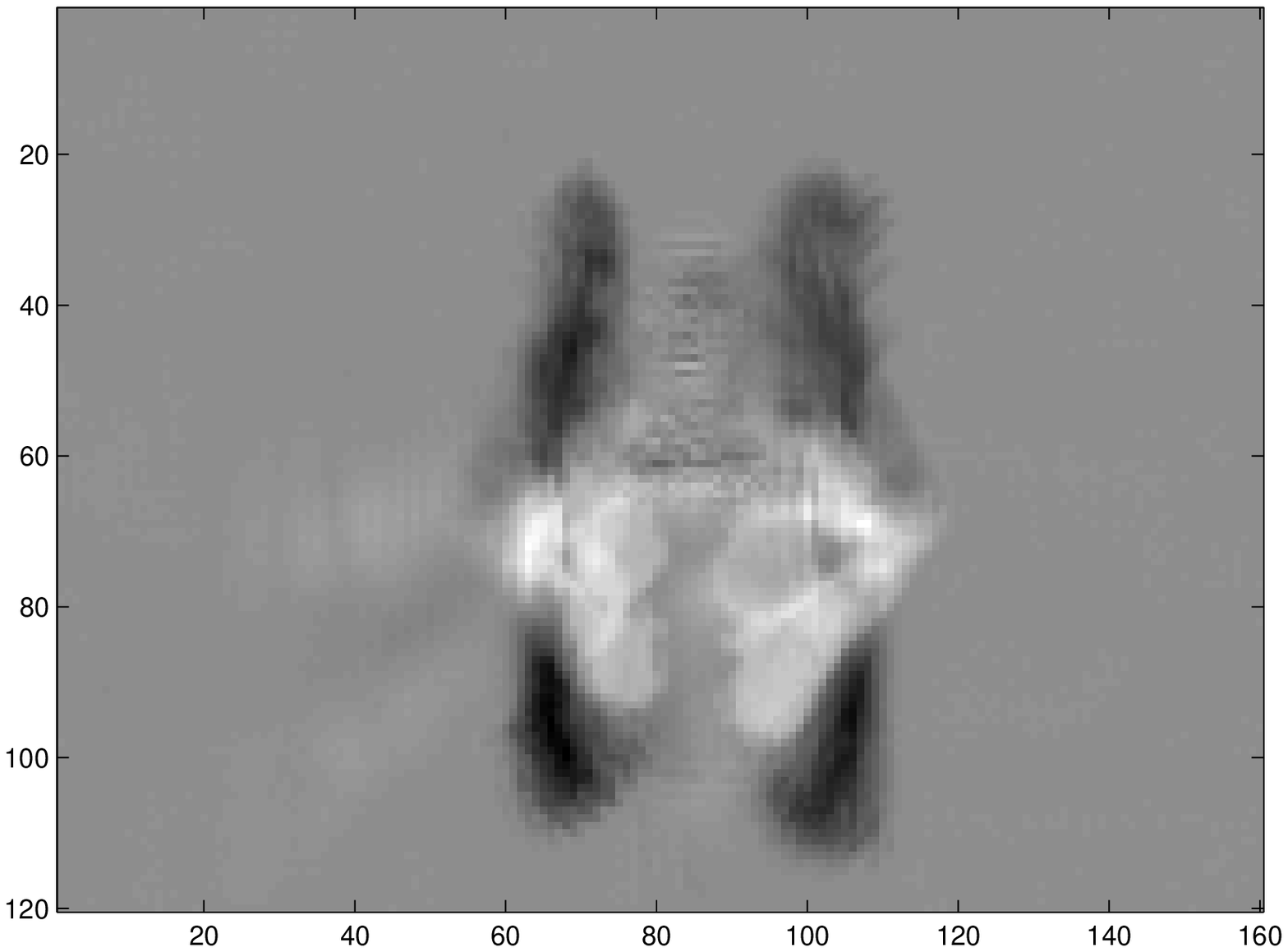}
    \includegraphics[width=2.1cm]{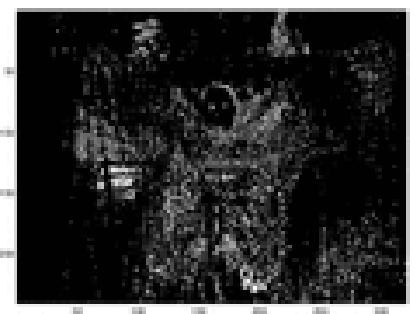}
    \includegraphics[width=2.1cm]{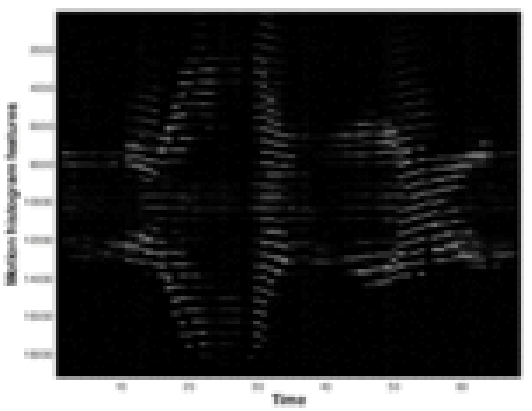}
  \includegraphics[width=2.1cm]{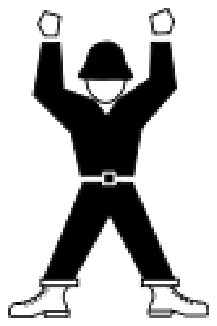}\\
  \caption{Principal motion components for the gesture vocabulary: \emph{``Helicopter signals''}. Each row is associated with a different gesture, the first 3 columns of each row display top-3 principal motion components of the gesture; columns 4-6 show the MHI, motion maps and a visual description of the corresponding gesture, respectively.}\label{fig:pcs}
\end{figure}

Figure~\ref{fig:pcs} shows the principal motion components for a particular gesture vocabulary, the figure illustrates the benefits of the proposed approach. We can appreciate that the principal motion components indeed capture the intrinsic dimensions of motion of each gesture.  By comparison, informative motion is not as clearly captured by competing motion-based representations,
e.g., MHI (column 4) and the sequence of motion maps (column 5). For this particular vocabulary, the principal motion components can be easily associated by visual inspection with the image that visually describe the gesture (column 6).

A test video $\mathcal{V}_T$, depicting a single gesture\footnote{We assume each video to be processed depicts a single gesture. Gesture segmentation is an open problem by itself that we do not approach in this paper, although we evaluate the performance of our method using gestures manually and automatically segmented with a basic technique.} that needs to be classified is processed similarly as training videos, thus it is represented by a matrix of motion maps $\textbf{H}_T$. Matrix  $\textbf{H}_T$ is projected into each of the $K-$ spaces induced by the training PCA models $(\textbf{S}_c, \textbf{V}_c)_{\{1, \ldots, K\}}$, where the projection of $\textbf{H}_T$ under the $i^{th}$ PCA model is obtained as follows~\citep{pca}:
\begin{equation}
\hat{\textbf{H}}^i_{T} =  \big( \textbf{H}_T -  \textbf{H}^i_{\mu} \big)  \textbf{V}_{c,i} \textbf{S}^{-\frac{1}{2}}_{c,i}
\end{equation}
where $\textbf{H}^i_{\mu}$ is a matrix with each row being the average of $\textbf{H}_i$, and subscript $i$ in $\textbf{S}_{c,i}$ and $\textbf{V}_{c,i}$ indicates the index of the associated PCA model.
Next projections are reconstructed back, the reconstruction of $\textbf{H}_{T}$ under the $i^{th}-$PCA model is given by:
\begin{equation}
\textbf{R}_i = \hat{\textbf{H}}^i_T \big(\textbf{S}^{-\frac{1}{2}}_i \textbf{V}^T_{c,i} \big)    + \textbf{H}^i_{\mu}
\end{equation}
where superscript $T$ indicates the transpose of a matrix.

We can measure the reconstruction error for each $\textbf{R}_i$ as follows:
\begin{equation}
\epsilon(h) = \frac{1}{q} \sum_{i=1}^q  \sqrt{\sum_{j=1}^m (\textbf{R}_{ij} - \textbf{H}_{T,ij})^2}
\end{equation}
where $q$ and $m$ are the number of rows and columns of $\textbf{H}_T$, respectively and with $h = 1, \ldots,K$. Finally, we assign $\mathcal{V}_T$ the gesture corresponding to the PCA model that obtained the lowest reconstruction error, that is: $\arg\min_h  \epsilon(h)$.

Similar reconstruction-error approaches have been adopted for one-class classification~\citep{Tax}, where instances of the target class are used to generate the PCA model and a threshold on the reconstruction error is used for classification. Reconstruction error has been also used for spam filtering~\citep{Gomez}, face recognition~\citep{eigenfaces} and pedestrian detection~\citep{PCAOBDET}, see Section~\ref{sec:rw}. One should note that in previous work a set of labeled instances have been used to generate the PCA model of each class, whereas under the proposed approach the elements of a single-instance (the amount of motion in the frame differences under the bag-of-frames representation) are used. Besides the granularity, the main difference stems in that, in previous work, one can assume each instance is representative of the category, while in our setting the set of motion maps associated to a gesture  are
not necessarily representative of the gesture (e.g., similar motion maps may be shared by different gestures). 

\begin{figure}
  \includegraphics[width=1.25cm]{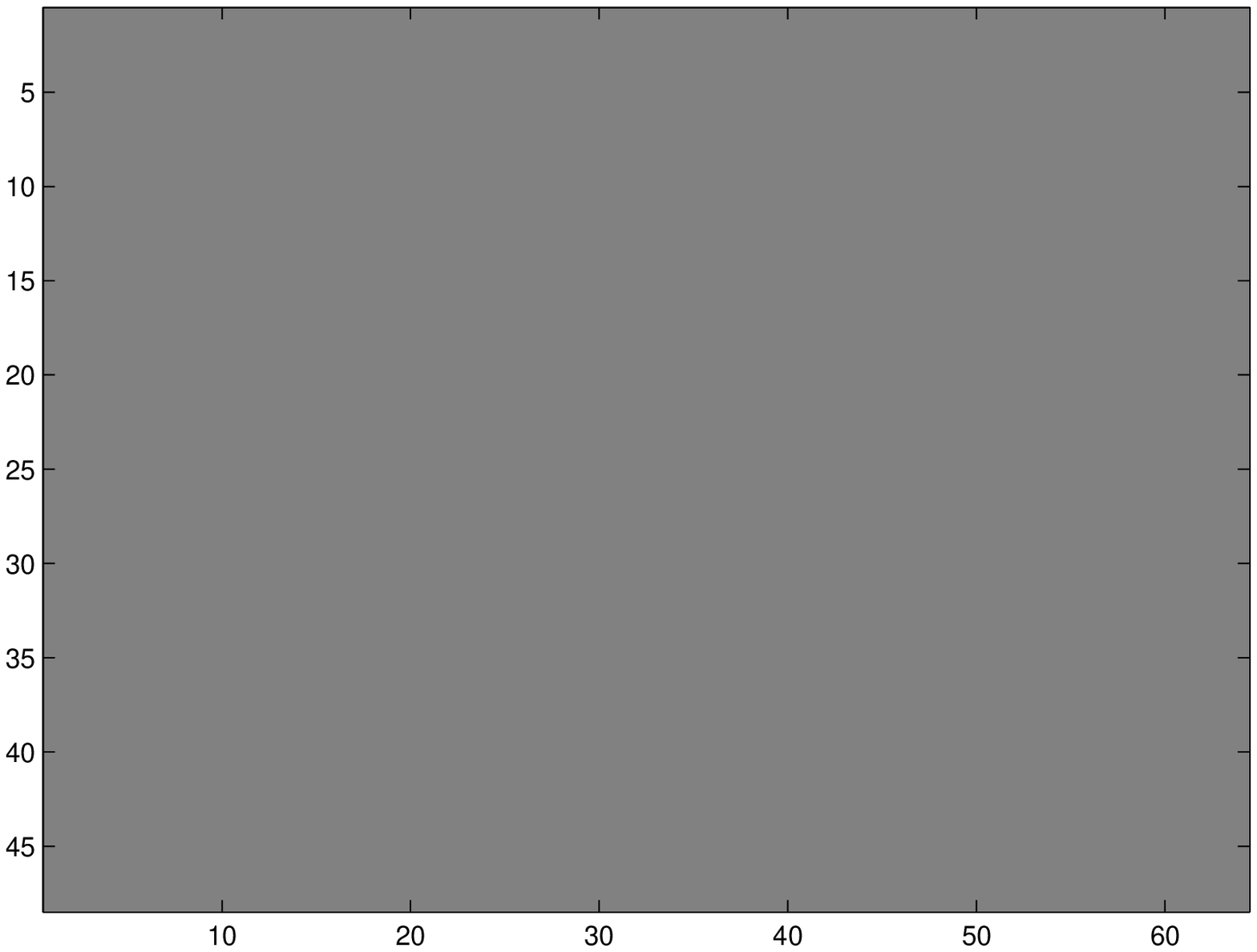}
    \includegraphics[width=1.25cm]{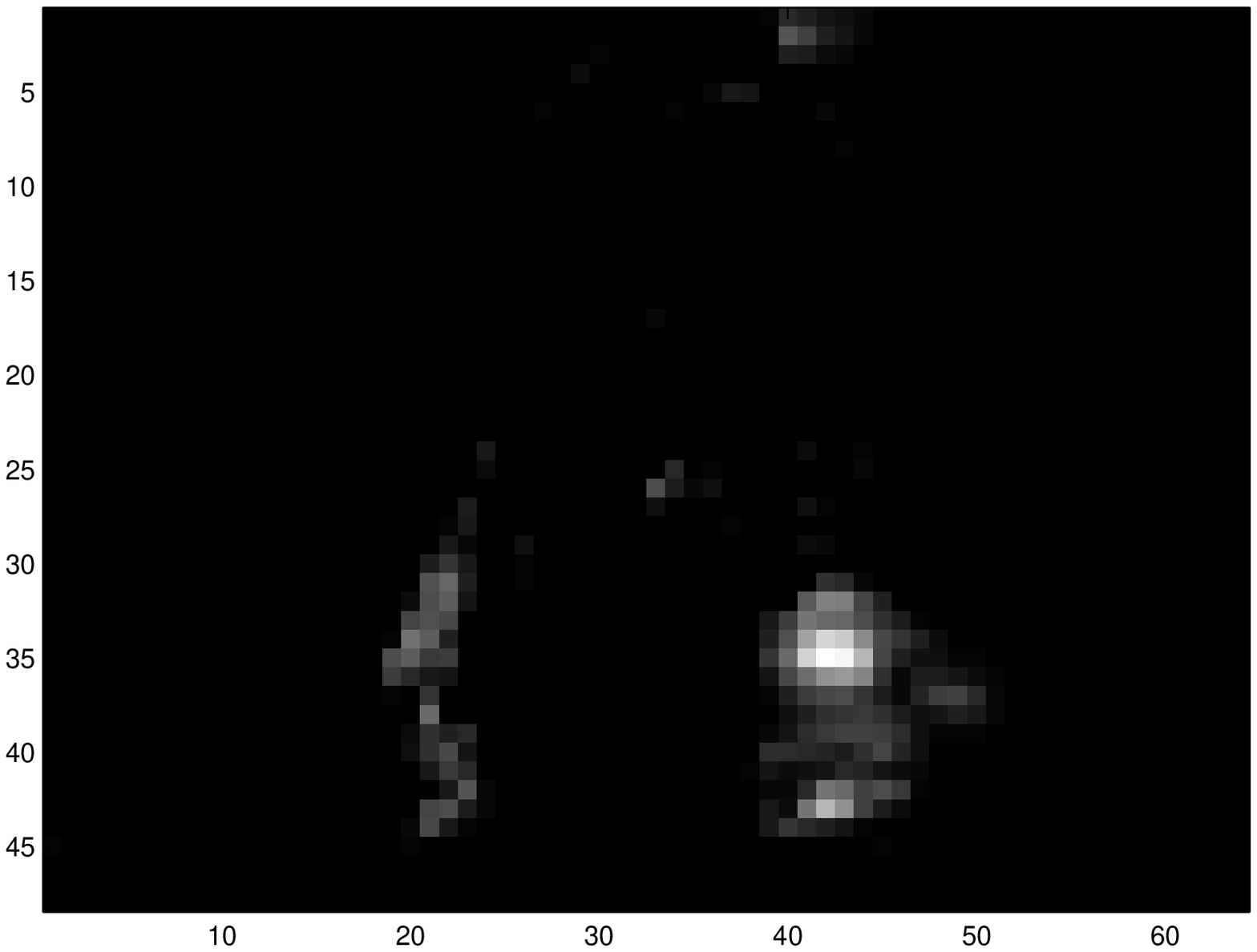}
      \includegraphics[width=1.25cm]{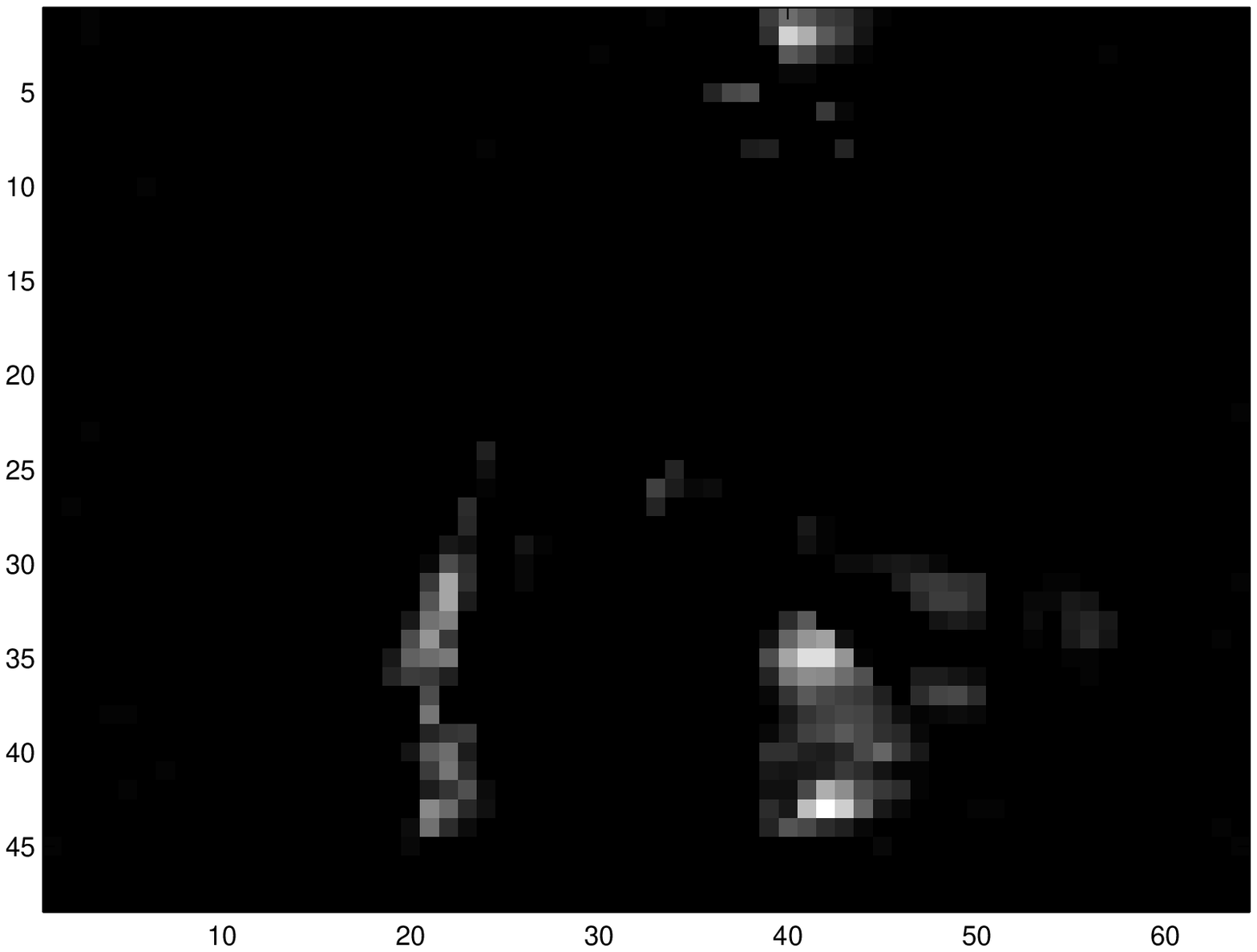}
        \includegraphics[width=1.25cm]{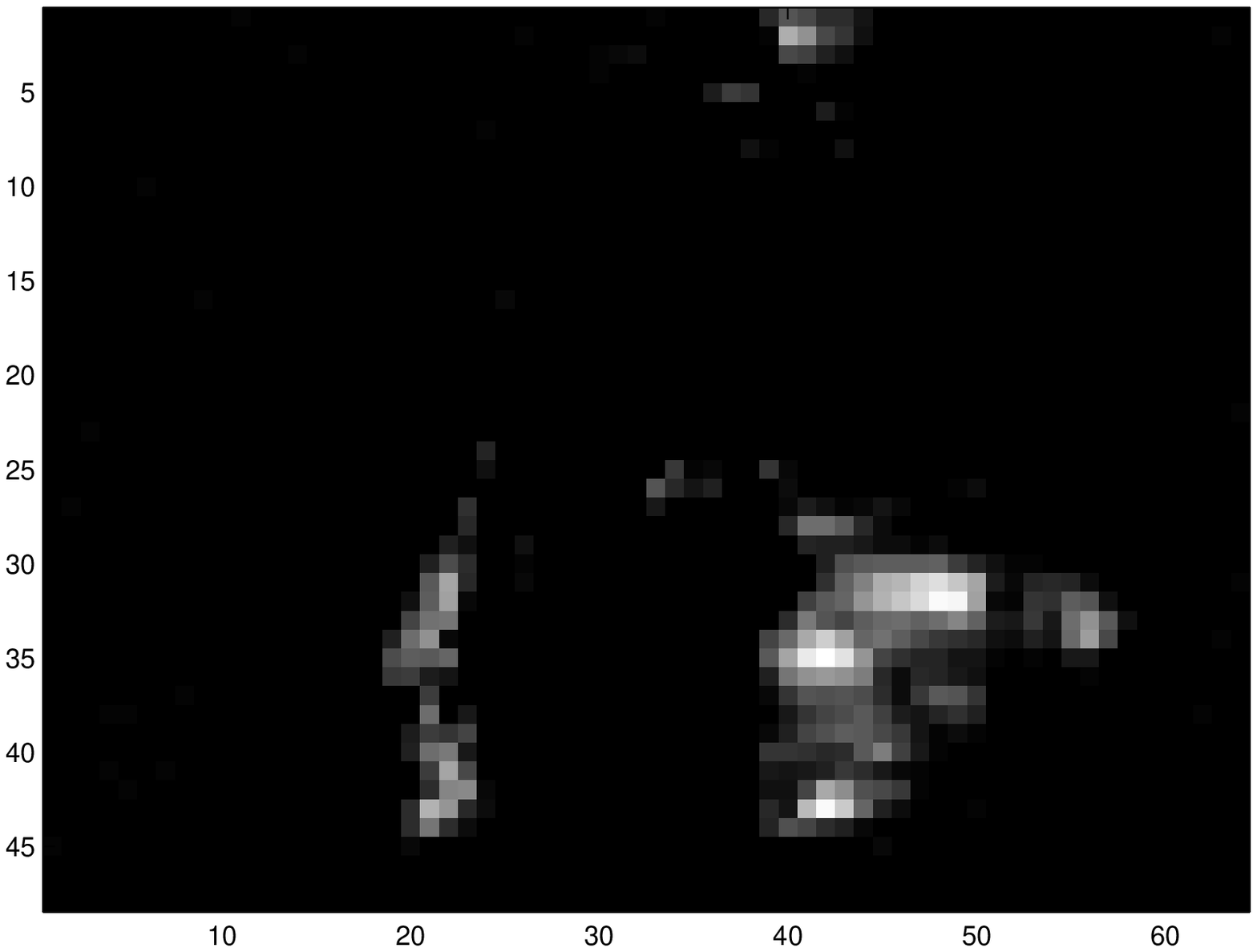}
          \includegraphics[width=1.25cm]{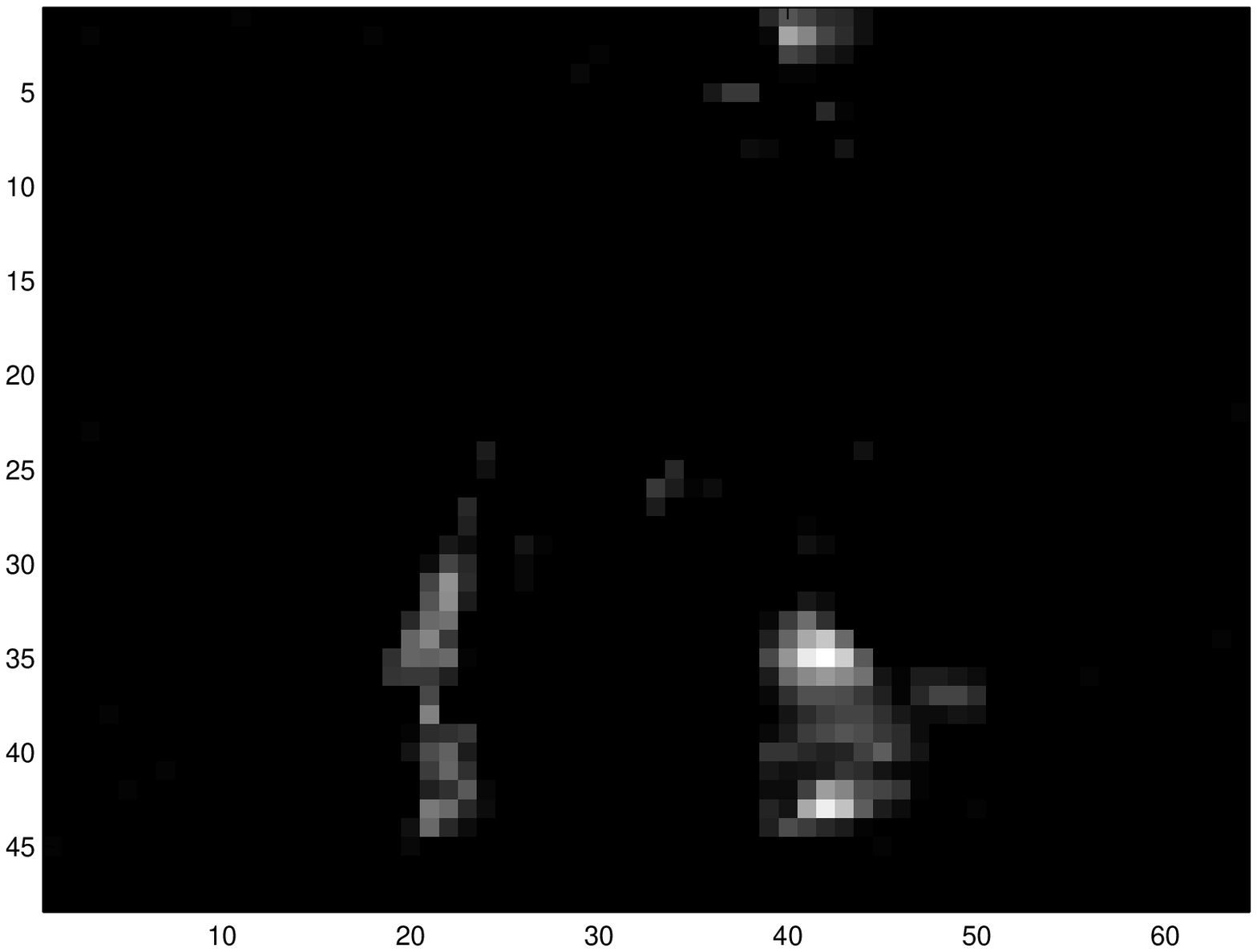}
            \includegraphics[width=1.25cm]{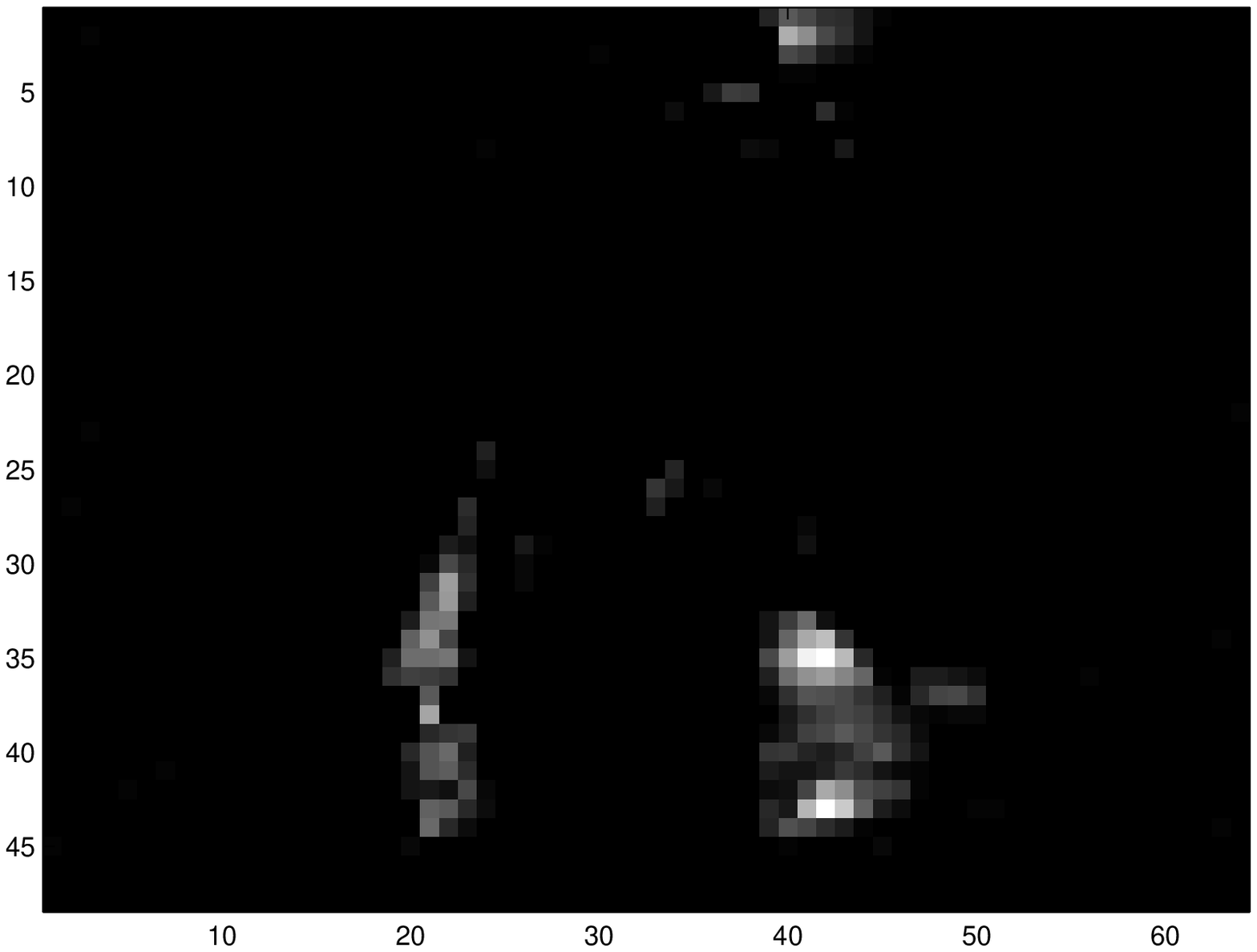}
              \includegraphics[width=1.25cm]{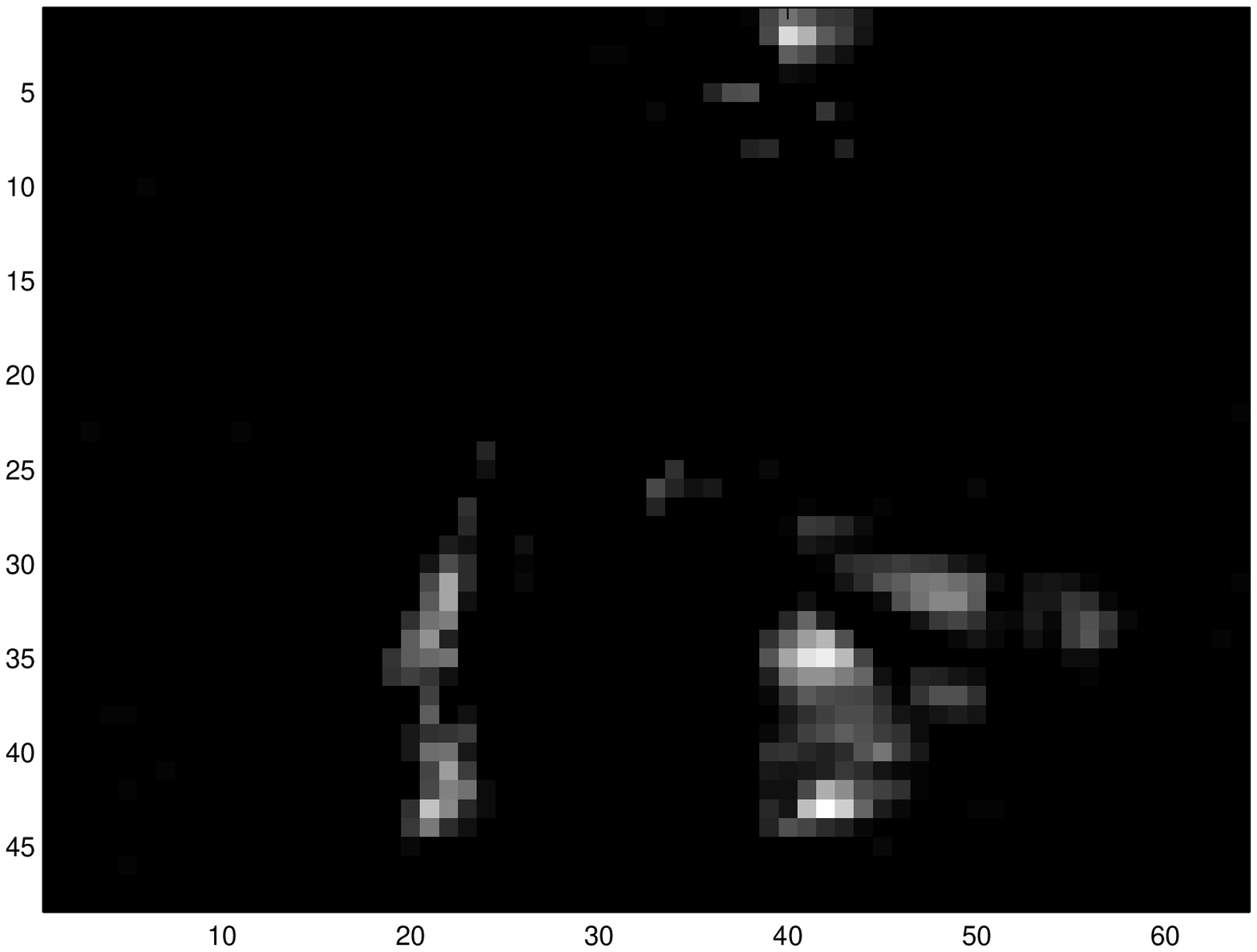}
                \includegraphics[width=1.25cm]{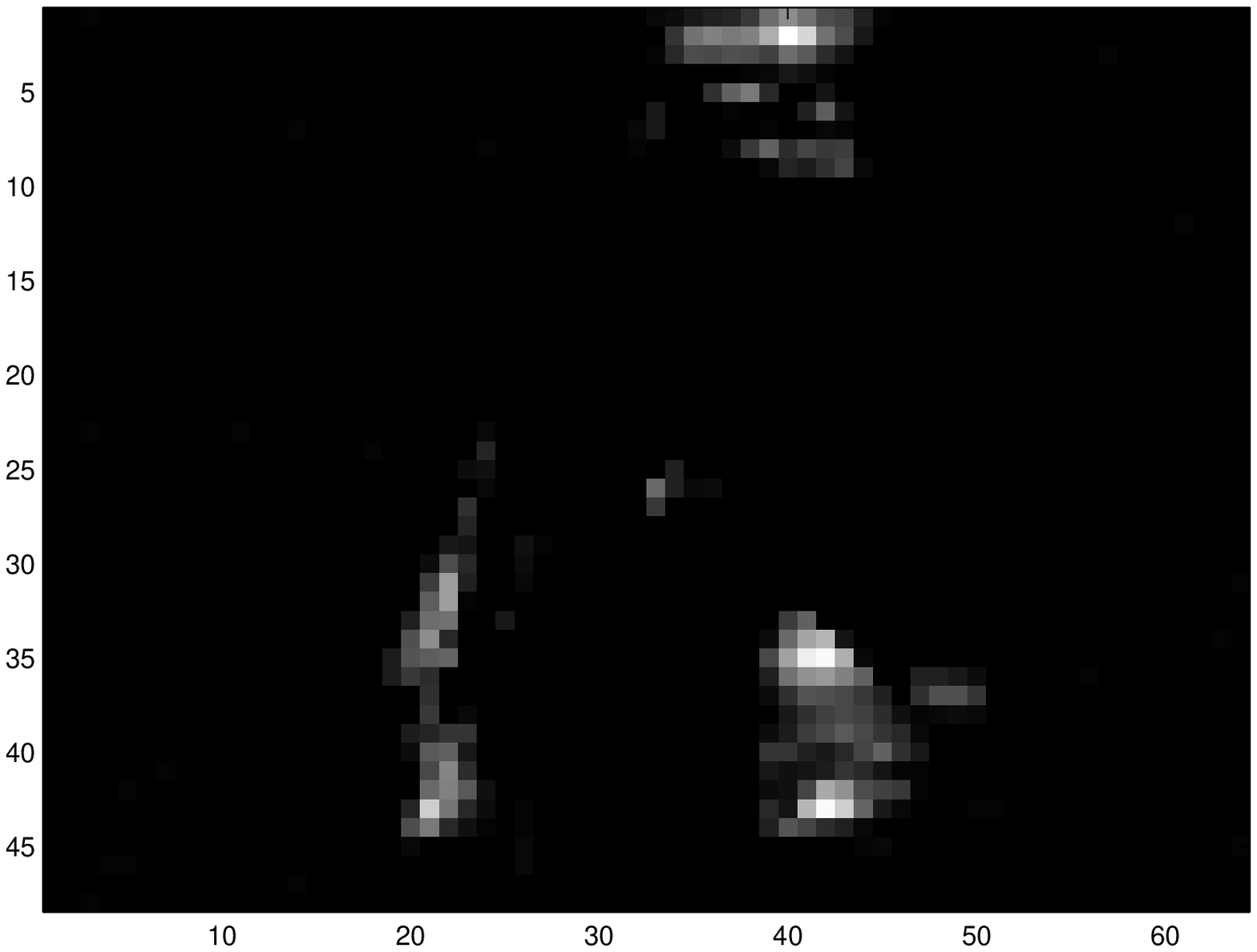}
                \includegraphics[width=1.25cm]{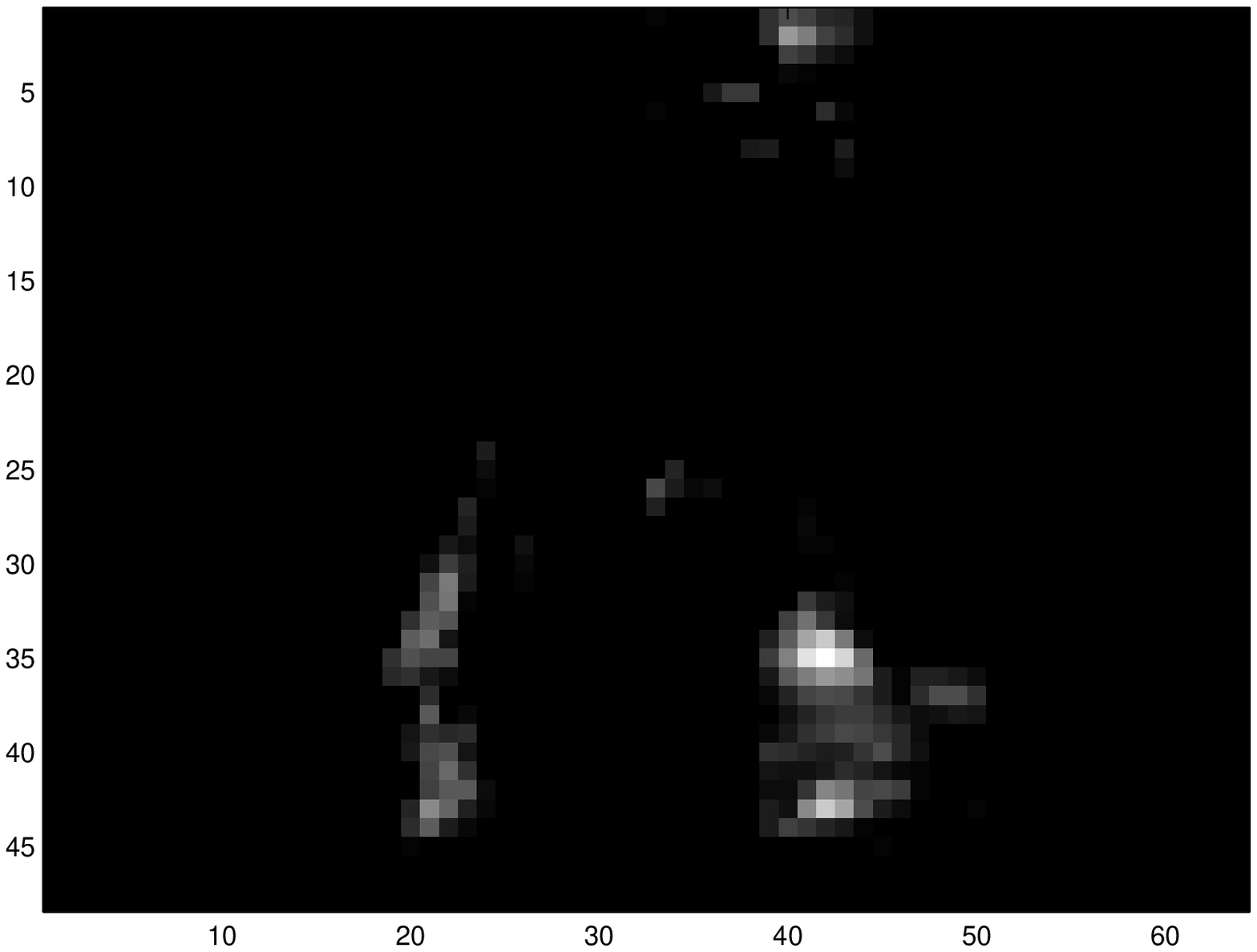}
                \includegraphics[width=1.25cm]{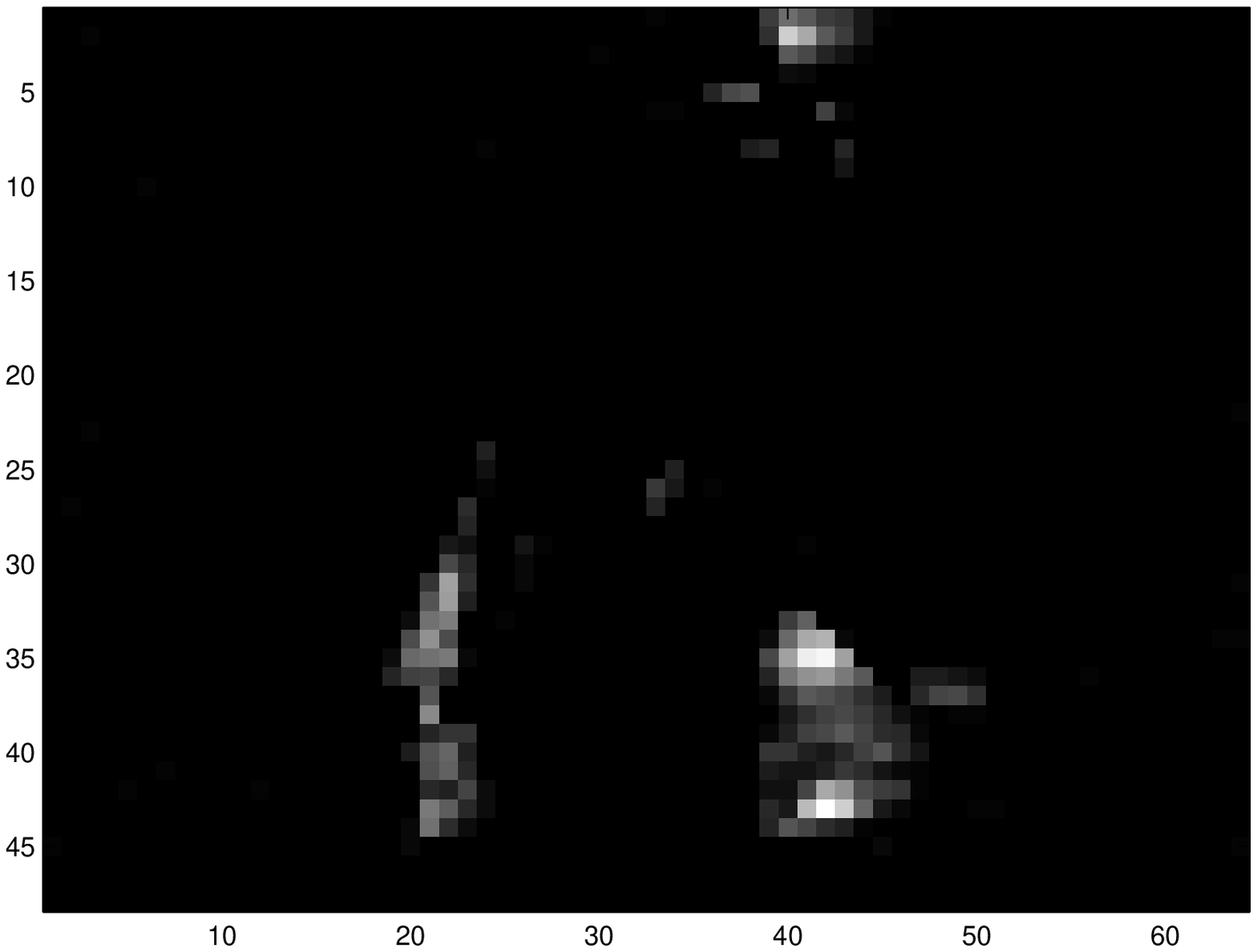}\\
  \includegraphics[width=1.25cm]{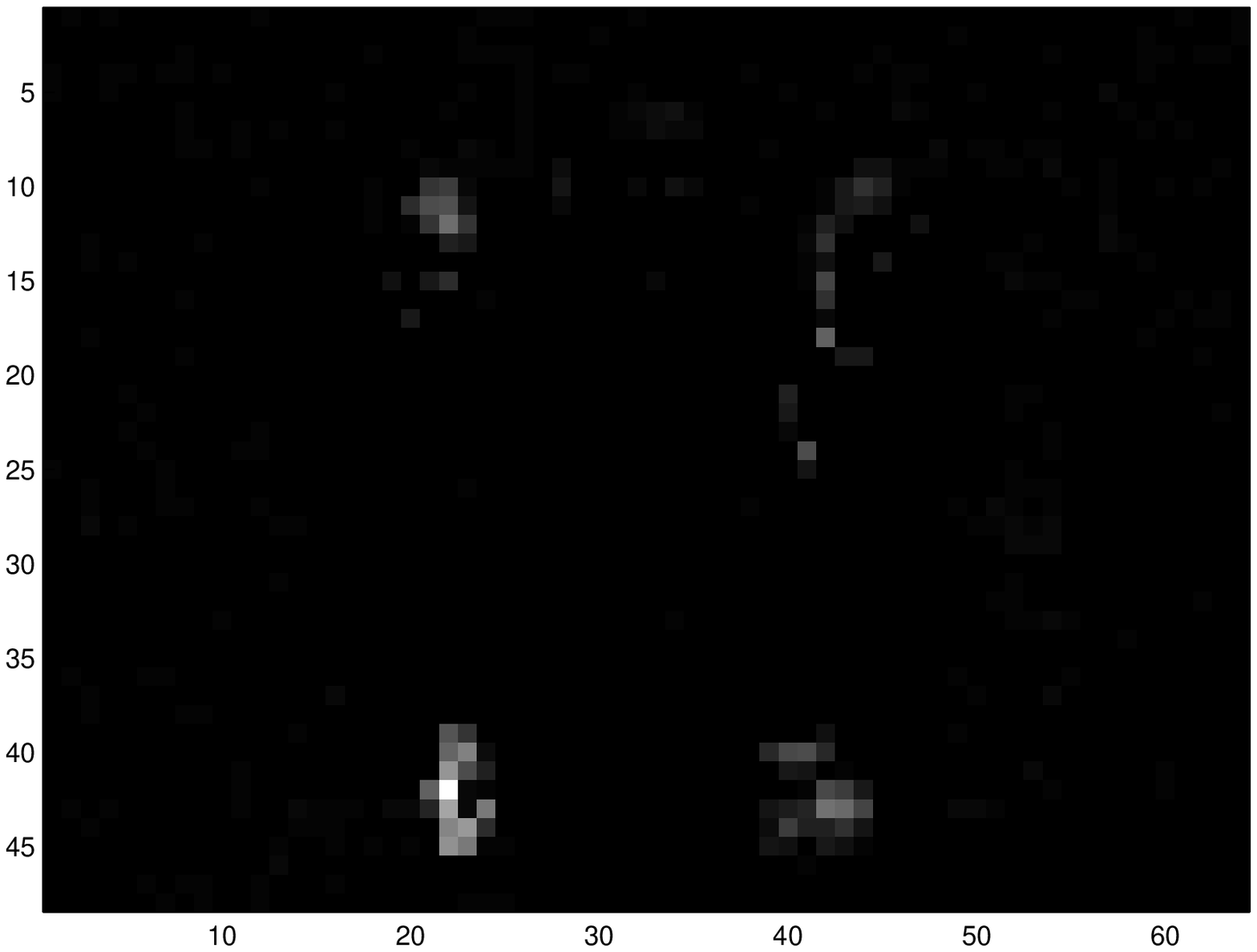}
    \includegraphics[width=1.25cm]{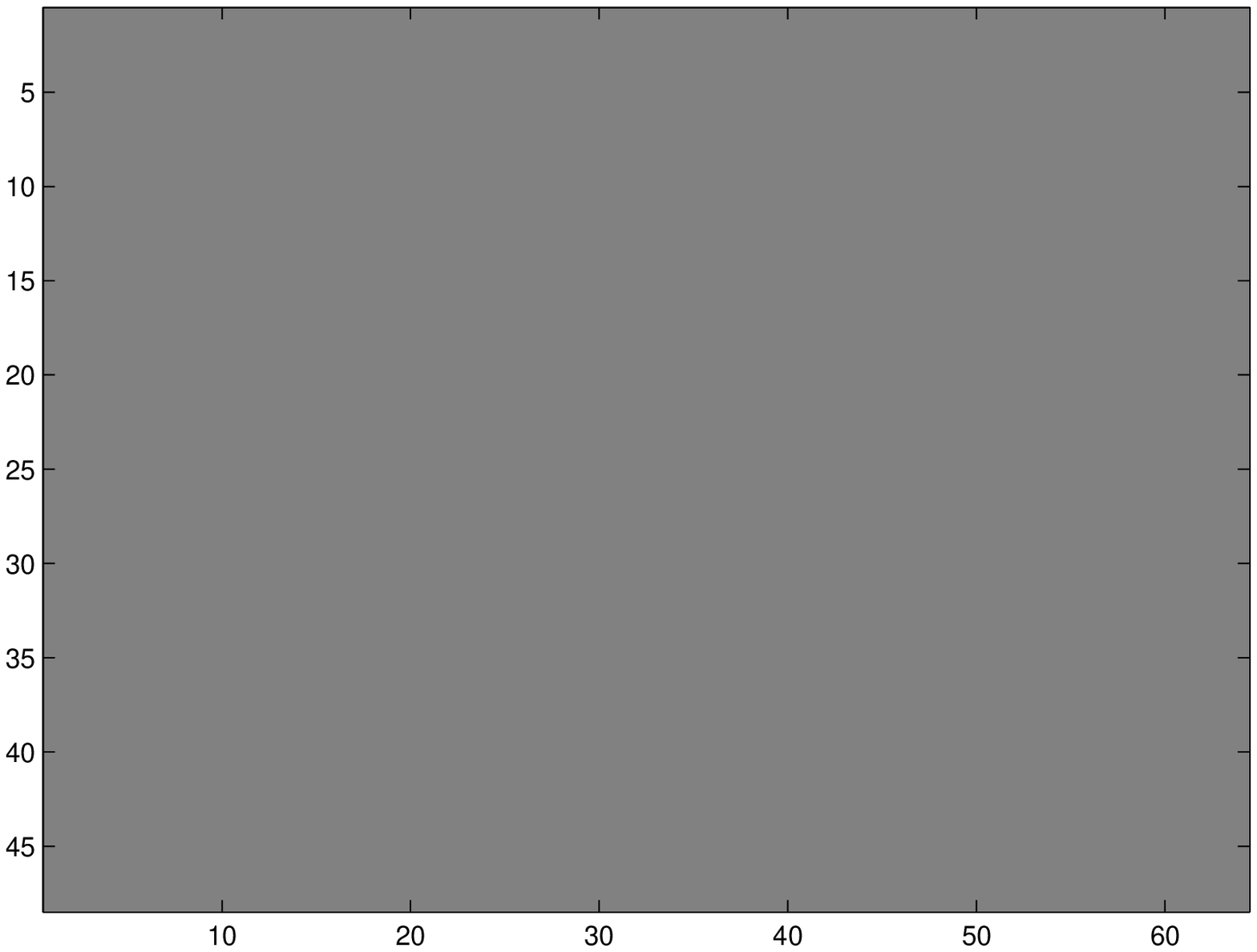}
      \includegraphics[width=1.25cm]{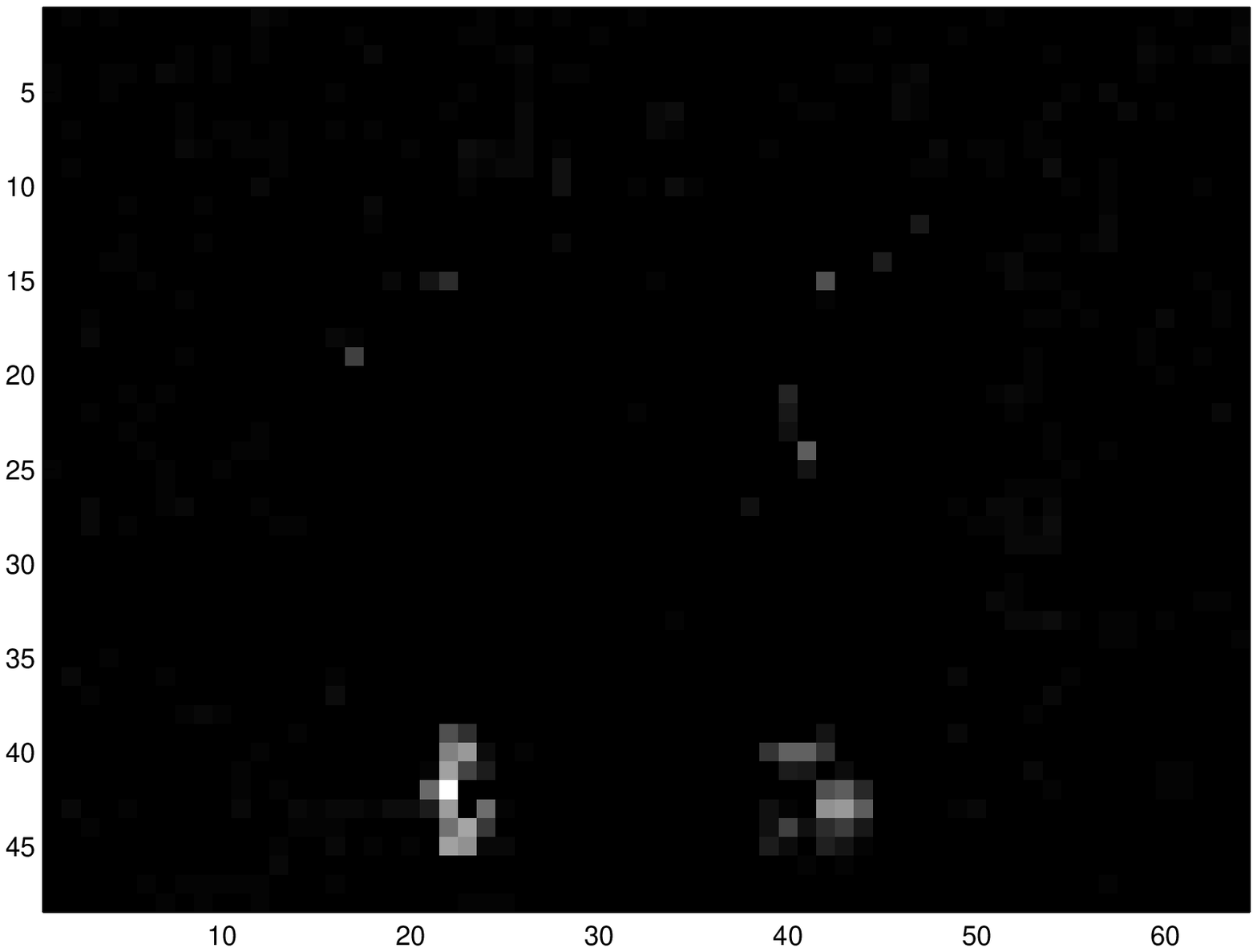}
        \includegraphics[width=1.25cm]{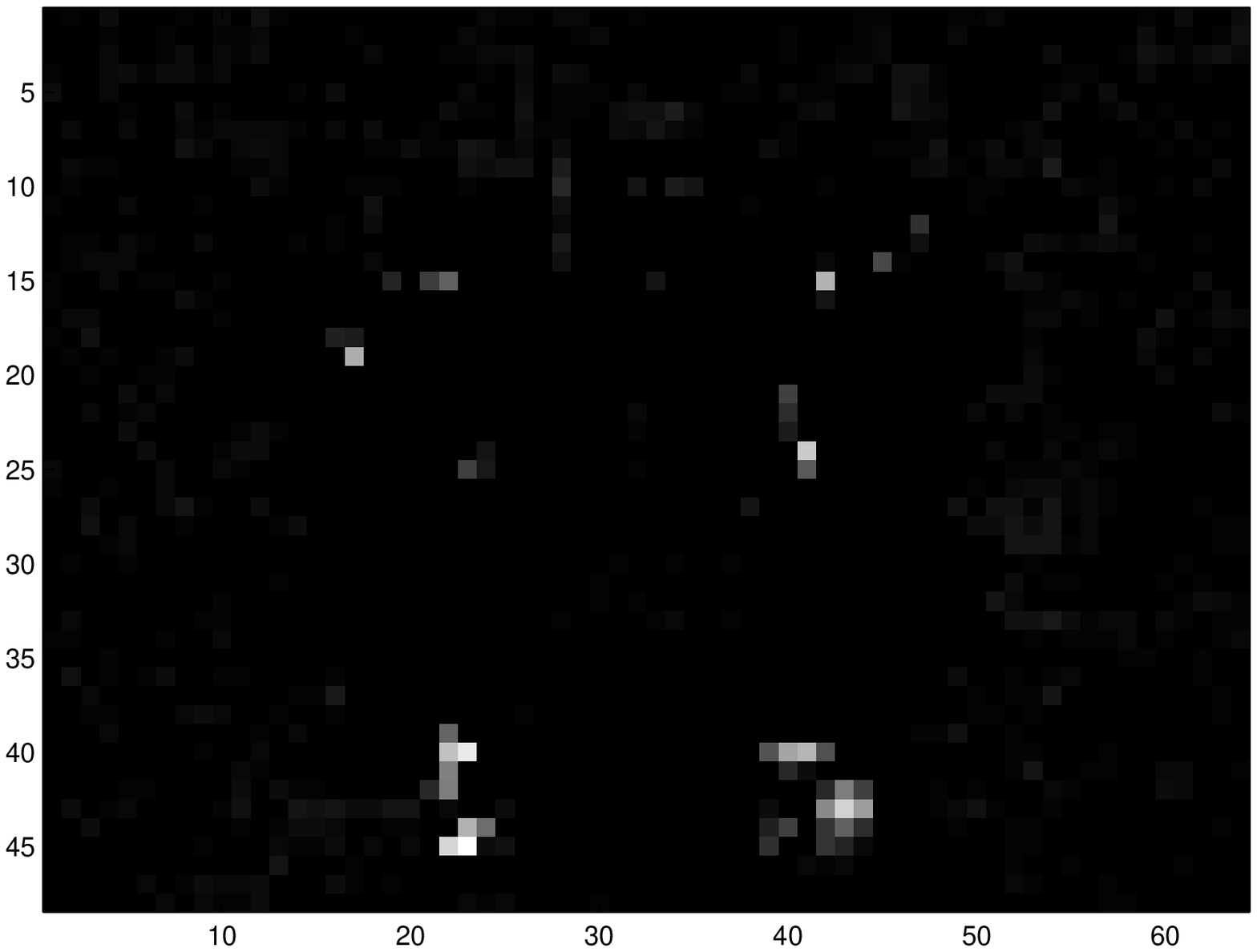}
          \includegraphics[width=1.25cm]{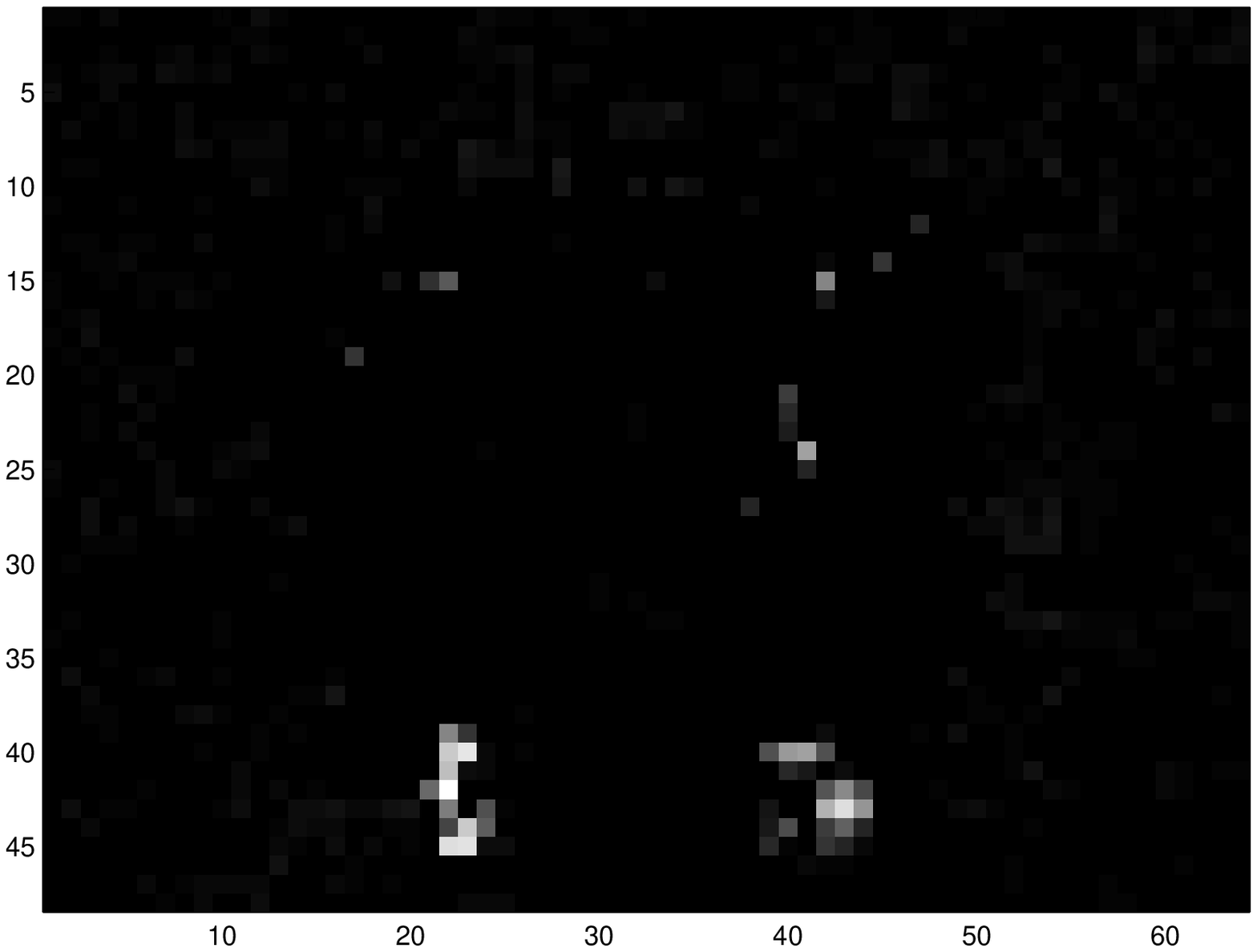}
            \includegraphics[width=1.25cm]{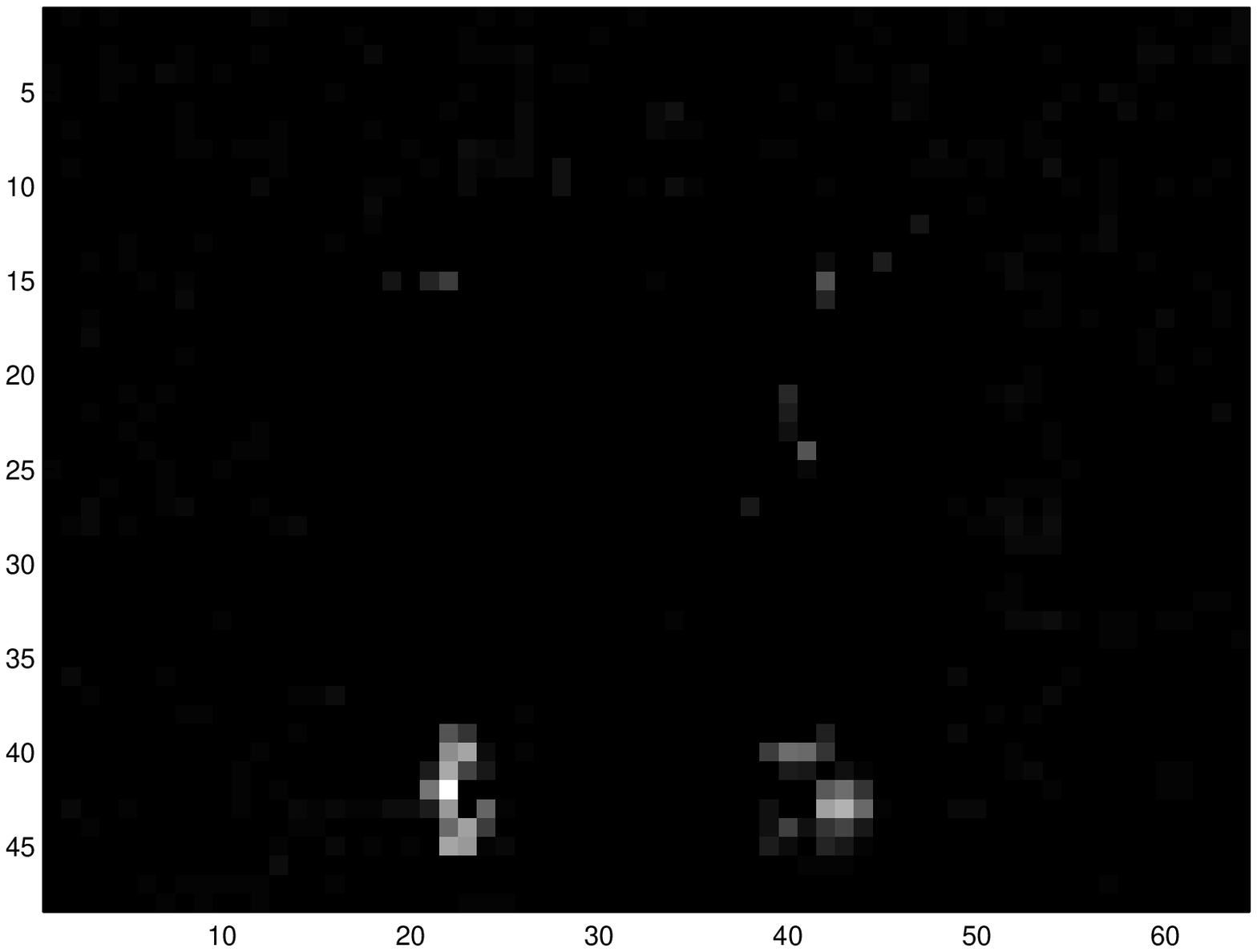}
              \includegraphics[width=1.25cm]{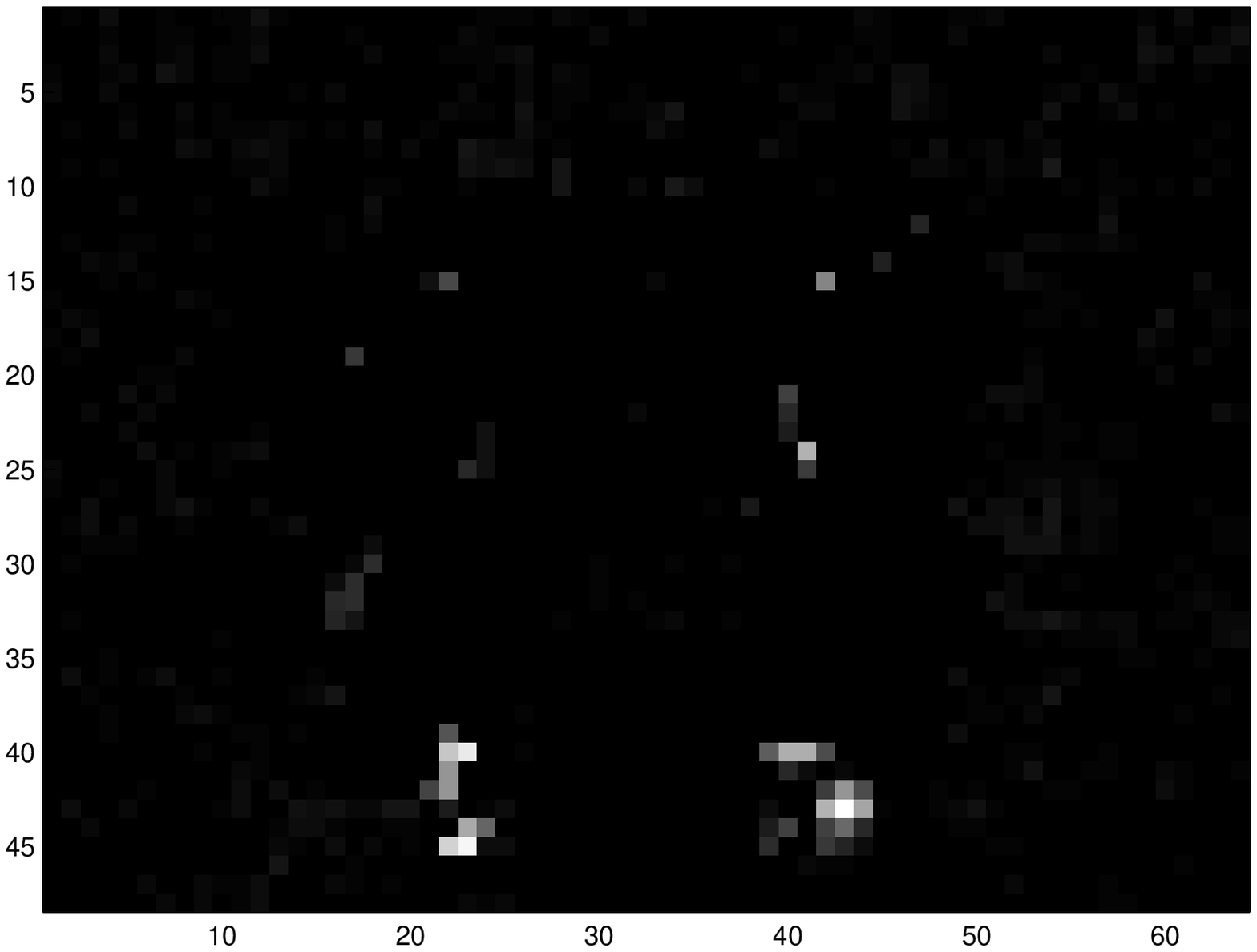}
                \includegraphics[width=1.25cm]{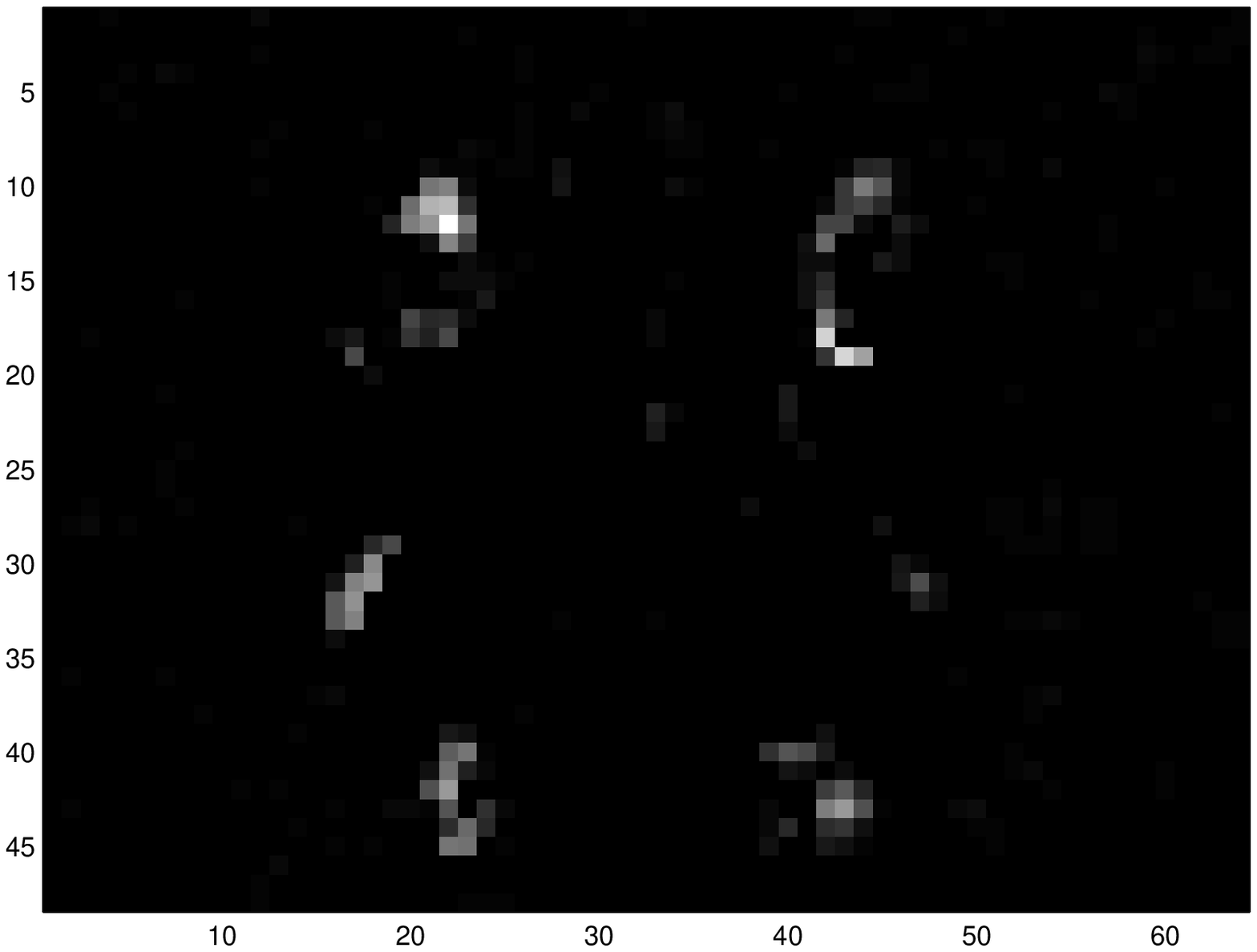}
                                \includegraphics[width=1.25cm]{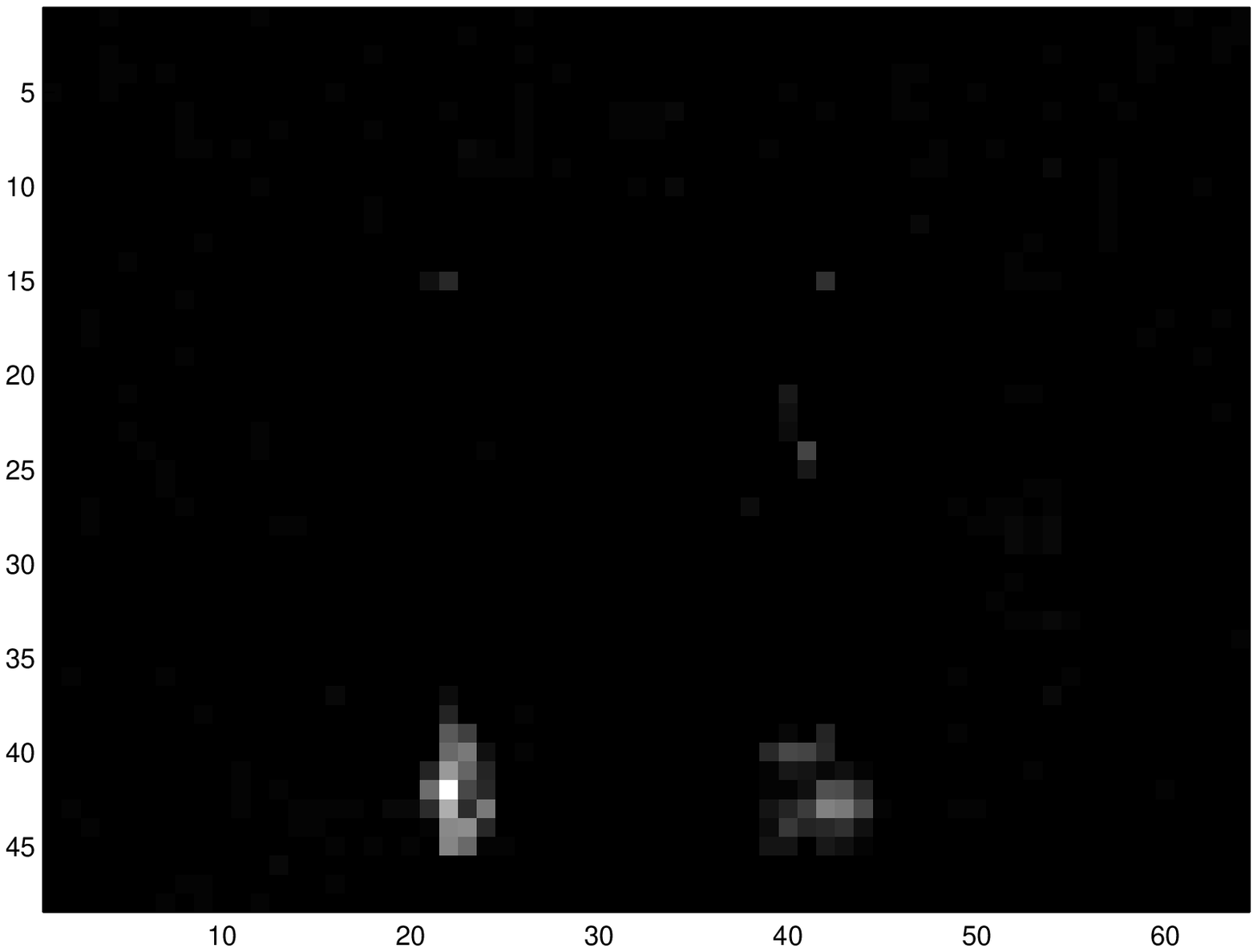}
            \includegraphics[width=1.25cm]{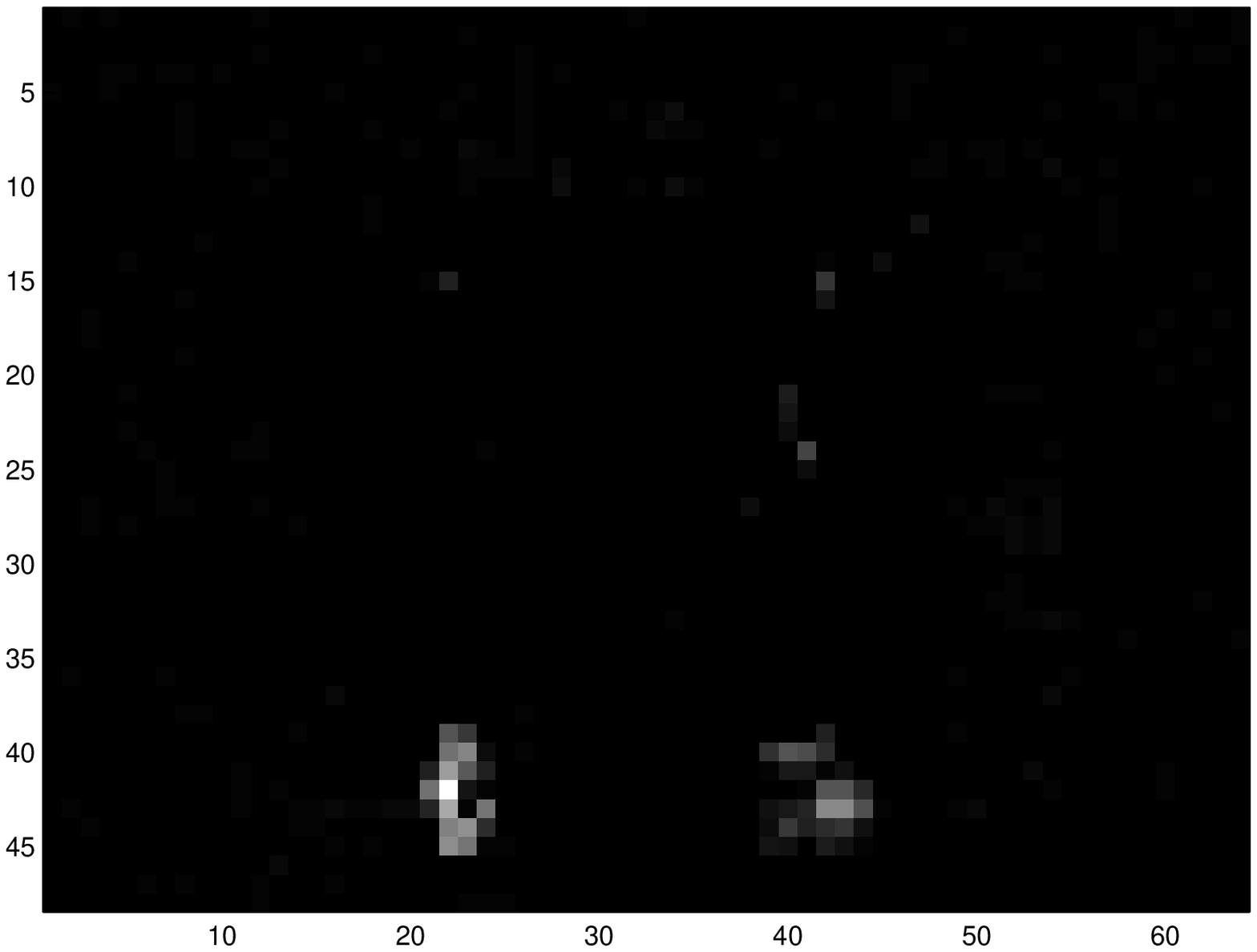}\\
  \includegraphics[width=1.25cm]{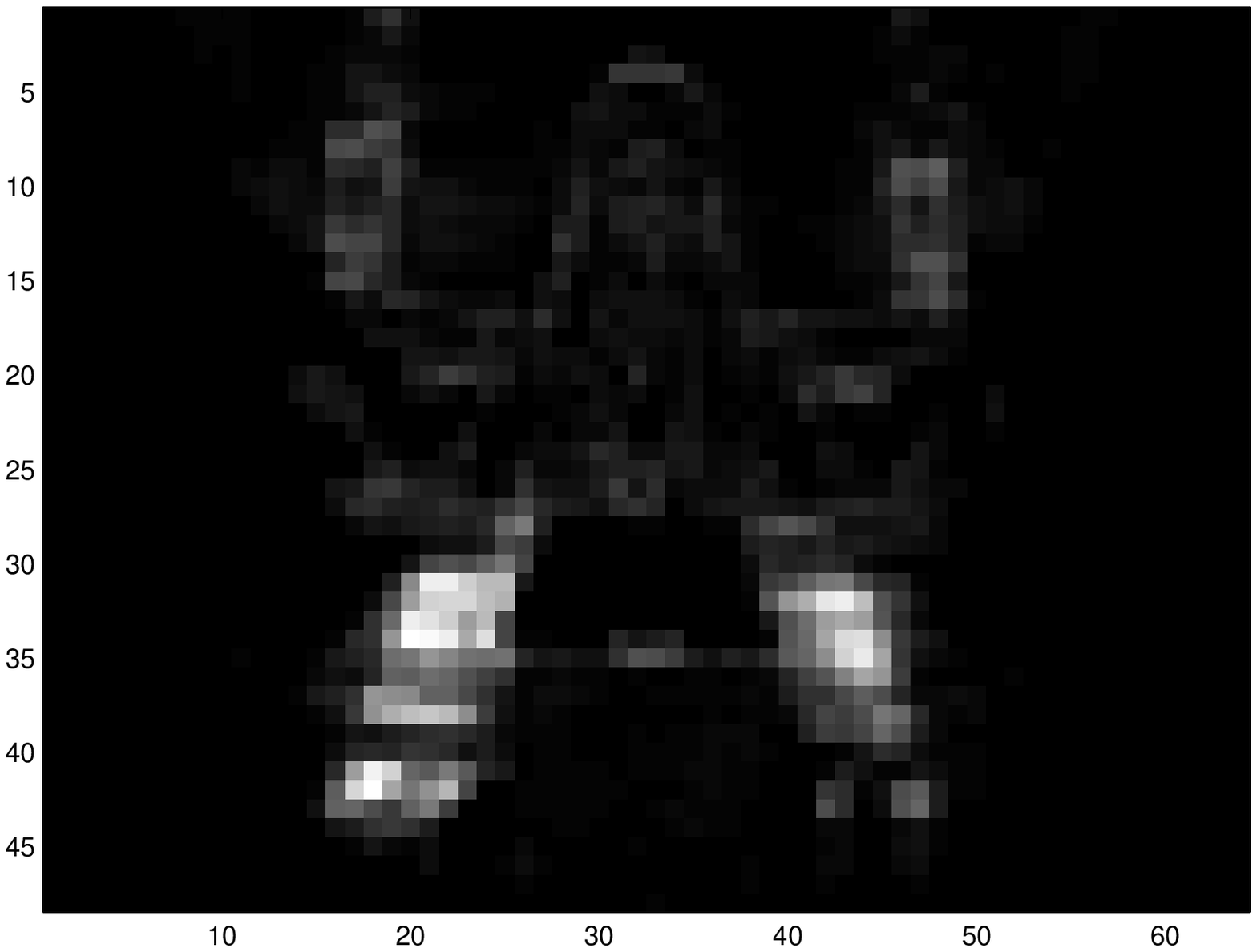}
    \includegraphics[width=1.25cm]{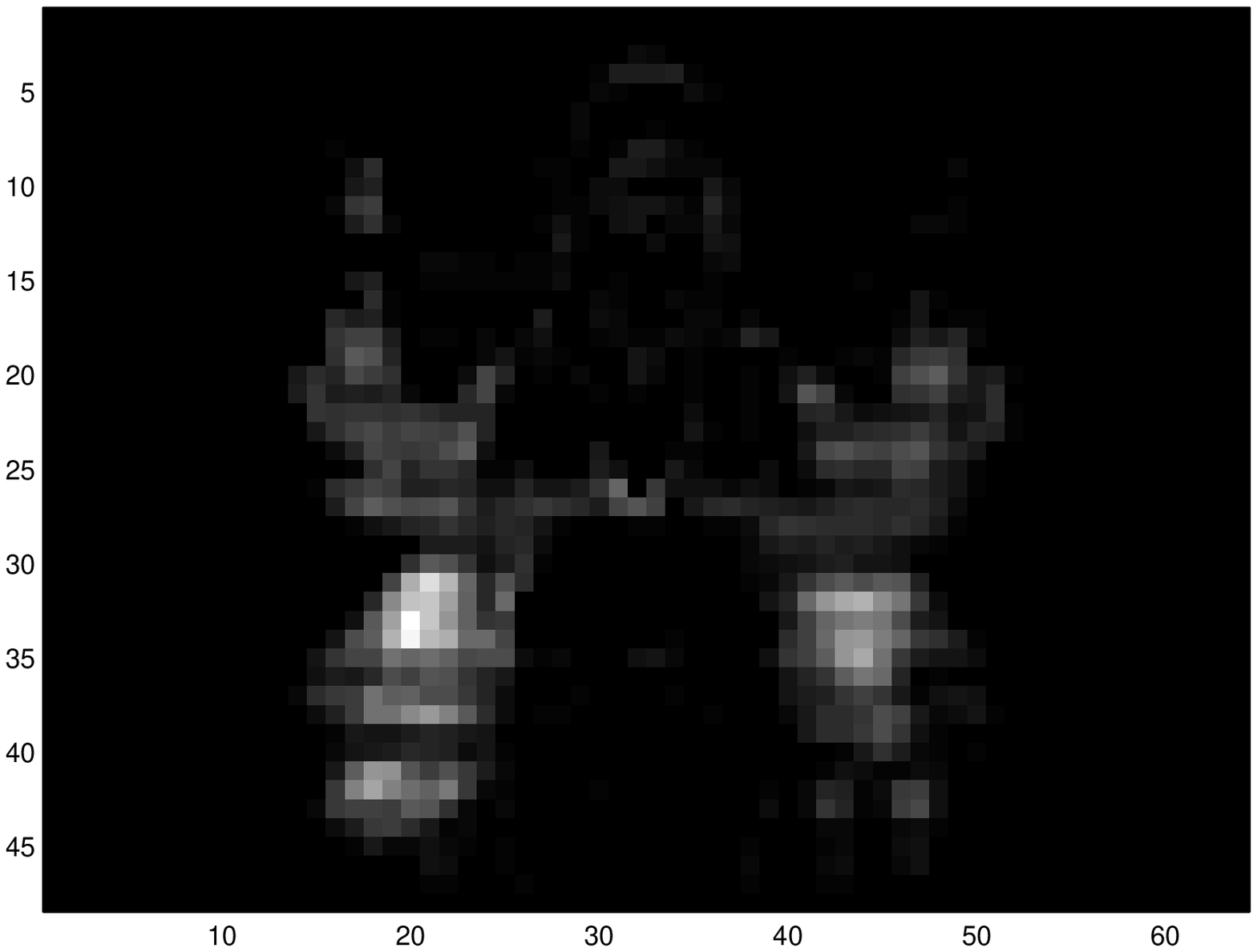}
      \includegraphics[width=1.25cm]{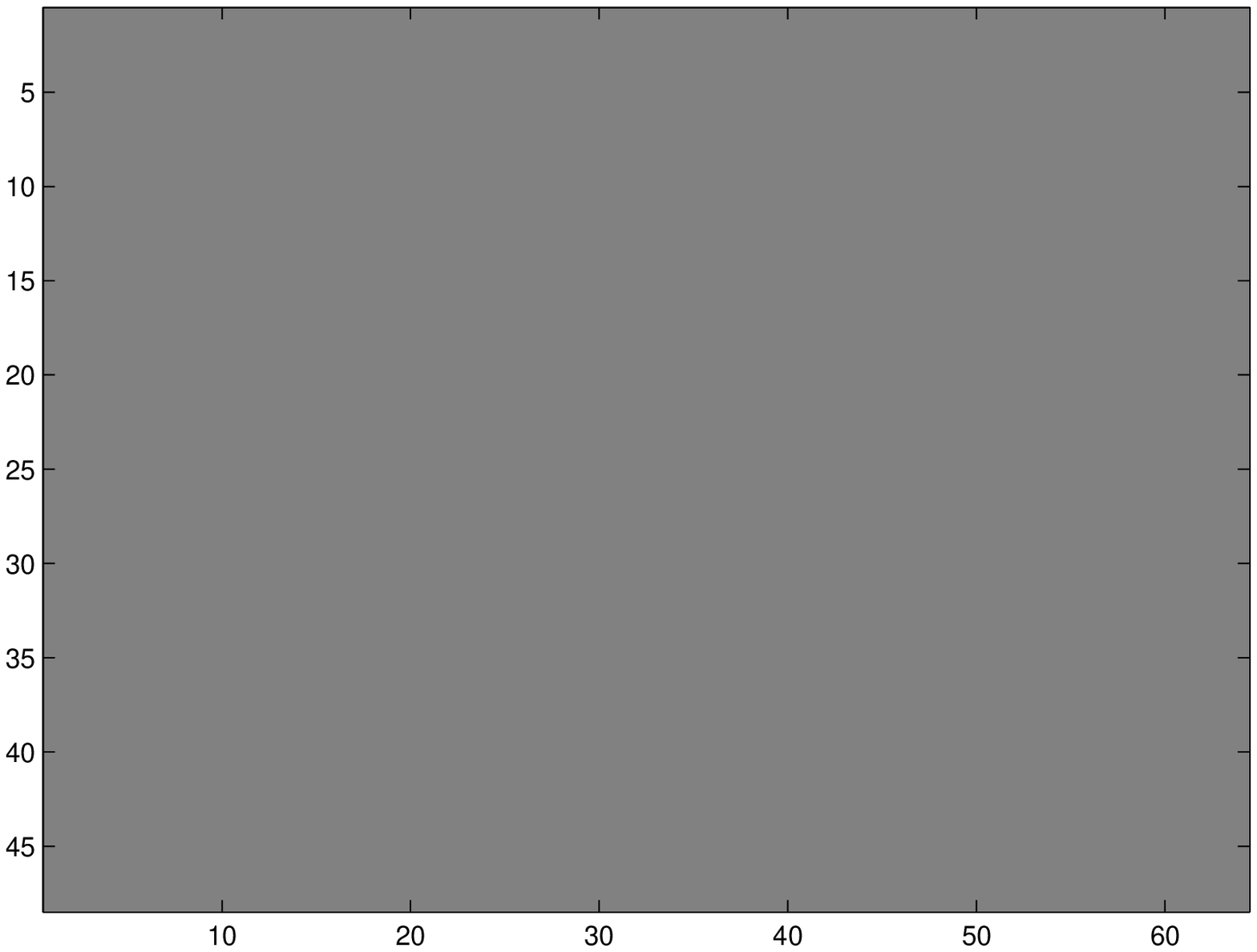}
        \includegraphics[width=1.25cm]{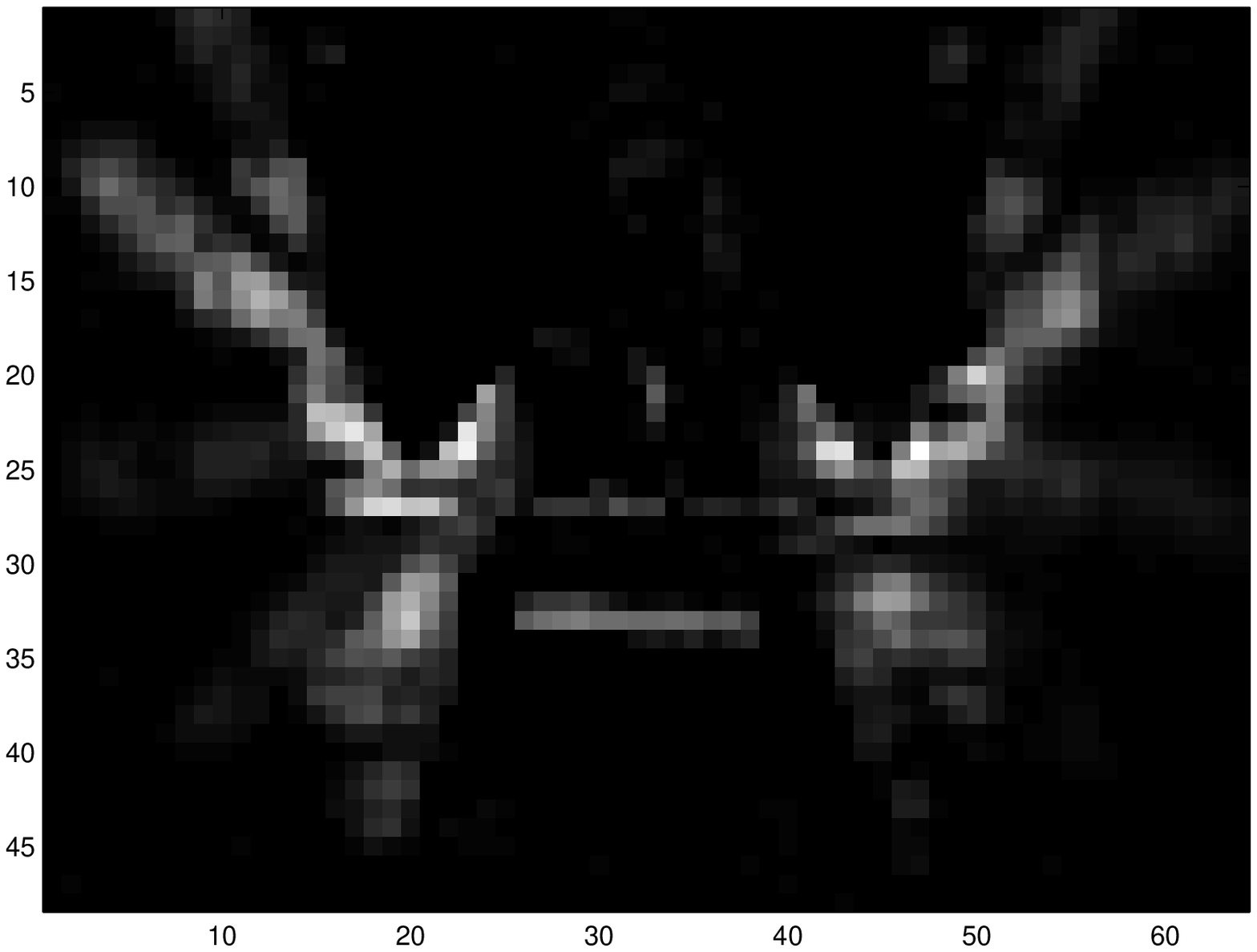}
          \includegraphics[width=1.25cm]{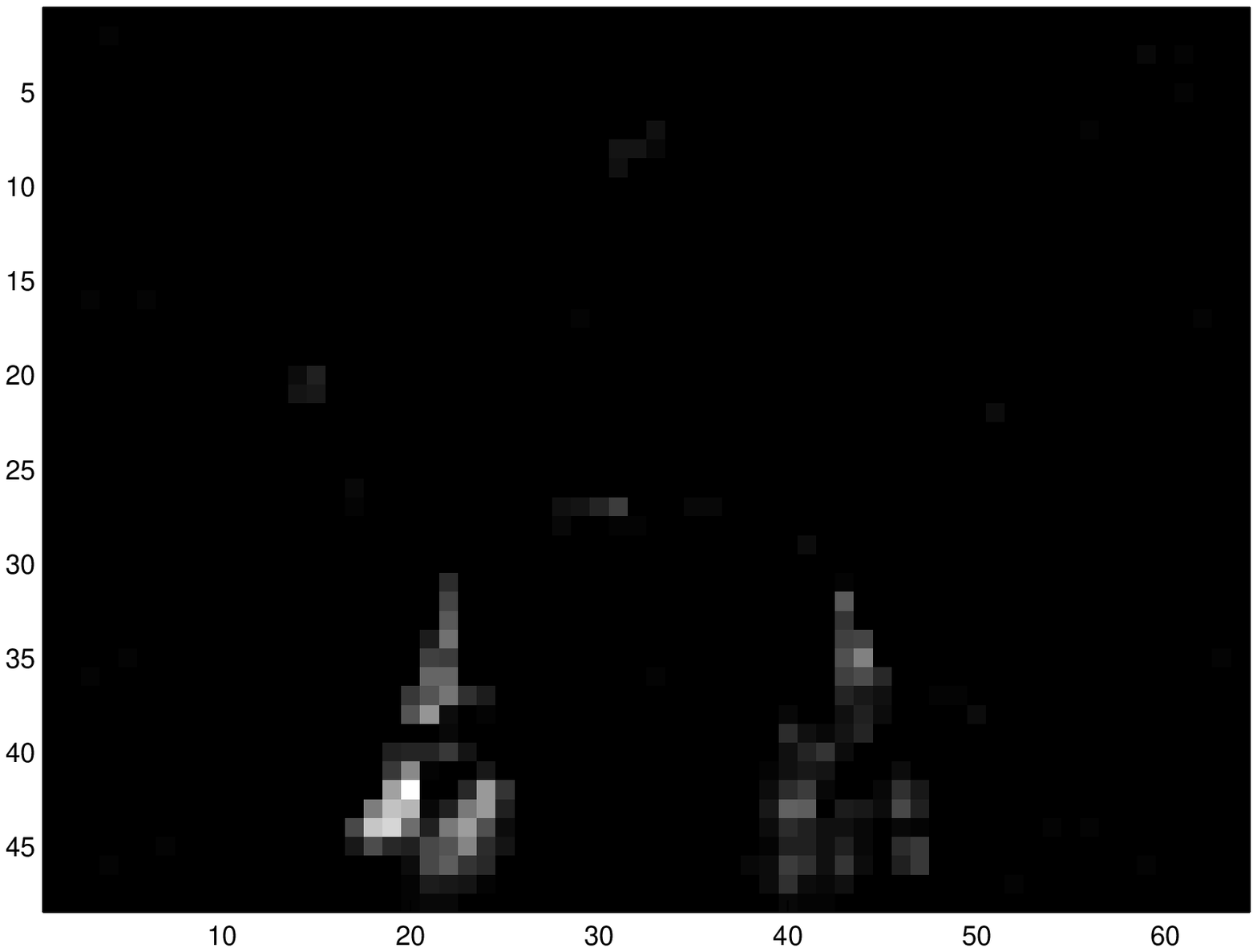}
            \includegraphics[width=1.25cm]{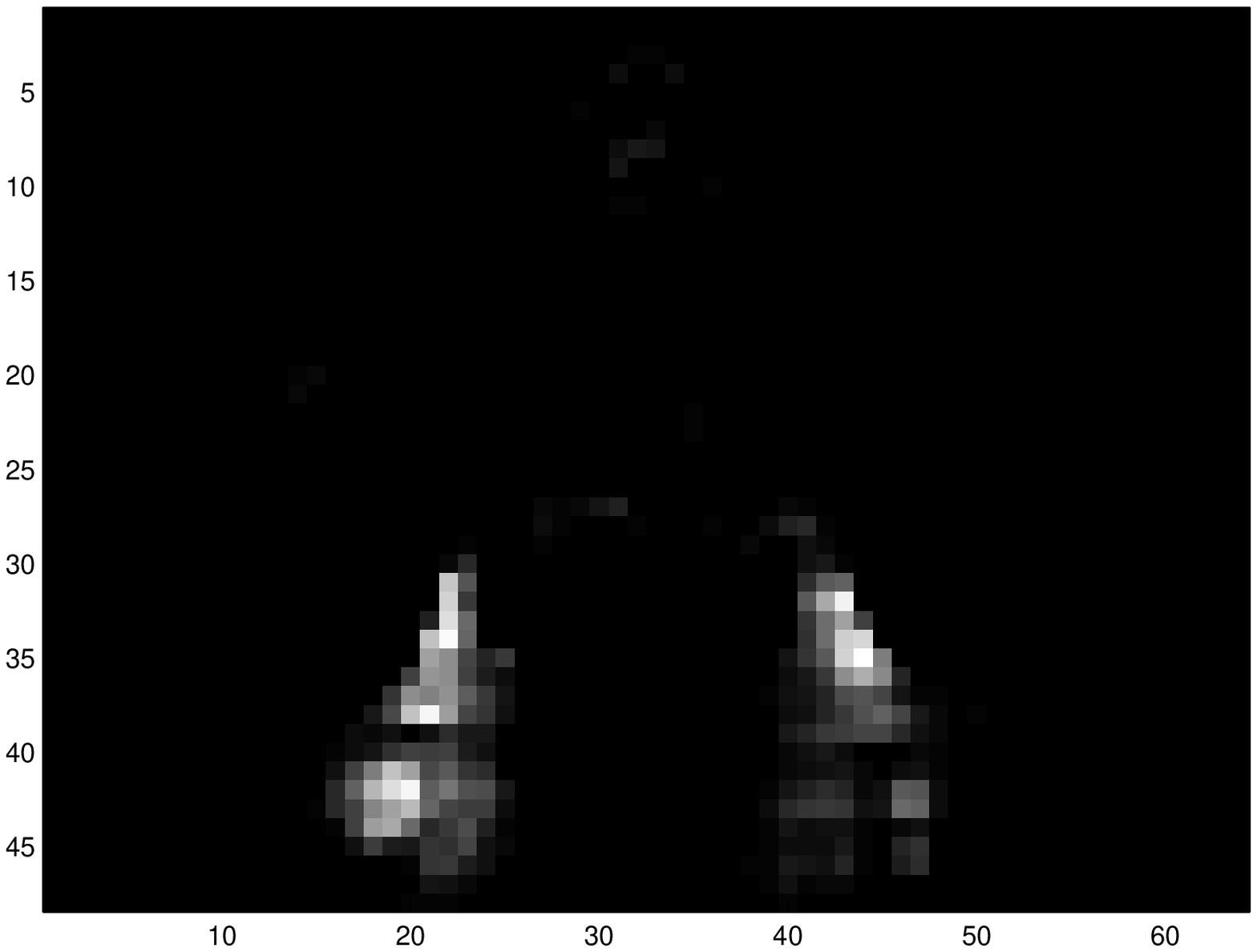}
              \includegraphics[width=1.25cm]{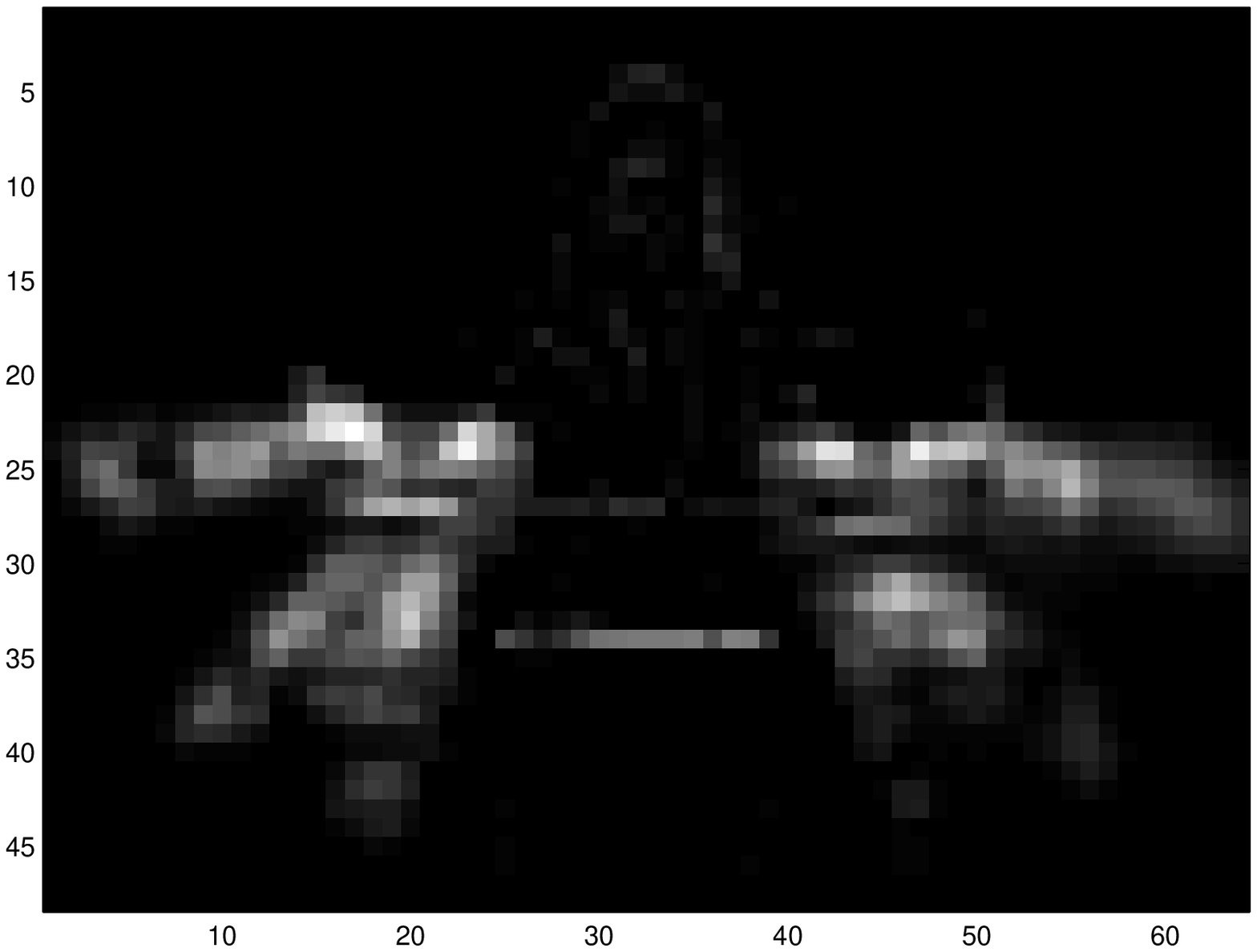}
                \includegraphics[width=1.25cm]{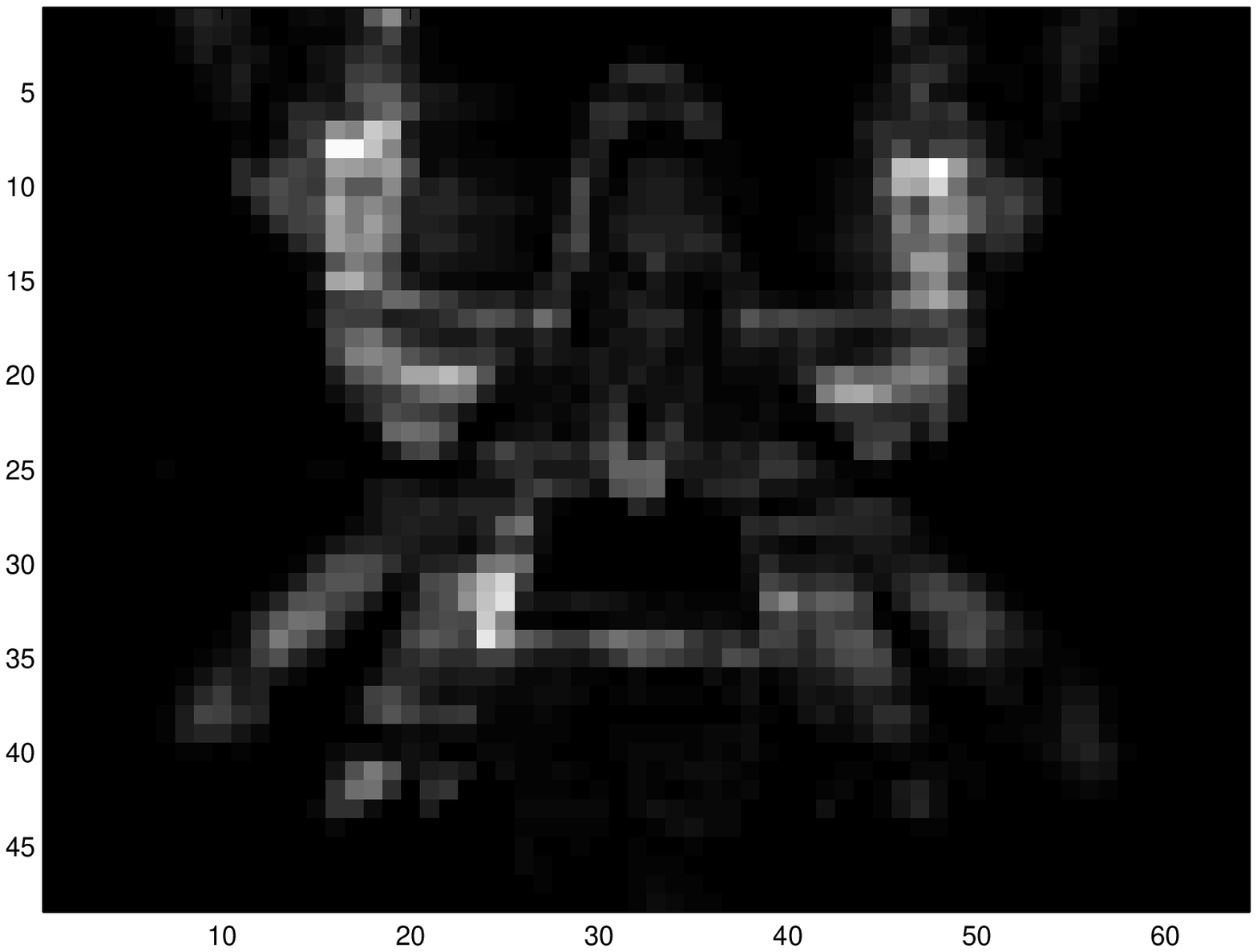}
                \includegraphics[width=1.25cm]{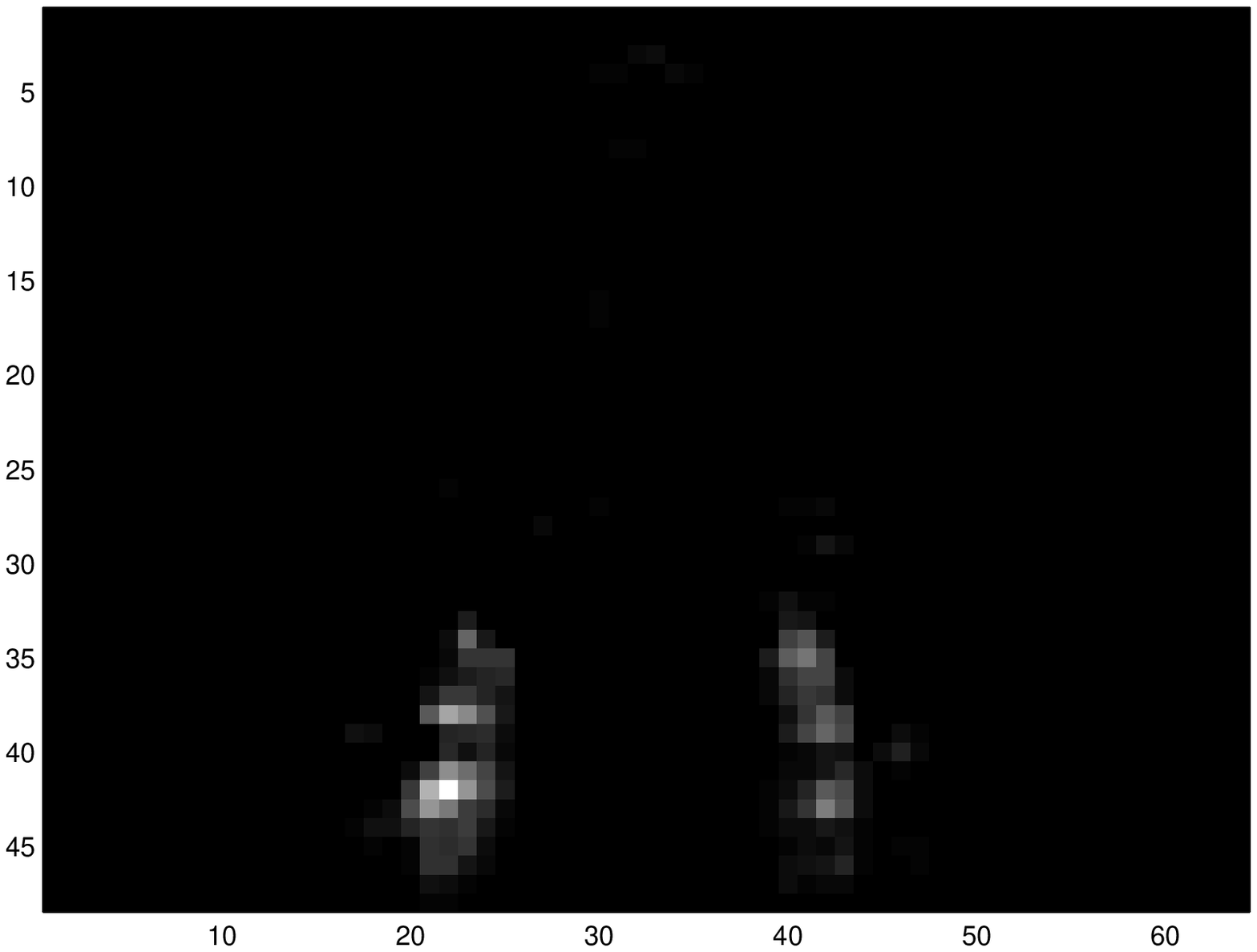}
                \includegraphics[width=1.25cm]{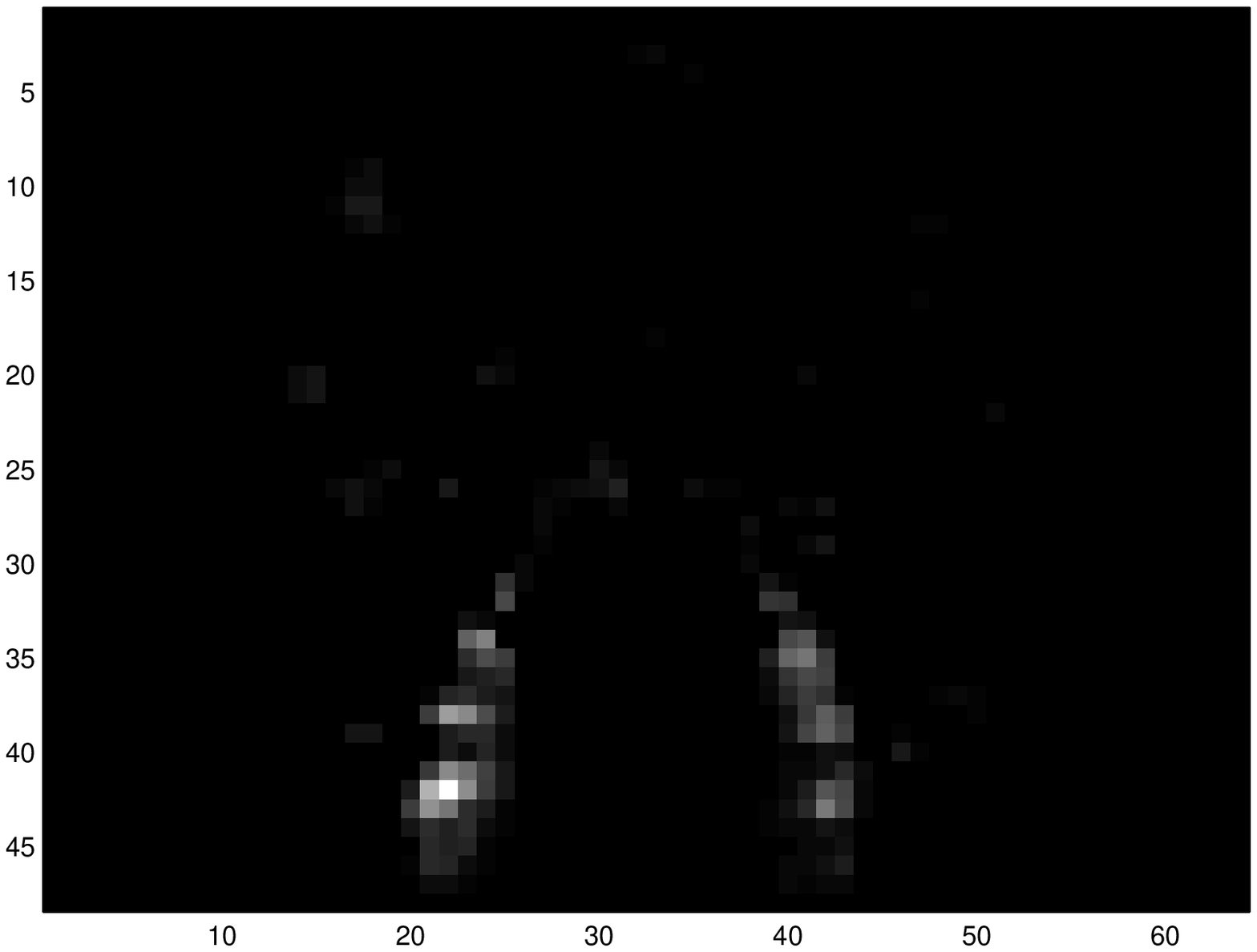}\\
  \includegraphics[width=1.25cm]{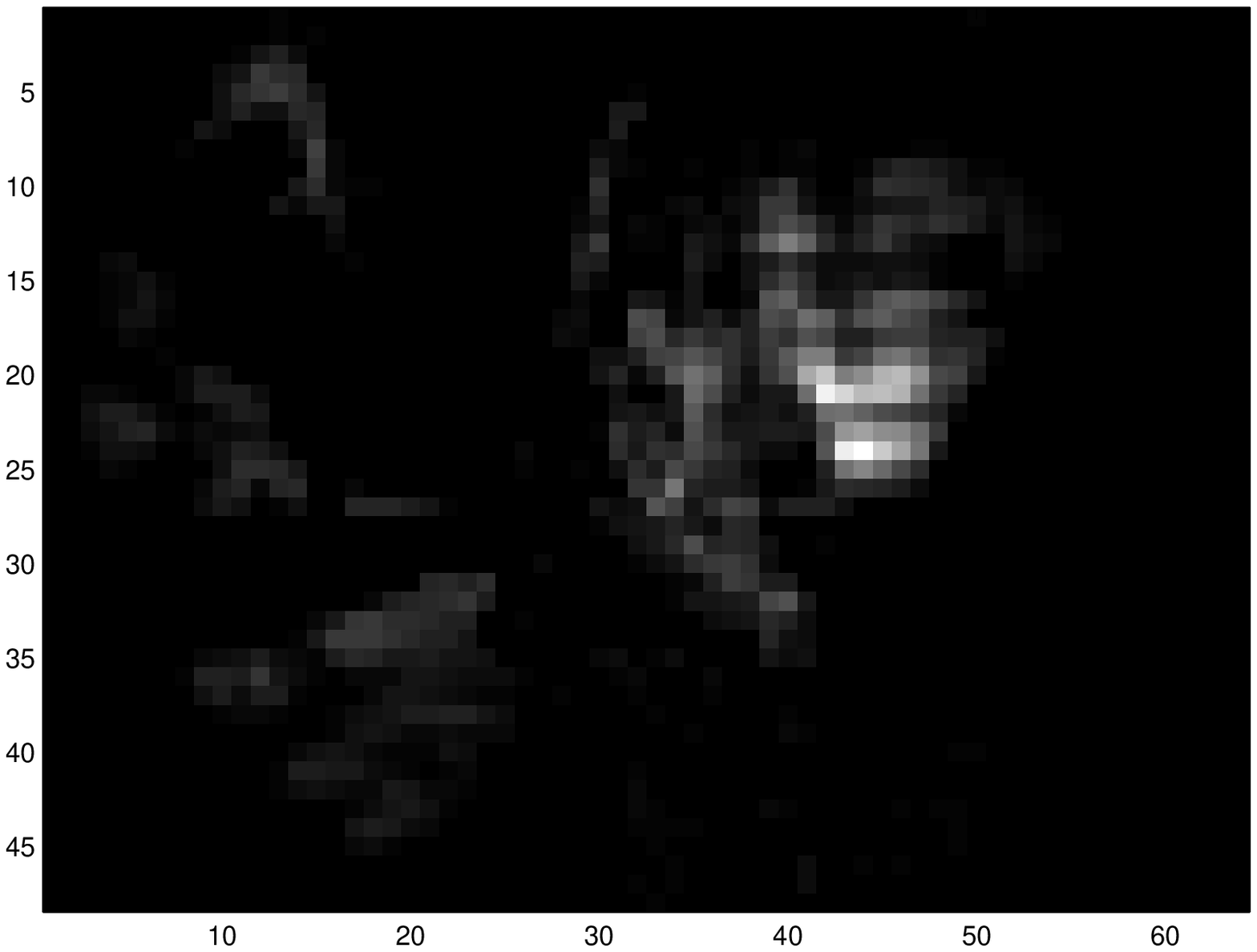}
    \includegraphics[width=1.25cm]{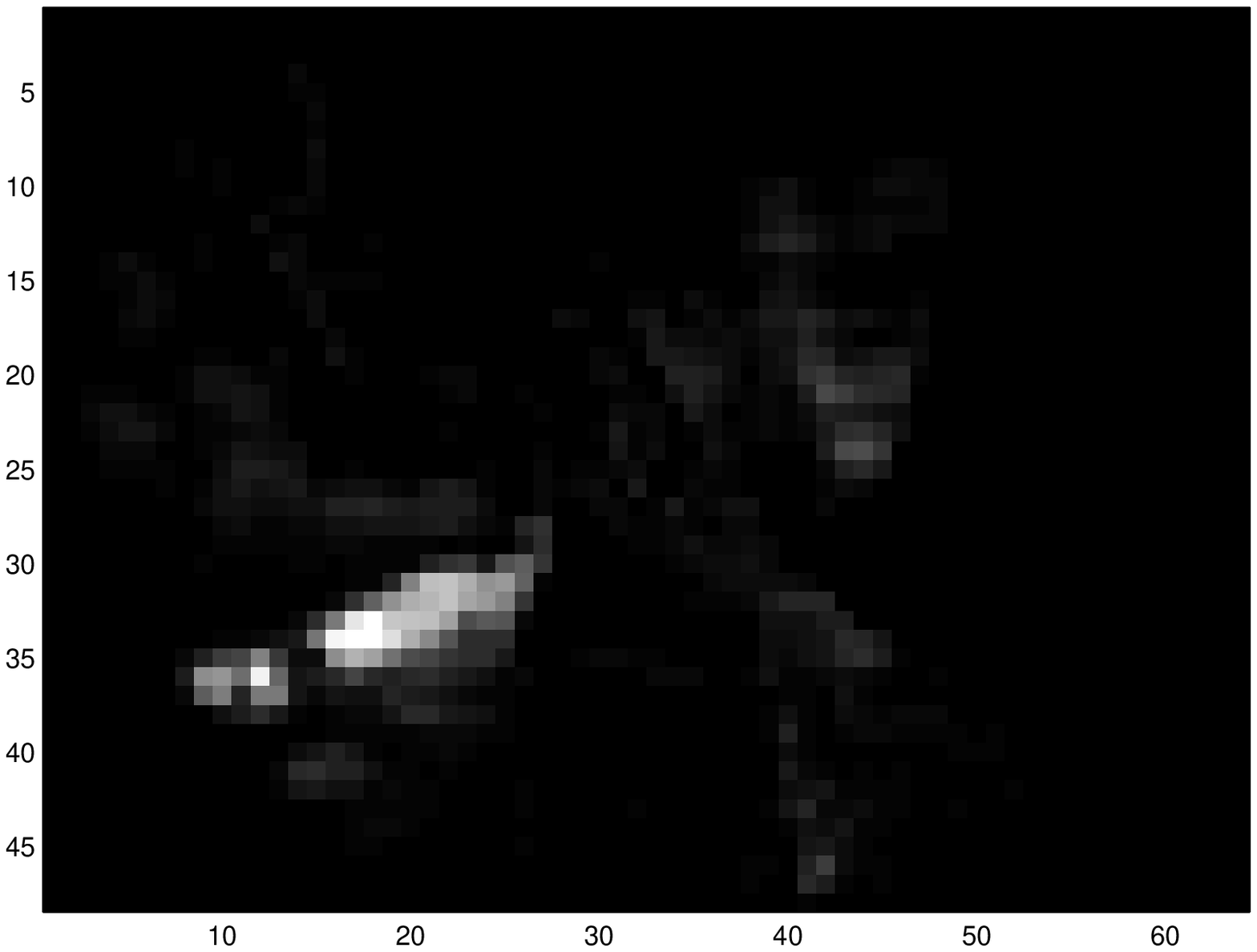}
      \includegraphics[width=1.25cm]{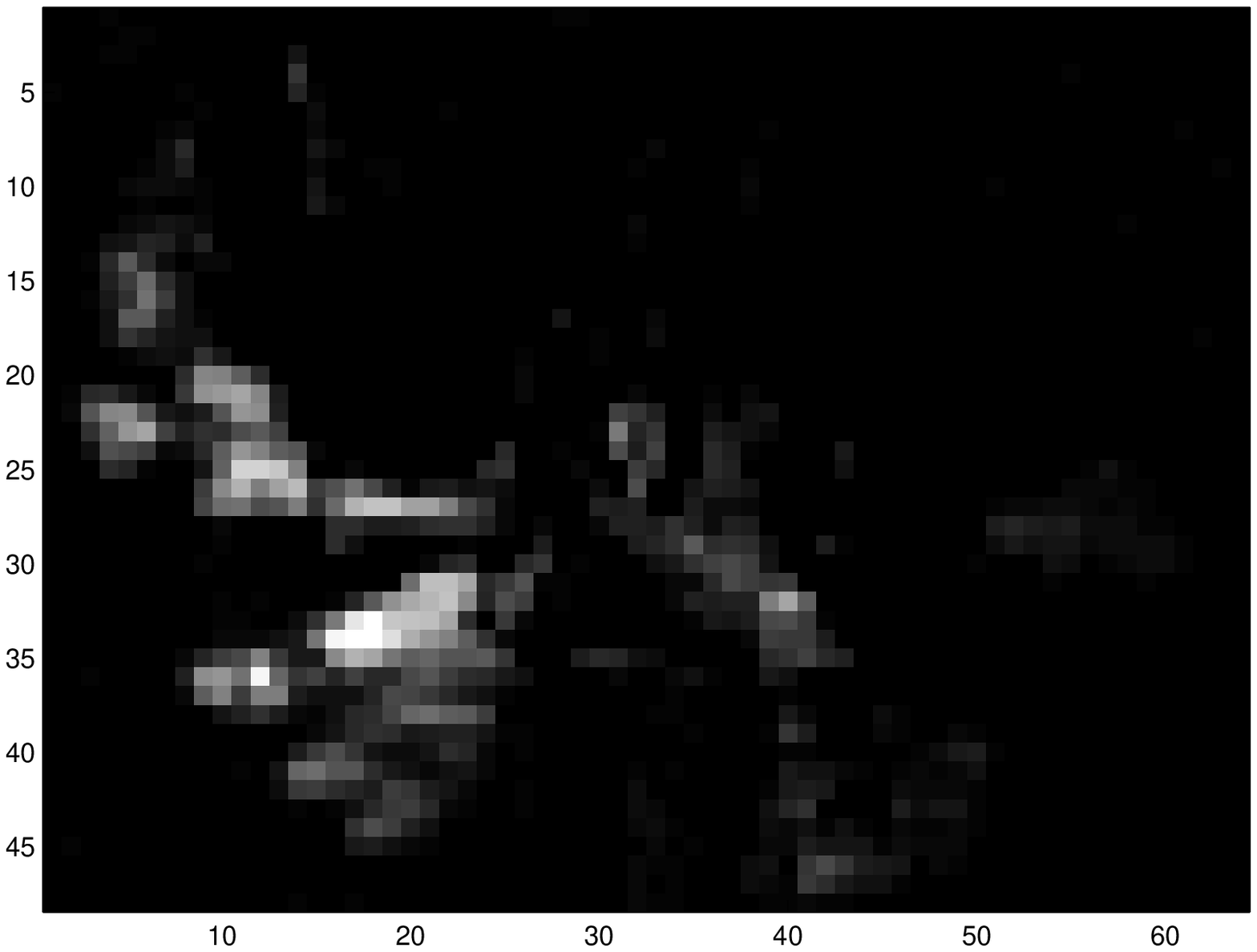}
        \includegraphics[width=1.25cm]{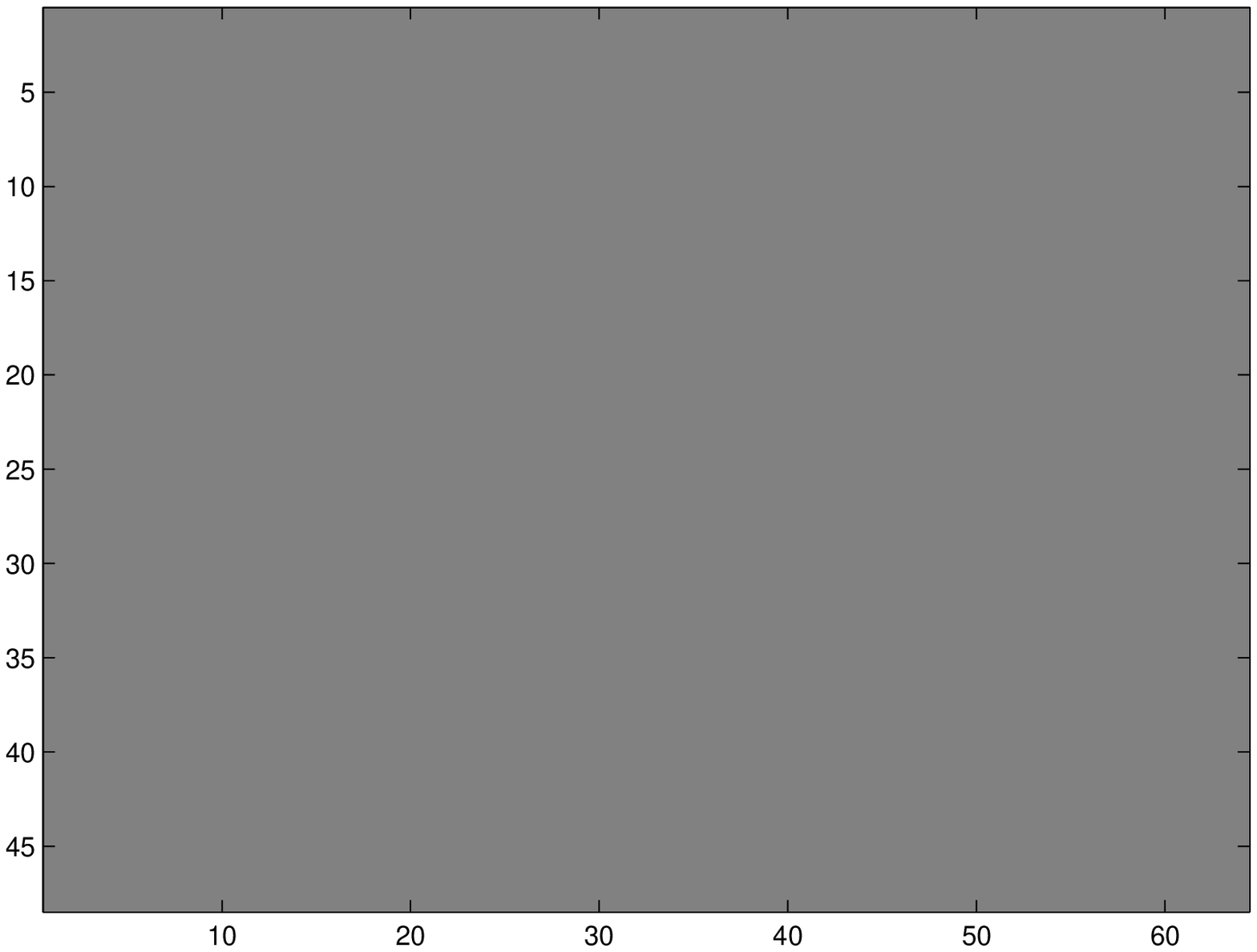}
          \includegraphics[width=1.25cm]{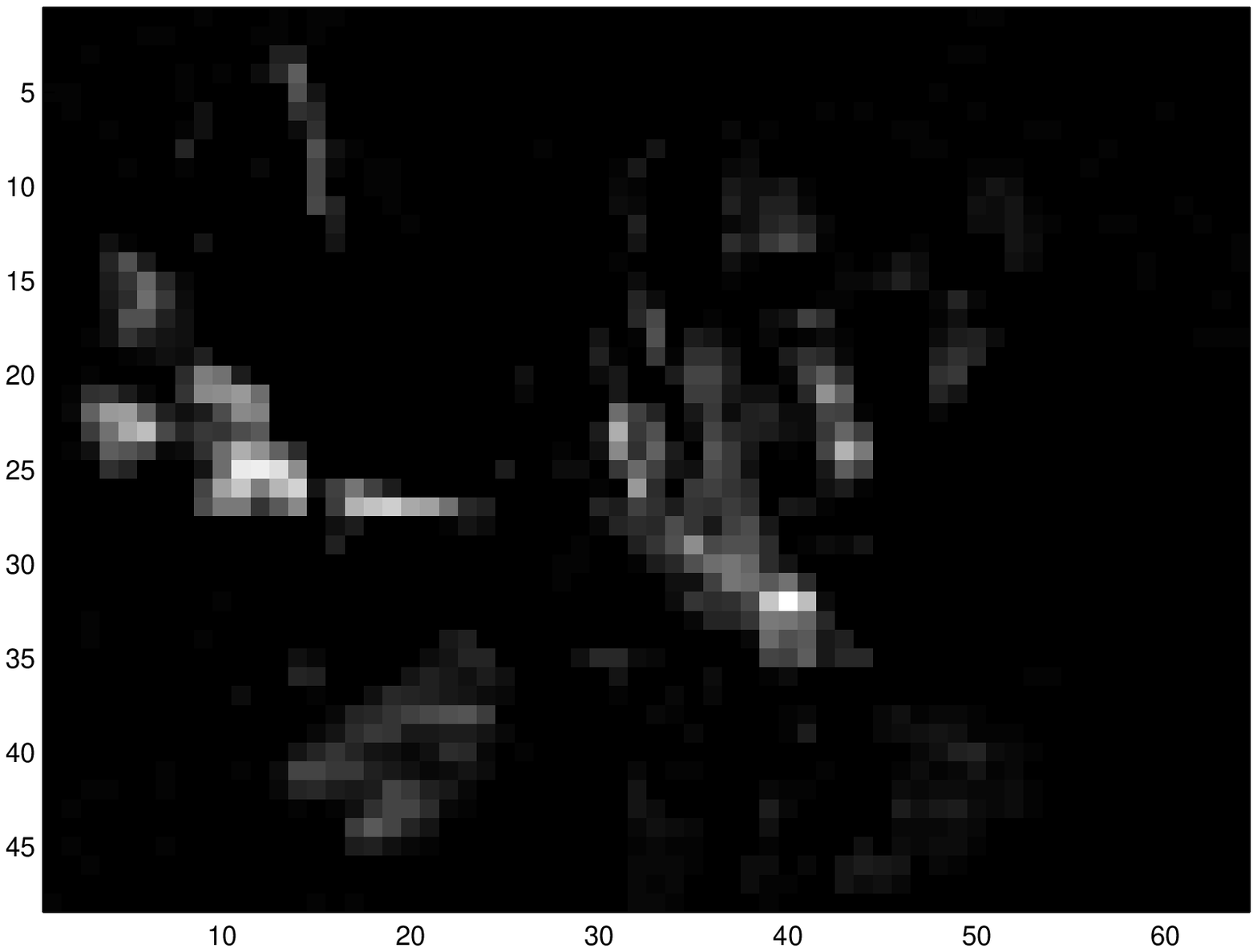}
            \includegraphics[width=1.25cm]{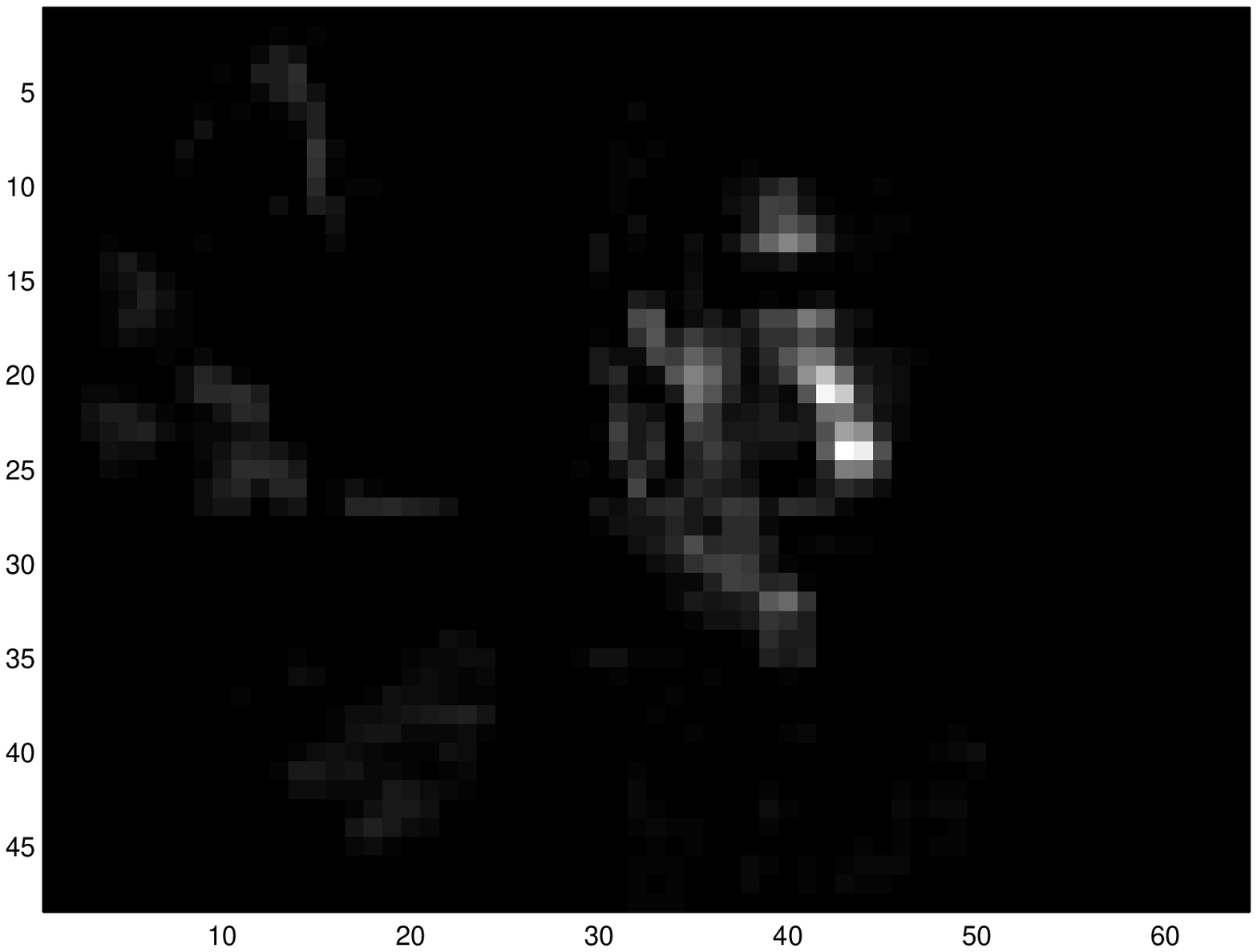}
              \includegraphics[width=1.25cm]{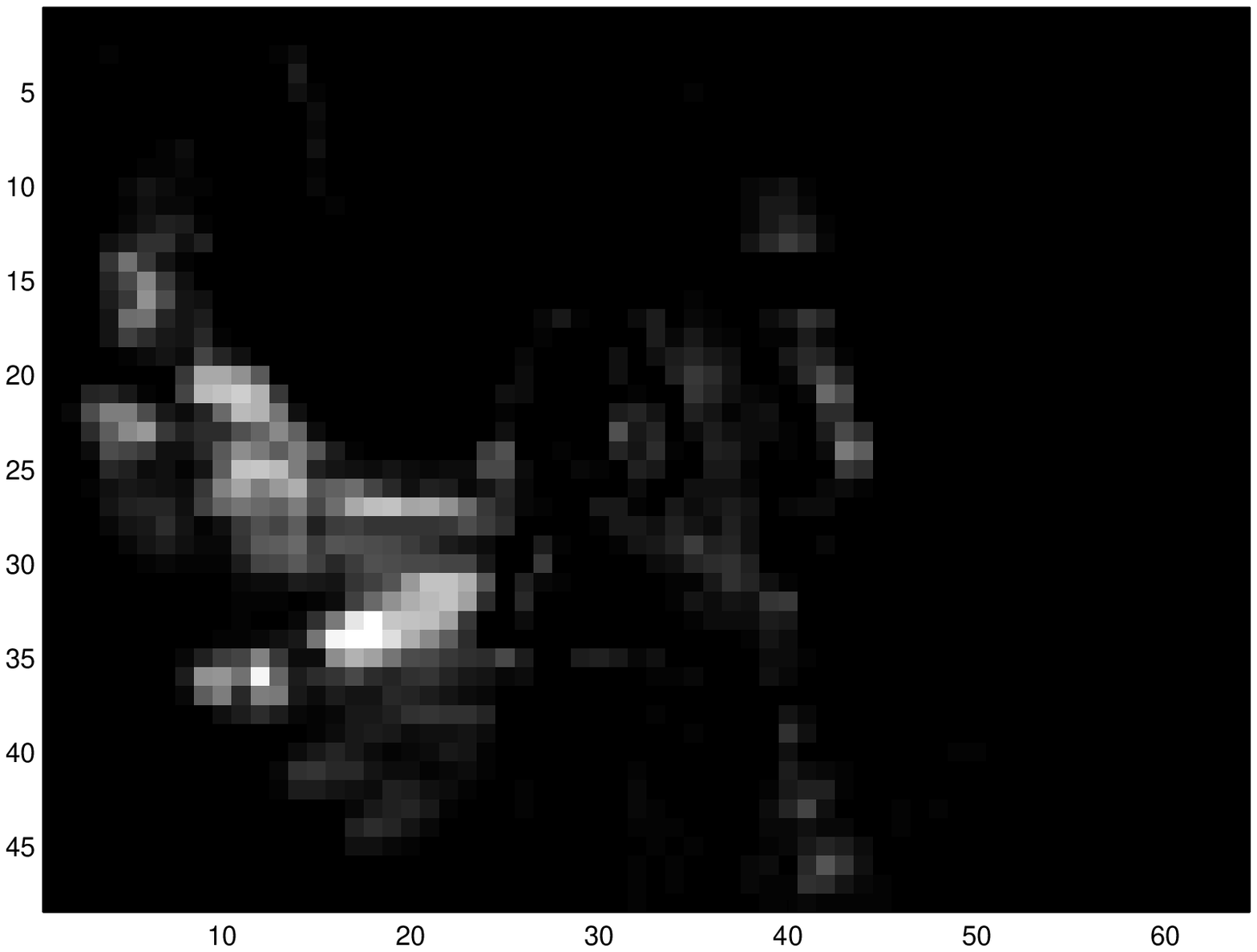}
                \includegraphics[width=1.25cm]{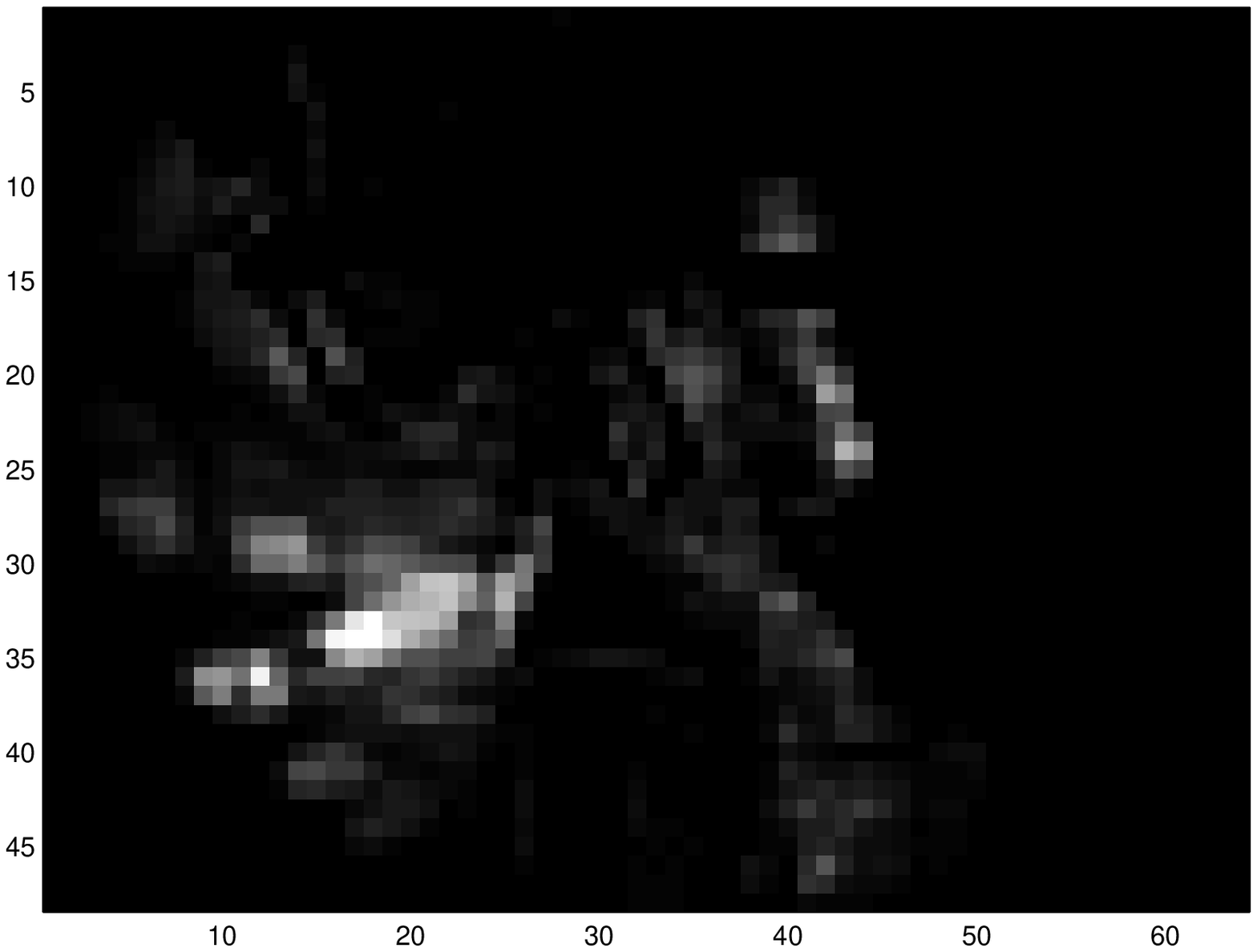}
                \includegraphics[width=1.25cm]{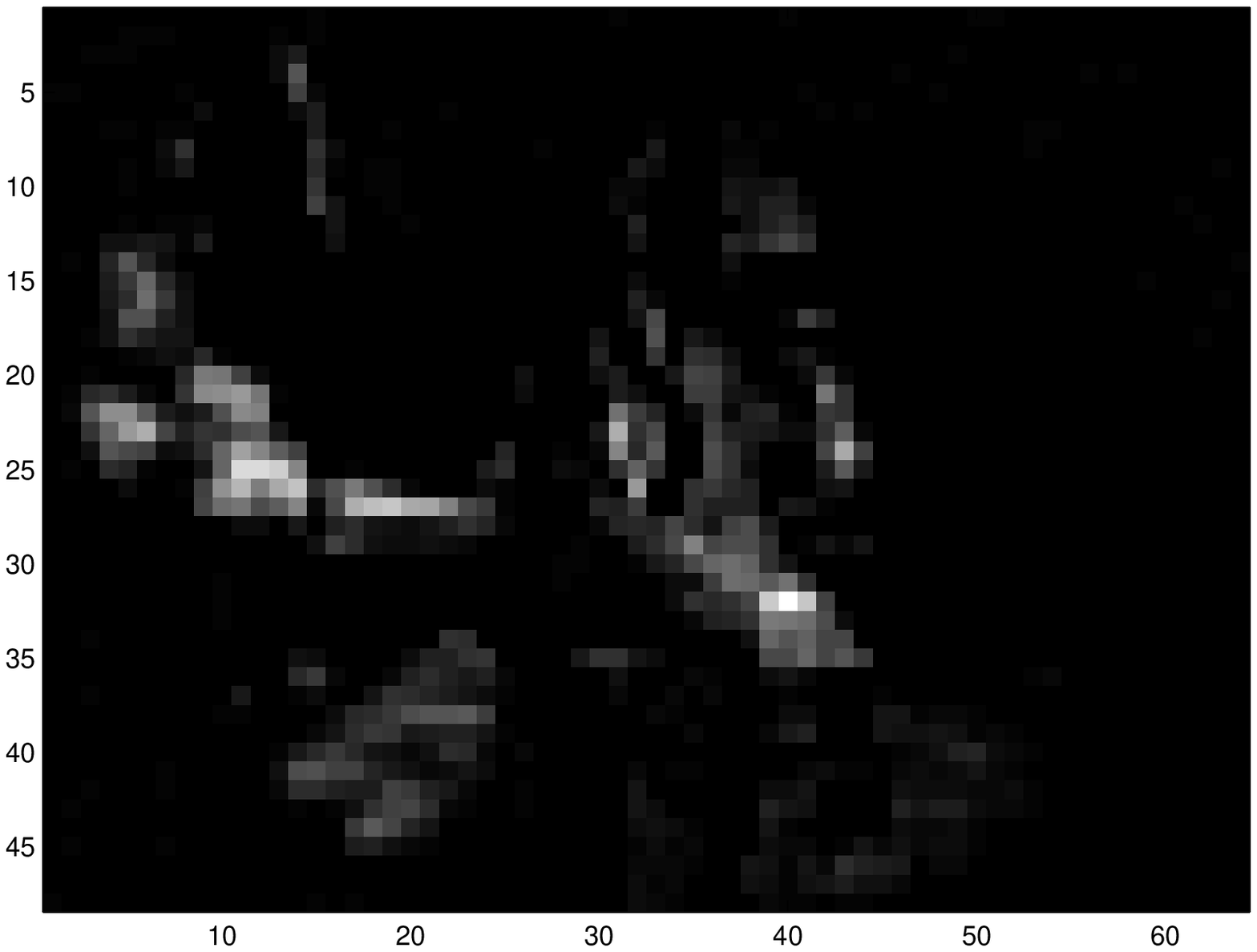}
                \includegraphics[width=1.25cm]{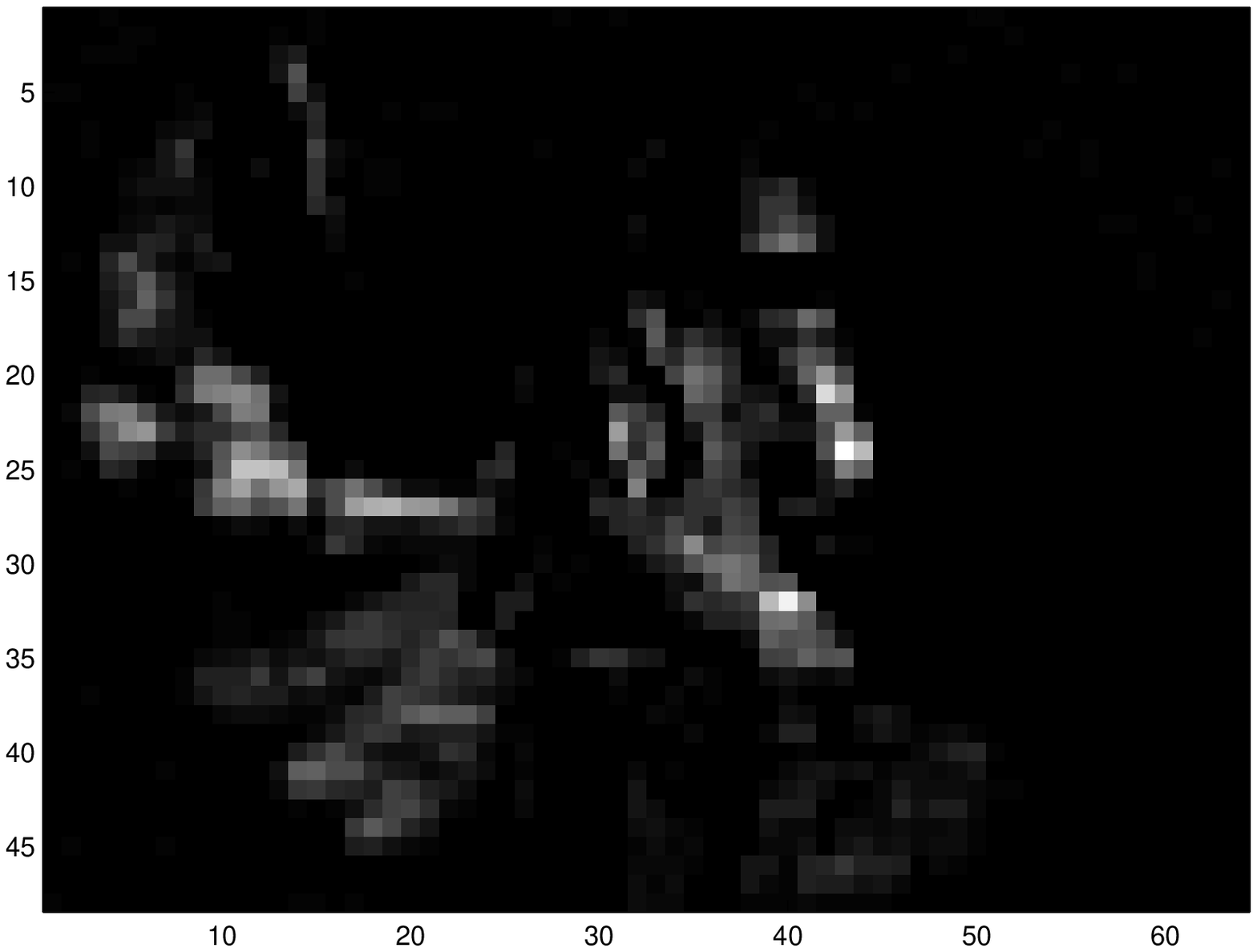}\\
  \includegraphics[width=1.25cm]{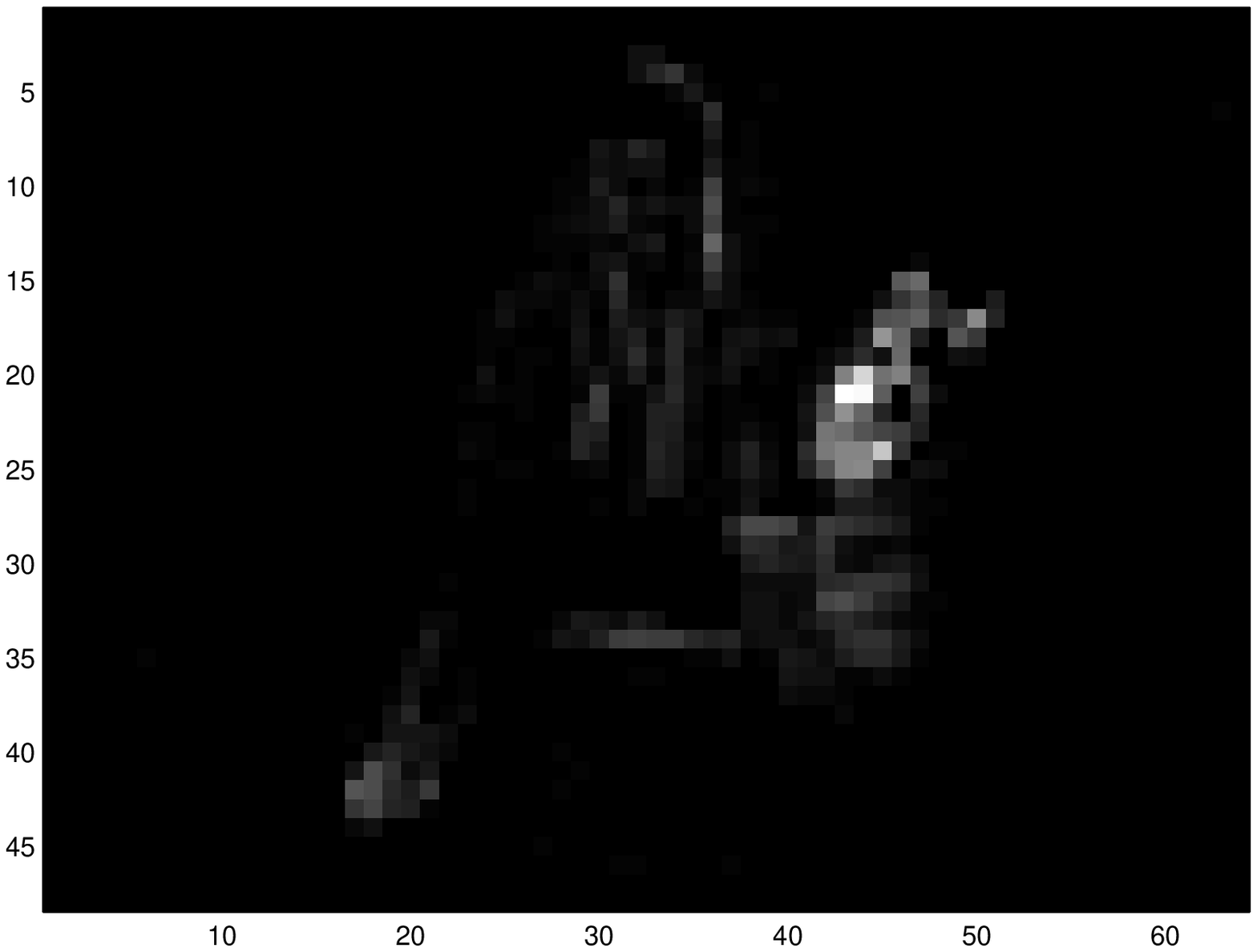}
    \includegraphics[width=1.25cm]{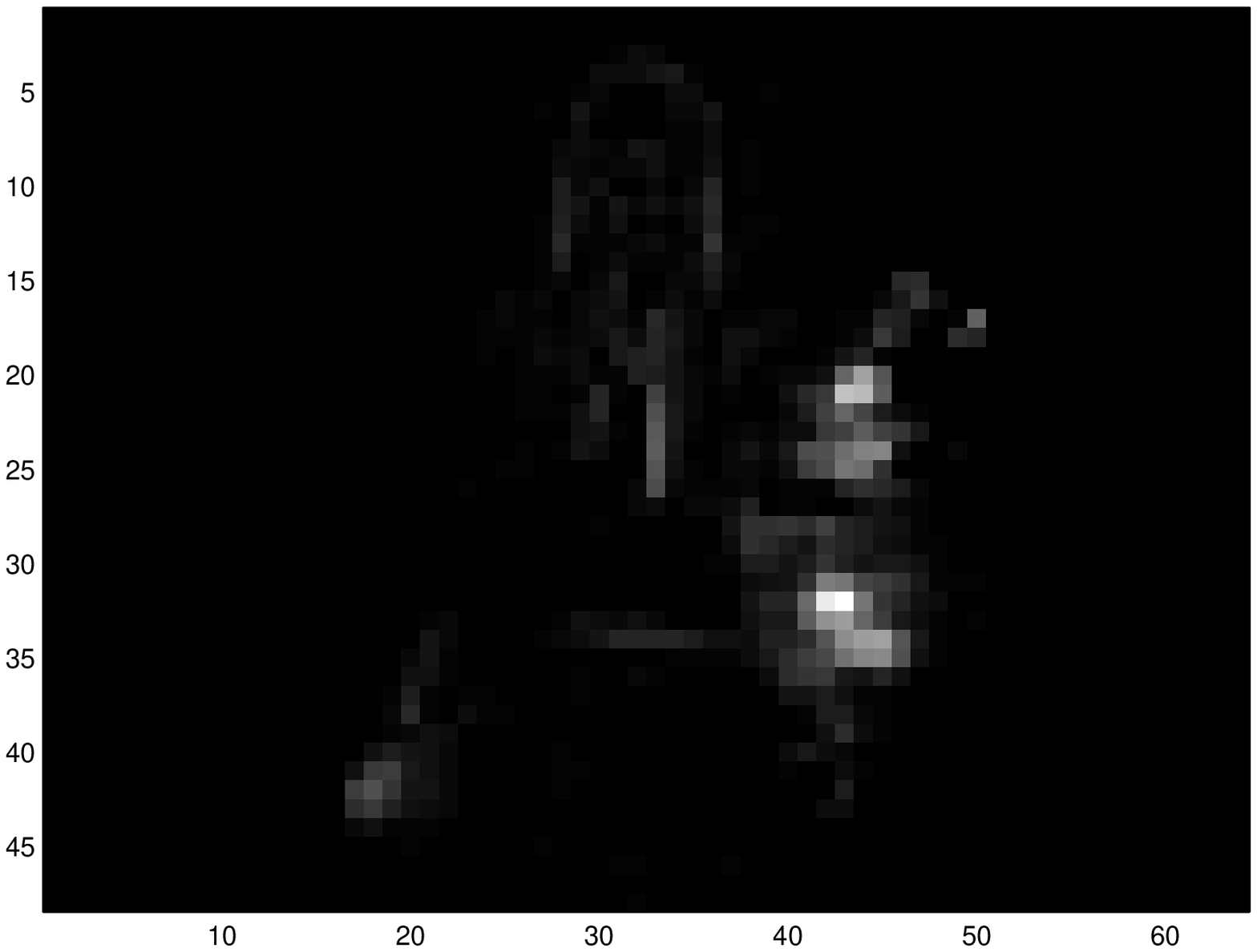}
      \includegraphics[width=1.25cm]{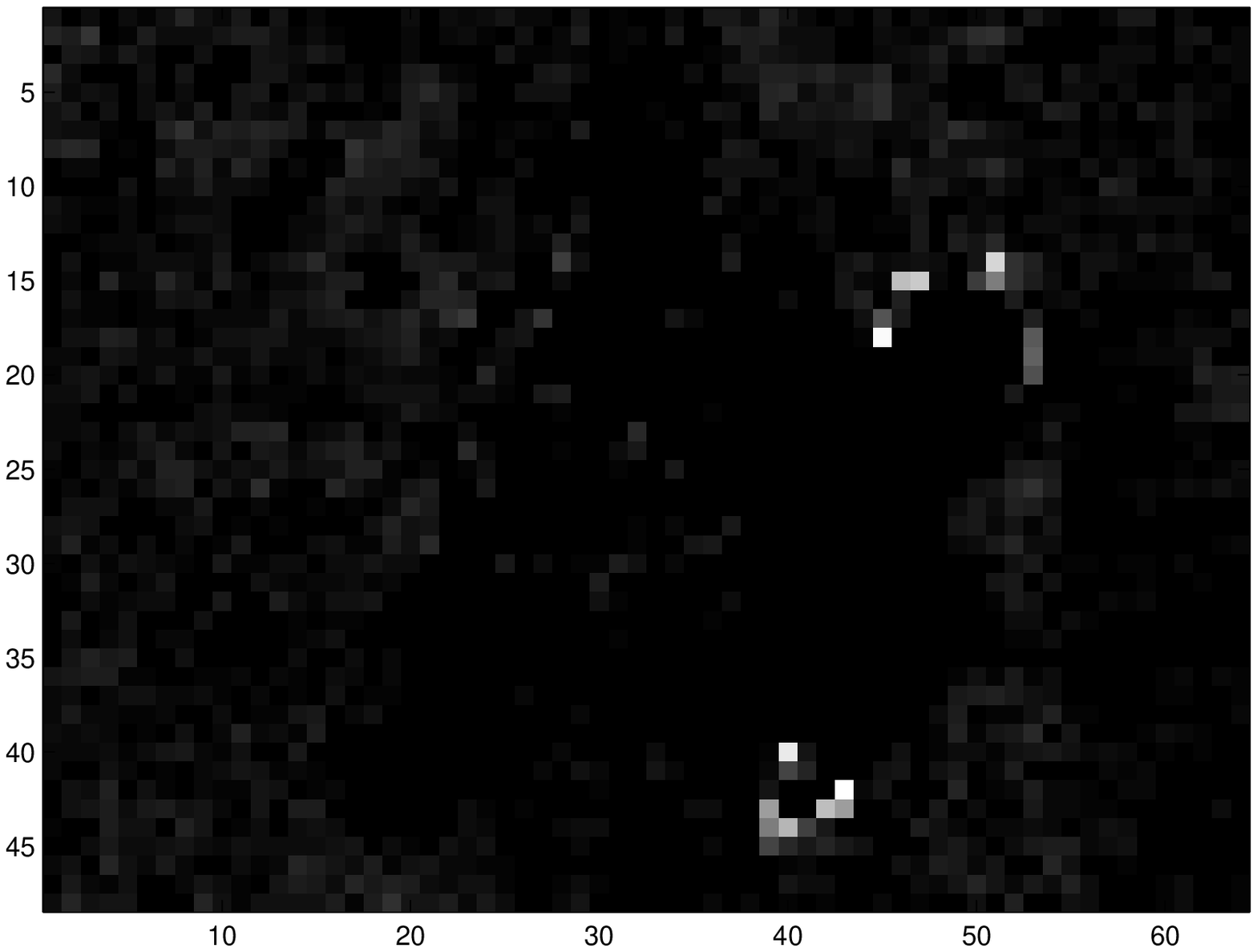}
        \includegraphics[width=1.25cm]{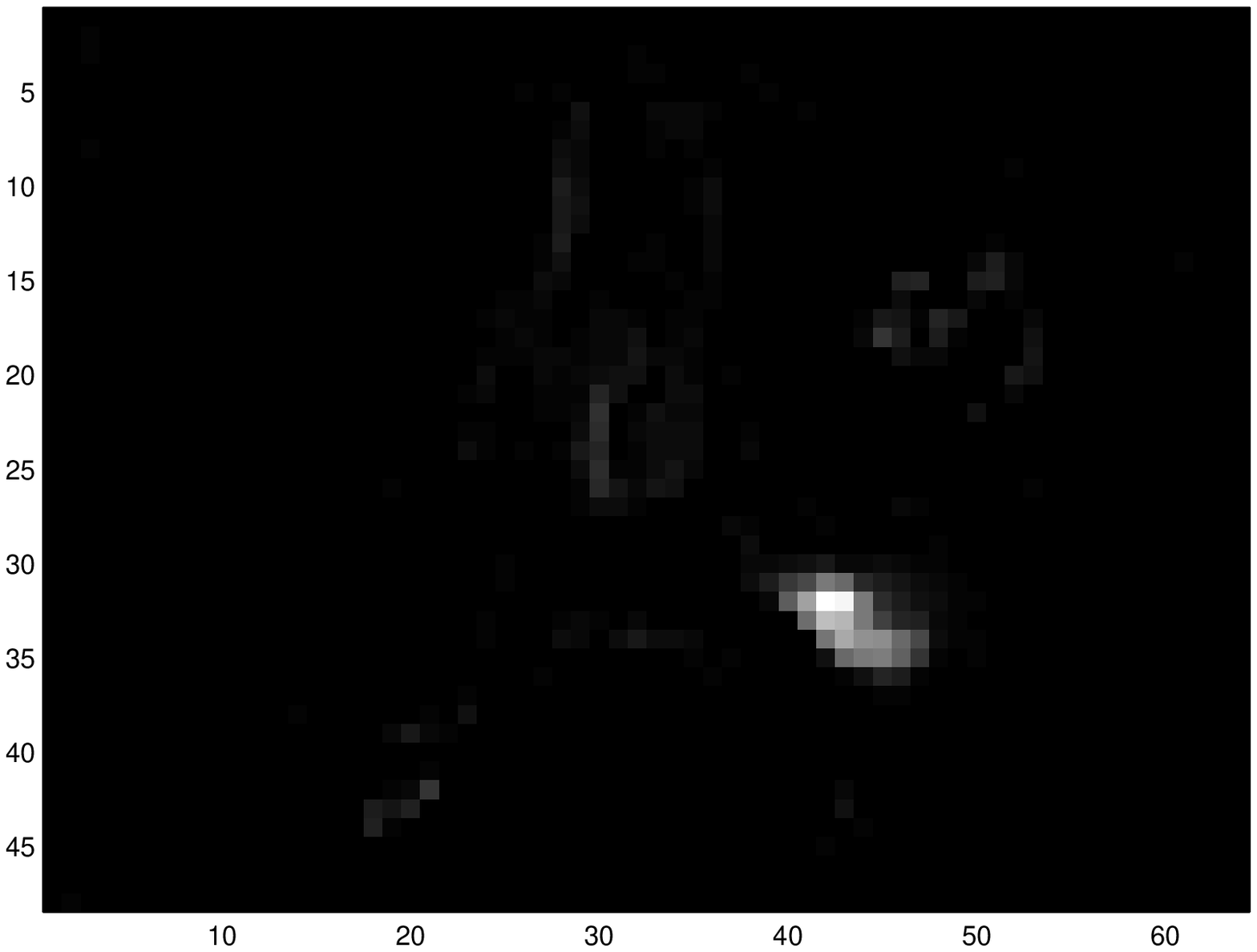}
          \includegraphics[width=1.25cm]{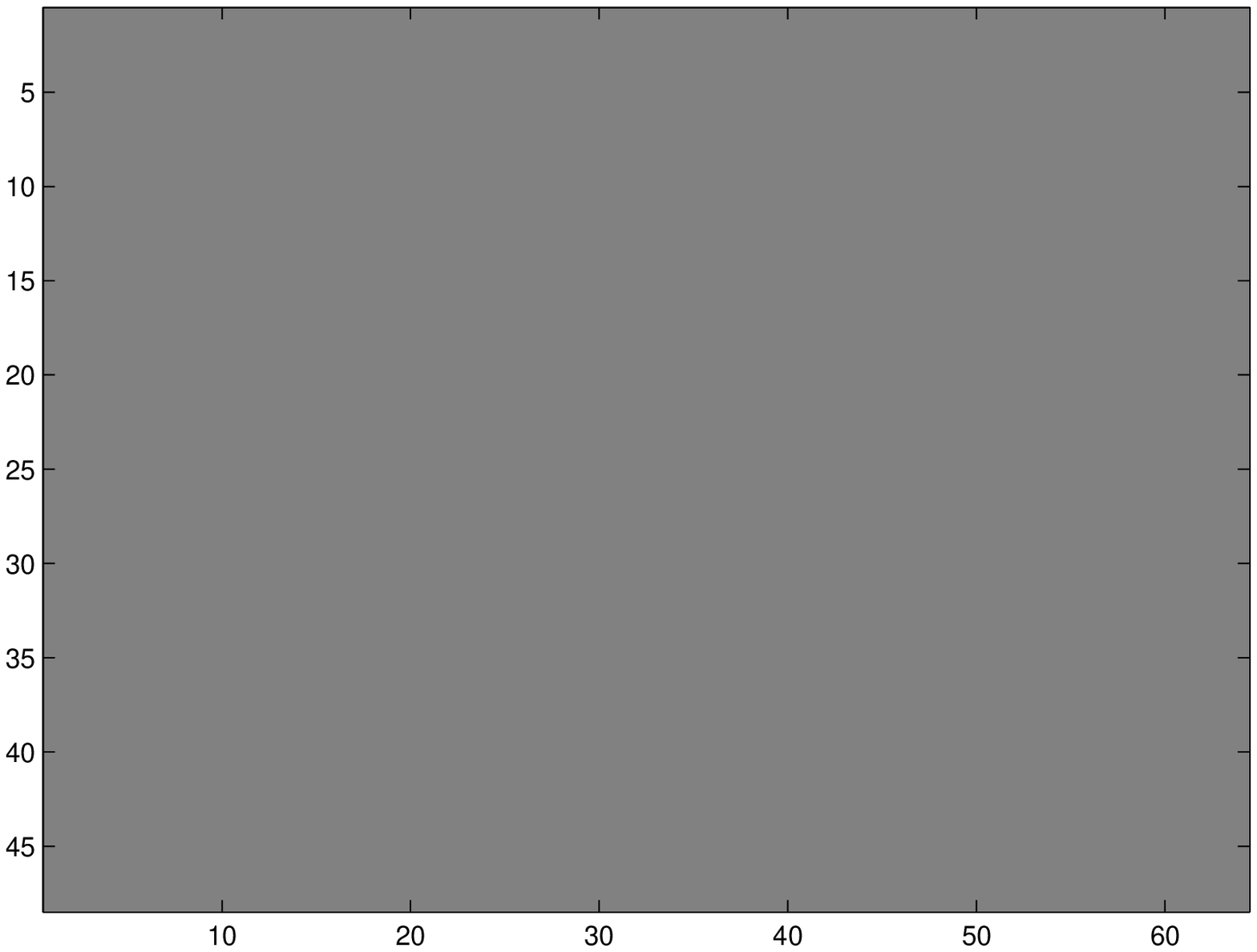}
            \includegraphics[width=1.25cm]{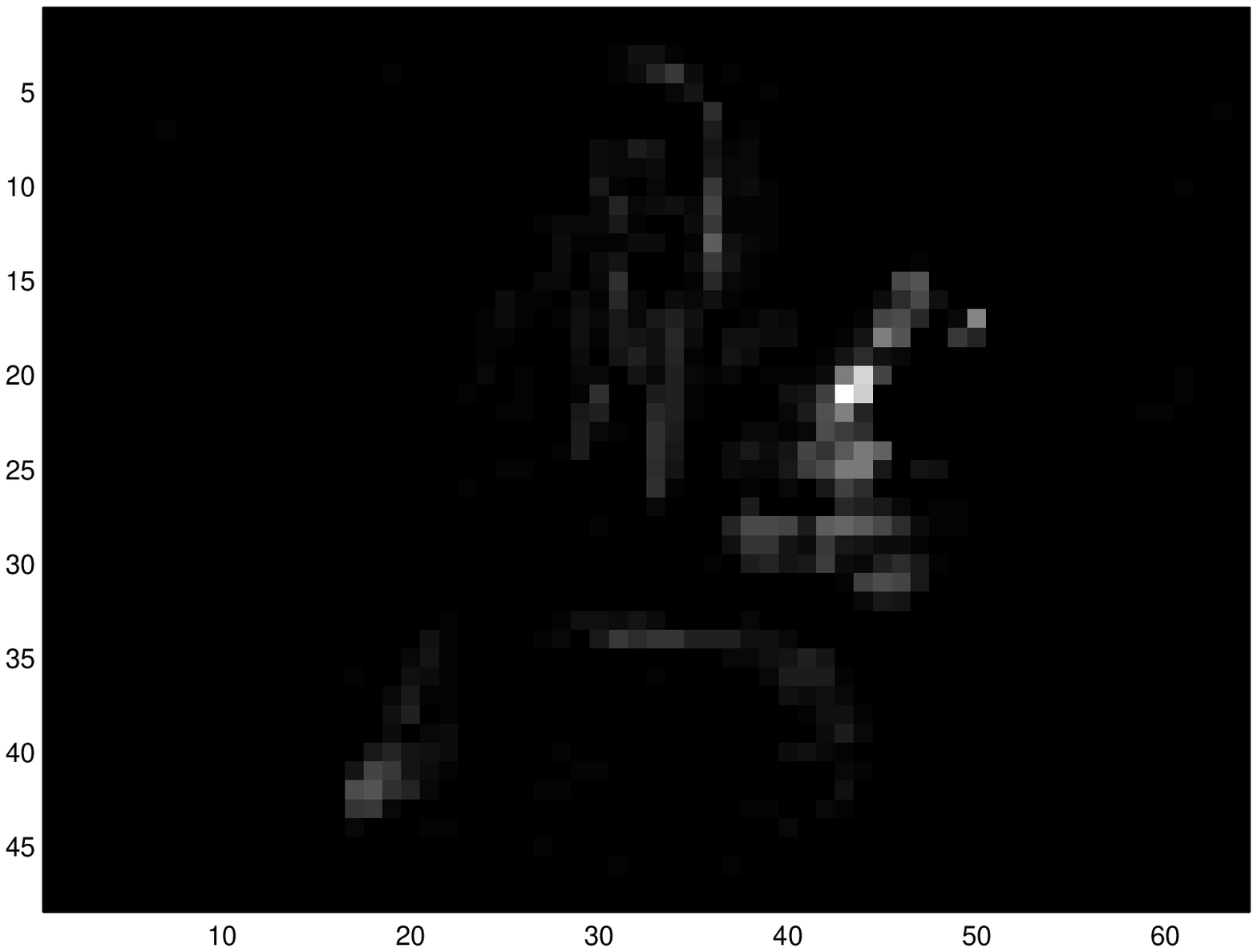}
              \includegraphics[width=1.25cm]{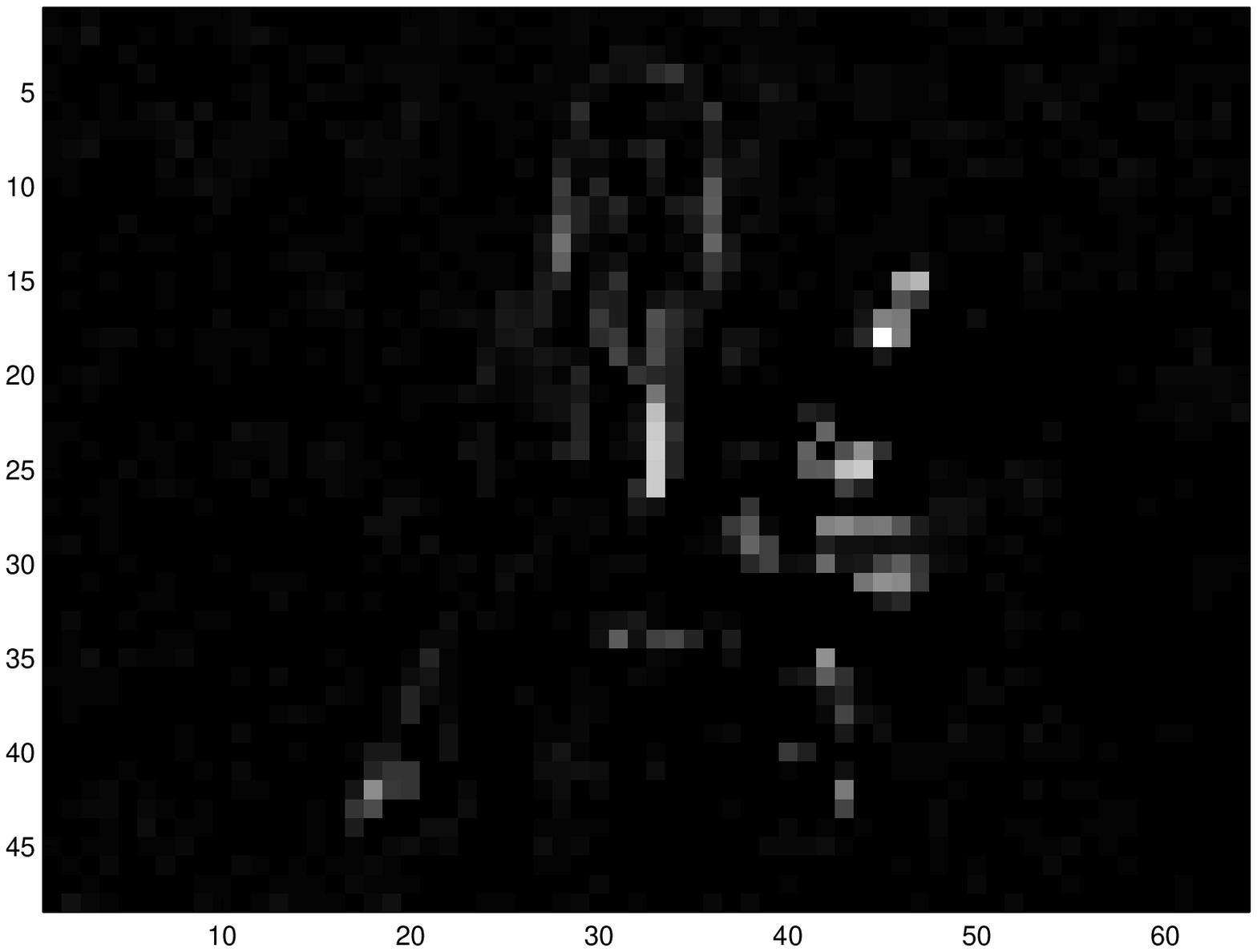}
                \includegraphics[width=1.25cm]{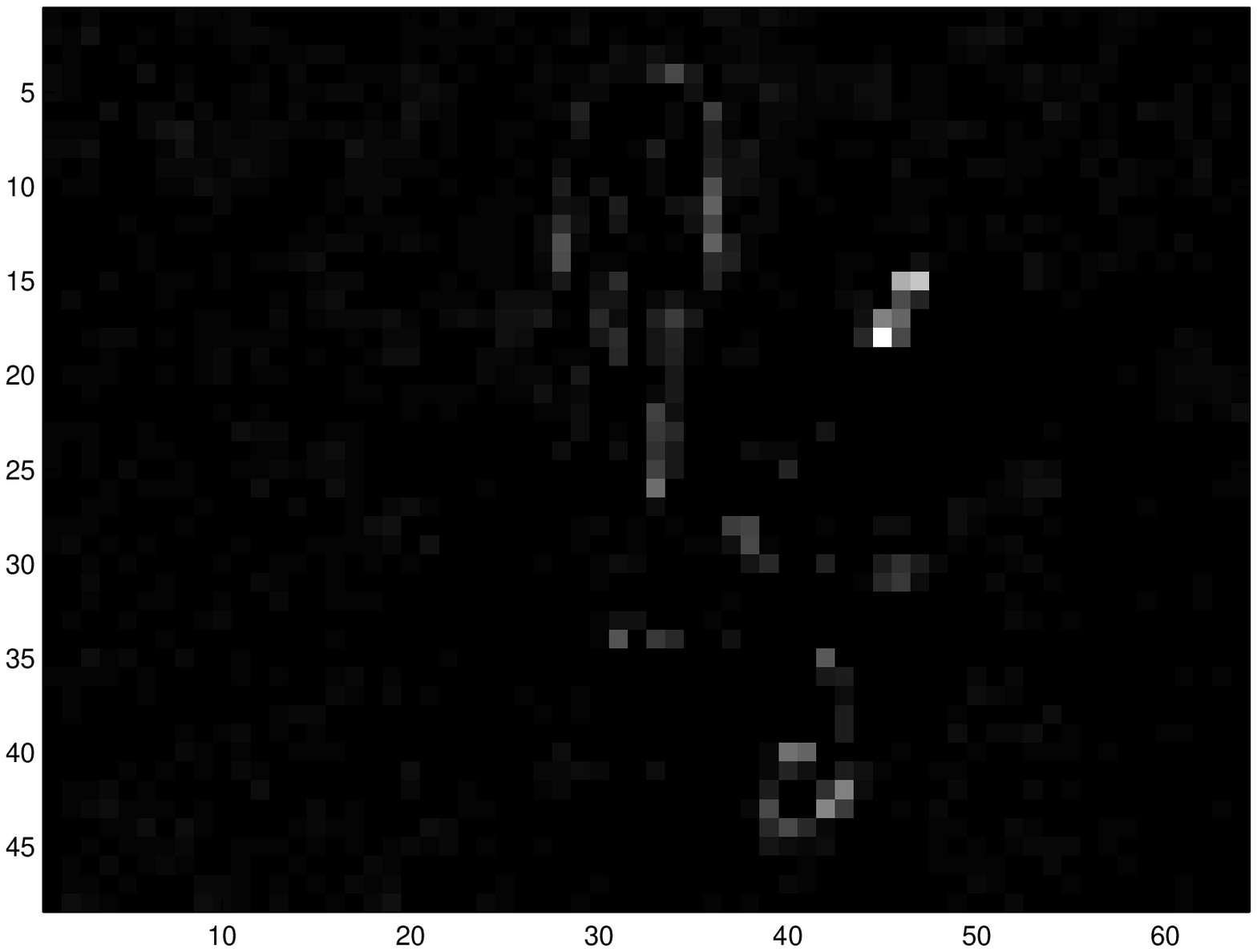}
                                \includegraphics[width=1.25cm]{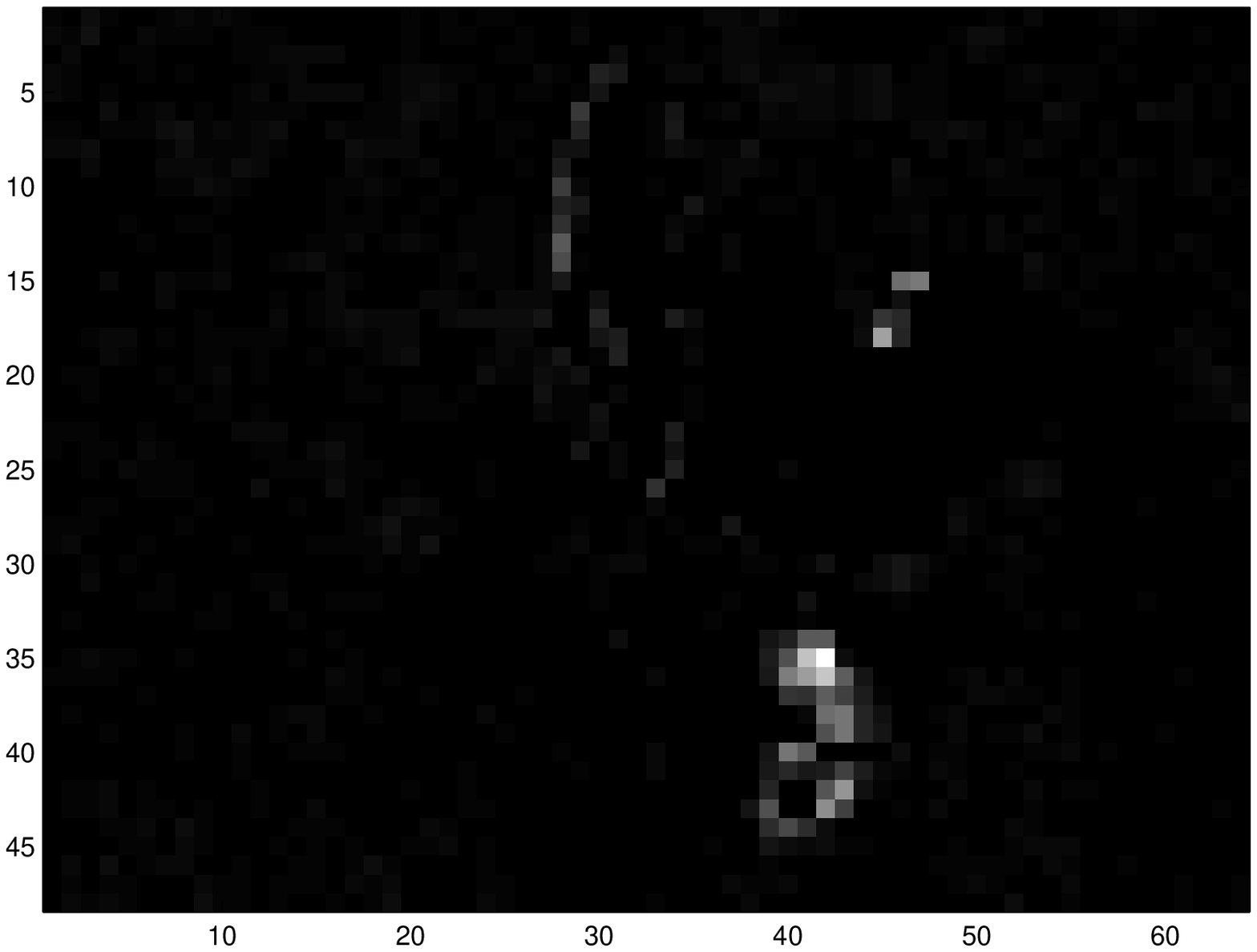}
                                                \includegraphics[width=1.25cm]{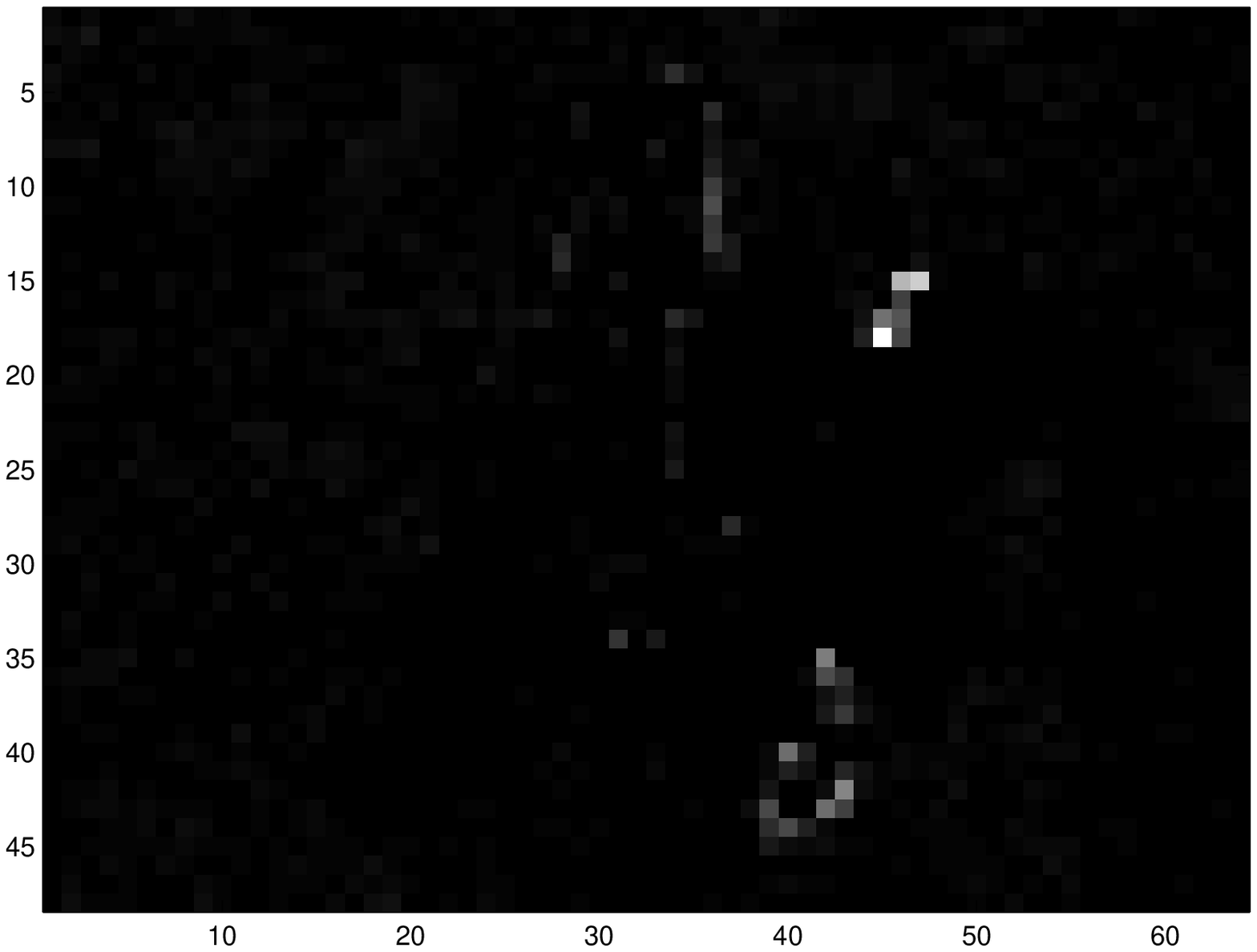}\\
                  \includegraphics[width=1.25cm]{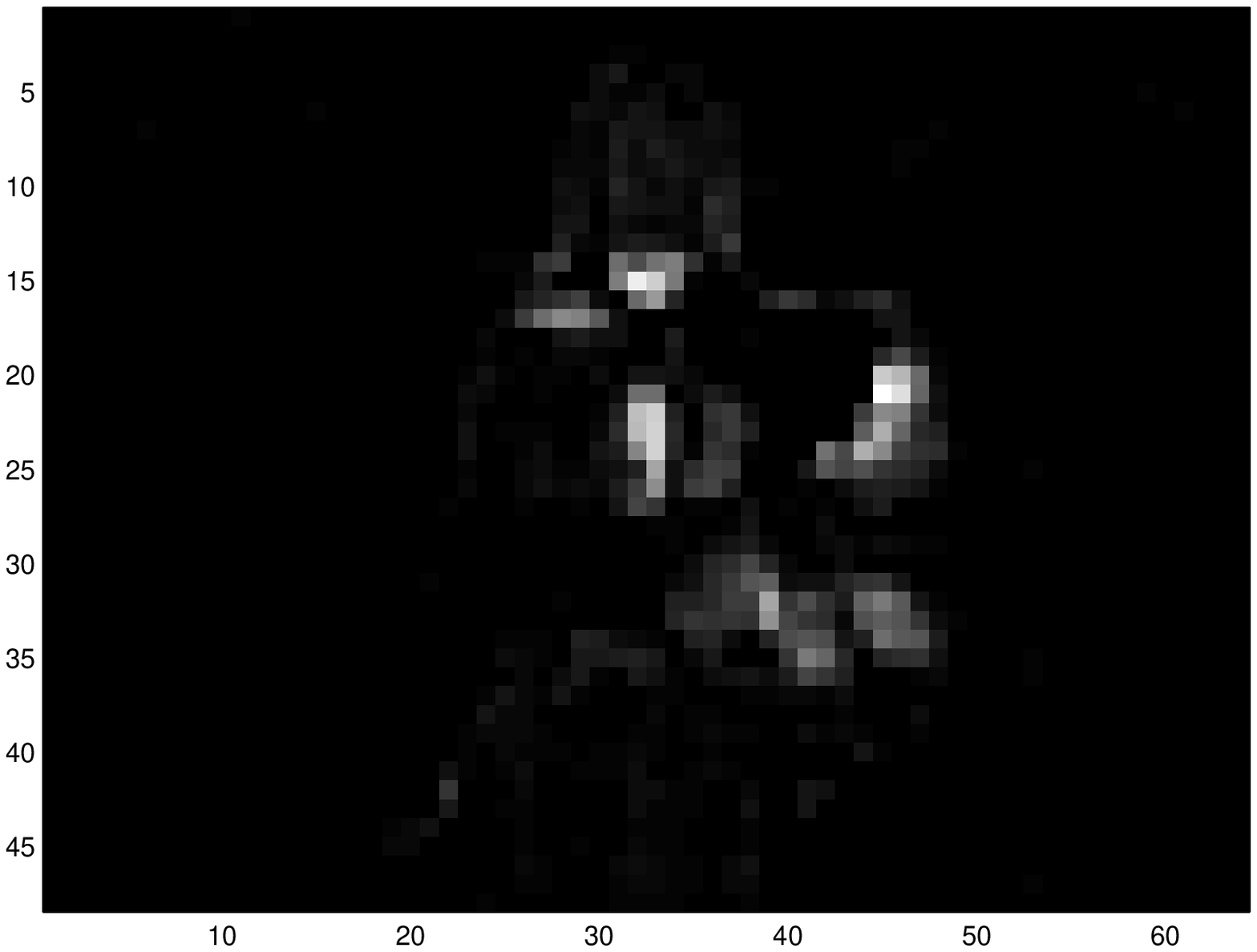}
    \includegraphics[width=1.25cm]{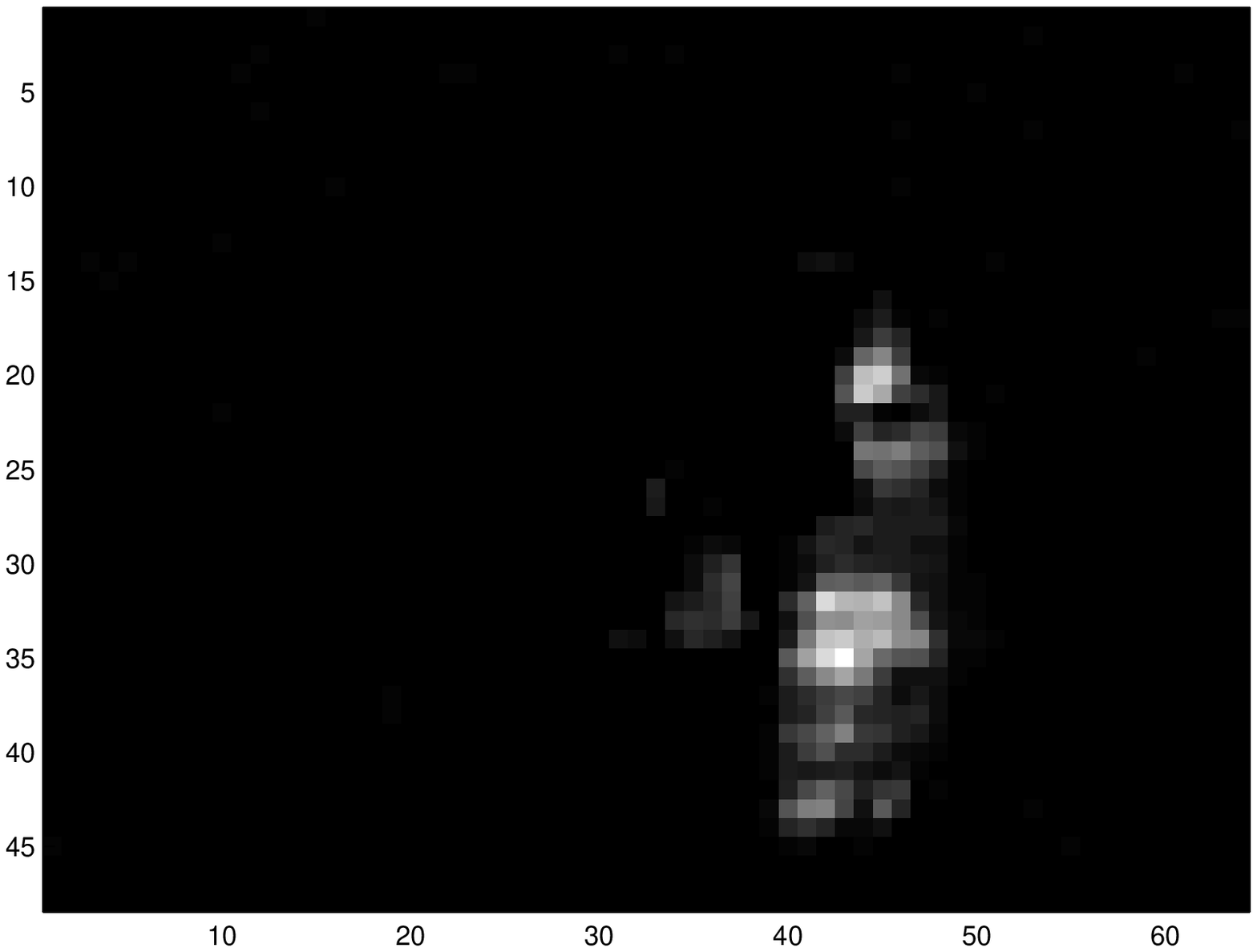}
      \includegraphics[width=1.25cm]{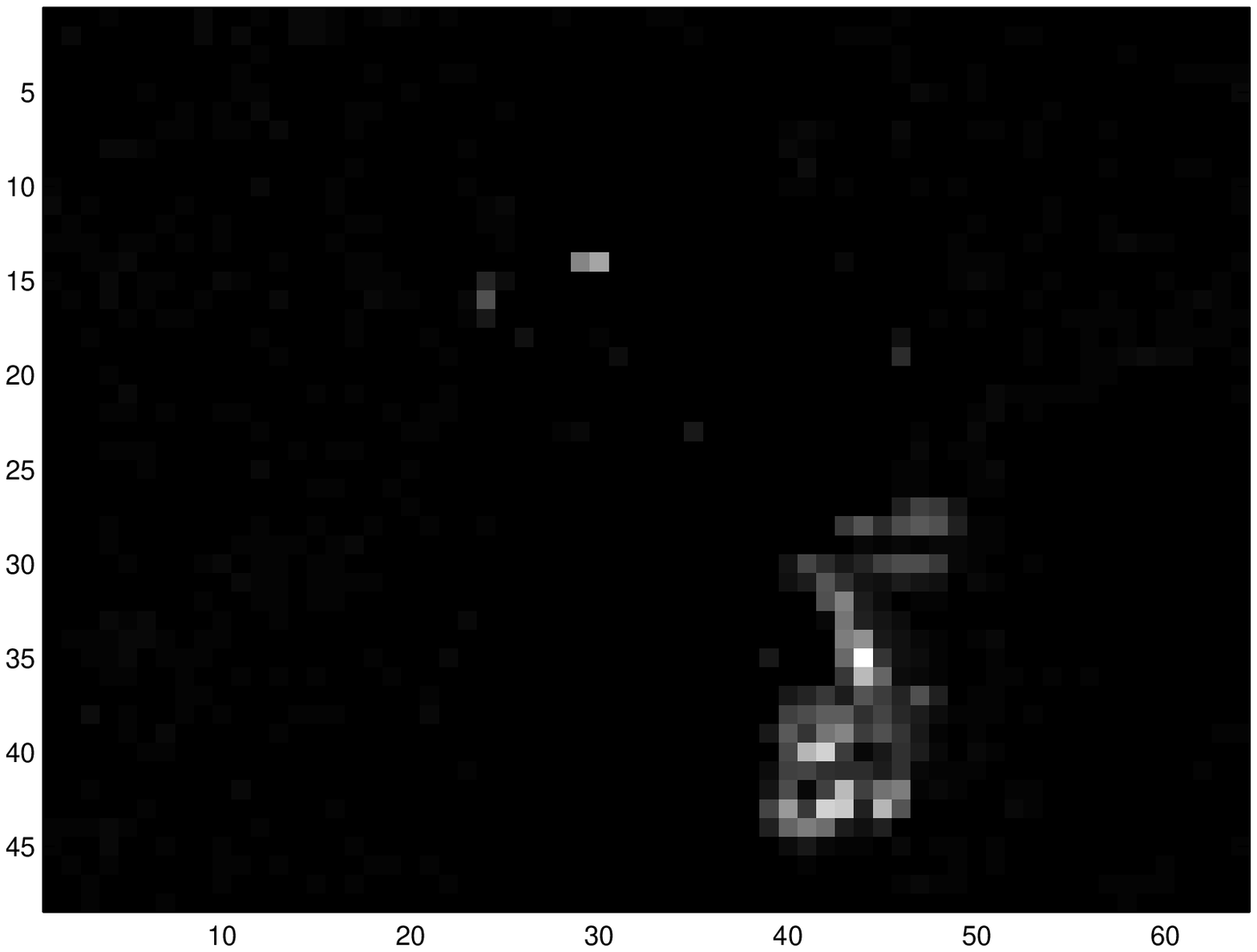}
        \includegraphics[width=1.25cm]{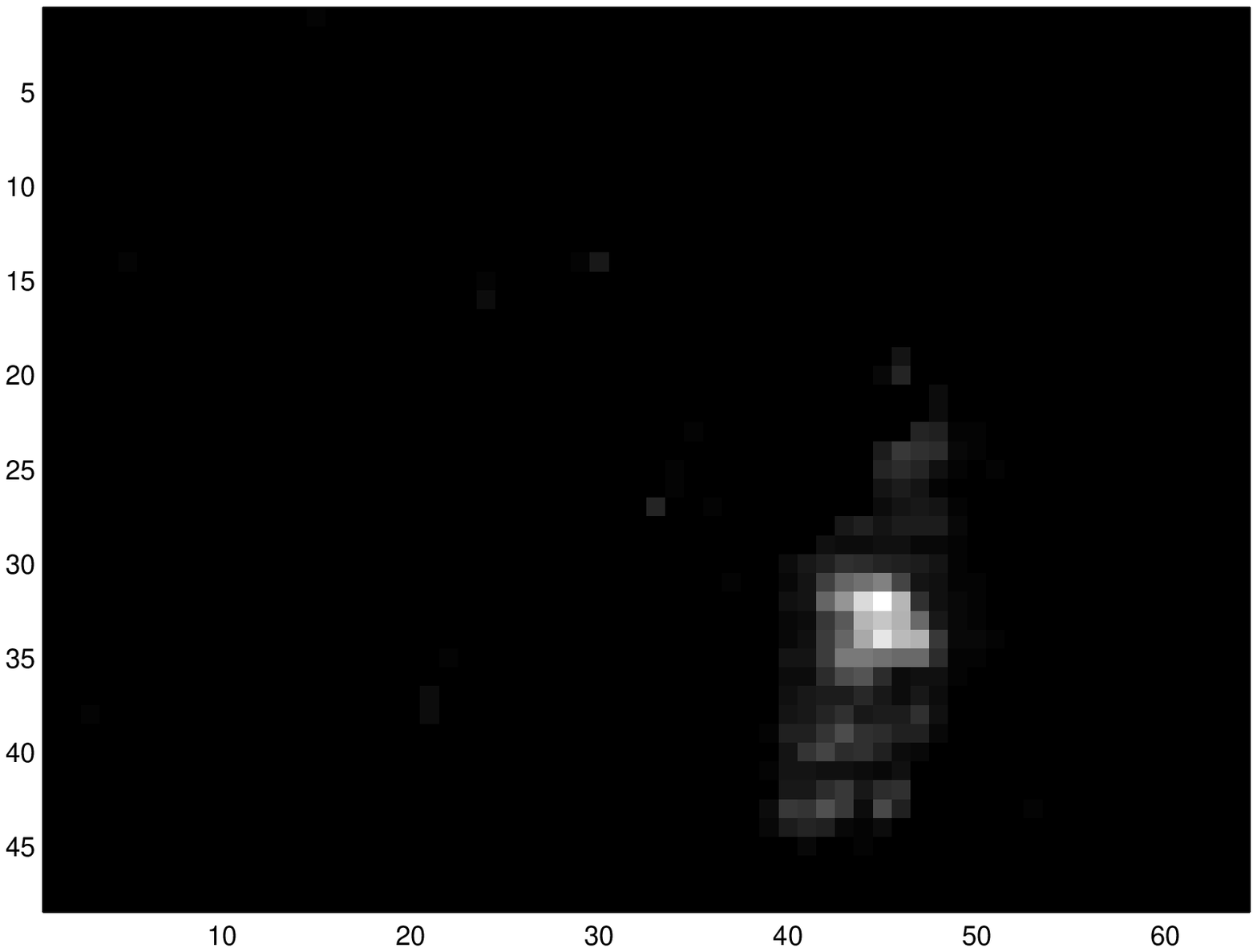}
          \includegraphics[width=1.25cm]{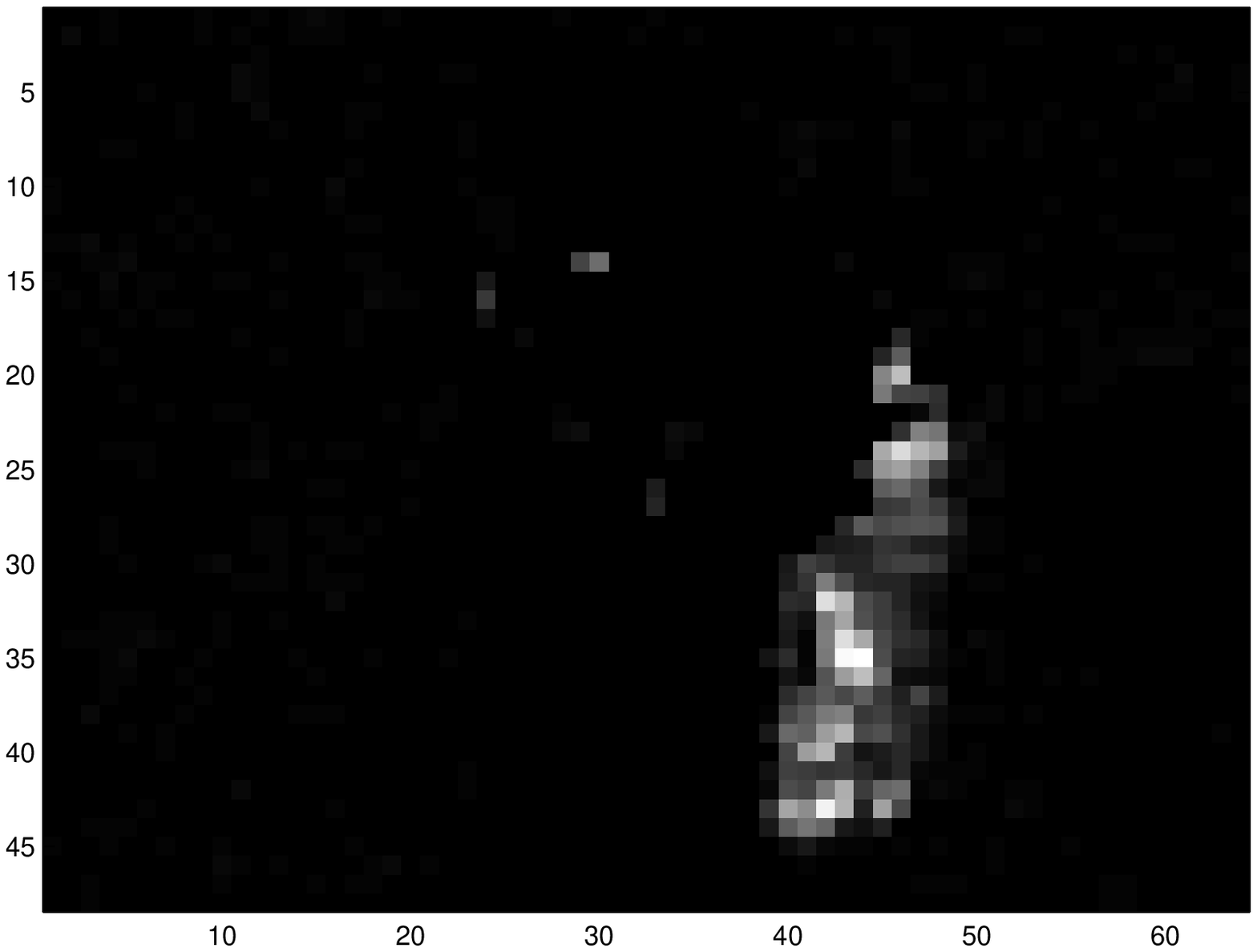}
            \includegraphics[width=1.25cm]{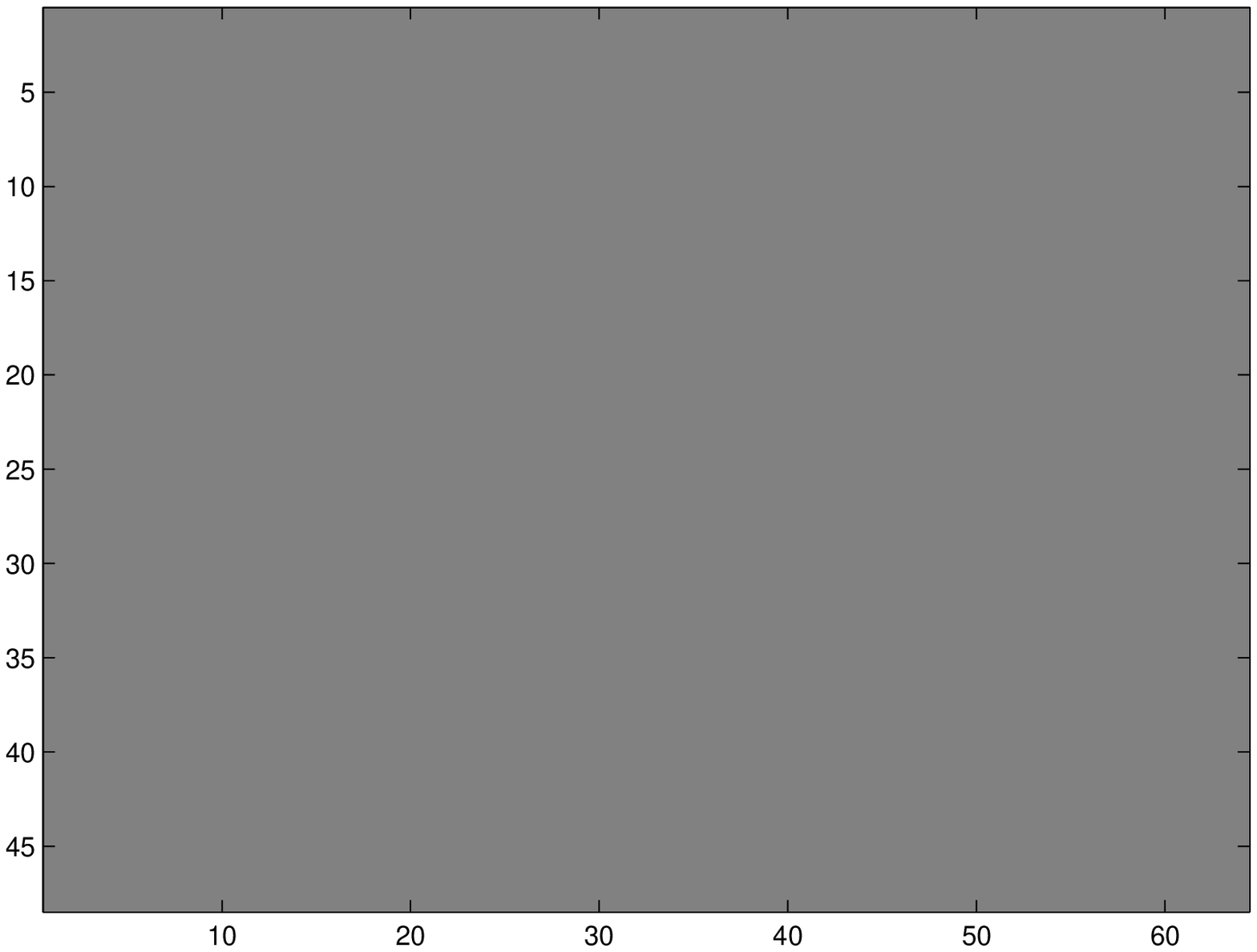}
              \includegraphics[width=1.25cm]{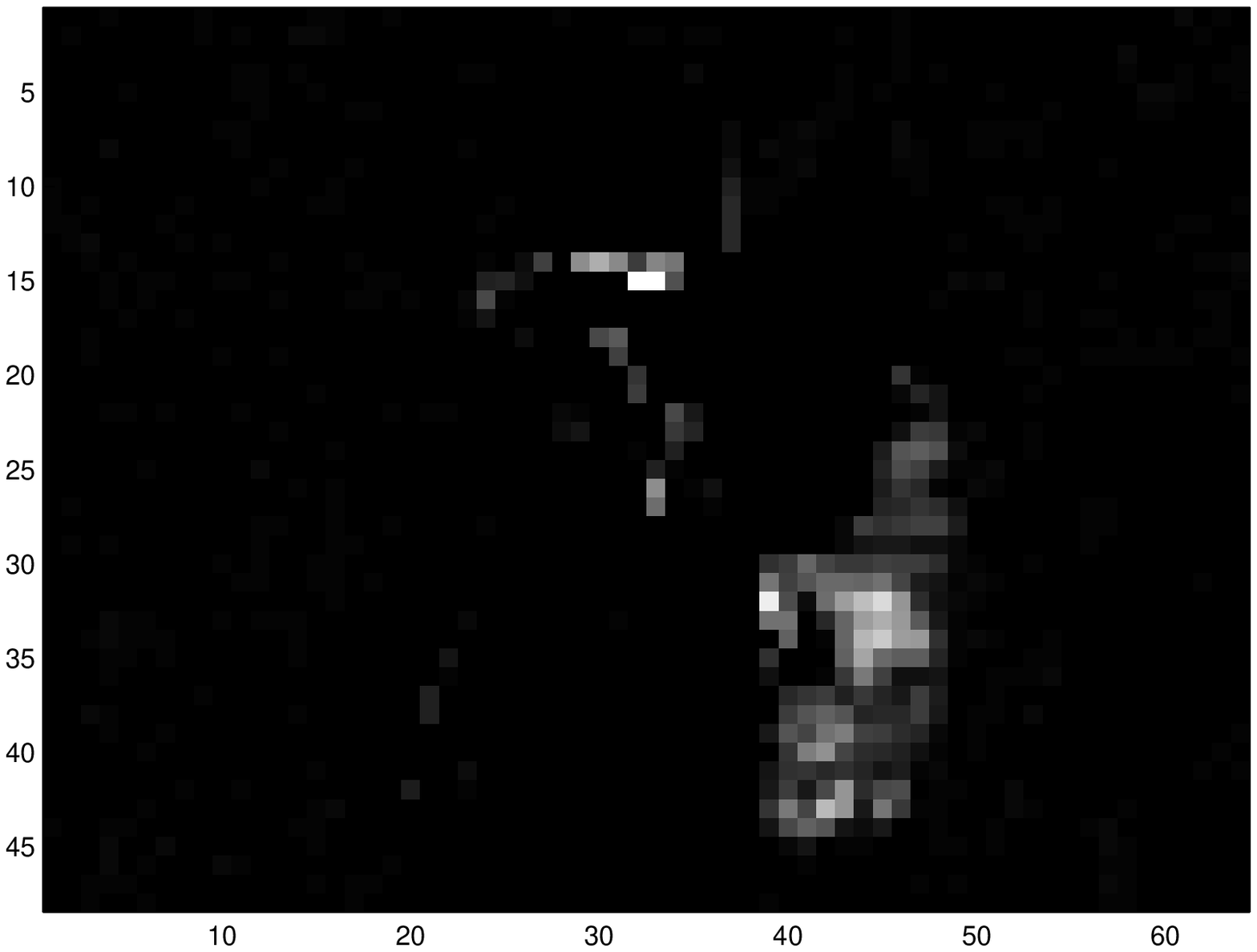}
                \includegraphics[width=1.25cm]{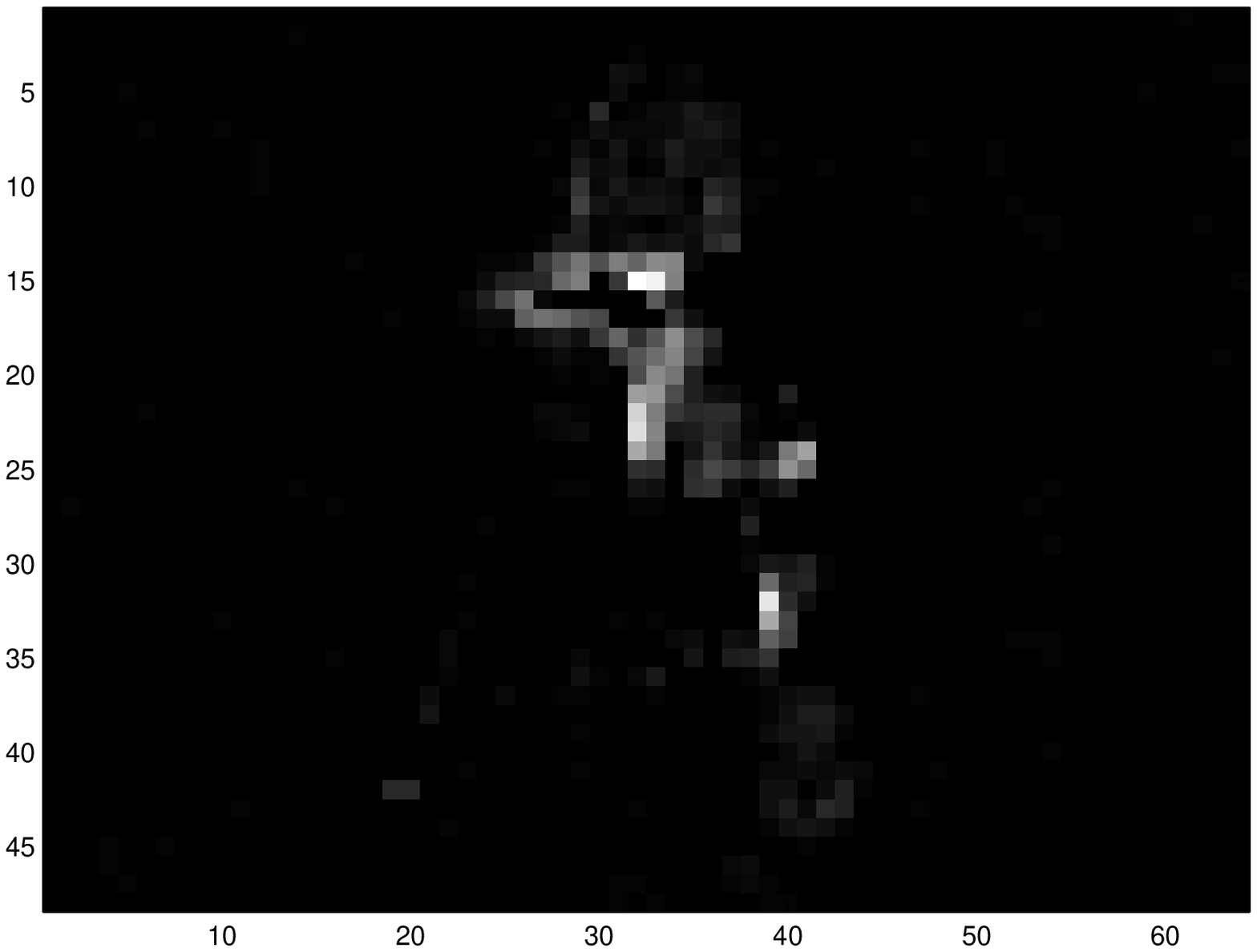}
                                \includegraphics[width=1.25cm]{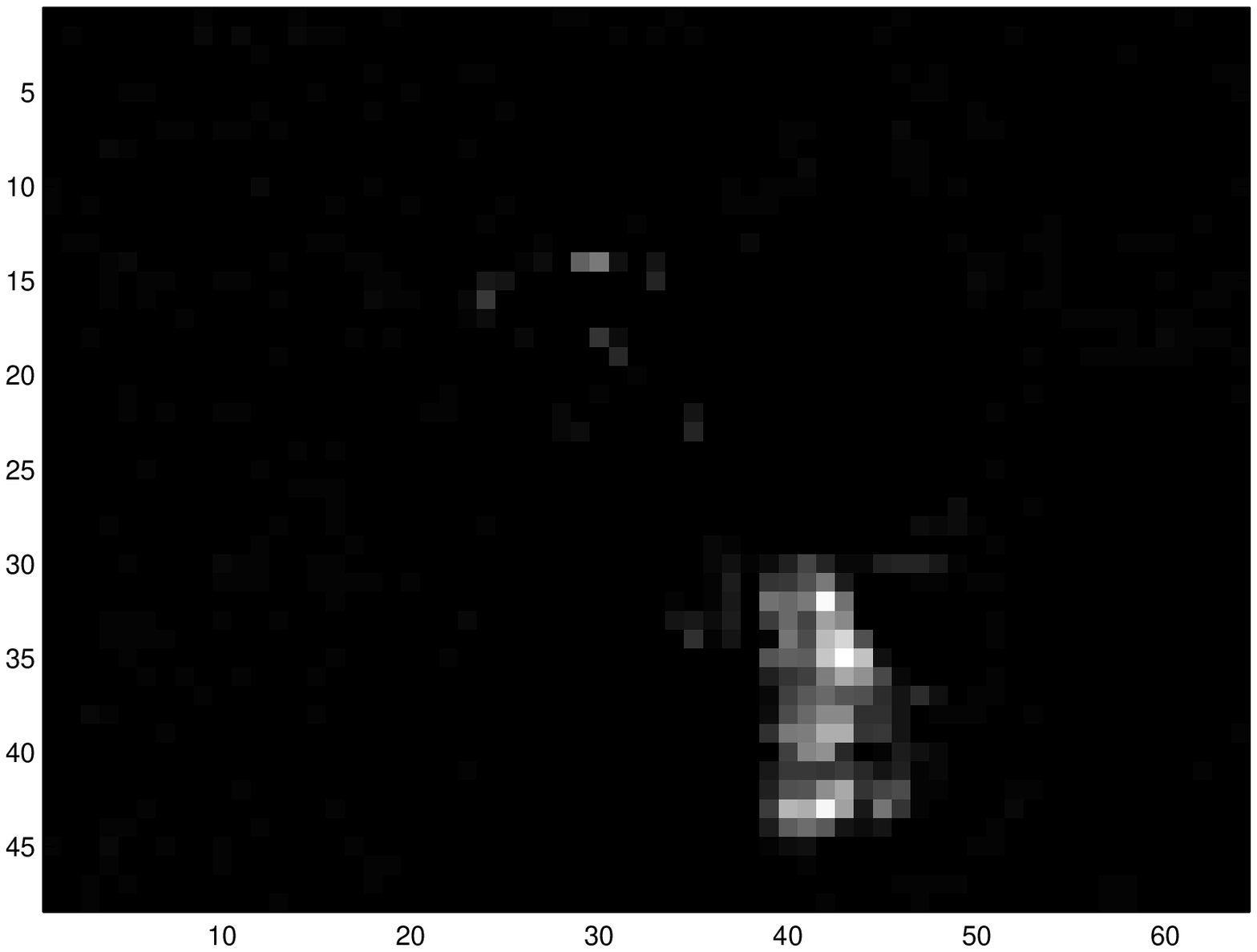}
                                                \includegraphics[width=1.25cm]{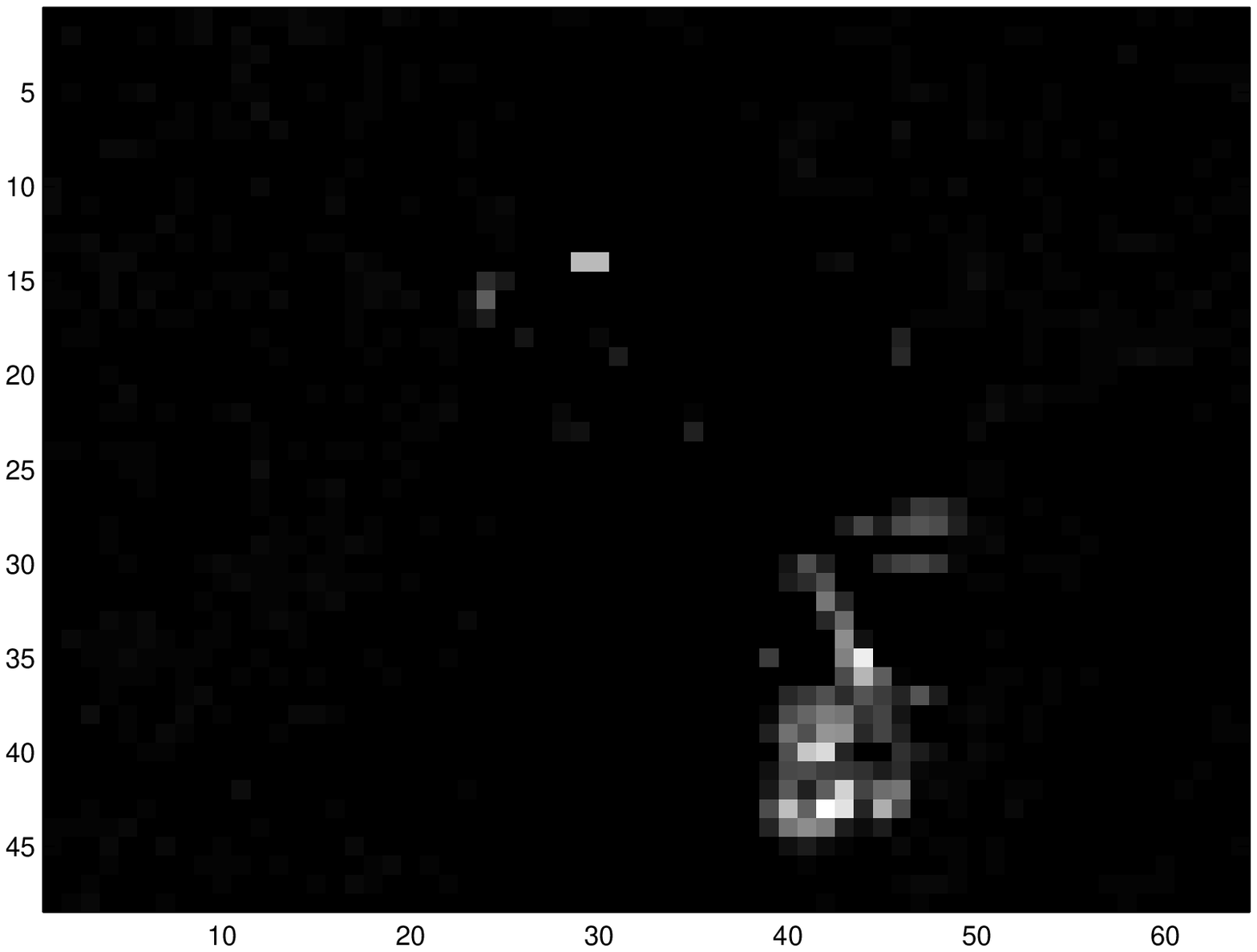}\\
  \includegraphics[width=1.25cm]{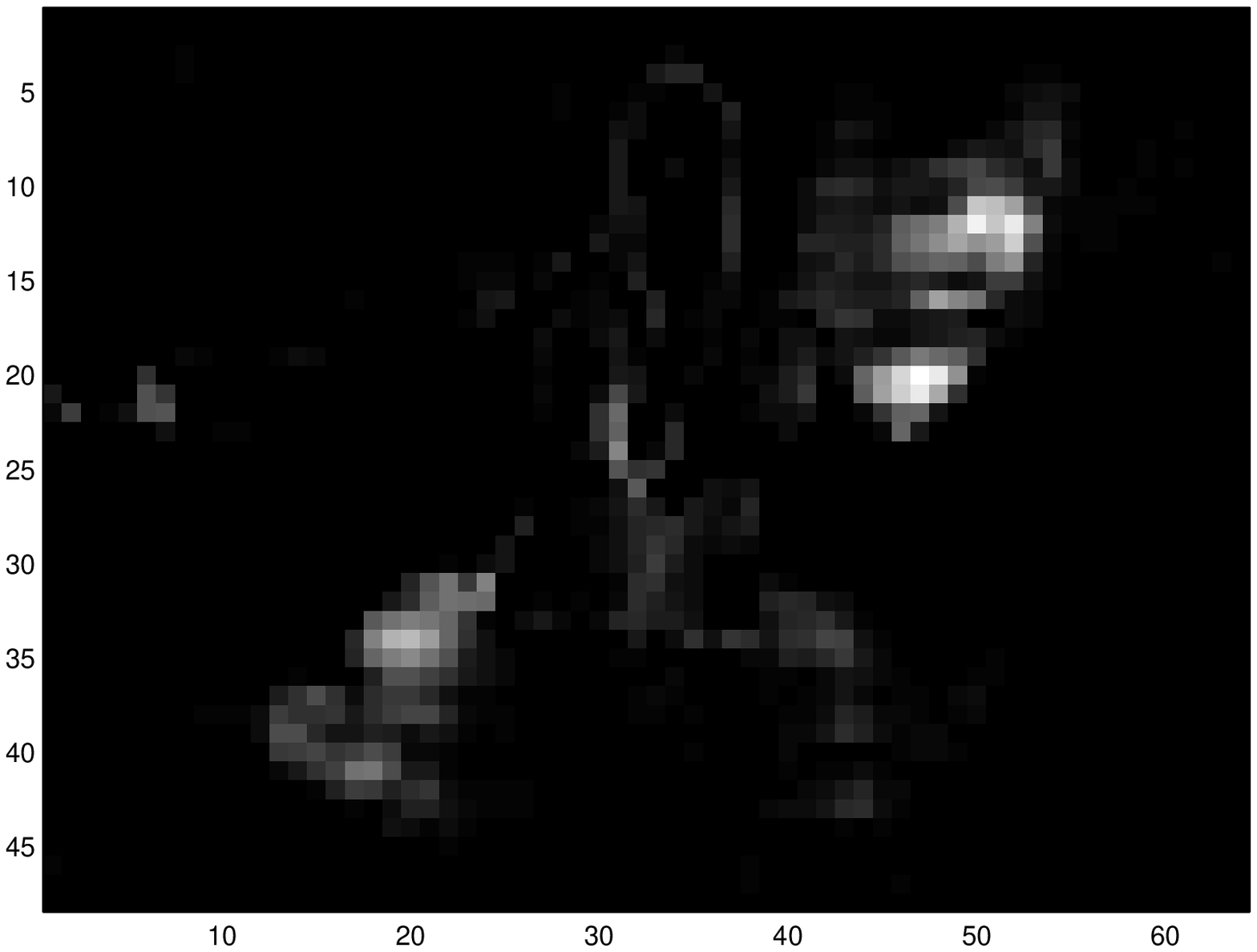}
    \includegraphics[width=1.25cm]{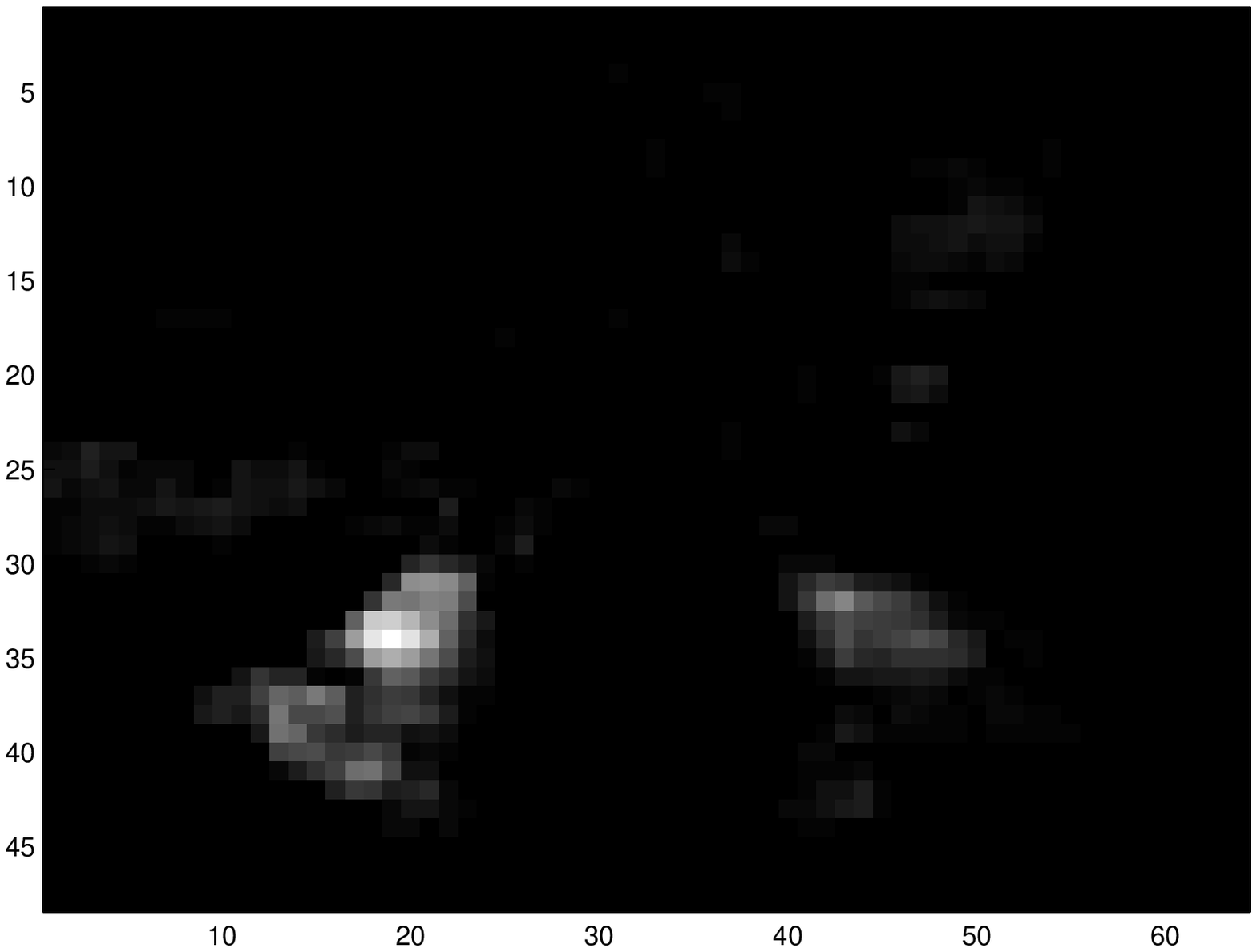}
      \includegraphics[width=1.25cm]{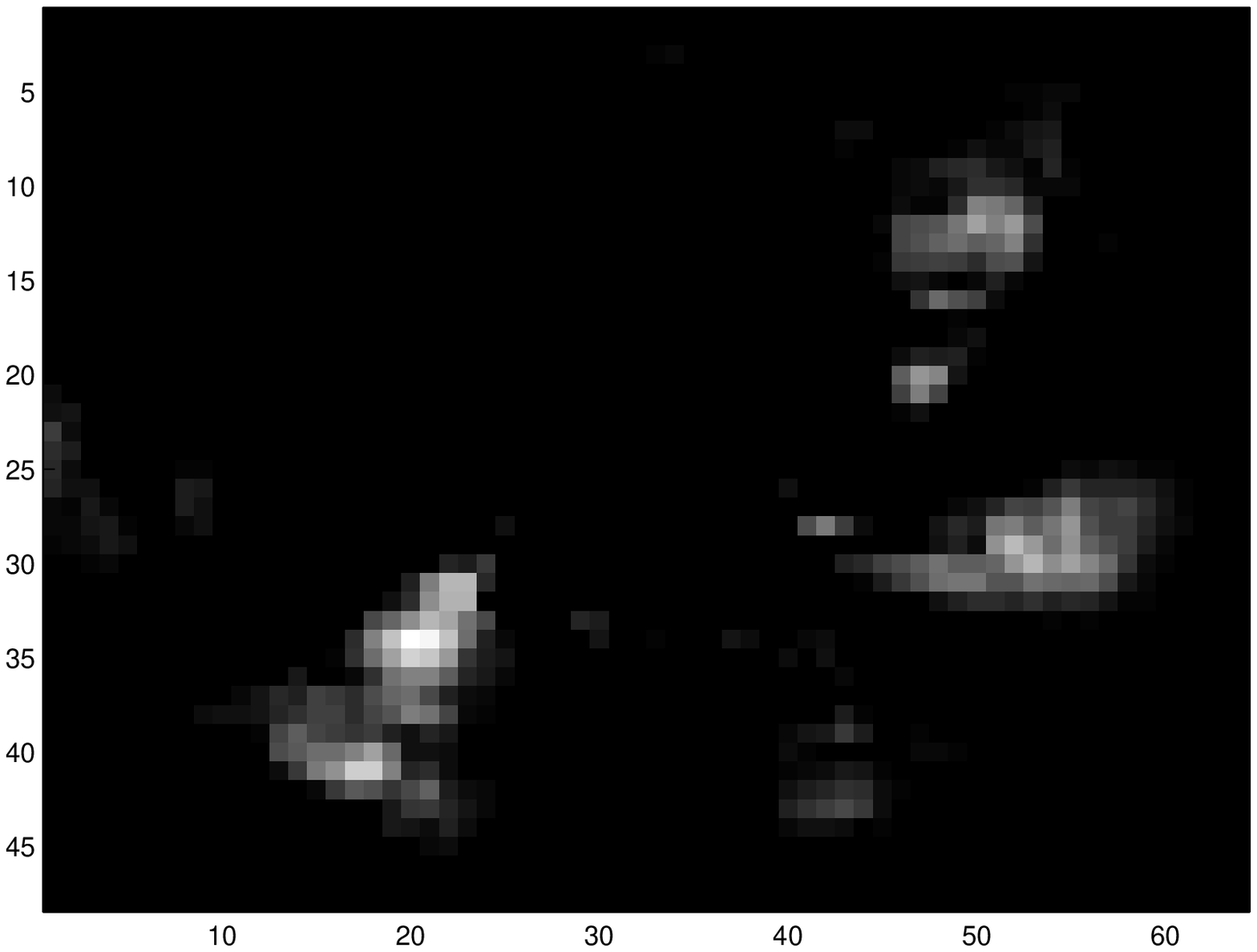}
        \includegraphics[width=1.25cm]{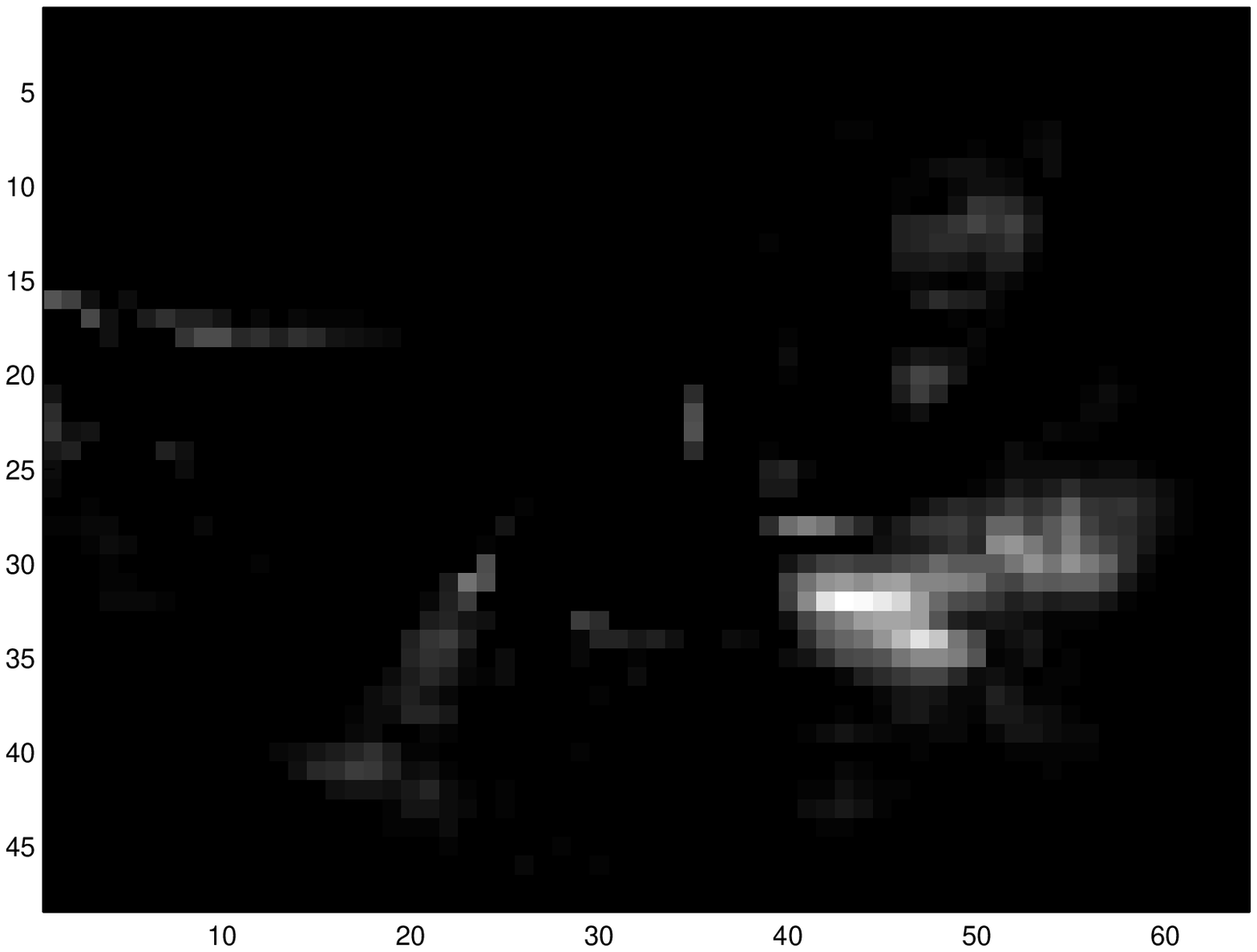}
          \includegraphics[width=1.25cm]{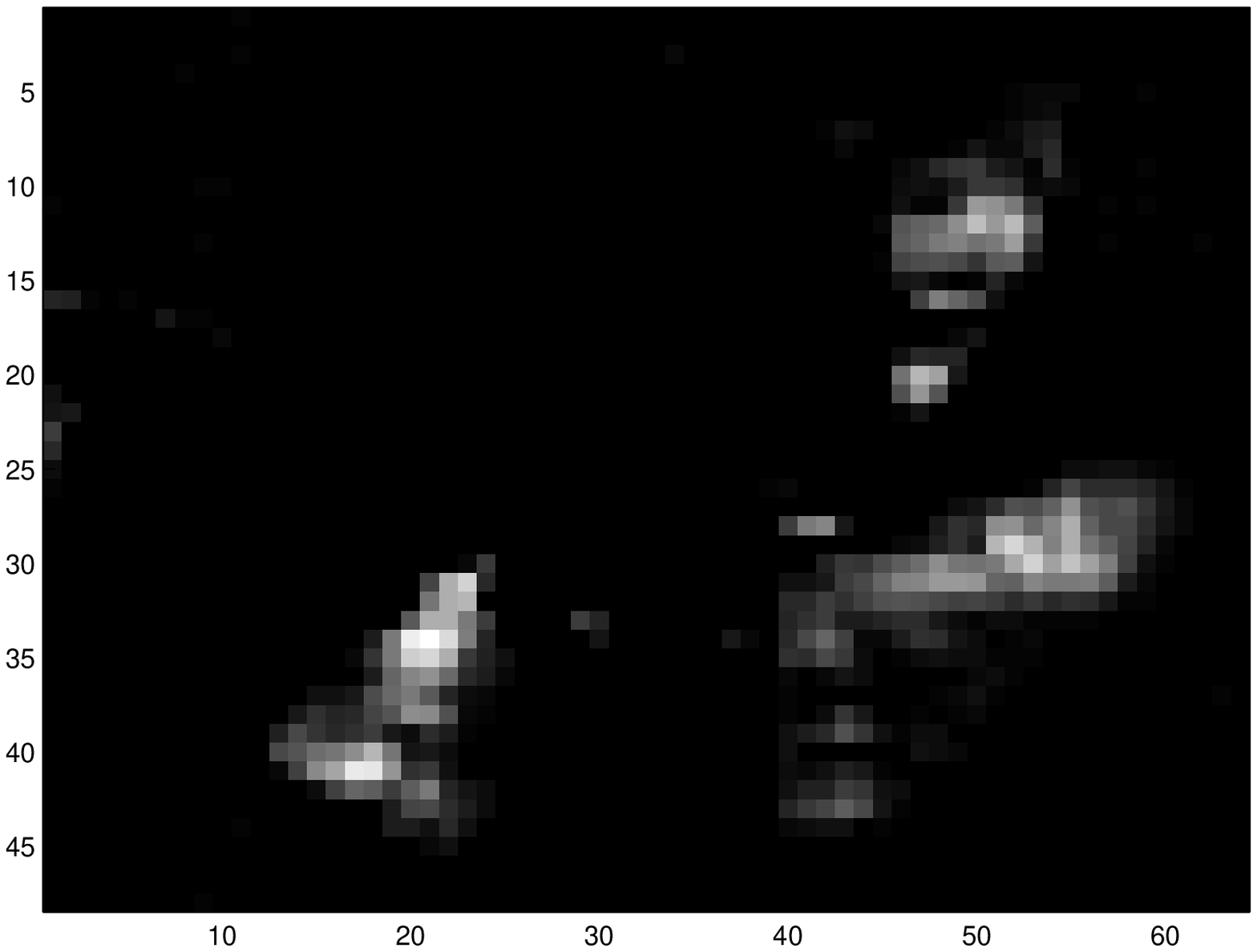}
            \includegraphics[width=1.25cm]{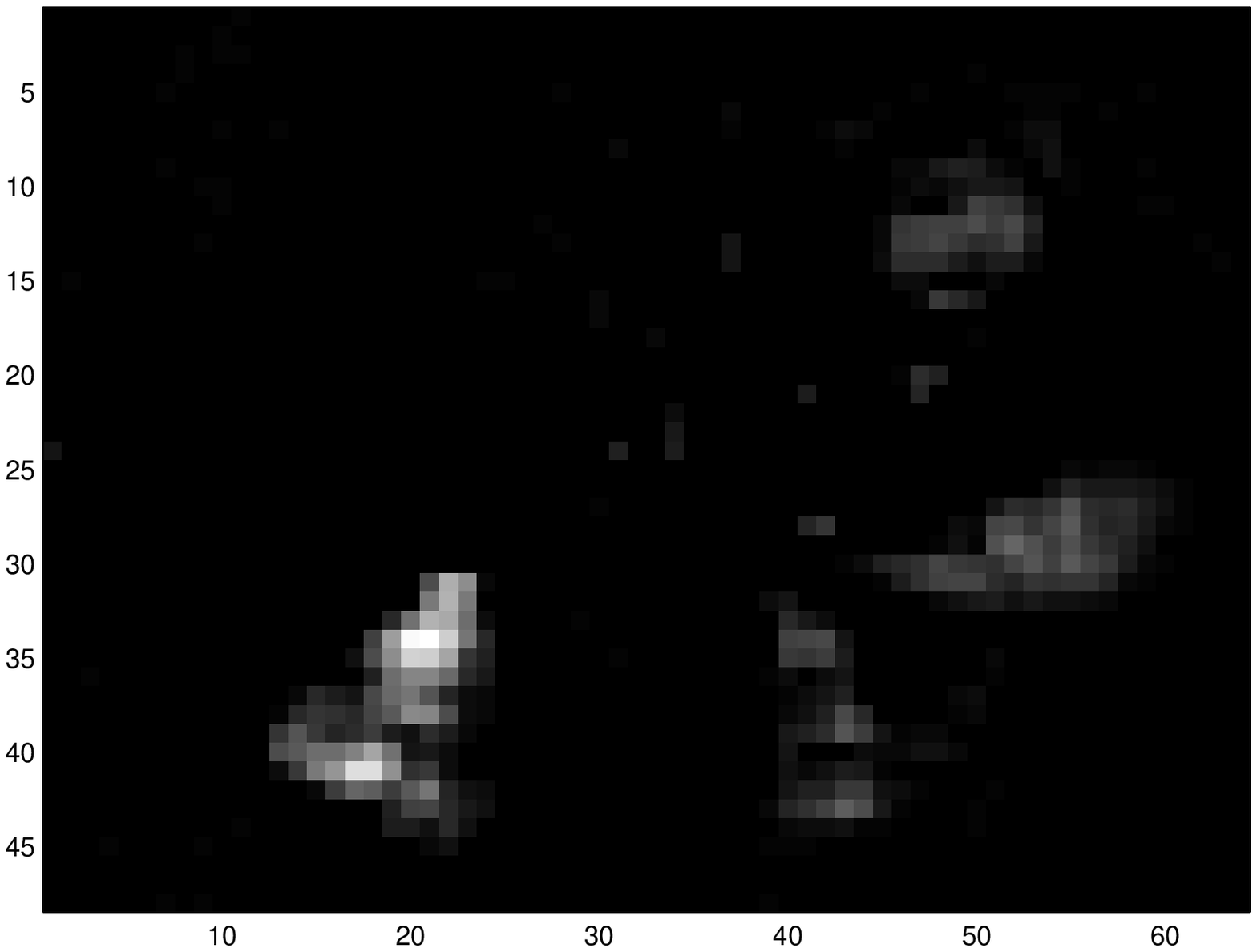}
              \includegraphics[width=1.25cm]{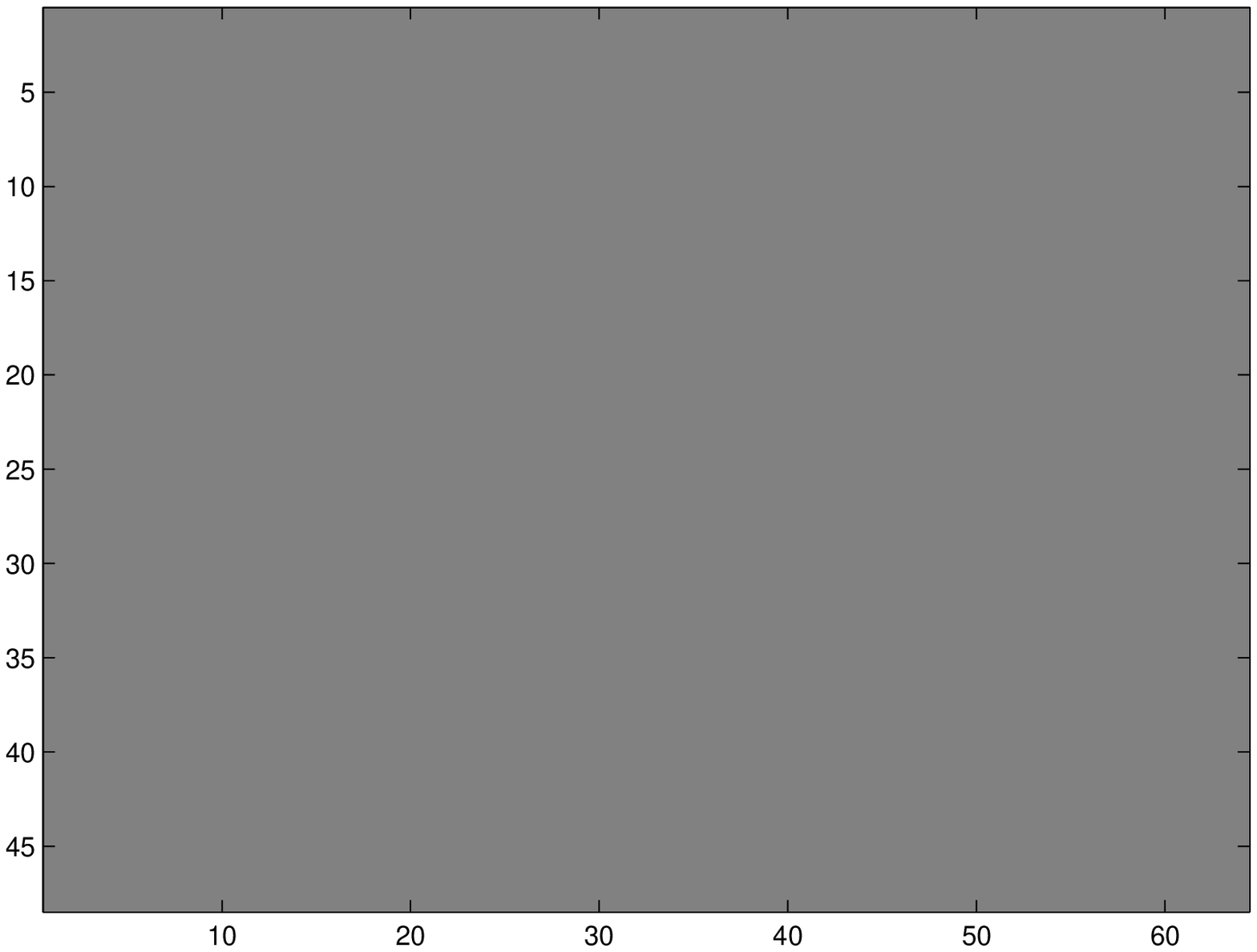}
                \includegraphics[width=1.25cm]{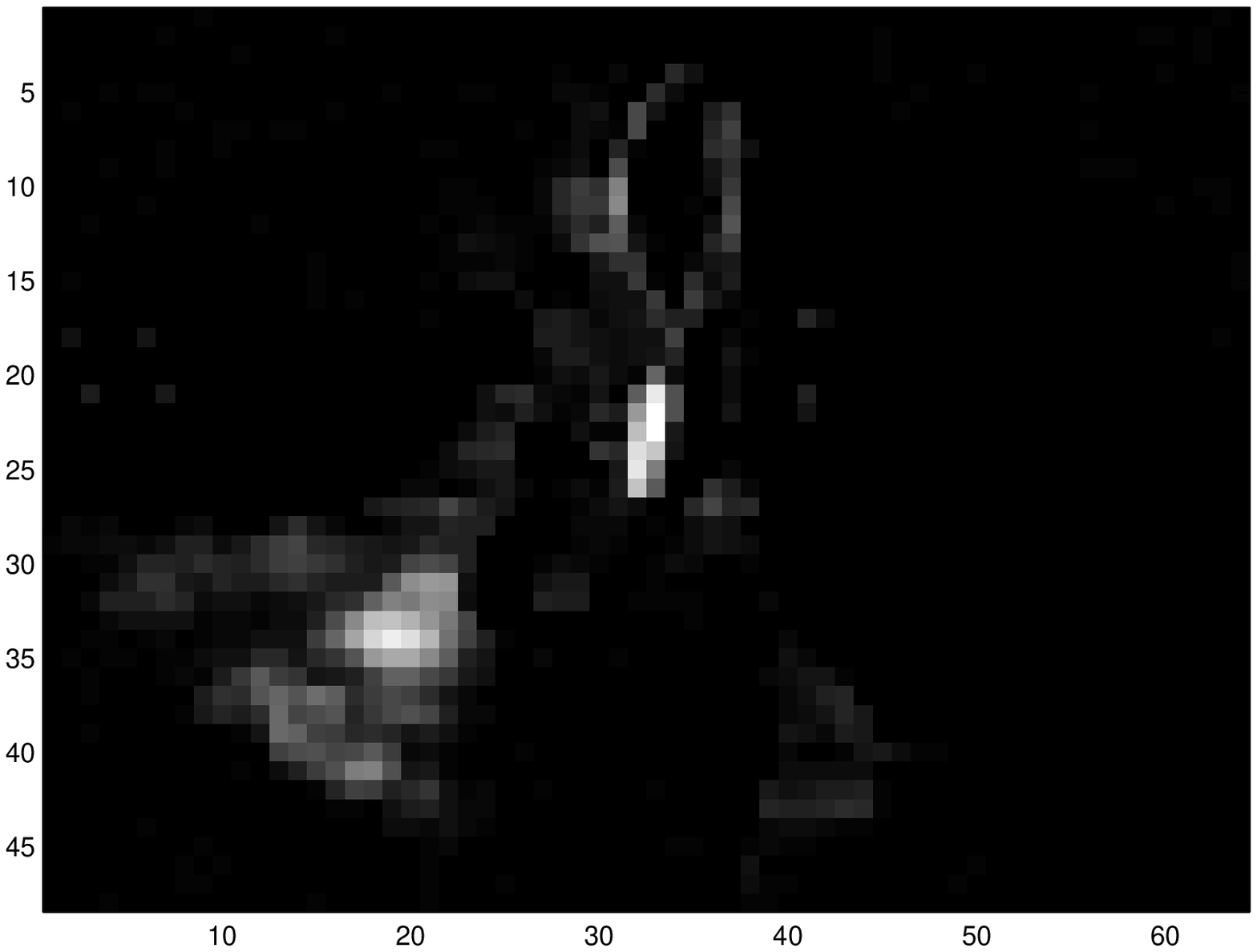}
                \includegraphics[width=1.25cm]{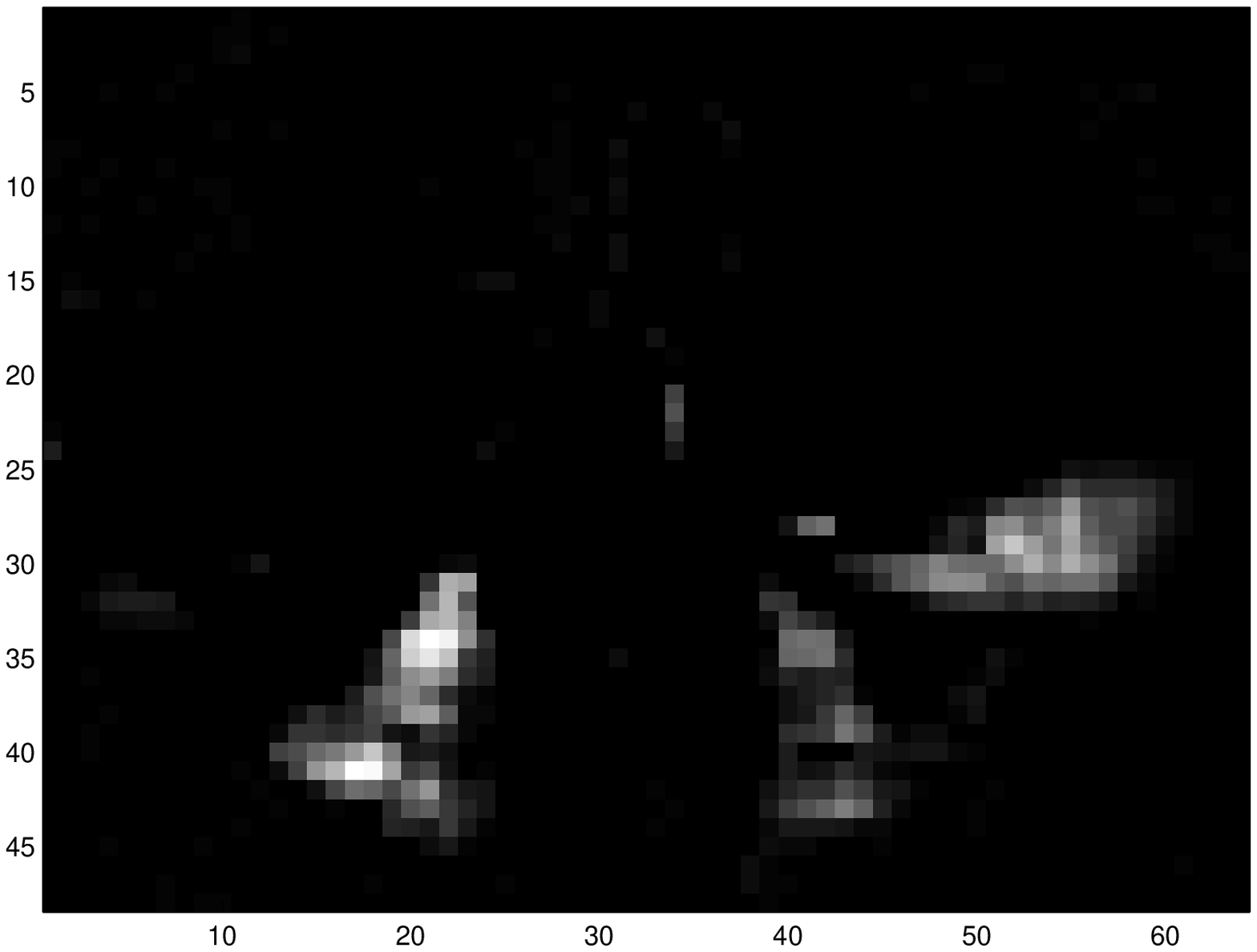}
                \includegraphics[width=1.25cm]{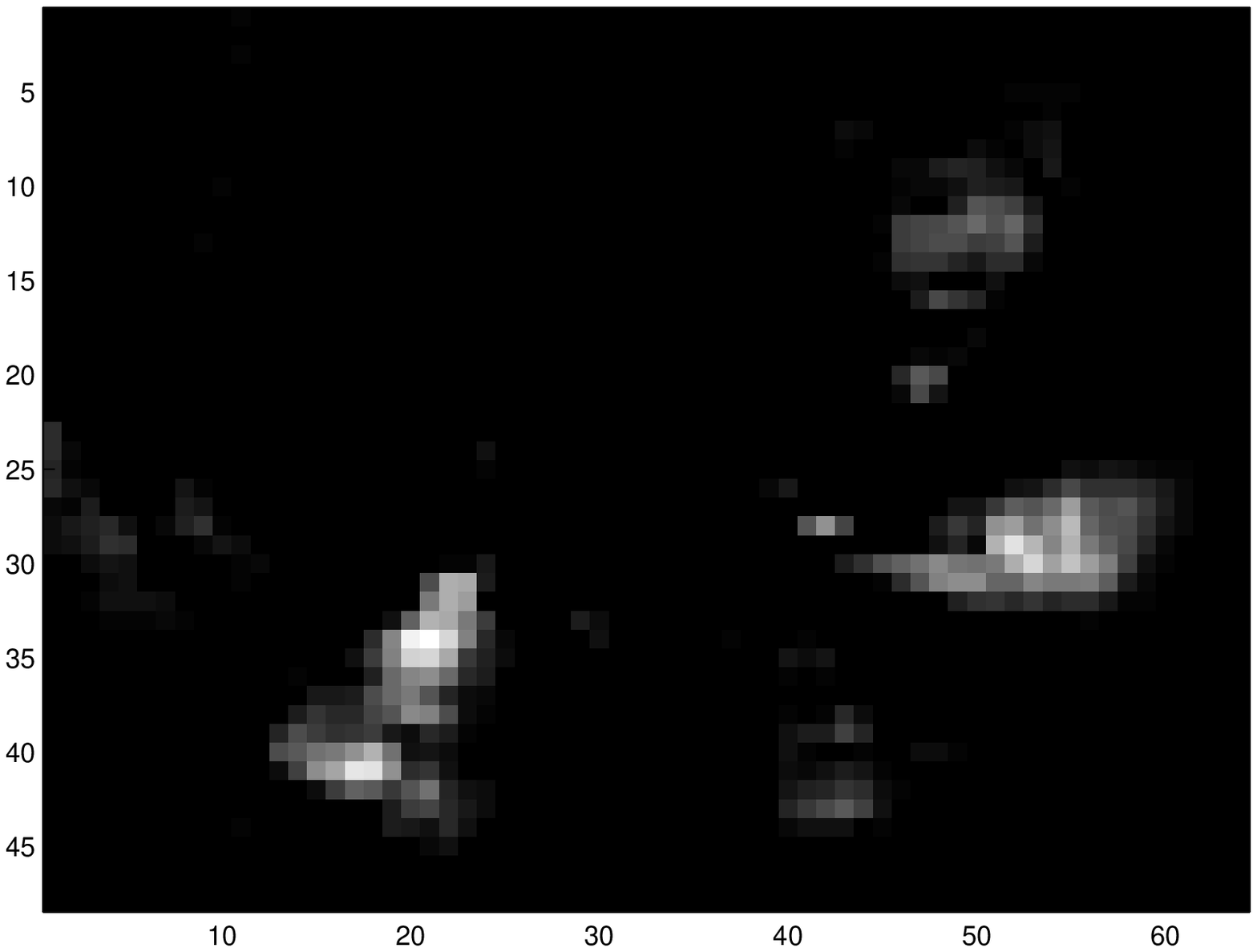}\\
  \includegraphics[width=1.25cm]{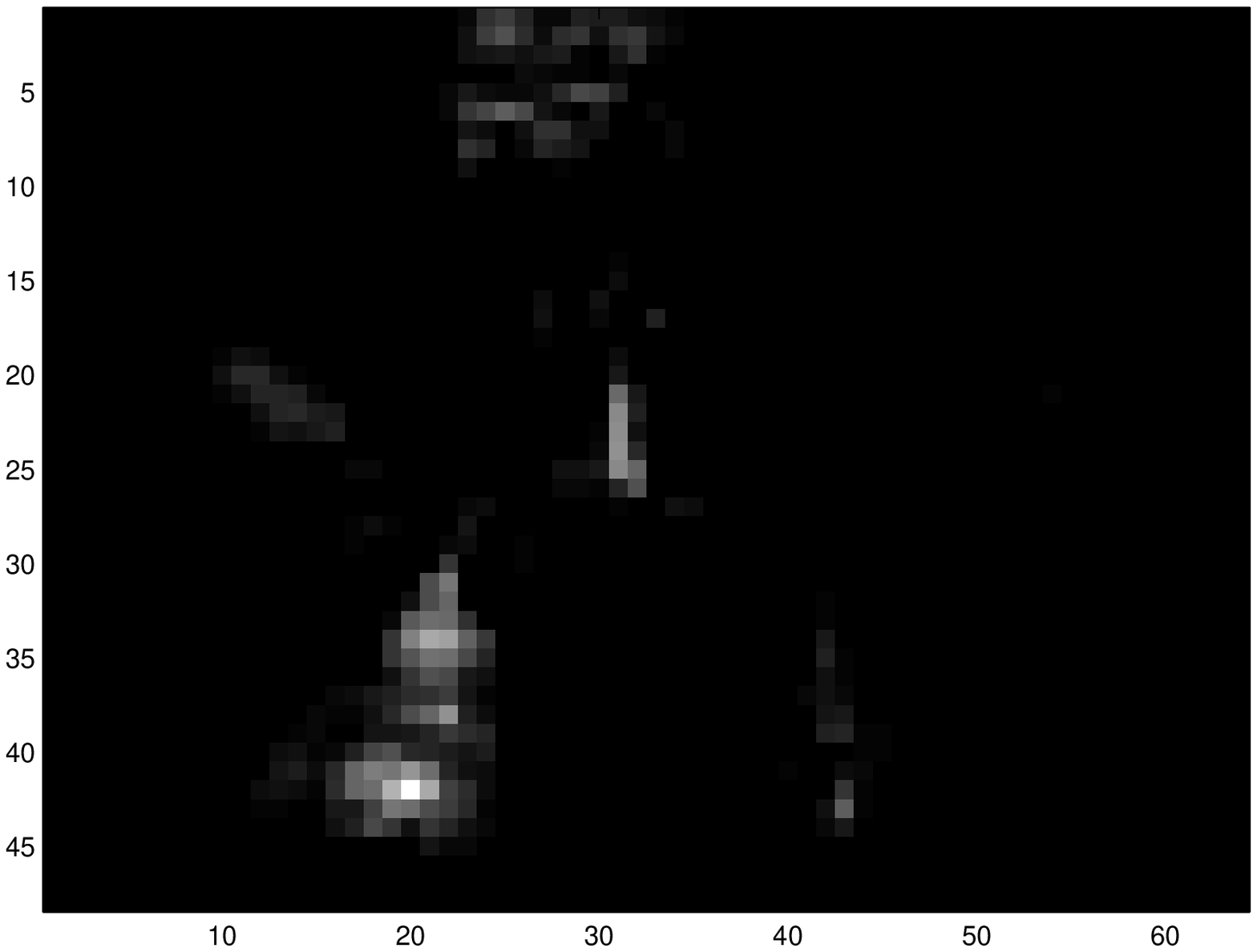}
    \includegraphics[width=1.25cm]{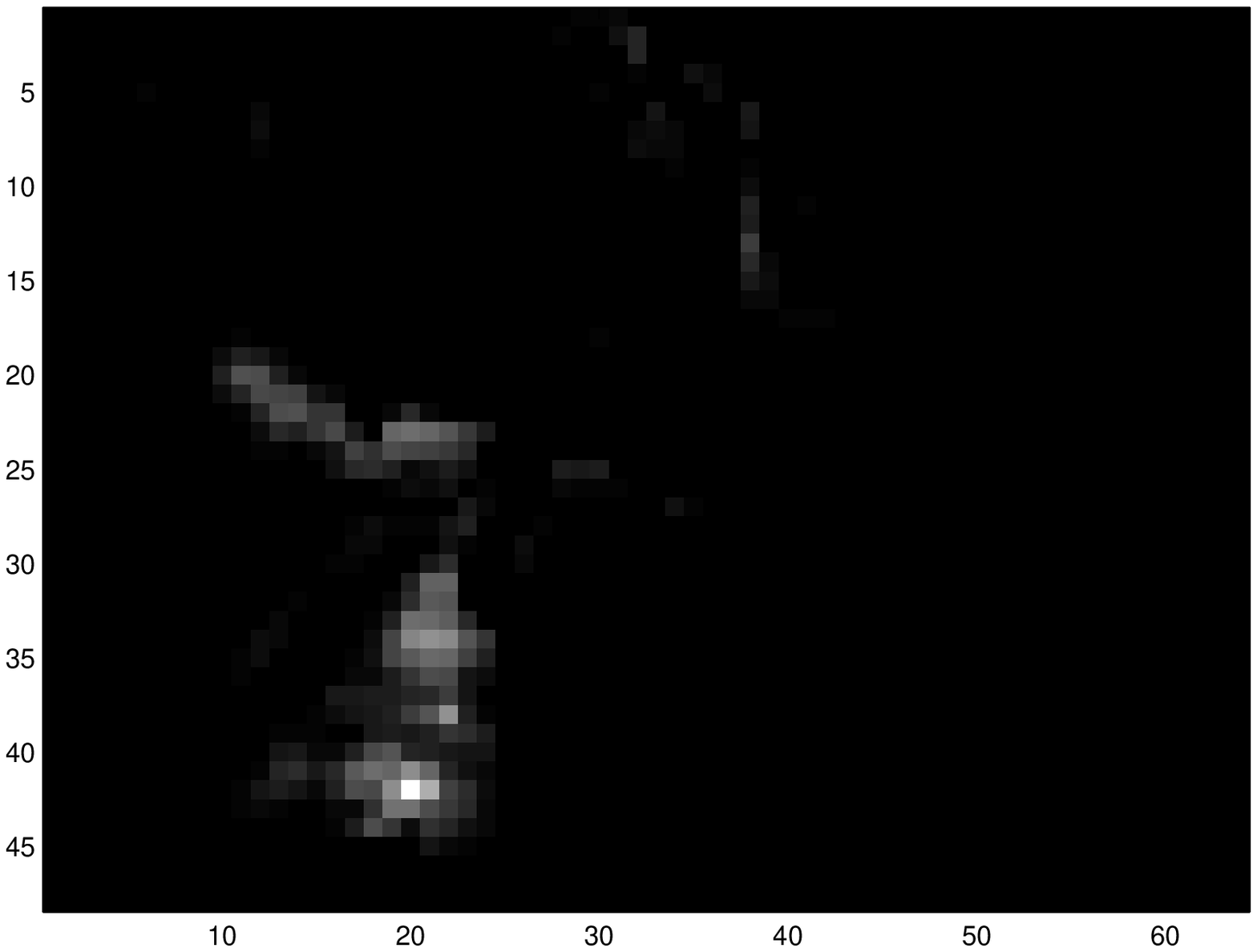}
      \includegraphics[width=1.25cm]{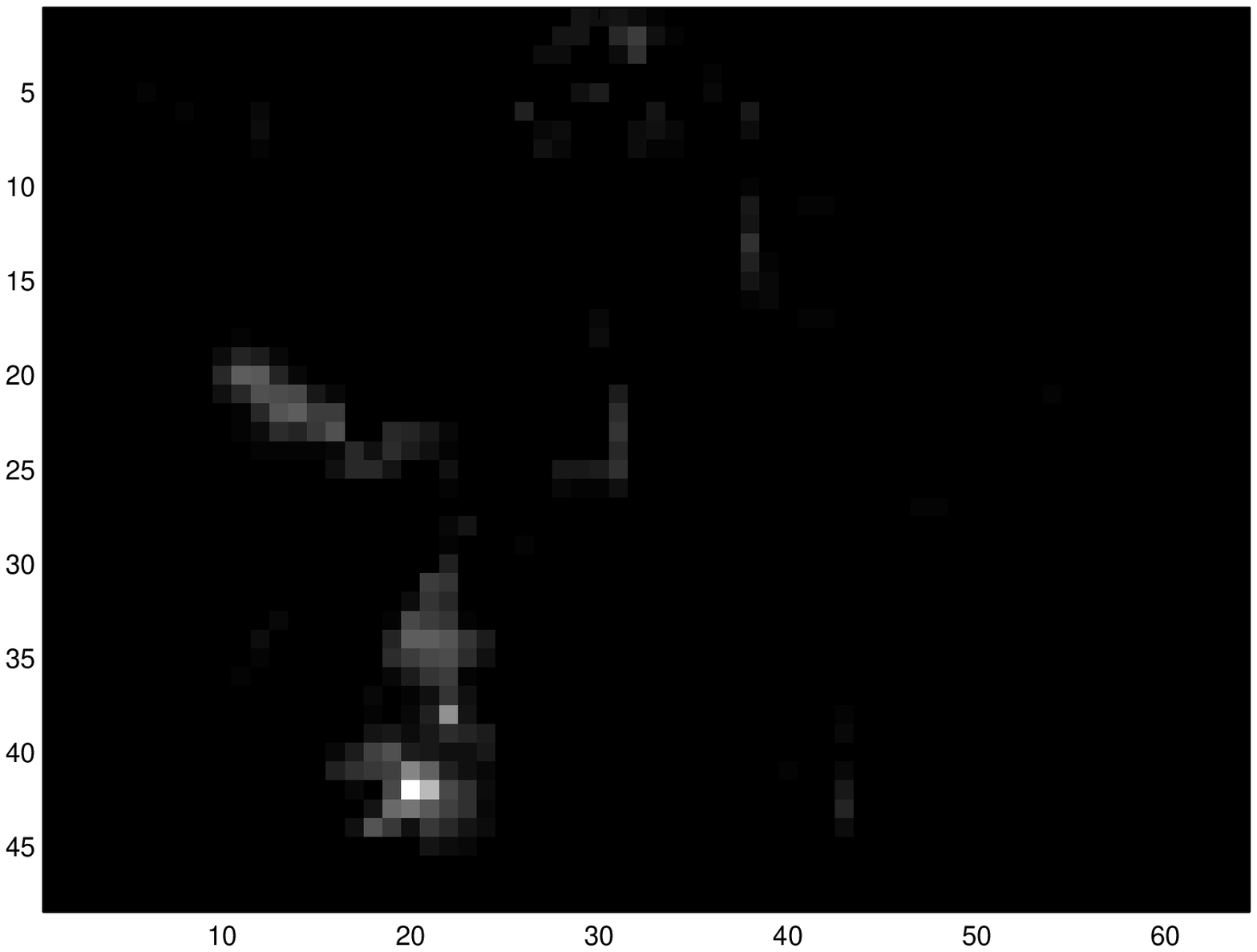}
        \includegraphics[width=1.25cm]{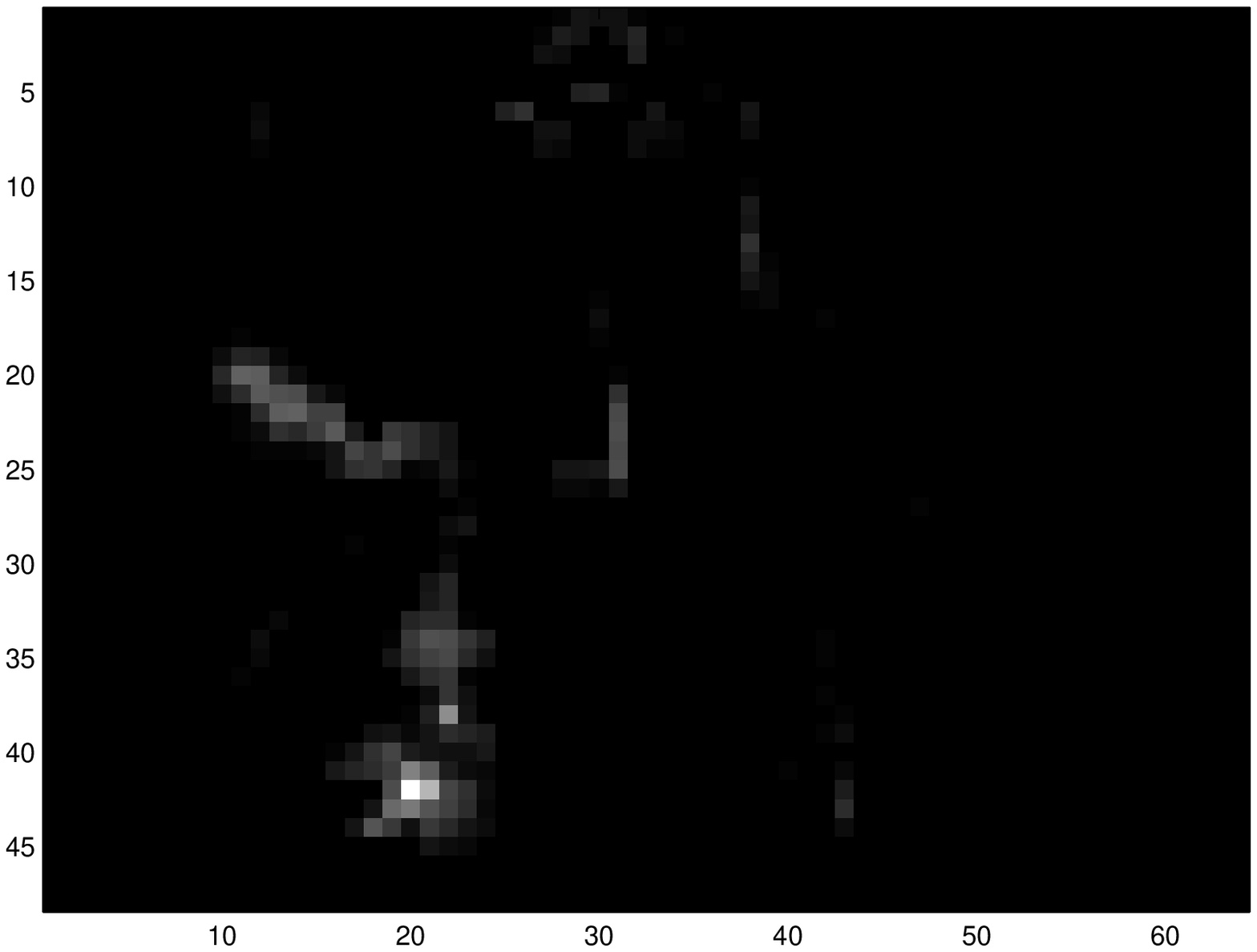}
          \includegraphics[width=1.25cm]{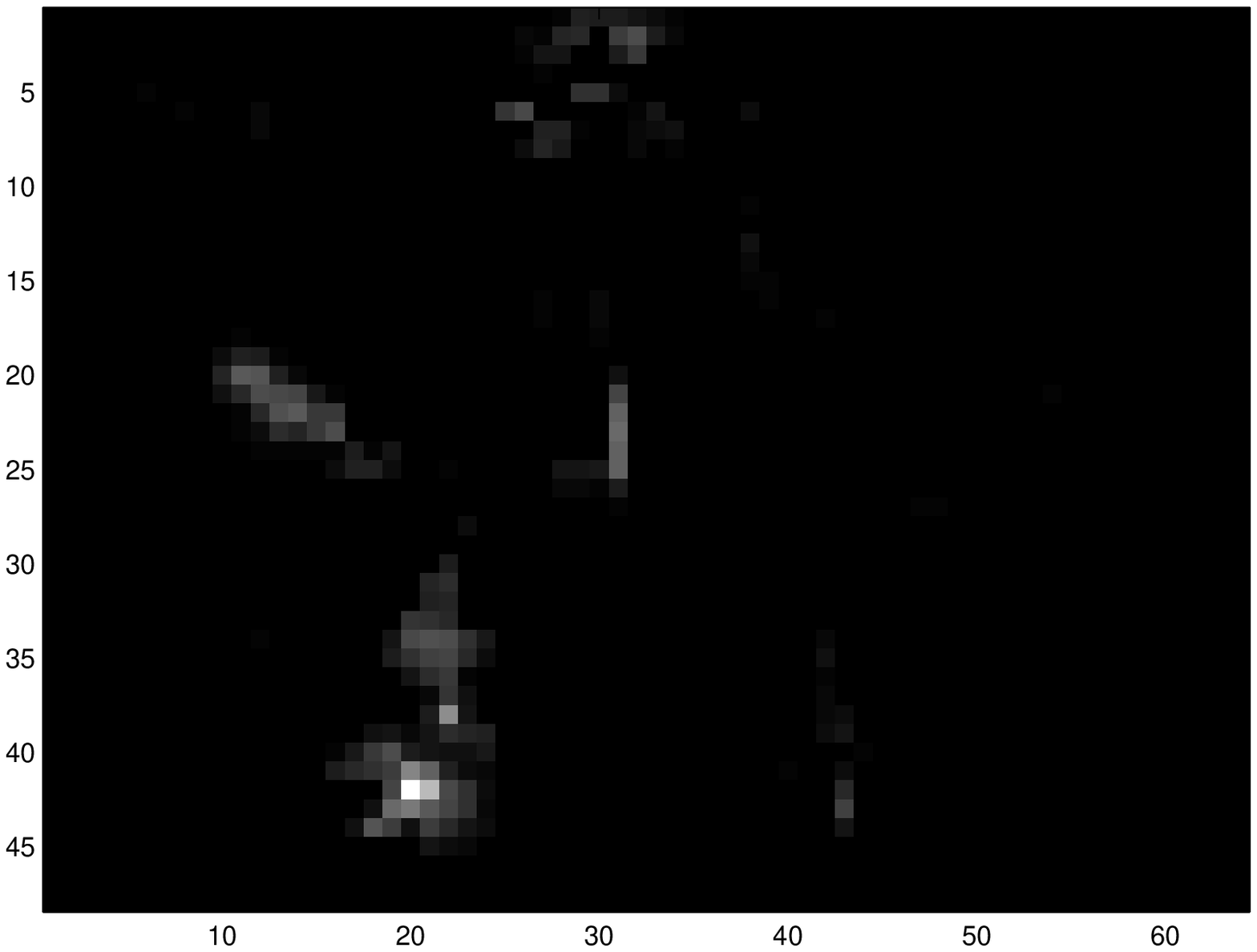}
            \includegraphics[width=1.25cm]{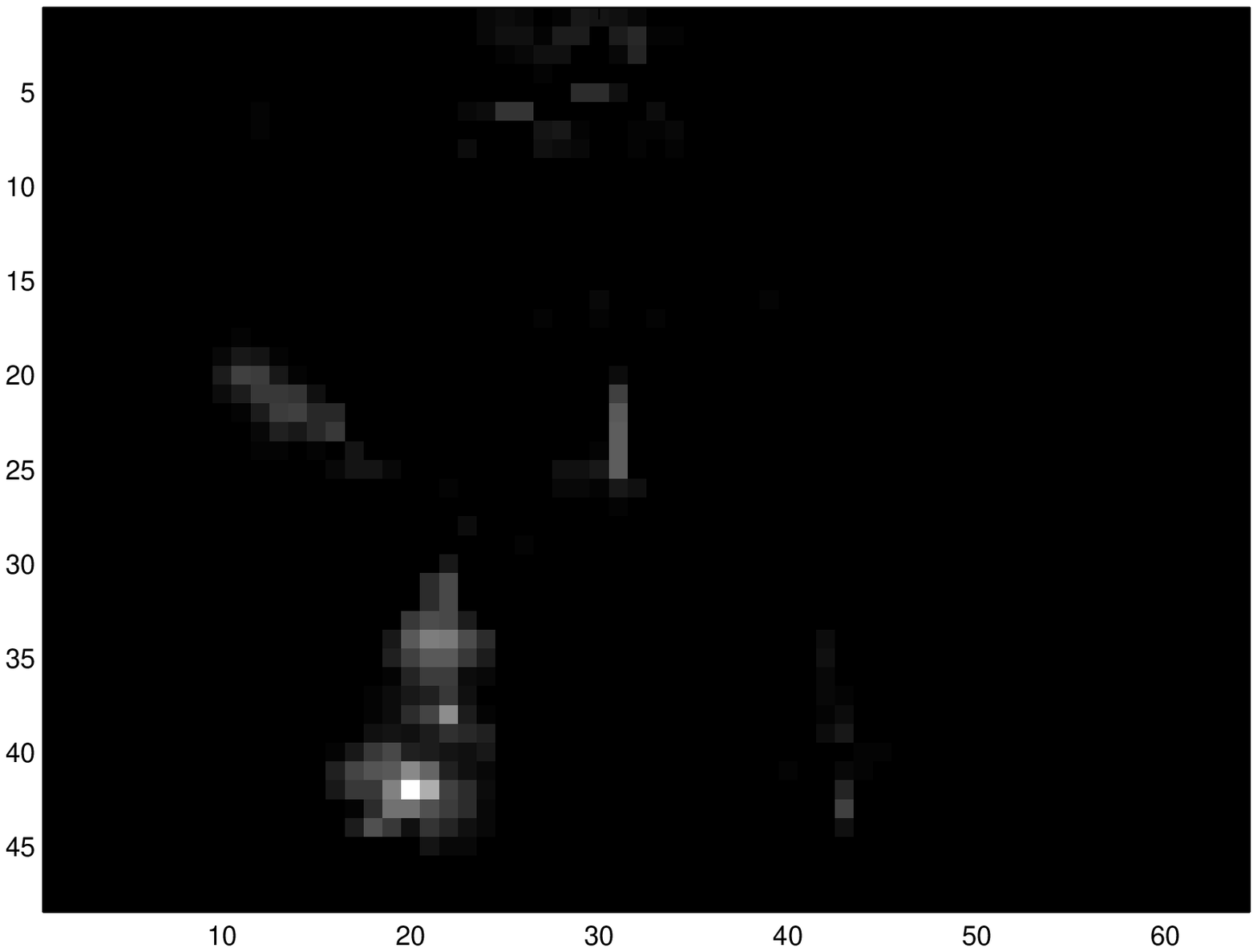}
              \includegraphics[width=1.25cm]{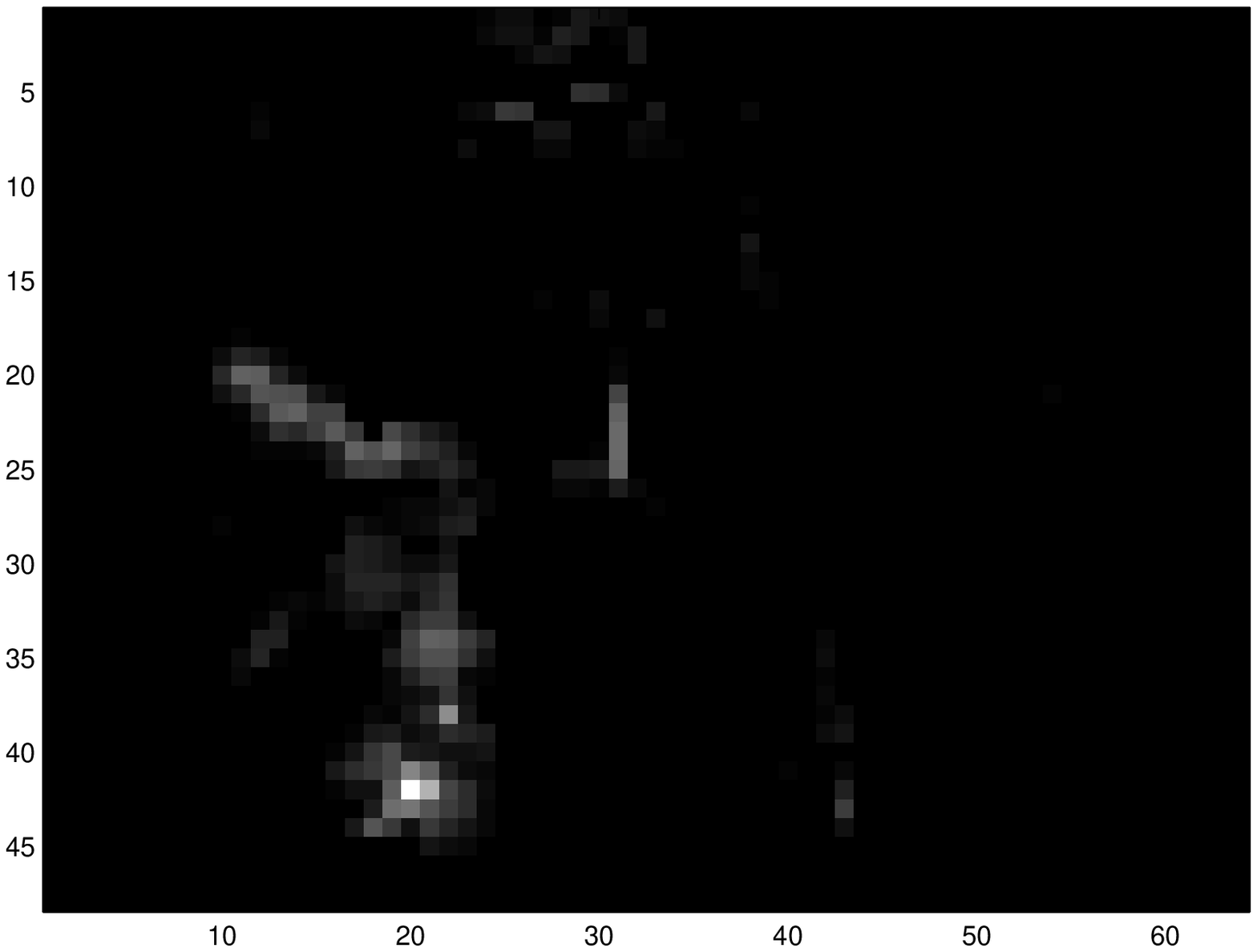}
                \includegraphics[width=1.25cm]{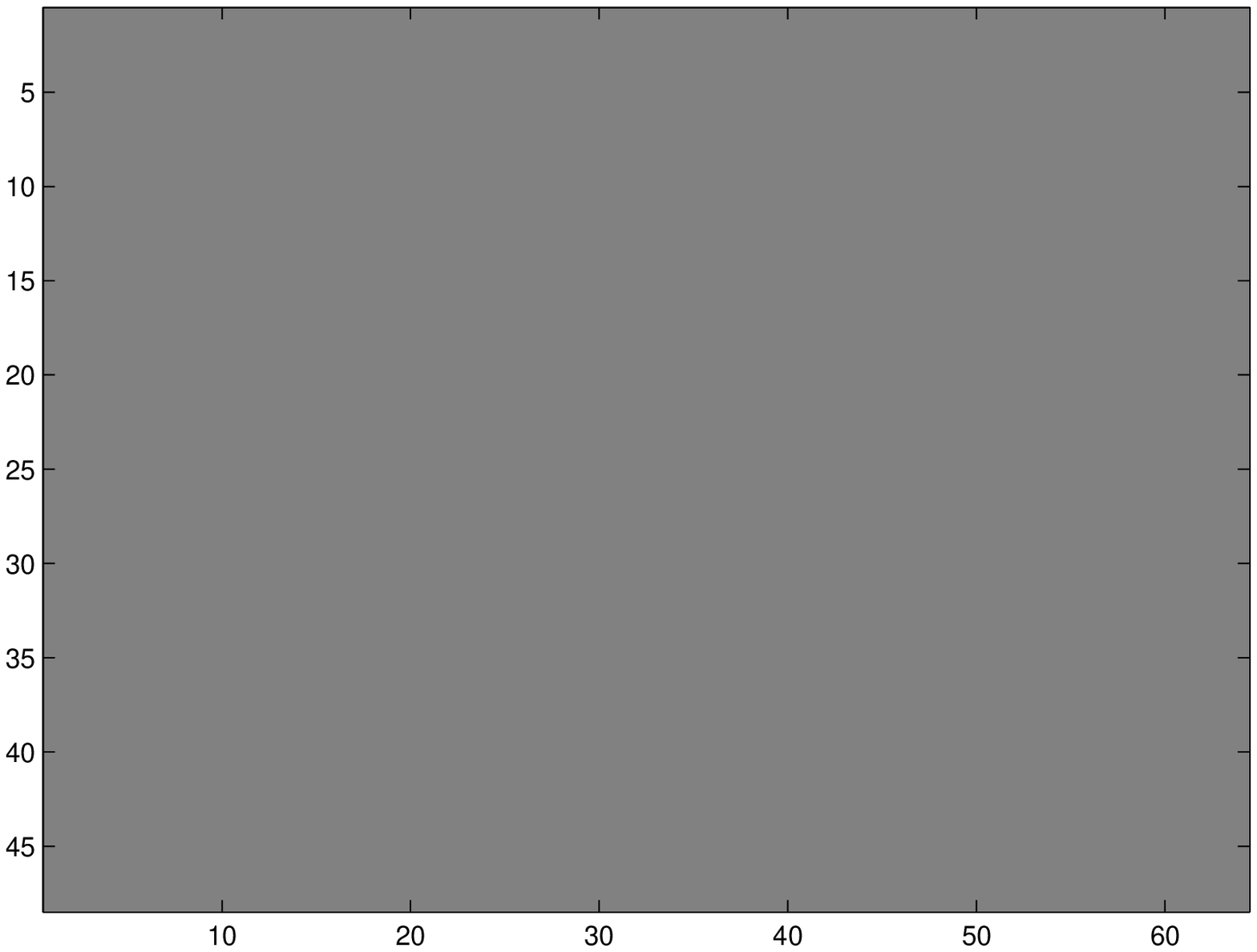}
                \includegraphics[width=1.25cm]{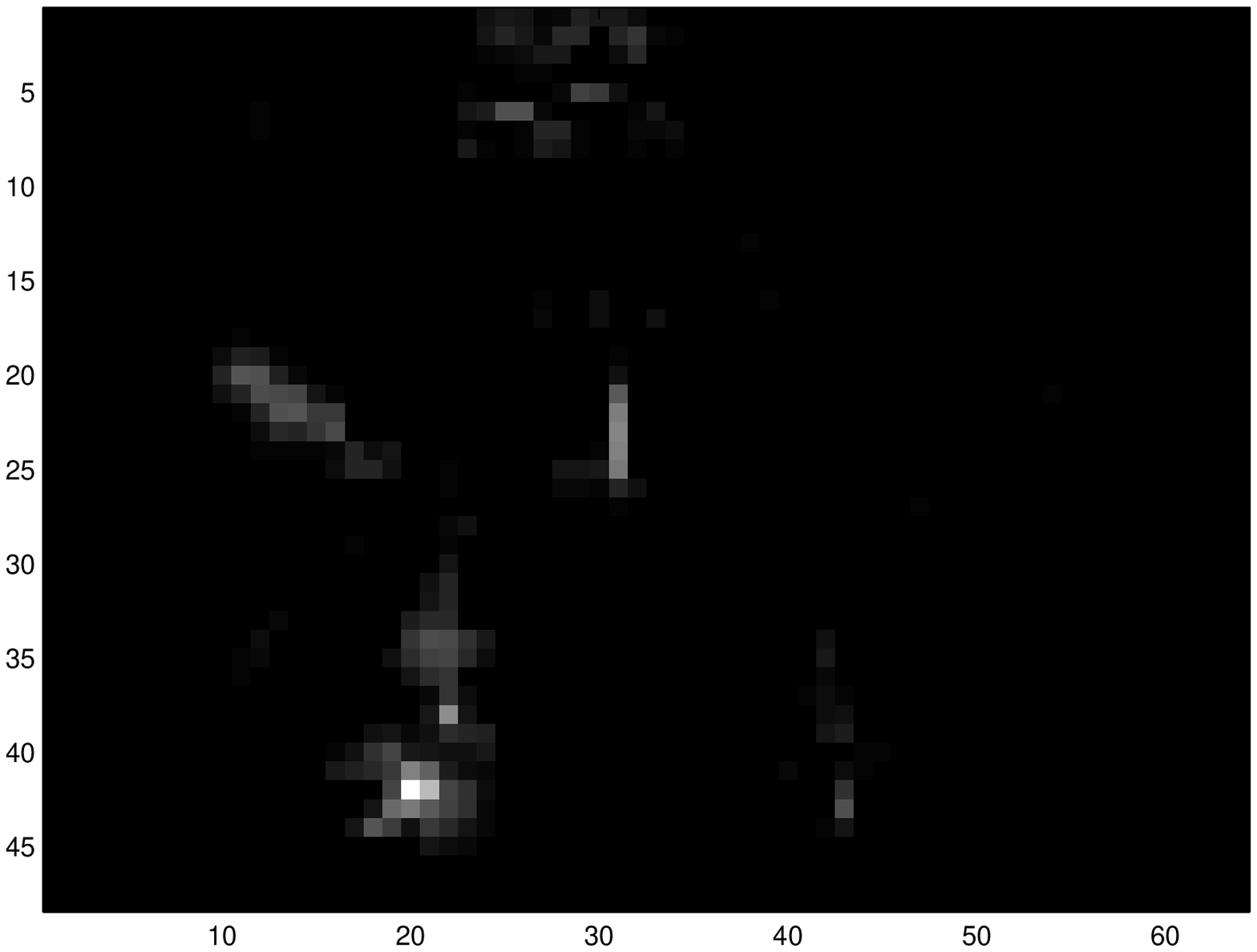}
                \includegraphics[width=1.25cm]{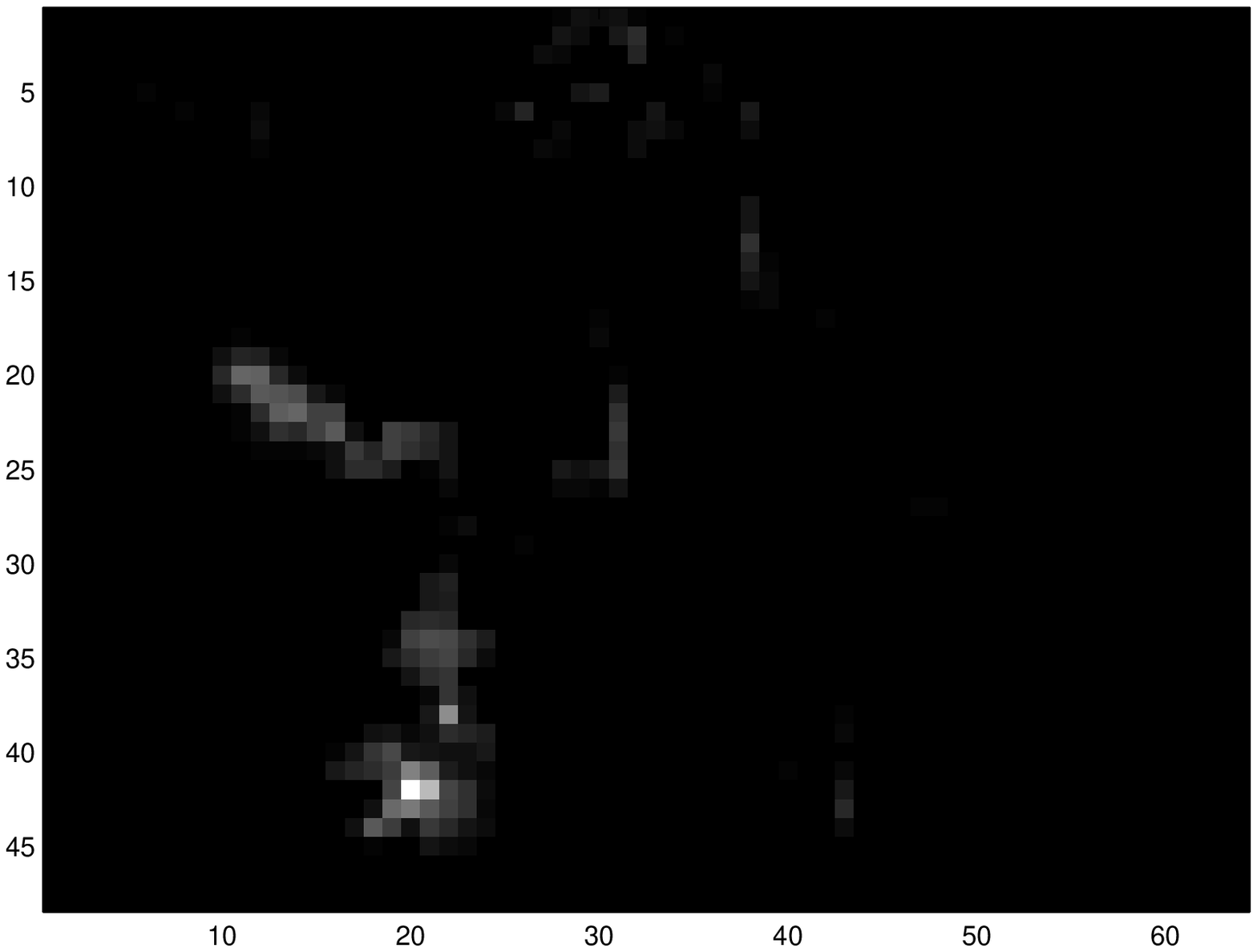}\\
                  \includegraphics[width=1.25cm]{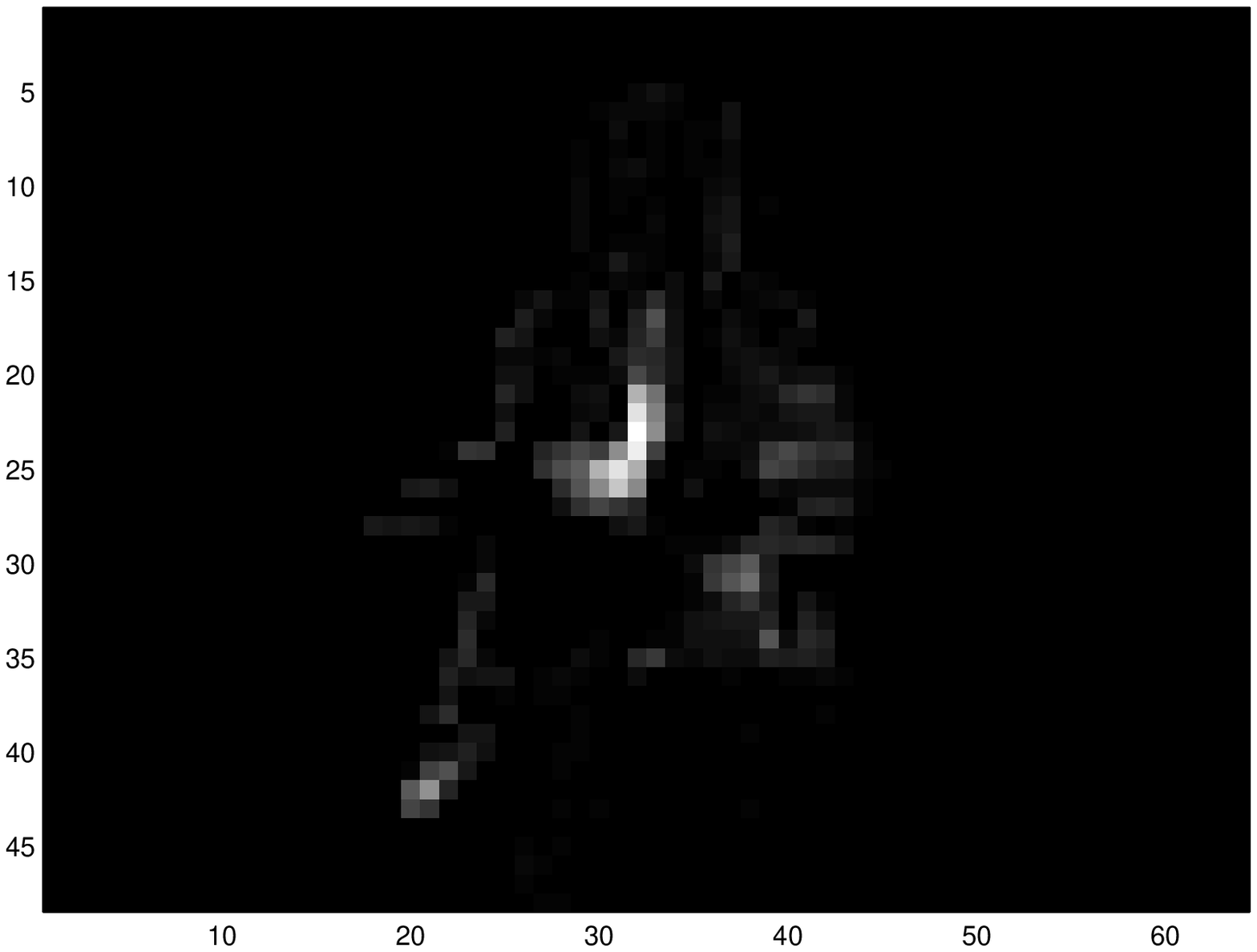}
    \includegraphics[width=1.25cm]{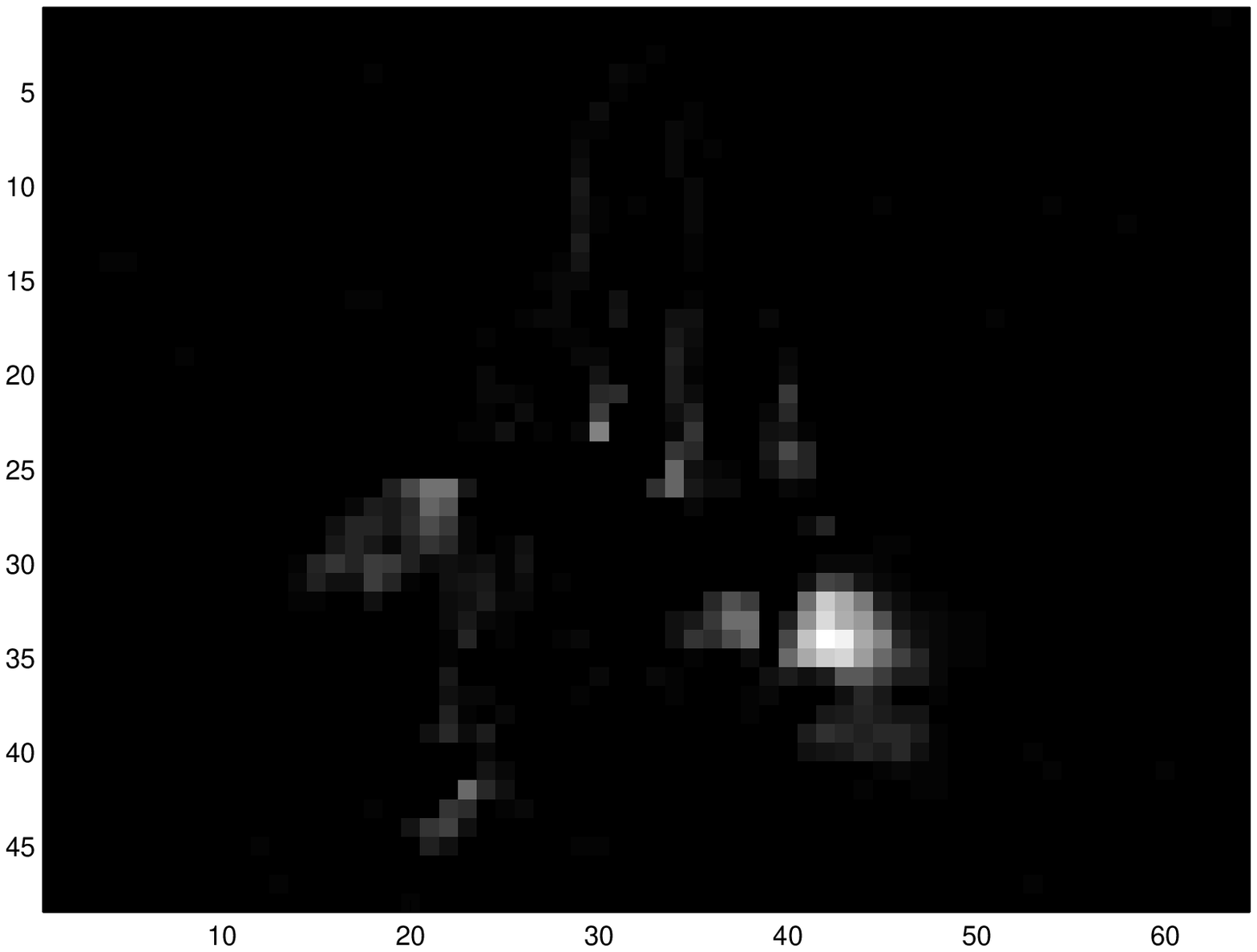}
      \includegraphics[width=1.25cm]{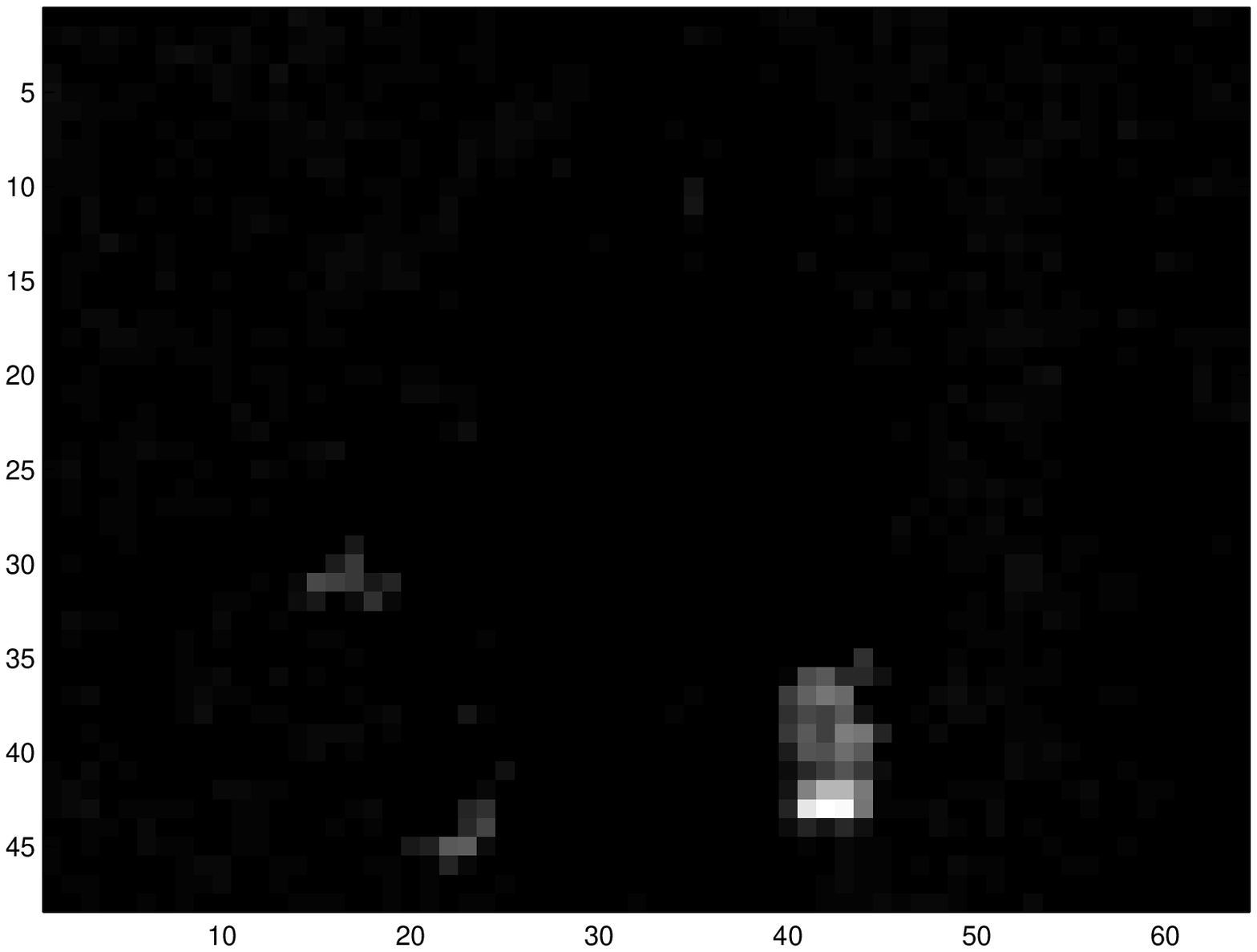}
        \includegraphics[width=1.25cm]{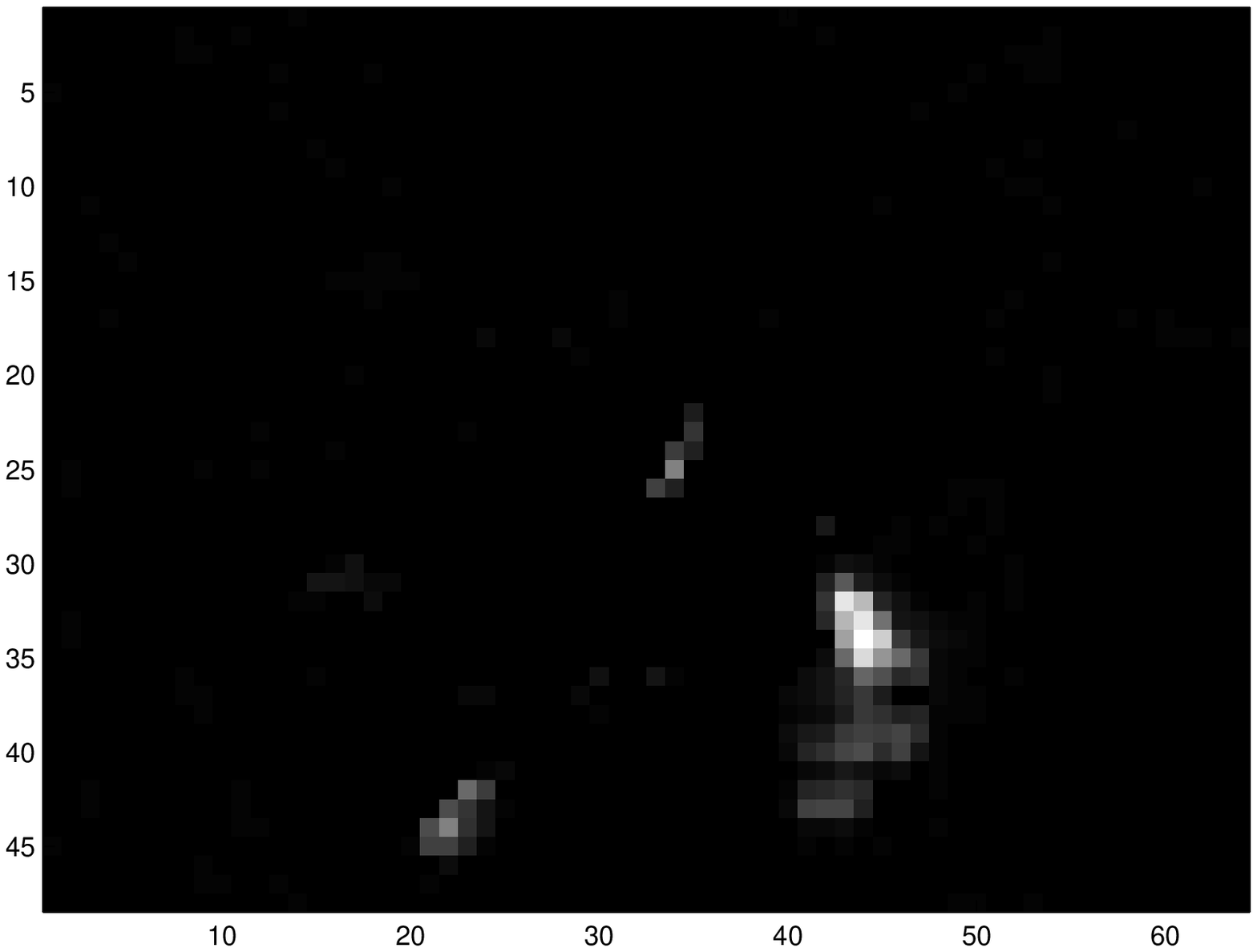}
          \includegraphics[width=1.25cm]{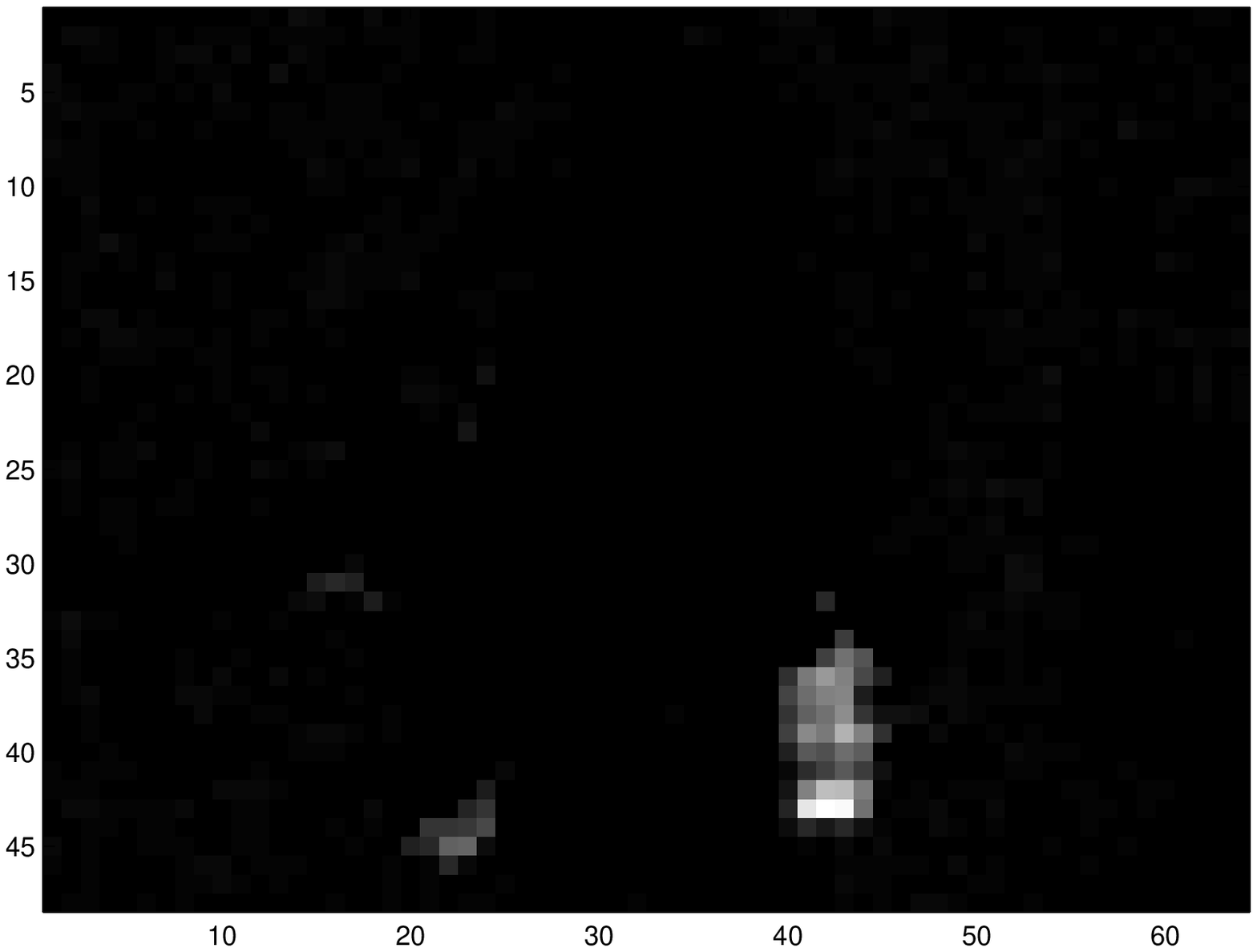}
            \includegraphics[width=1.25cm]{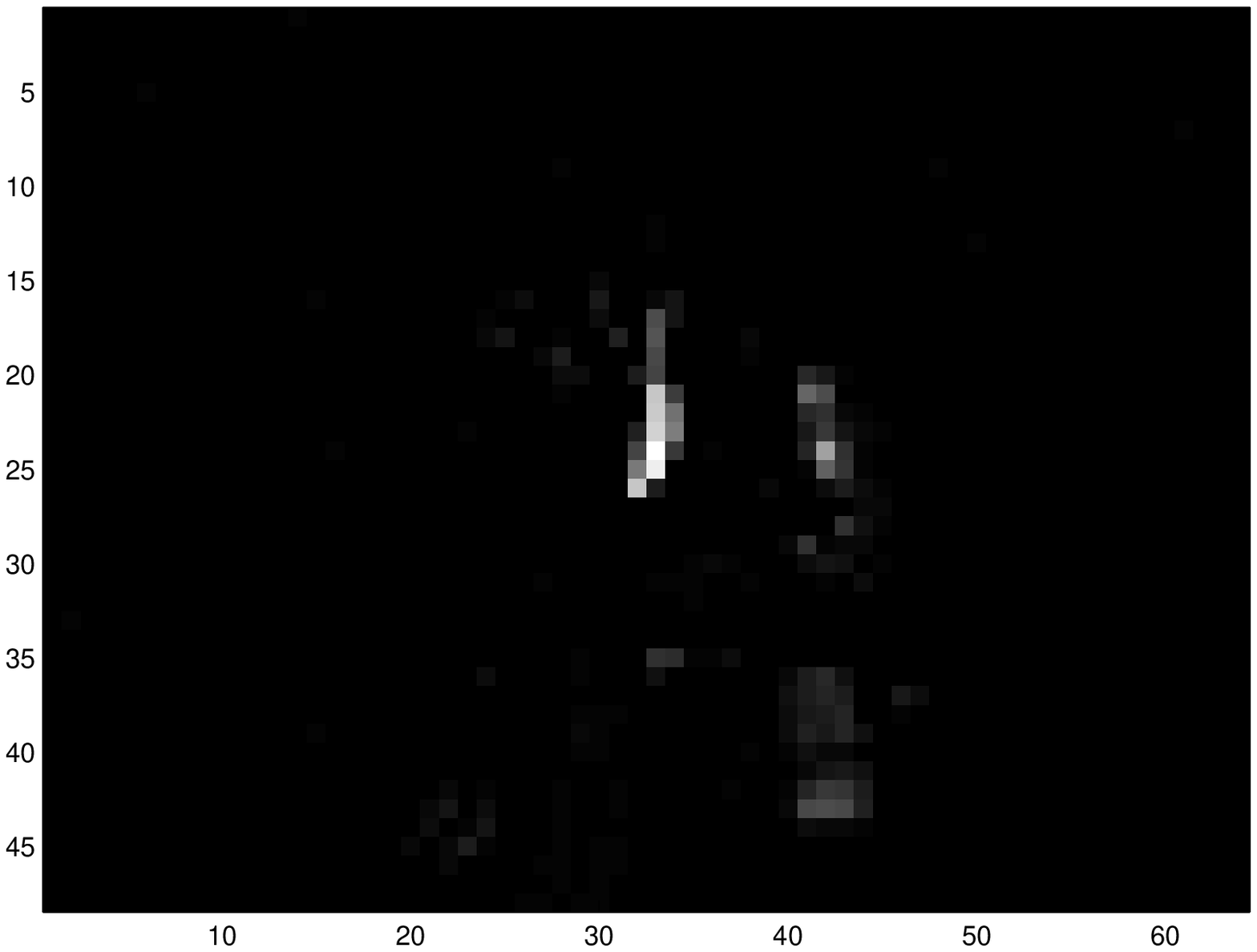}
              \includegraphics[width=1.25cm]{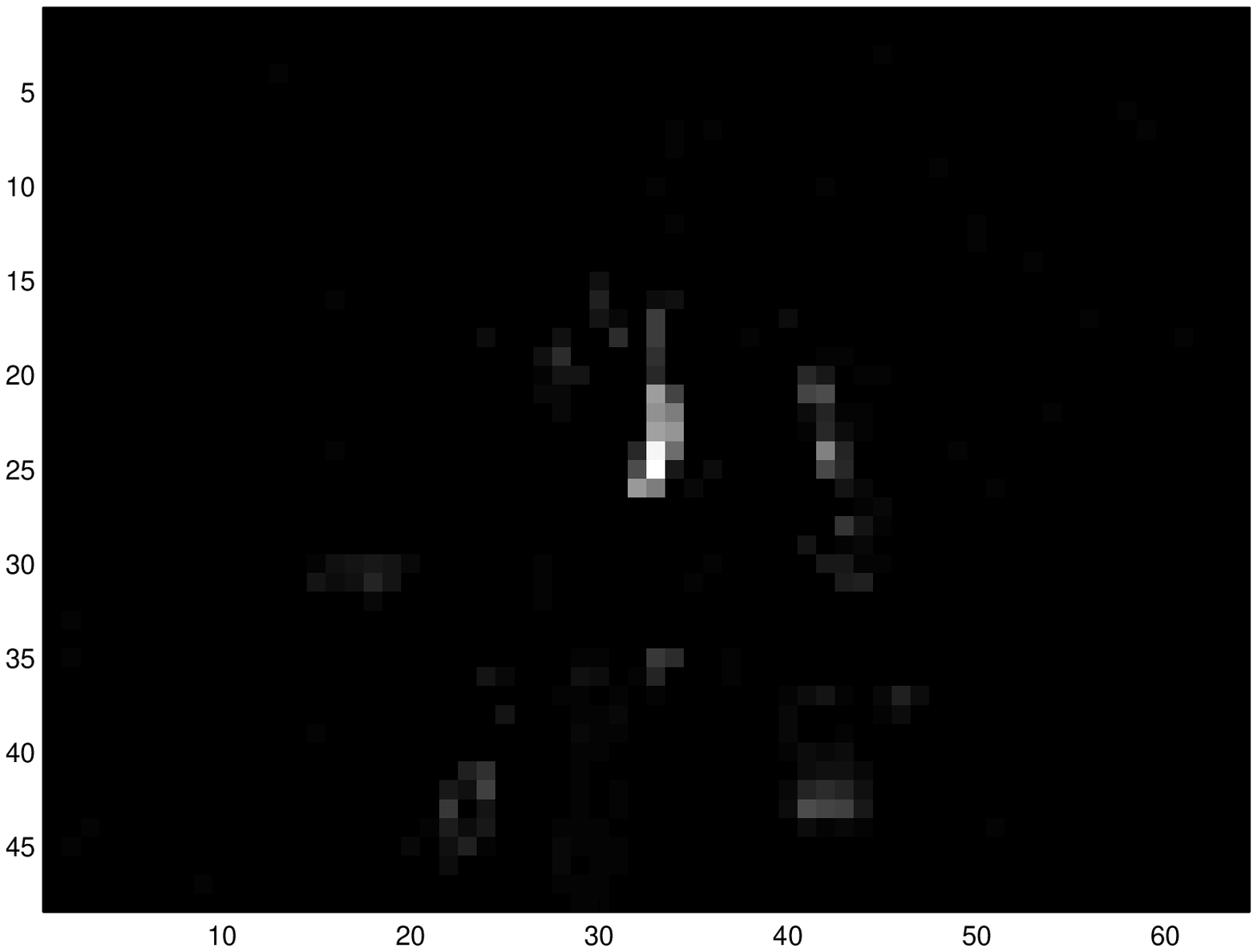}
                \includegraphics[width=1.25cm]{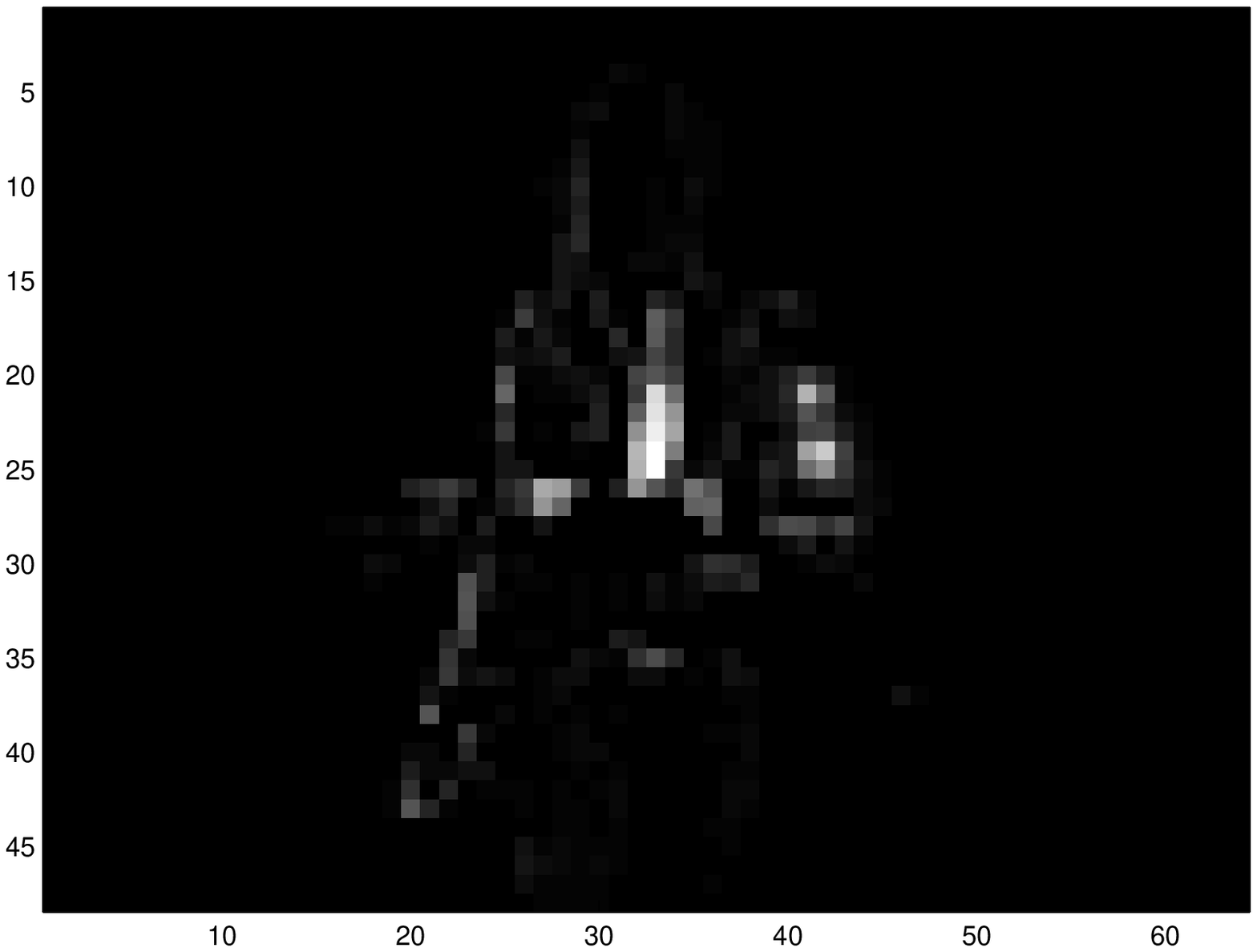}
                \includegraphics[width=1.25cm]{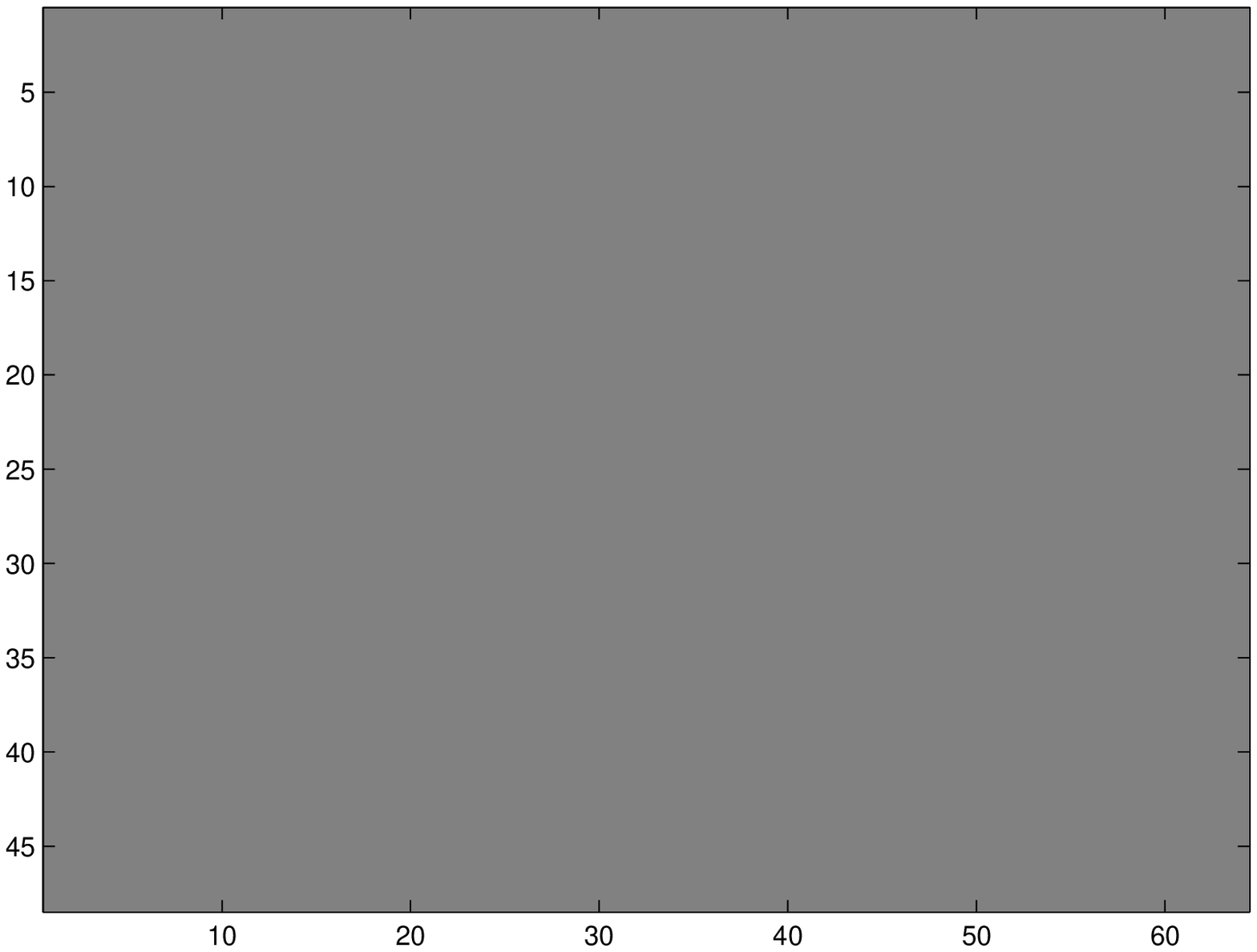}
                \includegraphics[width=1.25cm]{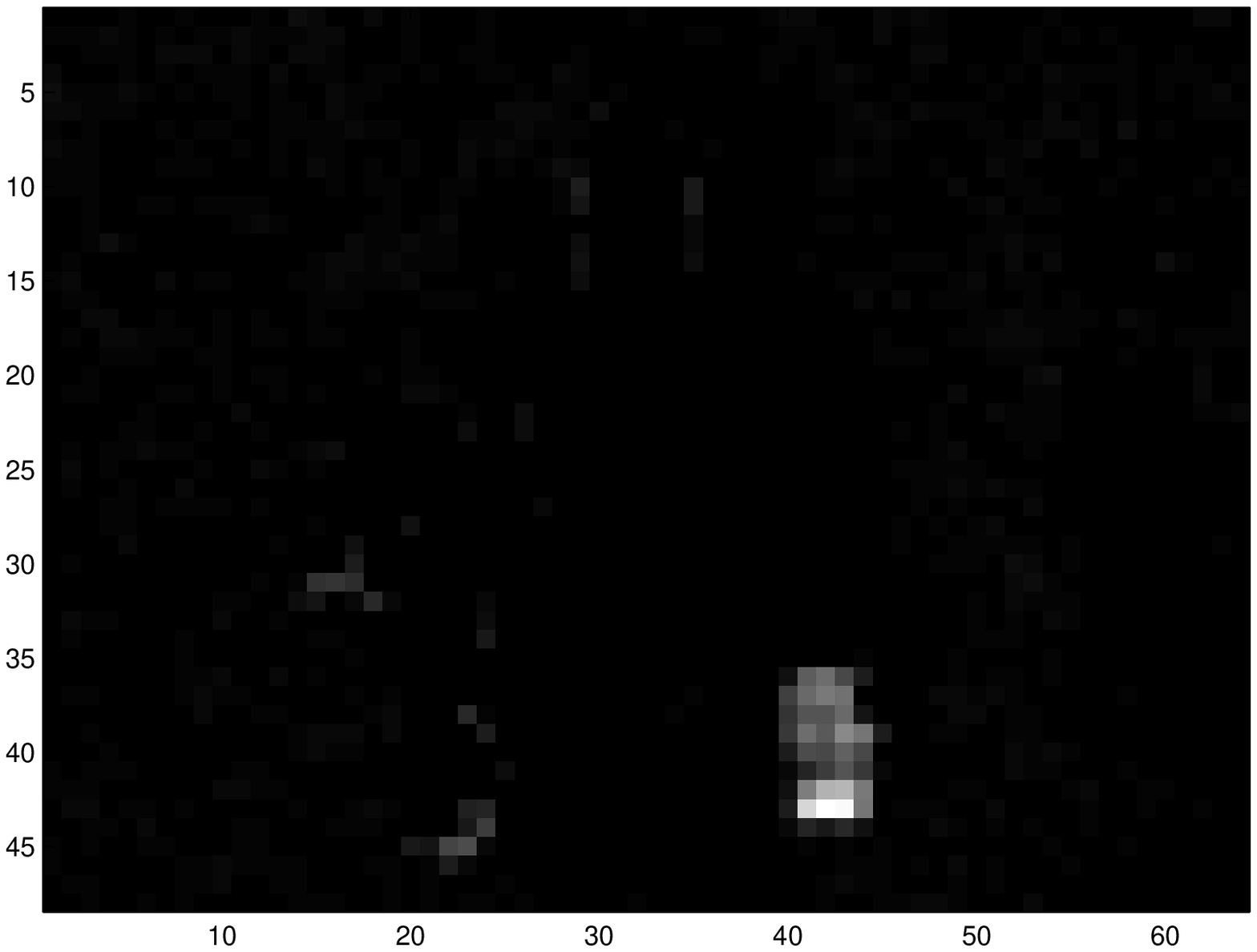}\\
                  \includegraphics[width=1.25cm]{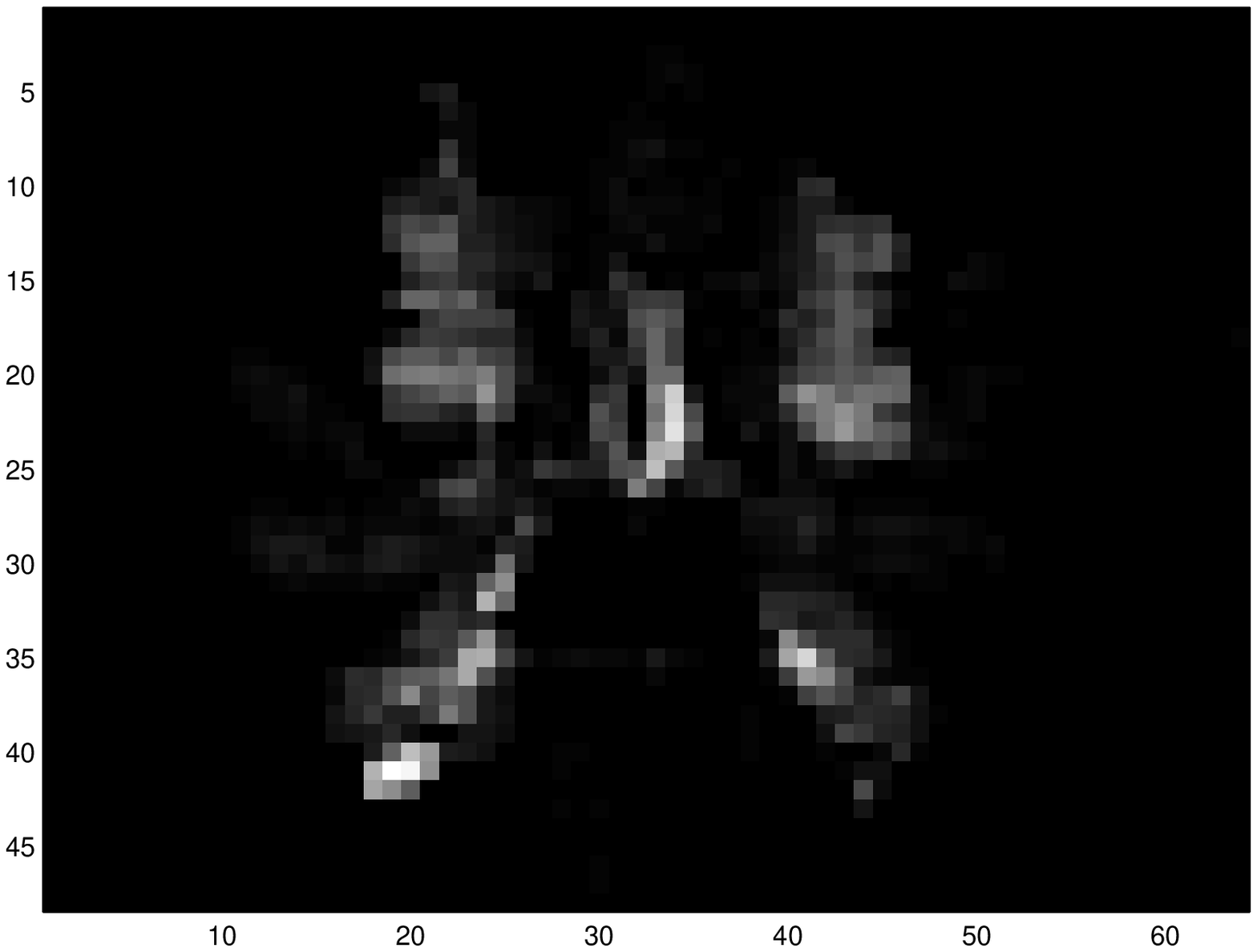}
    \includegraphics[width=1.25cm]{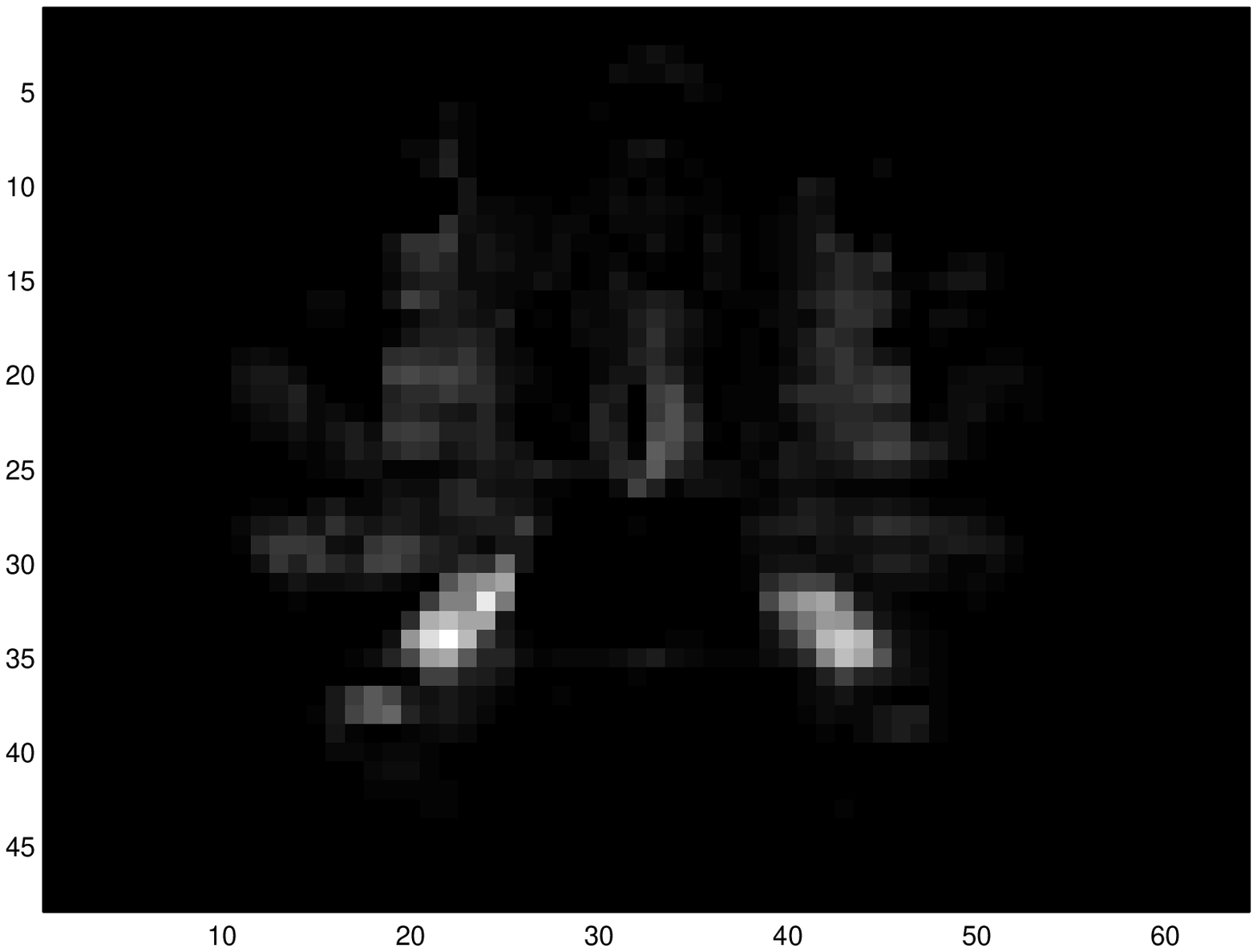}
      \includegraphics[width=1.25cm]{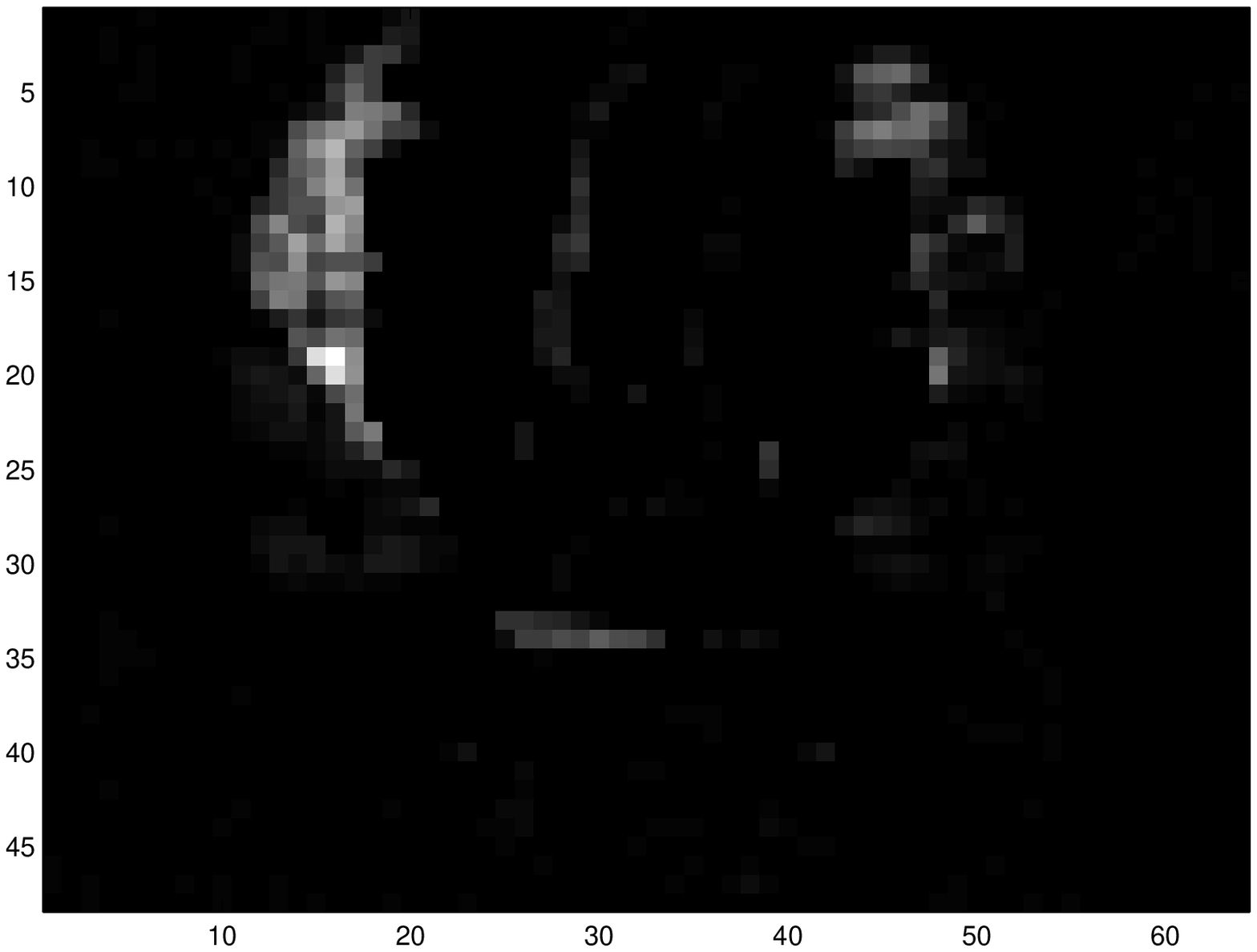}
        \includegraphics[width=1.25cm]{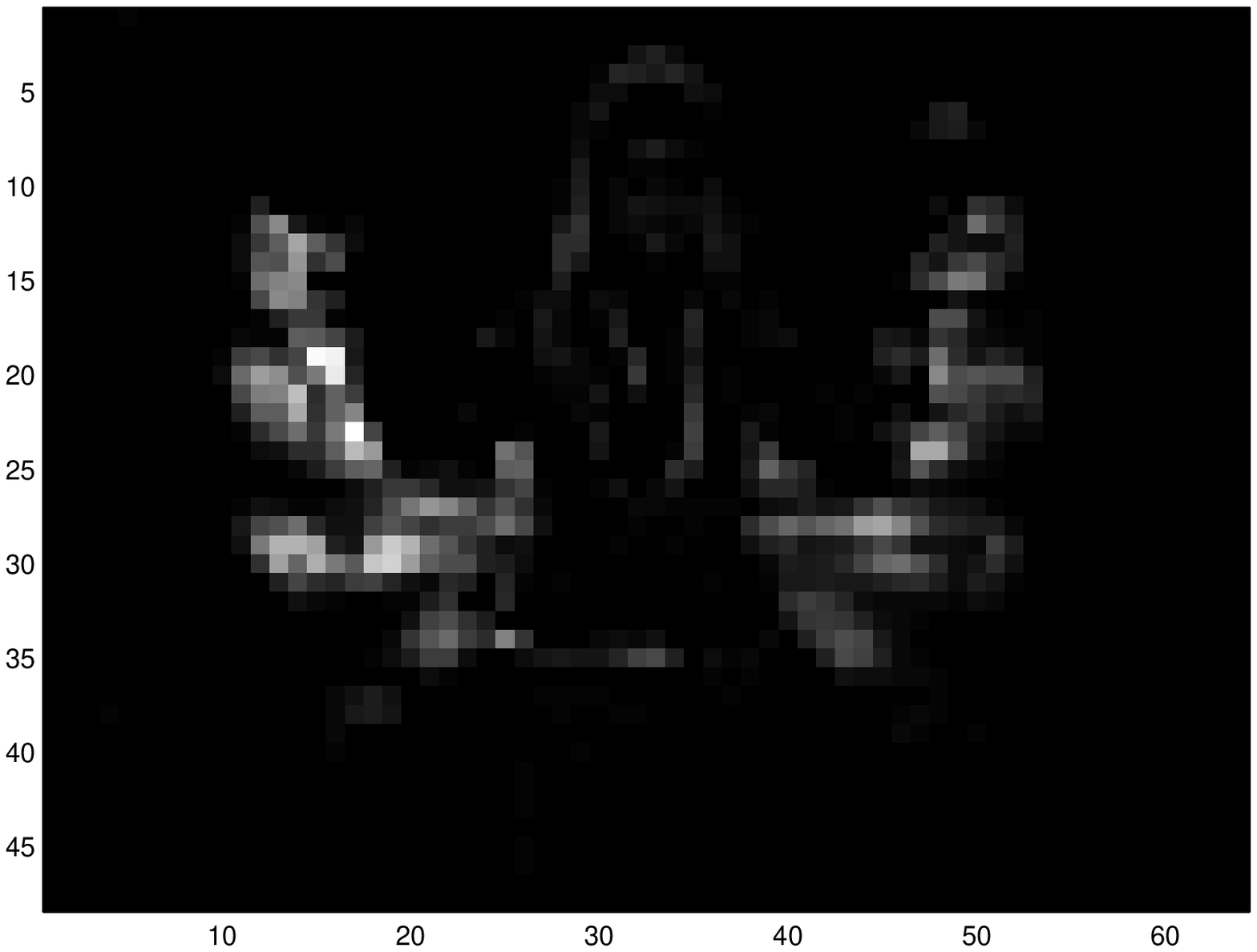}
          \includegraphics[width=1.25cm]{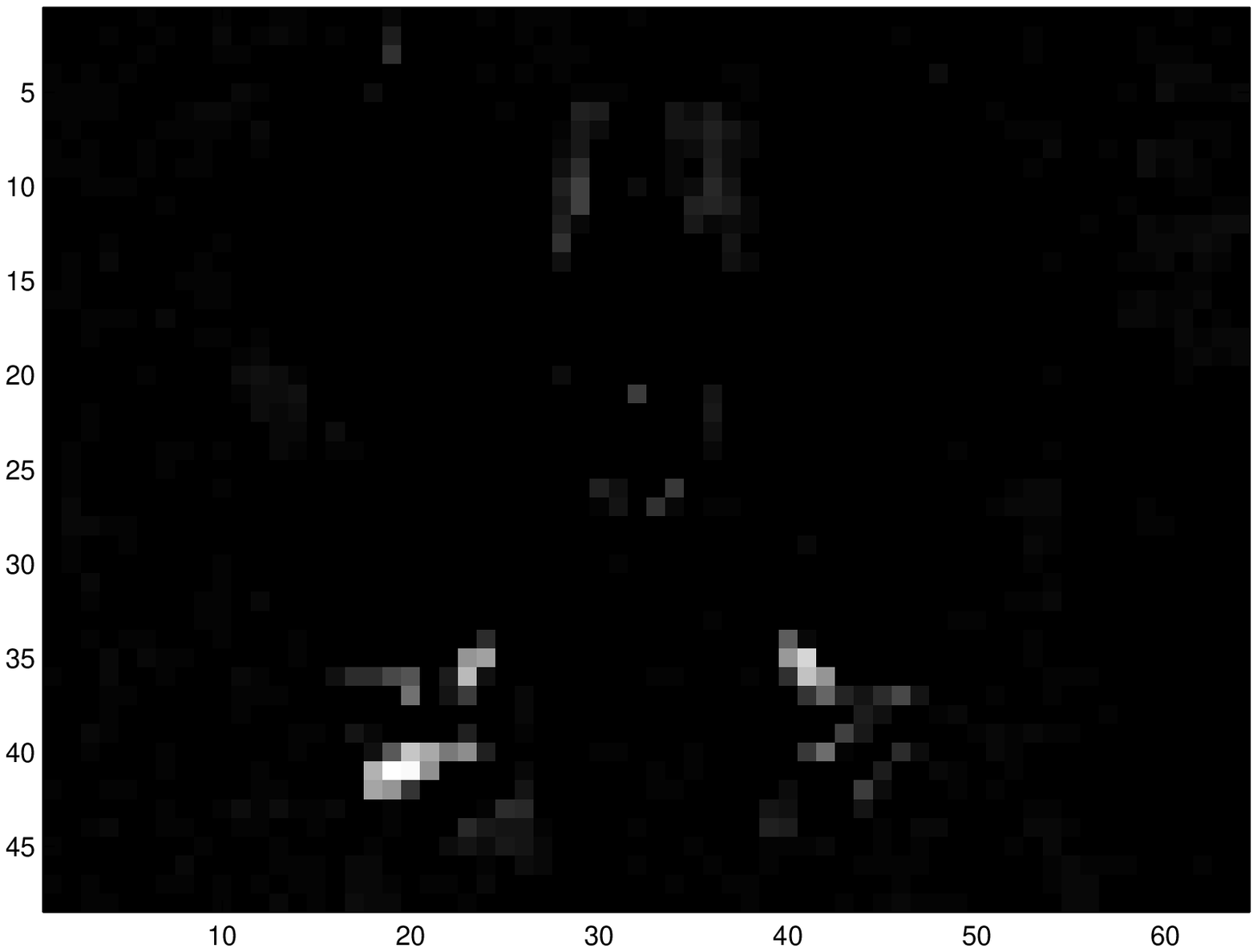}
            \includegraphics[width=1.25cm]{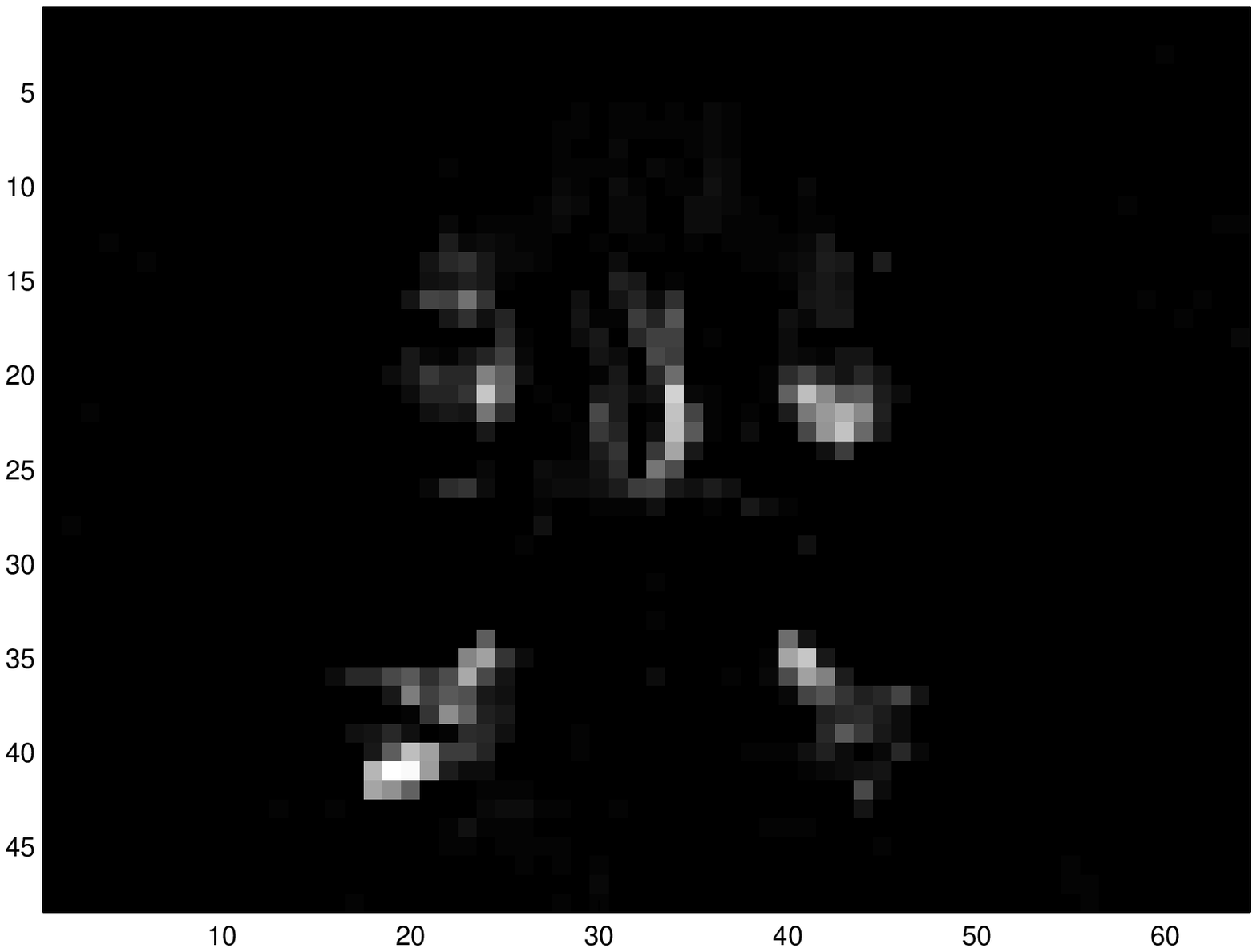}
              \includegraphics[width=1.25cm]{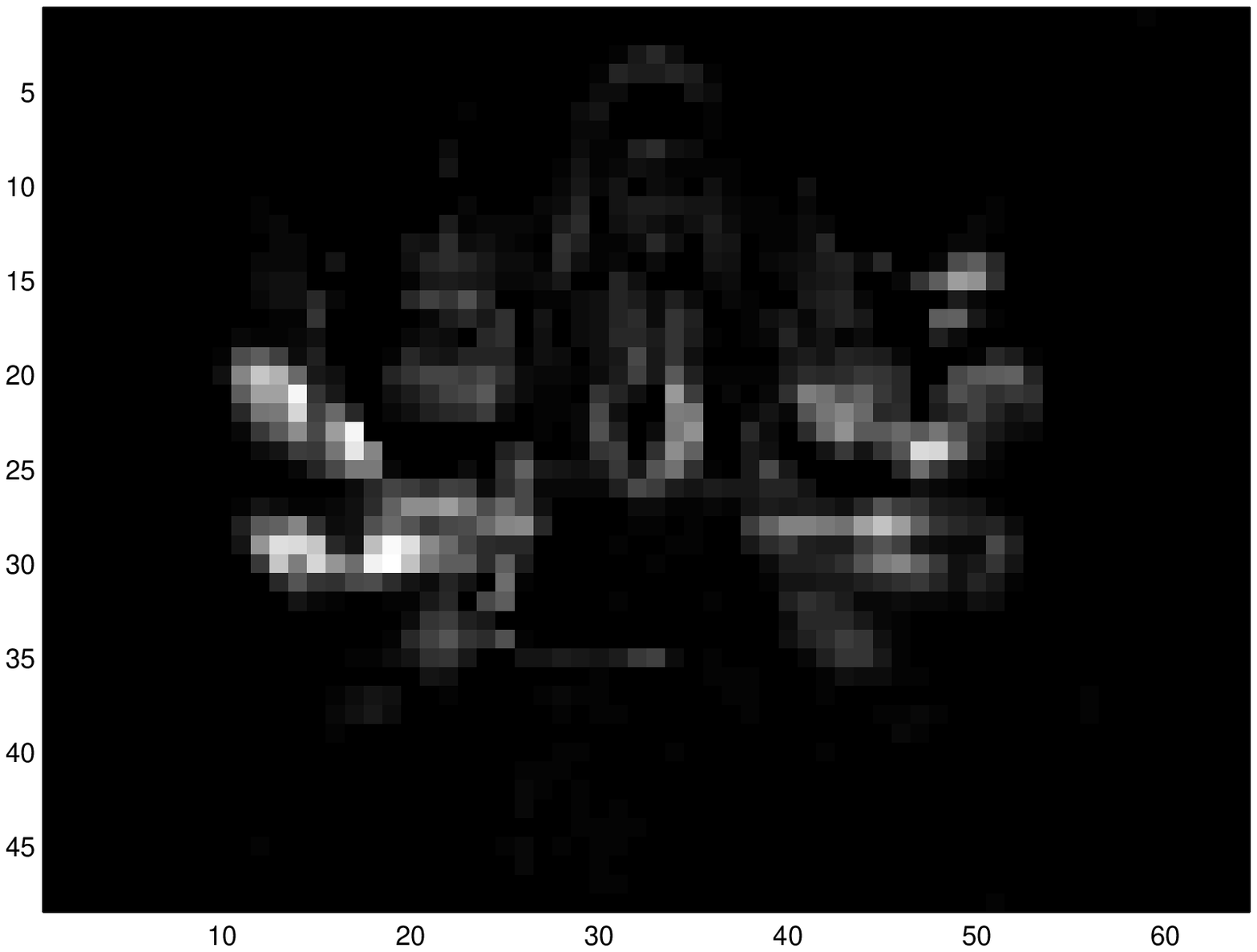}
                \includegraphics[width=1.25cm]{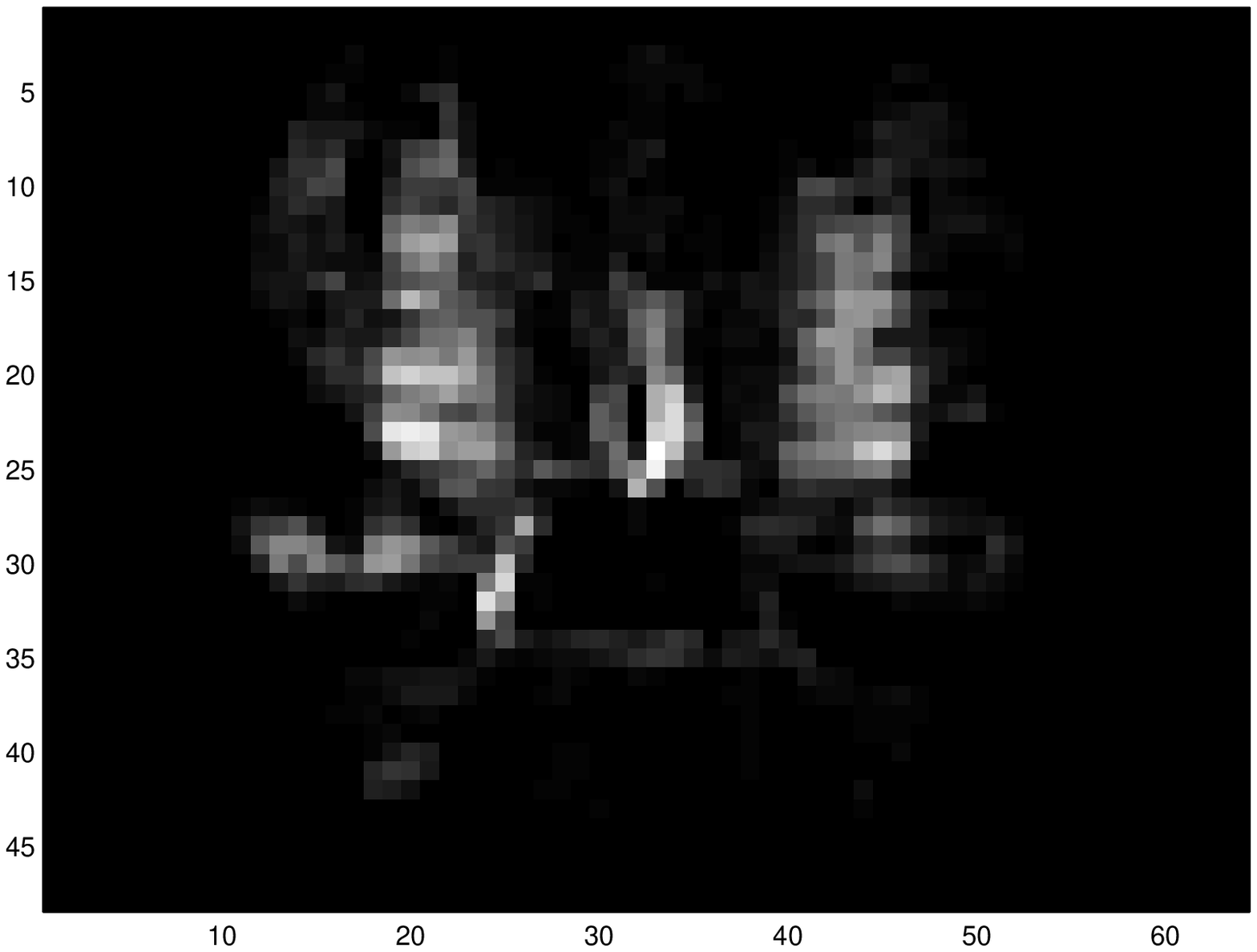}
                \includegraphics[width=1.25cm]{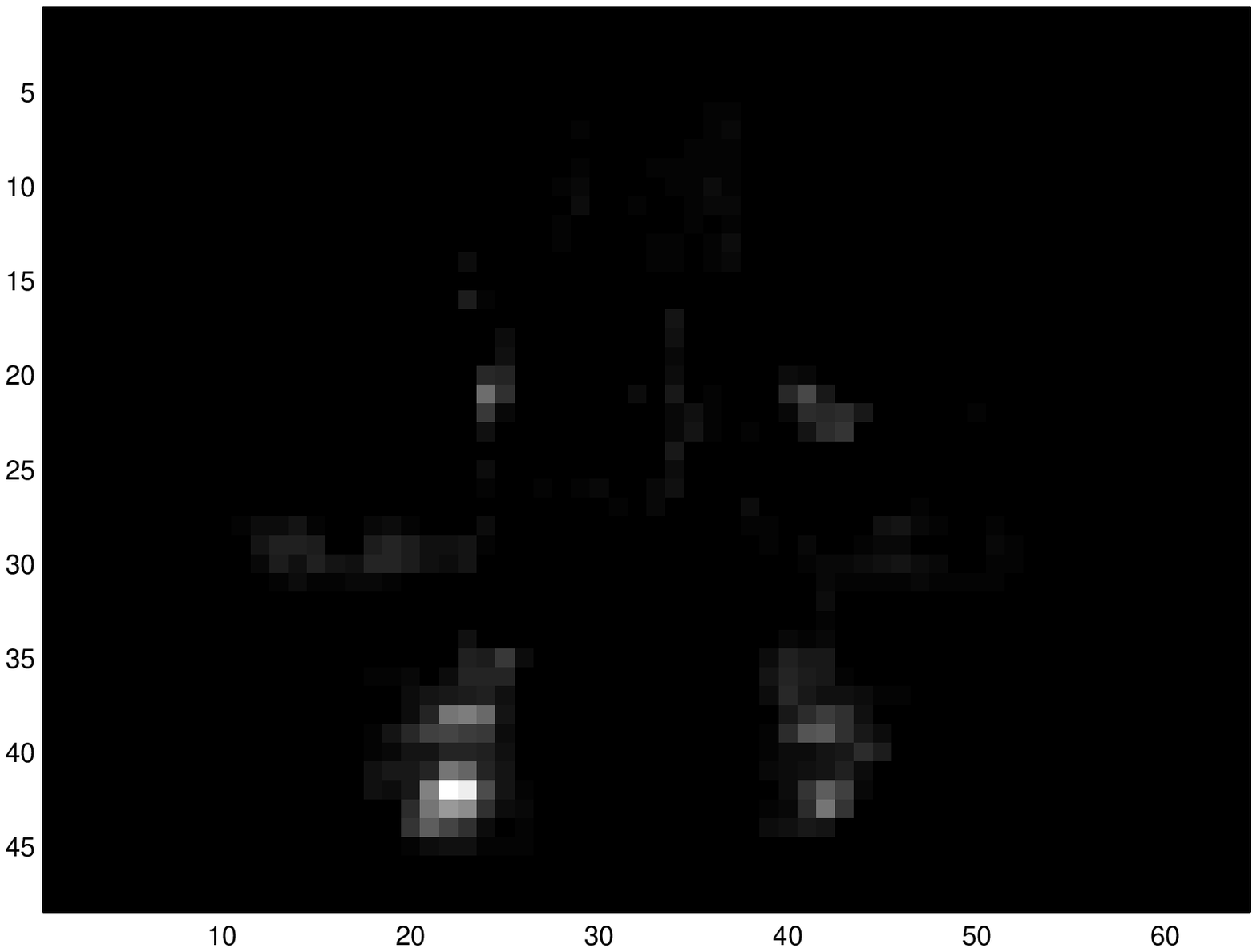}
                \includegraphics[width=1.25cm]{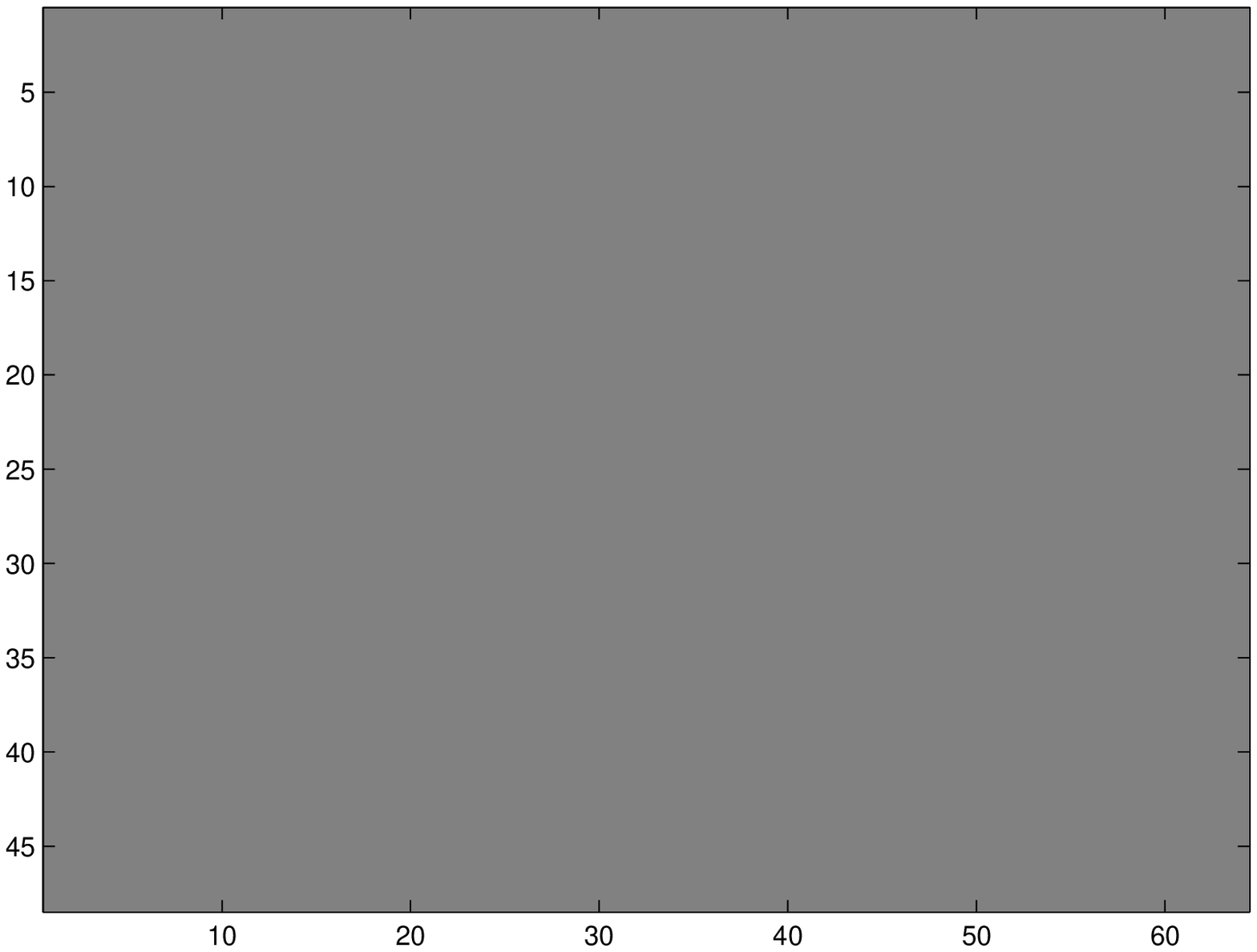}\\
  \caption{Differences of the cumulative sum of reconstructed gestures and the original data.  }\label{fig:reconst3}
\end{figure}

Figure~\ref{fig:reconst3} shows the difference image obtained by subtracting original from reconstructed motion maps for a particular vocabulary (\emph{``helicopter''}). Specifically, image $i, j$ in the array of images, depicts the difference between: the average of motion maps for image $i$, minus the average of motion maps for image $i$ reconstructed with PCA model $j$ (e.g., images in the diagonal show the difference image obtained by subtracting original representations from the reconstruction with the \emph{correct} model). Only differences exceeding the value of $1\times10^{-10}$ are shown in the images.
As expected, gestures reconstructed with the correct PCA model, obtain lower differences than the threshold, while the reconstruction of gestures using other models results in large differences across the whole 2D space.

The main motivation for our recognition technique is the fact that
principal components minimize the reconstruction error when projecting the data into the components' space; it can be show that this is equivalent to finding the directions that maximize the variance of the data, which is the most known derivation of PCA, see e.g.,~\cite{bishop,pca}. Since the PCA model for a gesture is the one that minimizes the average reconstruction error for motion maps belonging to the corresponding video, this model should be the one (among the PCA models for other gestures) that better reconstructs new motion maps belonging to the same gesture. 
Clearly this is not a discriminant classifier, since the PCA model for a gesture is generated independently of the models for other gestures, hence no inter-gesture information is captured by the PCA approach. Nevertheless, our experimental study from Section~\ref{sec:results} reveals that even with this limitation the proposed approach performs better than supervised methods 
that use the bag-of-frames representation. 

\section{Experimental settings}
\label{sec:experiments}

We evaluate the performance of the principal motion components approach in the ChaLearn Gesture Dataset (CGD)~\citep{guyon_cgd}.
CGD comprises $54,000$ different gestures divided into $540$ batches of 100 gestures each, gestures were recorded in RGB and depth video using a Kinect$^{TM}$ camera. 
The data set was divided into development (480 batches), validation (20 batches) and additional batches for evaluation (40 batches, referred to as final batches). Each batch is associated to a different gesture vocabulary, and it contains exactly one video from each gesture in the vocabulary for training and several videos containing sequences of gestures taken from the same vocabulary for testing. Each batch contains 100 gestures, the number of training videos/gestures ranges from 8 to 12, depending on the vocabulary. There are 47 videos for testing in each batch containing sequences from 1 to 5 gestures each; hence, a gesture segmentation method has to be applied before recognition. The number of test gestures in each batch ranges from 88 to 92.
About 20 different users contributed for the generation of gestures and there are about 30 different gesture vocabularies.
See~\citep{guyon_cgd} for a comprehensive description of the CGD. It is important to mention that gesture vocabularies are quite diverse and come from many domains, e.g., see those mentioned in Table~\ref{tab:sampledomains}. 
\begin{table}[bht]
\begin{center}
\caption{A few same vocabularies from the different batches.}\label{tab:sampledomains}
\begin{tabular}{|l|l|l|}
\hline
Referee wrestling signals&Motorcycle signals&Diving signals\\
Surgeon signals&Taxi South Africa&Gang hand signals\\
Tractor operation signals&Chinese numbers&Mudra signals\\\hline
\end{tabular}
\end{center}
\end{table}

The CGD was developed in the context of Chalearn gesture challenge\footnote{http://gesture.chalearn.org/}, an academic competition that focused in the development of gesture recognition systems under the one-shot-learning scenario~\citep{guyon_grc,guyonwdai}. During the challenge, participants had access to the labels of all of the development batches (1-480), although most participants used only twenty batches (1-20) when developing their systems. This can be due to the fact that for those batches additional information was provided by the organizers (e.g., manual segmentation of test videos, hand tracking information, body-part estimates, etc.). Validation data was used by the organizers to provide immediate (on-line) feedback on the performance of participants' methods. Final batches were used to evaluate the performance of the different methods. 
See~\citep{guyonwdai} for more details on the Chalearn gesture challenge.

The evaluation measure used in the challenge was the Levenshtein's distance (normalized by the length of the truth labeling), which accounts for the number of edits that must be performed for taking a sequence of predictions into the ground truth labeling for a gesture. In the next section we report experimental results on the CGD benchmark to evaluate the effectiveness of the principal motion components approach.

\section{Experimental results}
\label{sec:results}

In this section we report results from experiments that aim at evaluating different aspects of the proposed approach. First, we evaluate the performance of our method in the whole CGD collection. Next, we evaluate the method under different parameter settings. Then  
we compare the proposed approach to a number of related techniques we implemented. Finally, we compare the performance of the principal motion components technique to other methods developed in the context of Chalearn's  gesture challenge.

As explained previously, videos must be segmented in order to isolate gestures prior to recognition. We report results of experiments using both: manually segmented (batches 01-20 for development and validation only) and automatically segmented (all of the batches) videos. For automatic segmentation we used a simple method based on dynamic time warping, which is also based on the motion maps representation (a time ordered version at a very coarse resolution). 
This method was provided by the organizers of the Chalearn gesture challenge; it is  publicly available from the challenge website. 
\subsection{Performance over the whole collection}

In a first experiment we applied the principal motion components approach to the whole GRC database of $54,000$ gestures using both RGB and depth video. Results in terms of the Levenshtein score are shown in Table~\ref{tab:r1}. For this experiment all of the videos were automatically segmented. The translation gap was set to $\tau=5$ pixels, the scale for image downsizing  was fixed to $\gamma=0.1$, while the number of principal components was set to $c=10$; our choices were based on the results obtained in a preliminary study, see section~\ref{sec:parsel}. 
\begin{table}[!htb]
\centering
\begin{center}
\caption{Average (and standard deviation) of performance obtained by the proposed approach on the development ($48,000$ gestures), validation ($2,000$ gestures) and final batches ($4,000$ gestures). 
}\label{tab:r1}
\begin{tabular}{|l|l|l|}\hline
\textbf{Data set / Type}& \textbf{RGB} & \textbf{DEPTH}\\\hline
Devel01-480&  0.4079 (0.2387) & 0.4103 (0.2068) \\
Valid01-20& 0.3178 (0.2030)&  0.3189 (0.1891)\\
Final01-20& 0.2747 (0.1842)& 0.2641 (0.1971)\\
Final21-40& 0.2124 (0.1404)&0.2263 (0.1362)\\\hline
\end{tabular}
\end{center}
\end{table}
%

The performance of our method in the 480 development batches was worst than that obtained in the final and valid batches. This can be due to the difference in number of batches and the diversity of their vocabularies. In development batches, results using depth video are slightly worse than those obtained with RGB video, nevertheless, the difference in performance is not statistically significant according to a two-sample T-test ($p-value= 0.8713$). The corresponding differences for the validation ($p-value= 0.9879$) and final ($p-value= 0.8607$) batches were not statistically significant neither. Thus, can conclude that the proposed method performs similarly, regardless of the type of information used: either RGB or depth video. This is advantageous as we do not need a Kinect sensor to achieve acceptable recognition performance with our method;  one should note, however, that the standard deviation of performance is lower for depth video (in the 480 batches and for validation batches), hence, when available it would  be preferable to use it. 

The proposed approach took an average of $41.23$ seconds to entirely process a batch\footnote{Experiments were performed in a workstation with Intel$^R$ Core$^{TM}$i7-2600 CPU at 3.4 GHz and 8GB in RAM.} (i.e., training the PCA models from the training videos and labeling all of the test videos, the time includes feature extraction and gesture segmentation).  This means that a test video is processed in approximately one second, which makes evident the efficiency of our proposed method and can be used in real time applications.

The performance of our method in validation and final batches followed the same behavior as in development batches, although it is better. In fact the performance of our method on validation and final batches is competitive with methods proposed by participants of the GRC. For example, the results on the final batches 1-20 from Table~\ref{tab:r1} would be ranked $9^{th}$ 
for the first round of the challenge, whereas for the final batches see 21-40 would be ranked $7^{th}$, see Section~\ref{sec:compar_res}. One should note, however, that this method was not designed to handle all cases (e.g., static gestures). Competitive methods also used some handshape features to recognize static gestures, and that is beyond the scope of this paper.

We now evaluate the impact of gesture segmentation in the performance of the principal motion components technique. Table~\ref{tab:r2} compares the performance obtained in batches 1-20 for development and validation data when using manually segmented gestures and the automatic segmentation approach.  As expected, using manual segmentation improves the performance of our approach, nevertheless, the achieved improvements are modest. In fact, statistical tests did not reveal that the differences were statistically significant for both modalities (RGB and depth video) and batches (development and validation). Therefore, we can conclude that we can apply the principal motion approach using automated methods for gesture segmentation and still obtain competitive performance.
\begin{table}[bht]
\begin{center}
\caption{Performance of our method on batches 1-20 for the development and validation data sets using manual and automatic segmentation.}\label{tab:r2}
\begin{tabular}{|c|p{1.5cm}|p{1.5cm}|p{1.5cm}|p{1.5cm}|}
\hline
\emph{Segmentation}&\multicolumn{2}{c|}{\textbf{Manual}}&\multicolumn{2}{c|}{\textbf{Automatic}}\\\hline
\emph{Data set / Type}& \emph{RGB} & \emph{DEPTH}& \emph{RGB} & \emph{DEPTH}\\\hline
Devel01-20&0.2944&0.2741&0.3022&0.3016\\
Valid01-20&0.3151&0.3134&0.3178&0.3189\\
\hline
\end{tabular}
\end{center}
\end{table}

\subsection{Performance under different parameter settings}
\label{sec:parsel}
Recall the only parameters of the proposed formulation are $\gamma$ (the scale for downsizing the image, see Section~\ref{sec:mothis}), which is related to the size of the patches to generate motion maps, and $c$, the number of principal components used to generate PCA models, see Section~\ref{sec:recogpca}. In a third experiment we aimed to determine to what extent varying the values of such parameters affect the performance of the proposed approach. We proceeded by fixing the value of a parameter and then we evaluate the performance of our approach when varying the second parameter.

\begin{figure}[!htb]
  \includegraphics[width=7.5cm]{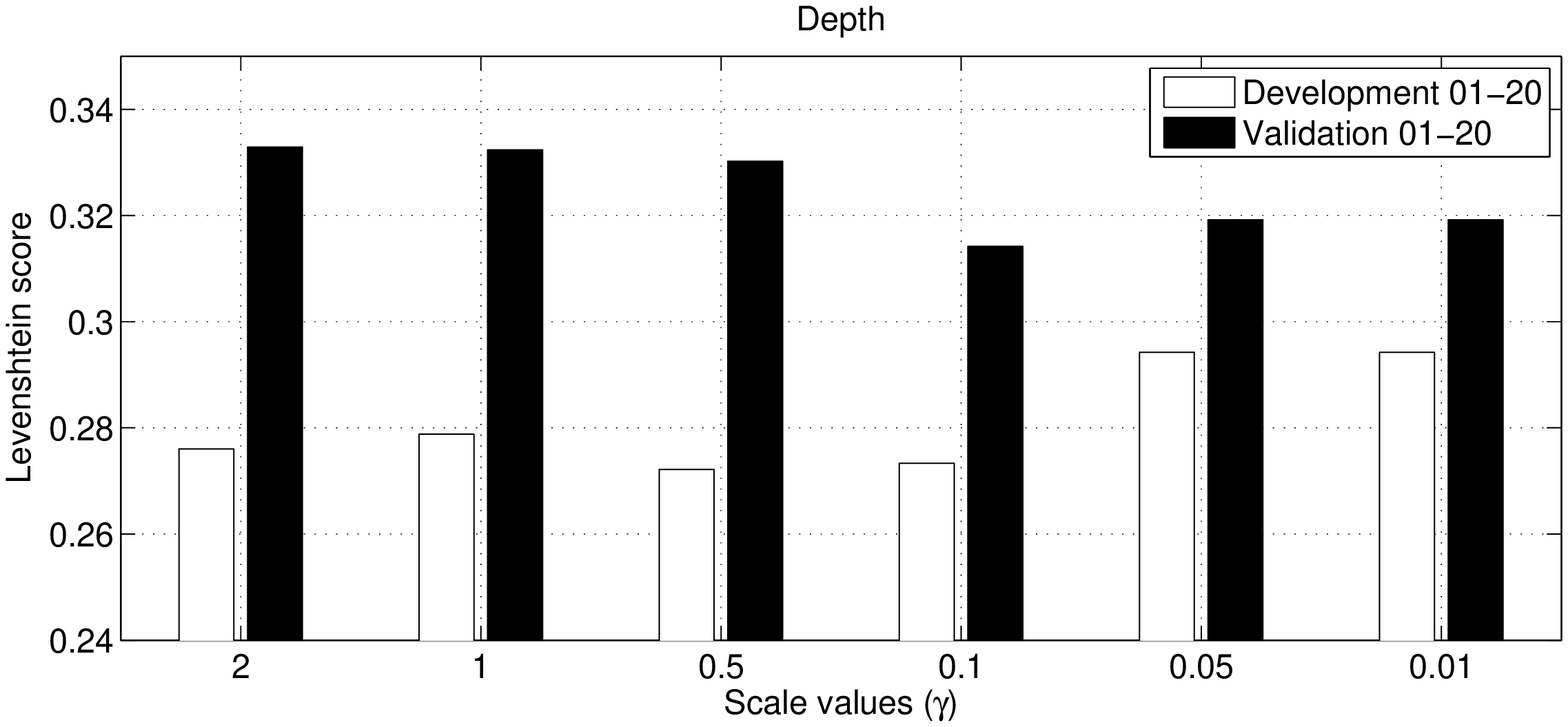}
  \includegraphics[width=7.5cm]{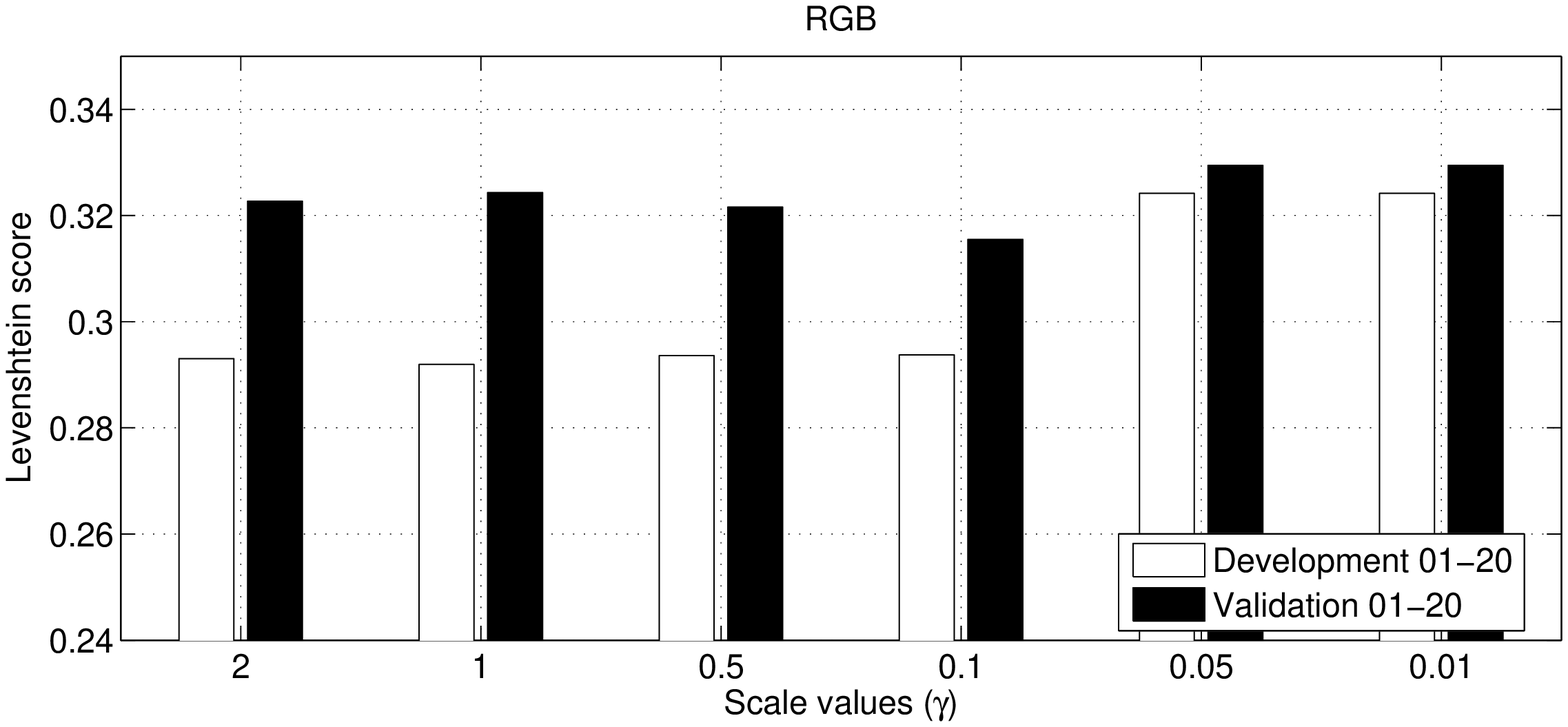}
 \includegraphics[width=7.5cm]{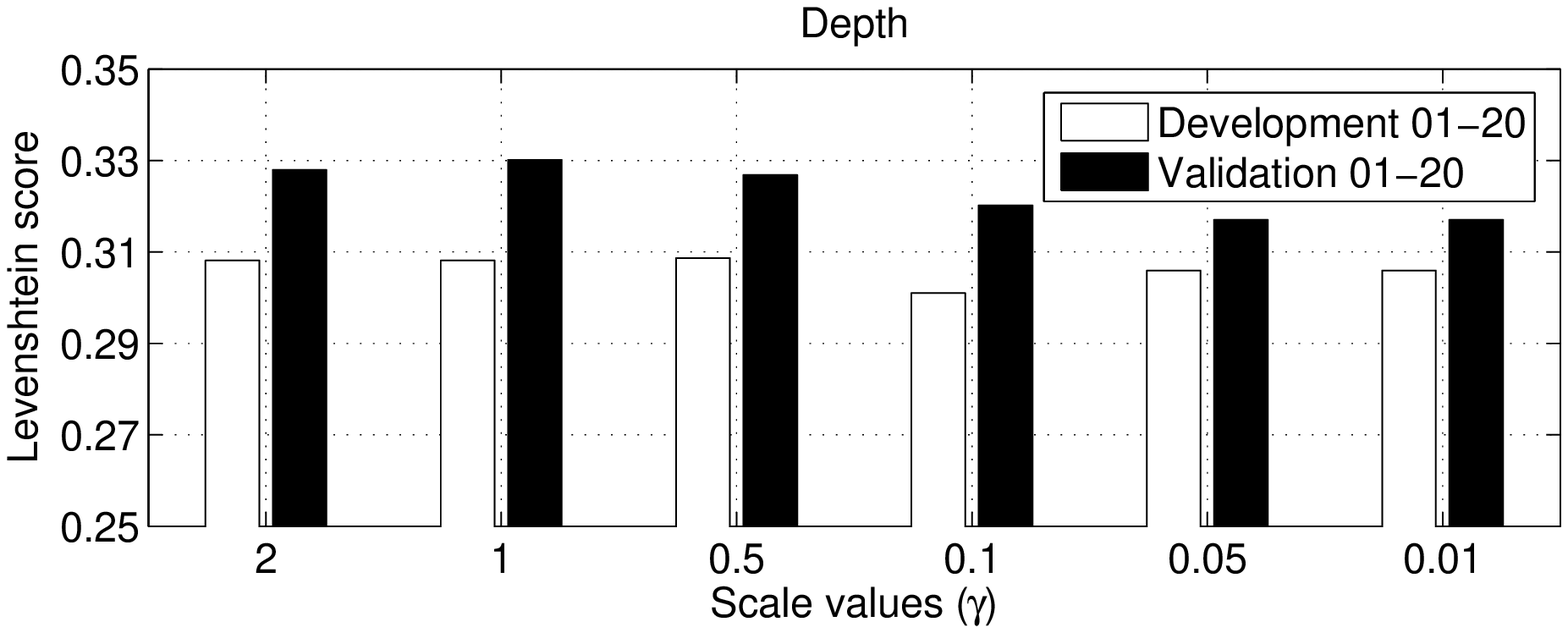}
   \includegraphics[width=7.5cm]{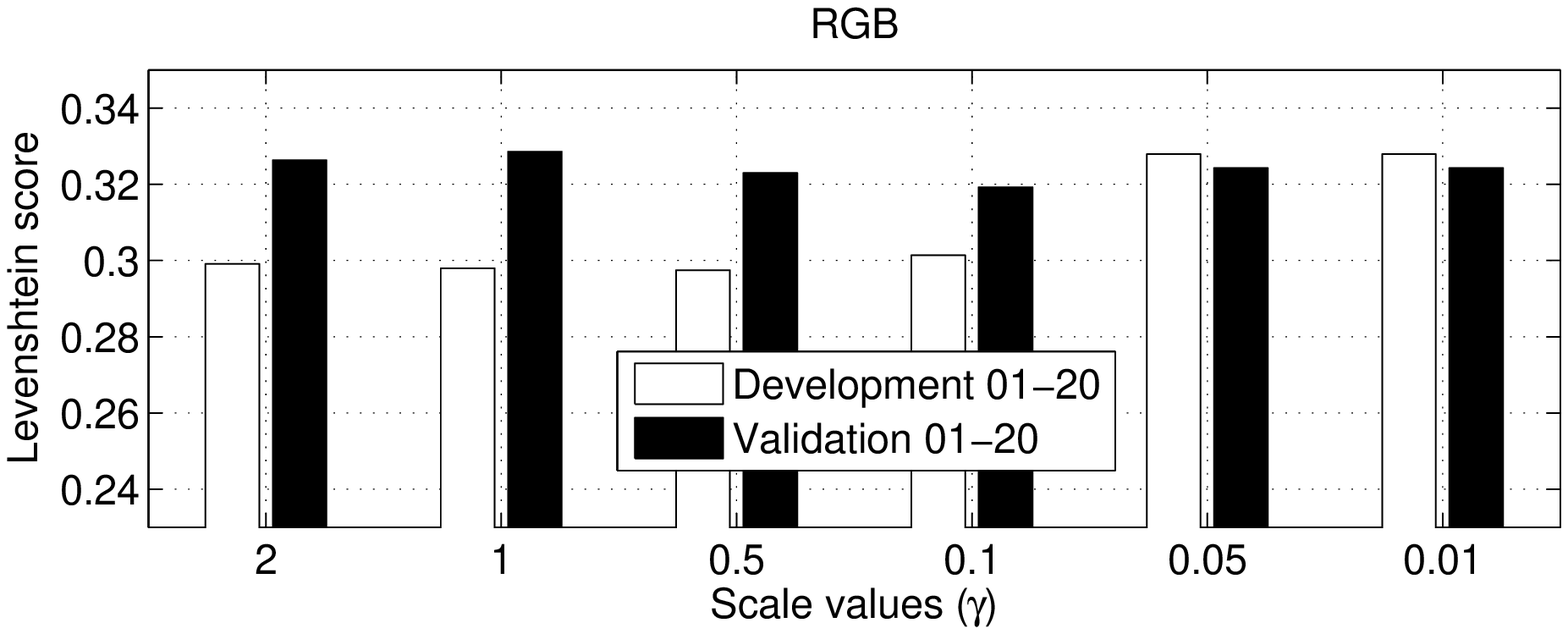}
  \caption{Average performance of principal motion components for different scale values. For the top plots
  manual segmentation was used, while for the bottom ones automatic segmentation was performed. 
  }\label{fig:scale}
\end{figure}

We start by analyzing the results in terms of the scale parameter ($\gamma$). For this experiment we fixed the number of principal components to $c = 10$. Results of this experiment are shown in Figure~\ref{fig:scale}. It can be seen that for both modalities there is not too much variation in the performance of the method for the different values we consider. This is due in part to the region growing preprocessing described in Section~\ref{sec:mothis}. The best results were obtained when $\gamma=\{0.5, 0.1\}$. Lower values of $\gamma$ are preferred because the dimensionality of the motion maps is reduced and the proposed approach can be applied faster. Besides, the smaller the value of $\gamma$ the larger the size of the patches for the motion maps and the more robust is the approach to variations in the position of the user with respect to the camera.  For instance, for $\gamma=0.1$ the dimensionality of the motion maps is of  $192$, the corresponding size of the patches is $\approx 15\times27$. Nevertheless, it can be seen from Figure~\ref{fig:scale} that for smaller values than $\gamma=0.1$ the performance of principal motion components is worse.

\begin{figure}[!htb]
  \includegraphics[width=15cm]{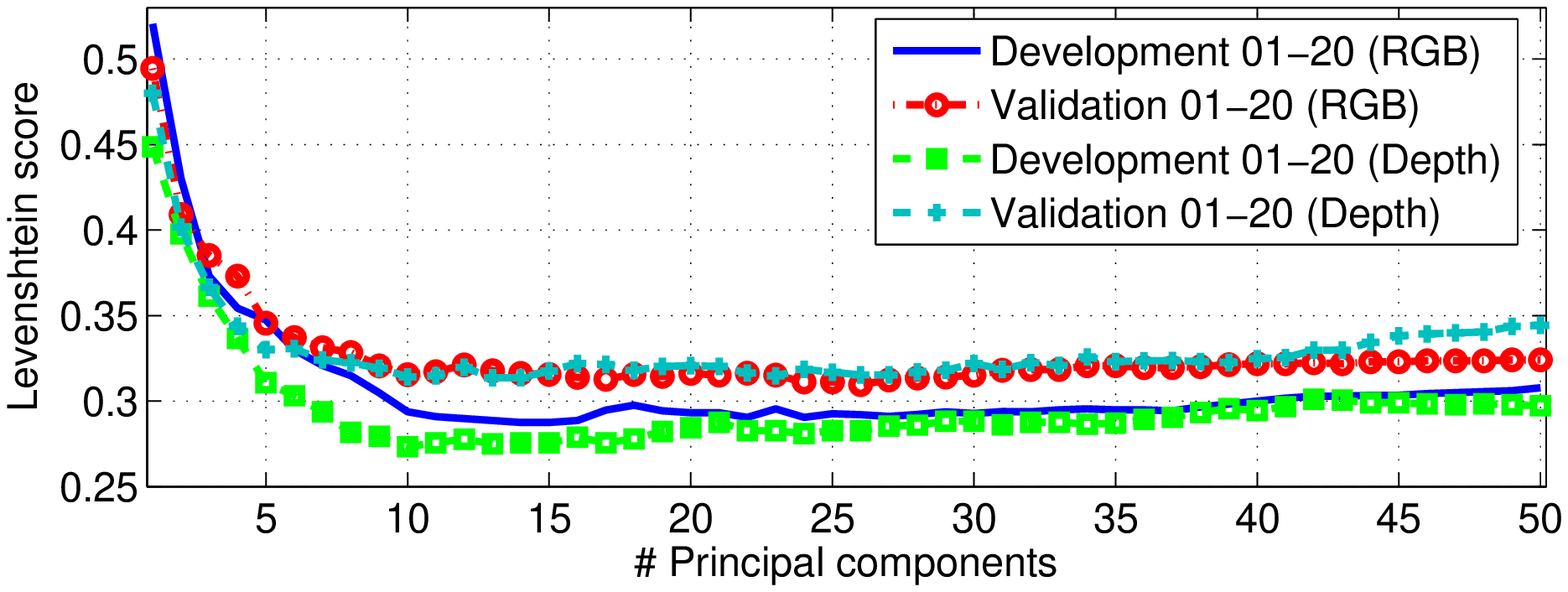}
    \includegraphics[width=15cm]{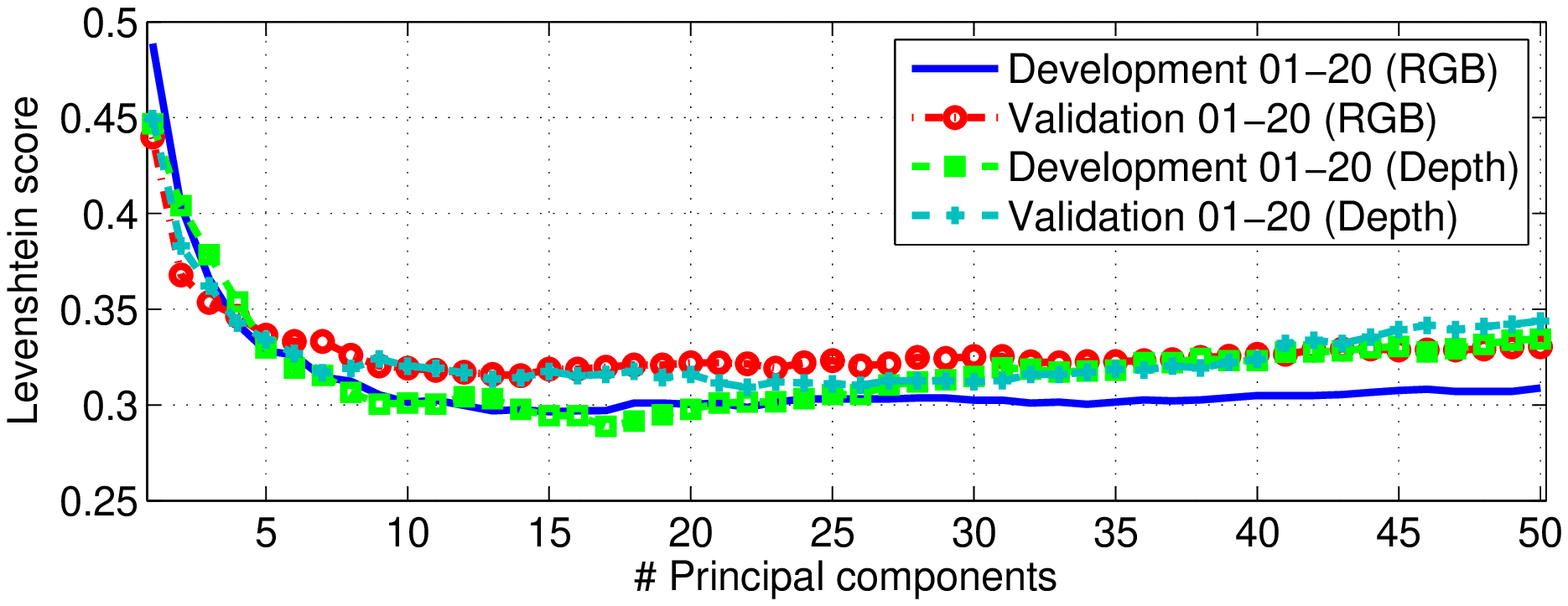}
  \caption{Average and standard deviation of the performance obtained by the proposed approach for different number of principal components.}\label{fig:components}
\end{figure}

For analyzing the influence of the number of components on the proposed technique we fixed the value of the scale to $\gamma=0.1$ and varied the number of principal components when building PCA models, experimental results are shown in Figure~\ref{fig:components}. It can be seen from these plots that, in general, the performance of principal motion components is poor when using few components, $c \in \{1, \ldots, 5\}$, for all the combinations of batches/modalities. The best performance for all of the batches/modalities was obtained when using a number of components $c \in \{10, \ldots, 15\}$; the performance is somewhat stable for $c \in \{10, \ldots, 25\}$ and then it decreases considerably.  This result may suggests the best value for $c$ is related to the number of gestures in the vocabularies ($\{8, \ldots, 12\}$). Actually, the average vocabulary lengths for development and validation batches are 9.7 and 9.5, respectively.   Nevertheless, we did not find significant correlation between the best value for $c$ and the size of the vocabulary ($\rho=-0.0529$).

We also evaluated the correlation between the best value of $c$ and the average and standard deviation of the length of training gestures, the minimum and maximum duration, the entropy on the duration of training gestures among other statistics. However, we
did not find a statistically significant correlation value either.
Thus, other aspects that have to do with the difficulty of vocabularies may have an impact into the optimal value for $c$. In this regard, Table~\ref{tab:caract_opt} shows information of the performance on each batch when using the optimal number of principal components for each of the development and validation batches (manual segmentation and RGB video were used).
\begin{table}[bht]
\begin{center}
\caption{Sorted results for batches 1-20 of the development and validation data sets along with some characteristics of each batch. Column \textbf{B} shows the id of the batch (either development, \emph{d}, or validation, \emph{v}) and its number. Column \textbf{M} indicates whether the body of the user moves significantly (\emph{D}) or not (\emph{S}) when performing the gesture. Column \textbf{T} specifies the type of the gesture, which can be either static (\emph{S}) or dynamic (\emph{D}). Column \textbf{c} indicates the number of principal components used for the corresponding batch.  Column \textbf{LS} shows the obtained Levenshtein score and column \textbf{V}, indicates the name of the vocabulary, see~\cite{guyon_cgd}.}\label{tab:caract_opt}
\scriptsize{
\begin{tabular}{|l|l|l|l|l|l|l|l|l|l|l|l|}
\hline
\multicolumn{6}{|c|}{\textbf{Devel01-20}}&\multicolumn{6}{|c|}{\textbf{Valid01-20}}\\\hline
\textbf{B}&\textbf{M}&\textbf{T}&\textbf{c}&\textbf{LS}&\textbf{V}&\textbf{B}&\textbf{M}&\textbf{T}&\textbf{c}&\textbf{LS}&\textbf{V}\\\hline
d05&S&D&4&1.09&Gestuno-Disaster&v02&D&D&5&1.10&Helicopter\\
d08&S&D&19&2.25&Gestuno-Topography&v16&S&S&6&3.26&Referee-Volleyball2\\
d01&D&D&17&4.44&Canada-Aviation&v05&D&D&7&3.37&Tractor-Operation\\
d13&S&S&26&6.82&Crane-Hand&v17&S&S&8&8.70&Body-Language-Dom.\\
d04&D&S&12&8.89&Diving2&v10&S&D&8&11.96&Pantomime-Objects\\
d09&D&S&3&12.09&Referee-Volleyball1&v11&S&D&5&14.44&McNeill-Gesticulation2\\
d14&S&D&10&16.85&Diving&v04&S&S&10&16.30&Swat-Hand2\\
d07&D&S&12&19.57&Referee-Volleyball1&v13&S&D&7&19.78&Gestuno-Small-Animals\\
d17&S&S&10&19.78&Gang-HandS-ignals2 &v06&D&S&32&21.11&Dance-Aerobics\\
d16&S&D&44&21.74&Gestuno-Landscape&v20&D&D&5&24.44&Canada-Aviation2\\
d20&S&S&7&22.99&Diving1&v12&S&D&22&31.11&Gestuno-Colors\\
d02&D&S&12&24.18&Referee-Wrestling1 &v07 &S&S&23&33.33&Referee-Wrestling2\\
d12&S&S&4&26.67&Italian-Gestures&v19&S&S&39&34.44&Taxi-SouthAfrica\\
d15&S&S&8&29.35&Swat-Hand1&v01&S&D&24&36.36&Motorcycle\\
d11&S&S&8&30.43&Music-Notes&v15&S&D&7&37.08&Italian-Gestures\\
d06&S&D&10&32.22&Diving3&v03&D&D&10&46.67&Diving2\\
d18&S&S&22&34.44&Taxi-SouthAfrica&v18&S&S&28&53.26&Music-Notes\\
d19&S&S&34&47.25&Mudra2&v08&D&D&11&54.35&Action-Objects\\
d10&S&S&11&48.35&Surgeon&v14&S&S&11&56.67&Mudra1\\
d03&S&S&8&60.87&Gang-Hand1 &v09&S&S&10&67.42&Chinese-Numbers\\\hline
\end{tabular}}
\end{center}
\end{table}

Along with the performance obtained in each batch it is shown the optimal value of $c$ and some characteristics about the dynamism of gestures in batches.   Interestingly, a few principal components are enough to obtain outstanding performance for some batches (e.g., \emph{``Referee-Volleyball1''} (3), \emph{``Gestuno-disaster''} (4), and \emph{``Helicopter''} (5)), while a large value for $c$ is used for some batches and yet the performance is poor (e.g., \emph{``Taxi-SouthAfrica''} (39),  and \emph{``Mudra2''} (34)). 
It seems that easier vocabularies (too much motion, movement across the whole image, small inter-class similarity) require of less components than difficult ones (little motion, motion happening in small regions of the image, large inter-class similarity). Although is not easy to define what an easy/difficult vocabulary is.

Other interesting findings can be drawn from the results of this experiment. First, it can be seen that the principal motion components approach is very effective for some gestures. For example, performance similar to that of humans was obtained for \emph{``Helicopter'',} \emph{``Gestuno-disaster'', ``Gestuno-topography'', ``Tractor-Operation''} and \emph{``Canada-Aviation''} vocabularies. These are highly dynamic gestures where motion happens in different regions of the image, thus proposed approach can effectively capture the differences among gestures in the same vocabulary.  In general, acceptable performance was obtained with the proposed approach when either the gesture is dynamic or the body of the user moves significantly when performing the gesture. The worst results were obtained when facing static gestures and users remained static when performed the gesture. This is a somewhat expected result as our approach attempts to exploit motion information. 

Table~\ref{tab:optimum} shows the average performance one one would obtain when selecting the optimal value for $c$ in each batch. The (hypothetical) relative improvements over the results reported in Table~\ref{tab:r2} range from $7.2 \%$ to $20 \%$. Hence, it is  worth pursuing research on methods for selecting the number of principal components for each particular batch or gesture. Although one should note that the raw differences in performance are small: an improvement of $20.1\%$ (RGB/Devel/MANUAL) corresponds to a raw difference of $\approx 0.06$ in Levenshtein score.  Development batches have a larger room for improvement than validation ones, the result is consistent with previous ones.
\begin{table}[bh]
\begin{center}
\caption{Optimum performance that can be obtained with principal motion components when selecting the optimal value for $c$ in each batch. It is shown between parentheses the relative improvement over the corresponding results from Table~\ref{tab:r2} (i.e., when using $c=10$ for all batches). }\label{tab:optimum}
\footnotesize{
\begin{tabular}{|c|l|l|l|l|}
\hline
\emph{Segmentation}&\multicolumn{2}{c|}{\textbf{Manual}}&\multicolumn{2}{c|}{\textbf{Automatic}}\\\hline
\emph{Data set / Type}& \emph{RGB} & \emph{DEPTH}& \emph{RGB} & \emph{DEPTH}\\\hline
Devel01-20$^*$&0.2351 (20.1\%)&0.2351 (14.2 \%)&0.2749 (9.1\%)&0.2635 (12.6\%)\\
Valid01-20$^*$&0.2876 (8.7\%)&0.2876 (8.23\%)&0.2949 (7.2\%)&0.2832 (11.1\%)\\
\hline
\end{tabular}}
\end{center}
\end{table}

Summarizing, the principal motion components approach is rather robust to parameter selection. The scale parameter set to $\gamma=0.1$ achieved the best results for most of the configurations we evaluated. Although, other values obtained competitive performance as well. Selecting the number of principal components remains a difficult challenge, yet acceptable performance can be obtained by fixing $c=10$. Finally, we showed evidence suggesting the principal motion components method is particularly well suited to vocabularies involving a lot of motion, and when motion happens in different locations of the image. 

\subsection{Comparison with alternative methods}
\label{sec:compar_res}
We now compare the performance of the principal motion approach to that obtained with alternative methods to solve the same one-shot learning problem. First we compare the performance of principal motion components to that of other techniques that are based on similar ideas/features. Next we compare the performance of the proposed technique to that obtained with other methods that were proposed during the Chalearn gesture challenge~\citep{guyon_grc,guyonwdai}.
\begin{table}[bh]
\begin{center}
\caption{Description of the alternative methods we implemented for one-shot gesture recognition. }\label{tab:comparisonimps}
\begin{tabular}{|l|l|l|}
\hline
\textbf{ID}&\textbf{Representation}&\textbf{Recog.}\\\hline
PMC&Motion maps& PCR\\
HOG-I&HOG features from frames & PCR\\
HOG-M&HOG features from difference of frames & PCR\\
HOF-I&HOF features from frames & PCR\\
HOF-M&HOF features from difference of frames & PCR\\
STIP-F&STIP-HOF features  & PCR\\
STIP-H&STIP-HOG features  & PCR\\
STIP-HF&STIP-HOG+HOF features  & PCR\\\hline
PMC-SVM& Motion maps & SVM\\
HOG-SVM& HOG features from difference frames & SVM\\
HOF-SVM& HOG features from difference frames & SVM\\\hline
STIP-BOW&STIP-HOG+HOF bag-of-features & KNN\\
MHI&Motion history image& TM\\
SMHI&Static-motion history image& TM\\\hline
\end{tabular}
\end{center}
\end{table}

For the first comparison we implemented the methods described in Table~\ref{tab:comparisonimps}. The goal of this comparison is assessing whether using different features to represent the video, under the bag-of-frames formulation,
could improve the performance of the one based on motion maps. We extracted the following (state-of-the-art) features widely used in computer vision: histograms of oriented gradients (HOG)~\citep{HoG}; histograms of oriented optical flow (HOF)~\citep{hof}; space-time interest points with 3D HOG and HOF features~\citep{stipf}; and motion history images~\citep{MHI}. 2D HOG and HOF features were extracted from the frames themselves (HOG-I, HOF-I) and from difference images (HOG-M, HOF-M). For STIP-based features we tried HOG-only, HOF-only and HOG+HOG 3D representations~\citep{stipf}. The variants of HOG, HOF and STIP-based features were represented under the bag-of-frames representation. Additionally, two variants of motion history images were implemented: the standard approach (MHI)~\citep{MHI}, and another version that accounted for non-motion (SMHI). The latter variant aimed to be helpful for highly static gestures.

\begin{figure}[!htb]
  \includegraphics[width=15cm]{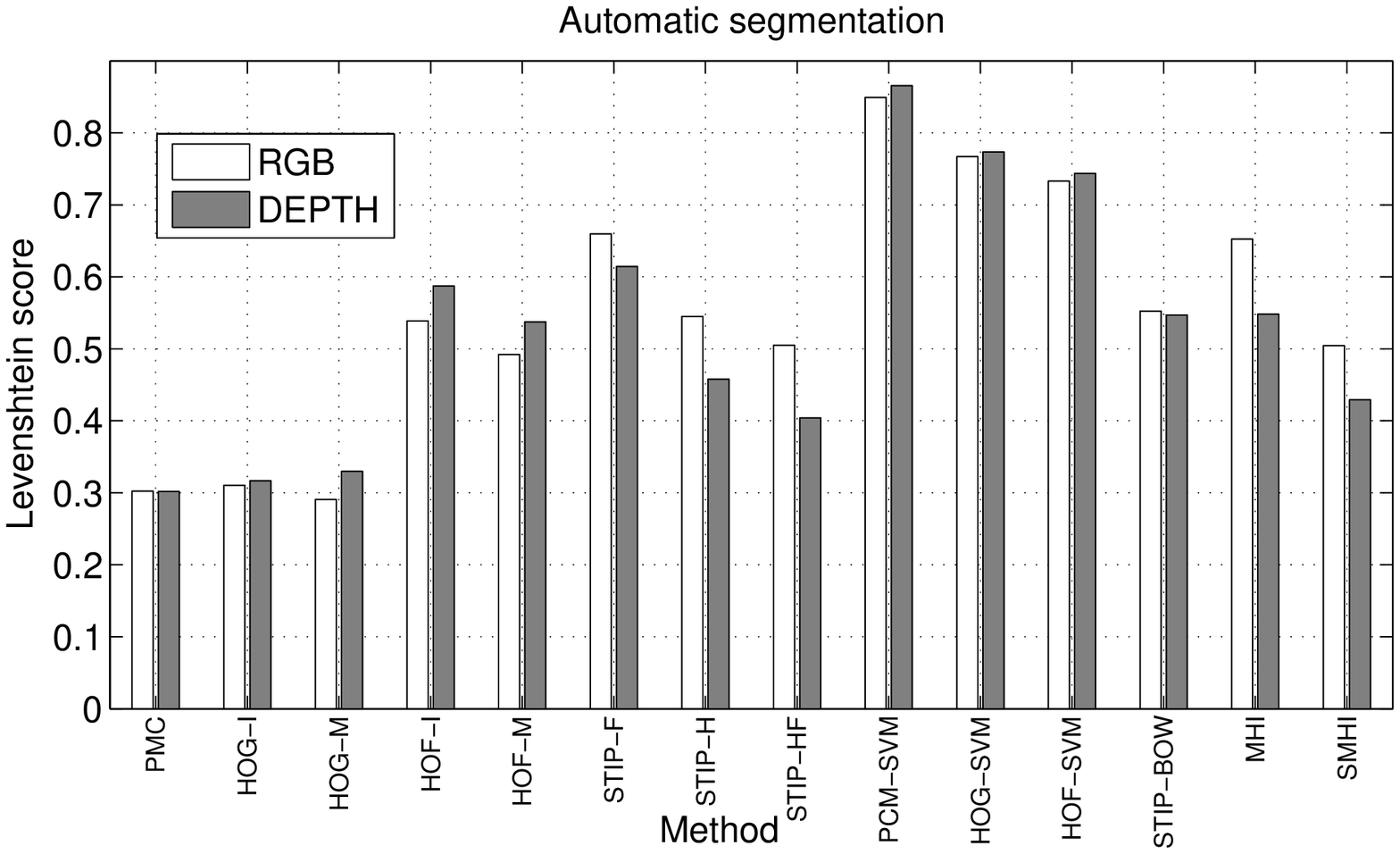}
    \includegraphics[width=15cm]{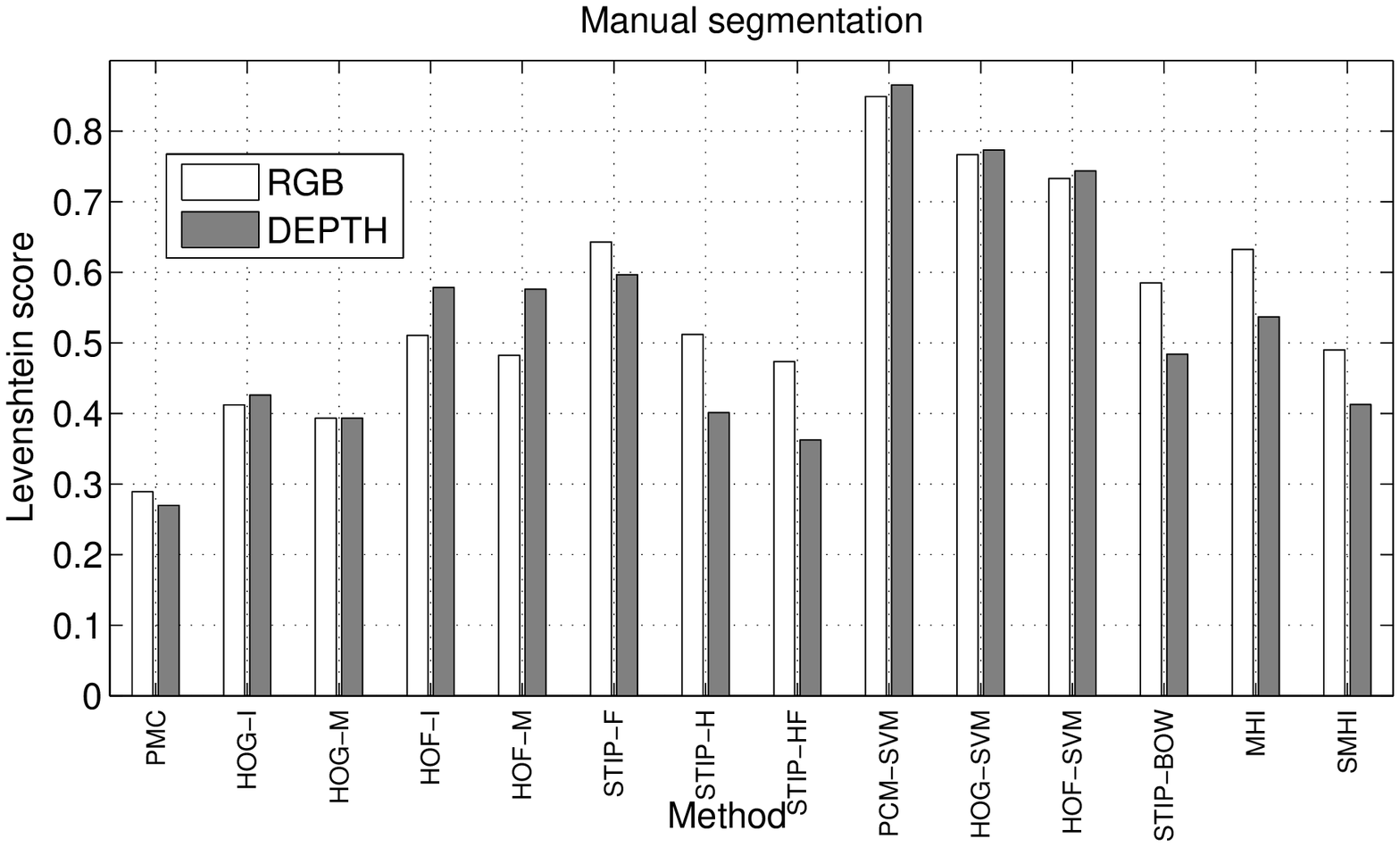}
  \caption{Levenshtein score for the methods of Table~\ref{tab:comparisonimps} in the Development01-20 data set.}\label{fig:comparison_1}
\end{figure}

The different bag-of-frames representations were used for gesture recognition under the proposed PCA-based reconstruction-error technique. Also, we evaluated the recognition performance of supervised approaches using the same representations. For these methods, each vector of features (either motion maps, HOG, HOF, of 3D-HOG/HOF) is treated as an instance of a classification problem, where the class of the instance is the gesture from which the corresponding vector was extracted. In preliminary experimentation we tried several classification methods including (linear discriminant analysis, neural networks, random forest, etc.), we report results for the best methods we found.  For motion and static-motion history images we used a template matching approach for recognition (correlation).
Experimental results obtained with the considered variants and with the principal motion components approach are shown in Figure~\ref{fig:comparison_1}.

From Figure~\ref{fig:comparison_1} it can be seen that principal motion components obtains the best performance for all but one of the configurations. HOG-M obtained the best results when using automatic segmentation and RGB video, the relative improvement was of $1.8\%$. This result indicates the suitability of the reconstruction approach for one-shot gesture recognition under the bag-of-frames representation, which is not tied to a particular type of features. In fact, when using automatic segmentation the three methods: HOG-M, HOG-I and PMC obtained very similar results.

When manual segmentation was used, our approach outperformed the other methods by a considerable margin. The improvement over the nearest technique in performance (HOG-M) was of $36.12\%$ and $45.9\%$ for RGB and depth video, respectively. The widely used STIP features were not very useful for gesture recognition under neither the bag-of-frames nor the bag-of-visual-words formulations. This can be due to the fact that a single video is not enough to capture discriminative features. Actually, none of the supervised approaches to one-shot-gesture recognition performed decently. This is not surprising as we are using as labeled samples to features that may have high overlap with several gestures. It is interesting that the static history images outperformed the standard MHI technique~\citep{MHI}.

Finally, we also compare the performance of principal motion components to that obtained by other authors that have used the ChaLearn Gesture Dataset~\cite{guyon_cgd}. We considered for this comparison methods that have been already described in a scientific publication. The performance of the considered methods as well as a brief description for each of them can be seen in Table~\ref{tab:comparisonRW2}.
\begin{table}[bht]
\begin{center}
\caption{Description of the published methods for one-shot gesture recognition we consider for comparison. }\label{tab:comparisonRW2}
\footnotesize{
\begin{tabular}{|l|l|l|l|}
\hline
\textbf{ID}&\textbf{Description}&\textbf{LS}&\textbf{Reference}\\\hline
\multicolumn{4}{|c|}{\textbf{Devel01-20}}\\\hline
MLS-Wu& Multi-layer Template+DTW & 0.1950 & \cite{wushen}\\
GM-MM&Graphical model & 0.2400 &\cite{manavender}\\
TM-Wu&Template matching & 0.2600 & \cite{wuetal}\\
PMC-M& PMC  / Manual segmentation & 0.2696 & - \\
MF-LIU&Manifold learning & 0.2873 & \cite{YML}\\
PMC-A& PMC / Automatic segmentation & 0.2890 & - \\
TM-Mahbub&Template matching & 0.3746 & \cite{mahbu}\\
TM-2-Mahbub&Template matching & 0.3125 & \cite{mahbu2}\\\hline
\multicolumn{4}{|c|}{\textbf{Valid01-20}}\\\hline
GM-MM&Graphical model & 0.2332 &\cite{manavender}\\
TM-Wu&Template matching & 0.2968 & \cite{wuetal}\\
PMC-A& PMC / Automatic segmentation & 0.3178 & - \\\hline
\end{tabular}
}
\end{center}
\end{table}

It can be observed from Table~\ref{tab:comparisonRW2} that the performance of the proposed approach is competitive with that obtained by the different methods. The best performance reported so far in a scientific publication is that reported by~\cite{wushen}. It is interesting that such method uses principal motion components as a preliminary step in their multi-layer architecture. Roughly, our method is used to determine if a gesture is dynamic or static. Dynamic gestures are treated with a method based on particle filtering and a tailored dynamic time warping; static ones, are processed with a novel method that incorporates contextual information.

The performance of our automatic approach is close to that obtained by~\cite{manavender} and~\cite{YML}. The former authors implemented a graphical model inspired in hidden-Markov models that have been used for keyword spotting, both modalities (RGB and depth video)  are used by the model. On the other hand, \cite{YML} represents videos with using a method based on higher-order singular value decomposition, recognition is done via least-squares regression for manifolds. Both approaches obtained outstanding performance in state-of-the-art data sets for human activity recognition and standard gesture recognition, besides they achieved acceptable results in data from ChaLearn Gesture challenge. The principal motion components approach obtained comparable performance to that techniques, hence, it is worth exploring the performance of our method on other closely-related tasks.

Regarding the ChaLearn Gesture Challenge, the latest version of principal motion components would be ranked $9^{th}$ and $7^{th}$ in stages one\footnote{http://www.kaggle.com/c/GestureChallenge/leaderboard} and two\footnote{http://www.kaggle.com/c/GestureChallenge2/leaderboard}, respectively. Principal motion components was proposed as a baseline method, whose simplicity and easy of implementation motivated participants to develop better methods. In this aspect we accomplished our goal and exceed it by motivating other researchers to build better methods on top of our proposal.
\section{Conclusions}
\label{sec:conclusions}
We introduced a novel gesture recognition approach for the one-shot learning setting called Principal Motion Components. The proposed approach represents the frames of a video by means of maps that account the amount of motion happening in spatial regions of the video. The bag of motion maps is used with a PCA-based recognition approach in which recognition error is used as a measure of gesture affinity.

We report experimental results in a large data set with $54,0000$ gestures, and two video modalities. Experimental results show that the proposed approach is very competitive, despite being simple and very efficient. The proposed method can work with RGB or depth video and obtain comparable performance. Likewise, the performance of the method does not degrade significantly when using manual or automatic gesture segmentation. We compare the performance of our approach to alternative methods we implemented ourselves and those reported by other researchers. Our approach compared favorably with some techniques and obtained close performance to others.
We analyze the performance of our approach under different parameter settings and show characteristics of gestures that can be effectively recognized with it. This study revealed that the proposed approach is well suited for highly dynamic gestures.

There are several future work directions we would like to explore. First, we would like to study the suitability of the principal motion components approach for related tasks, including gesture segmentation, keyframe extraction and motion-based retrieval. Also, we are interested in developing alternative recognition methods that use the bag-of-frames representation. Other interesting areas for research include developing a hierarchical principal motion components formulation, and extending the proposed representation to spatiotemporal features.

\bibliographystyle{elsarticle-harv}
\bibliography{principal_motion}







\end{document}